%% file: aaai24.tex
\documentclass[letterpaper]{article} 
\usepackage{aaai24}  
\usepackage{times}  
\usepackage{helvet}  
\usepackage{courier}  
\usepackage[hyphens]{url}  
\usepackage{graphicx} 
\usepackage{natbib}  
\usepackage{caption} 
\newif\ifarxiv
\arxivtrue
\usepackage[titletoc,page]{appendix}

\usepackage{algorithm}
\usepackage{algorithmic}
\usepackage{xcolor}    
\usepackage{amsmath}
\usepackage{amssymb}
\usepackage{relsize}
\usepackage{gensymb}
\usepackage{multirow}
\usepackage{paralist}     
\usepackage{siunitx}  
\usepackage{subcaption}
\usepackage{pifont}
\usepackage[algo2e,ruled,linesnumbered]{algorithm2e}

\usepackage{makecell}

\usepackage{newfloat}
\usepackage{listings}
\DeclareCaptionStyle{ruled}{labelfont=normalfont,labelsep=colon,strut=off} 
\floatstyle{ruled}
\newfloat{listing}{tb}{lst}{}
\floatname{listing}{Listing}
\newcommand{\cmark}{\ding{51}}%
\newcommand{\xmark}{\ding{55}}%

\renewcommand{\vec}[1]{\mathbf{#1}}
\newcommand{\set}[1]{\mathcal{#1}}
\newcommand{\mat}[1]{\mathbf{#1}}

\DeclareMathOperator*{\argmax}{arg\,max}
\DeclareMathOperator*{\argmin}{arg\,min}
\newcommand{\ie}{i.e.\ }

\newcommand{\eg}{e.g.\ }
\newcommand{\wrt}{w.r.t.\ }
\newcommand*{\tran}{^{\mkern-1.5mu\mathsf{T}}}
\newcommand*{\negtran}{^{\mkern-1.5mu\mathsf{-T}}}
\newcommand*{\inverse}{^{\mkern-1.5mu{-1}}}
\newcommand*{\crossmat}[1]{[{#1}]_{\times}}
\newcommand{\tssdagger}{\textsuperscript{\textdagger}}
\newcommand{\tssstar}{\textsuperscript{*}}

\newcommand{\tssddagger}{\textsuperscript{\textdaggerdbl}}
\newcommand{\tssstard}{\textsuperscript{**}}

\newcommand{\iverson}[1]{\left[#1\right]}

\newcommand{\aucone}{AUC @ $1\degree$}

\newcommand{\aucfive}{AUC @ $5\degree$}

\newcommand{\emptyseq}{(\,)}
\newcommand{\gt}[1]{\hat{#1}}

\newcommand{\obsv}{\vec{x}}
\newcommand{\obsvs}{\set{X}}
\newcommand{\numobsvs}{N}

\newcommand{\labels}{\set{Y}}
\newcommand{\gtlabels}{\set{\gt{Y}}}

\newcommand{\model}{\vec{h}}
\renewcommand{\models}{\set{M}}
\newcommand{\nummodels}{M}

\newcommand{\hypothesis}{\vec{h}'}
\newcommand{\hypotheses}{\set{S}}
\newcommand{\numhypotheses}{S}

\newcommand{\hypsets}{\set{H}}

\newcommand{\gtmodel}{\gt{\model}}
\newcommand{\gtmodels}{\gt{\models}}
\newcommand{\numgtmodels}{|\gtmodels|}

\newcommand{\pmodel}{\model^*}
\newcommand{\pmodels}{\models^*}
\newcommand{\numpmodels}{M^*}

\newcommand{\numsamples}{K}
\newcommand{\numhypsamples}{\tilde{K}}
\newcommand{\minsolver}{f_{\mathsf{S}}}
\newcommand{\minimalset}{\set{C}}

\newcommand{\nnw}{\vec{w}}

\newcommand{\psweight}{p(\obsv| j; \obsvs, \nnw)}

\newcommand{\sampleweight}{p(\obsv_i| j; \obsvs, \nnw)}
\newcommand{\inlierweight}{q(j,\obsv_i; \obsvs,\nnw)}
\newcommand{\outlierweight}{q(\varnothing,\obsv_i; \obsvs,\nnw)}
\newcommand{\inliercount}{I_{\text{w}}(\model_{j,k}';\obsvs,\nnw)}

\ifarxiv
\newcommand{\githubfootnote}{\footnote{Code, pre-trained models and datasets are available at: \texttt{https://github.com/fkluger/parsac}}}
\newcommand{\supp}{appendix~}
\else
\newcommand{\githubfootnote}{\footnote{Supplementary material, code and datasets are available at: \texttt{https://github.com/fkluger/parsac}}}
\newcommand{\supp}{supplementary~}
\fi

\title{PARSAC: \\Accelerating Robust Multi-Model Fitting with Parallel Sample Consensus}
\author{
Florian Kluger, Bodo Rosenhahn\\
}
\affiliations{
    Leibniz University Hannover\\
    {kluger@tnt.uni-hannover.de}
}

\begin{document}

\maketitle

\begin{abstract}
We present a real-time method for robust estimation of multiple instances of geometric models from noisy data.
Geometric models such as vanishing points, planar homographies or fundamental matrices are essential for 3D scene analysis.
Previous approaches discover distinct model instances in an iterative manner, thus limiting their potential for speedup via parallel computation.
In contrast, our method detects all model instances independently and in parallel.
A neural network segments the input data into clusters representing potential model instances by predicting multiple sets of sample and inlier weights.
Using the predicted weights, we determine the model parameters for each potential instance separately in a RANSAC-like fashion.
We train the neural network via task-specific loss functions, \ie we do not require a ground-truth segmentation of the input data.
As suitable training data for homography and fundamental matrix fitting is scarce, we additionally present two new synthetic datasets.
We demonstrate state-of-the-art performance on these as well as multiple established datasets,
with inference times as small as five milliseconds per image.
\end{abstract}

\section{Introduction}

In Computer Vision, we commonly aim to explain high-dimensional and noisy observations using low-dimensional geometric models.
These models can give important insights into the structure of our data, and are crucial for 3D scene analysis and reconstruction.
For example, by fitting vanishing points to a set of 2D line segments, we can deduce information about the 3D layout of a scene from a single view~\cite{zhou2019learning, zou2018layoutnet}.
Similarly, fitting homographies to point correspondences of image pairs allows us to reason about dominant 3D planes within a scene.
Robust estimation of fundamental matrices is indispensable for applications such as SLAM~\cite{orbslam22017} and two-view motion segmentation~\cite{ozbay2022fast}.
We typically obtain the features for model fitting using low-level algorithms, such as SIFT~\cite{Lowesift} and LSD~\cite{von2008lsd}.
As these methods are imperfect, model fitting algorithms must be able to identify erroneous features (outliers) in order to robustly determine model parameters using valid features (inliers) only~\cite{ransac1981}.
Given only one instance of a geometric model in our data, we thus have to segment our data into two groups: inliers and outliers.
However, as the number of model instances increases, the number of data groups also increases, and inliers of one model act as pseudo-outliers for all other models.
Consequently, robustly estimating the parameters of all models becomes more difficult, especially if the number of model instances is unknown beforehand.

\begin{figure}	
\centering
\includegraphics[width=\linewidth]{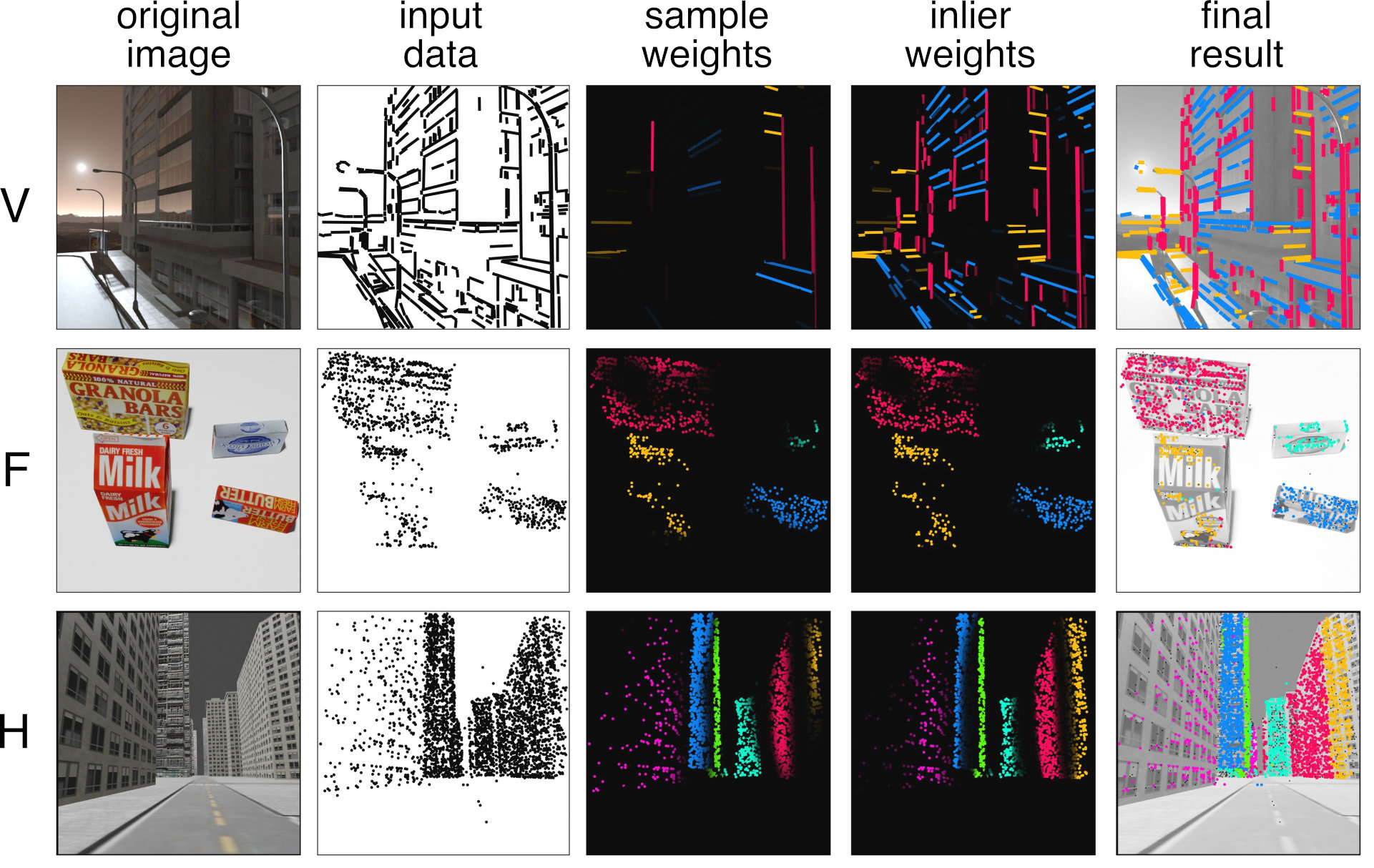}
\caption{ {Applications:} PARSAC estimates multiple vanishing points (V, top), fundamental matrices (F, middle) or homographies (H, bottom). We visualise distinct model instances using different colour hues. Brightness in columns three and four is proportional to the corresponding weight.
}
\label{fig:teaser}
\end{figure}   

Early approaches solve this problem by repeatedly applying a robust estimator like RANSAC~\cite{ransac1981}.
After each step, they remove all data points associated with previously predicted models~\cite{vincent2001seqransac}.
More recent methods instead generate model hypotheses beforehand, then use clustering and optimisation techniques in order to assign data points to a model or the outlier class~\cite{barath2018multi, barath2016multi, pham2014interacting, amayo2018geometric, isack2012energy,toldo2008robust,magri2014t,magri2015robust,magri2016multiple,magri2019fitting,chin2009robust}. 
Hybrid approaches combine the strengths of both methodologies by interleaving multiple sampling, clustering and optimisation steps~\cite{barath2019progressive, barath2021progressive}.
Being the first approach which applies deep learning to robust multi-model fitting, CONSAC~\cite{kluger2020consac} uses a neural network to guide hypothesis sampling in a sequential model fitting pipeline.
It achieves high accuracy but is computationally costly, since its sequential approach results in a large number of neural network forward passes.

In this paper, we propose a method that predicts sample weights for all model instances \emph{simultaneously}, thus overcoming the computational bottleneck of methods such as CONSAC.
In addition to sample weights for hypothesis \emph{sampling}, we also predict inlier weights for hypothesis \emph{selection}.
Using these weights, we recover model instances in a RANSAC-like fashion independently and \emph{in parallel}.
This translates into a significant speedup in practice: our method achieves inference times as small as five milliseconds per image.
It is considerably faster than its competitors while still achieving state-of-the-art accuracy.
In a nod to RANSAC, we call our method PARSAC: \emph{Parallel Sample Consensus}.
As shown in Fig.~\ref{fig:teaser}, PARSAC can be used for various applications.

Since PARSAC is a machine learning based approach, we require suitable training data.
For vanishing point estimation, we have seen an emergence of new large-scale datasets, such as SU3~\cite{zhou2019learning} and NYU-VP~\cite{kluger2020consac}, in recent years.
For fundamental matrix and homography estimation, however, AdelaideRMF~\cite{wong2011dynamic} is still the only publicly available dataset.
It consists of merely 38 labelled image pairs, with no separate training set.
This poses the danger that researchers inadvertently overfit their algorithms on this small dataset, because there is no other data to verify their performance on.
In order to alleviate this issue, we present two new synthetic datasets: \emph{HOPE-F} for fundamental matrix fitting and \emph{Synthetic Metropolis Homographies (SMH)} for homography fitting.

\noindent In summary, our \textbf{main contributions}\githubfootnote are:
\begin{itemize}
\item PARSAC, the first learning-based \emph{real-time} method for robust multi-model fitting. It uses a neural network to segment the data into clusters based on model instance affinity in a single forward pass. This segmentation allows for a parallelised discovery of geometric models. 
\item Two new large-scale synthetic datasets -- \emph{HOPE-F} and \emph{SMH} -- for fundamental matrix and homography fitting. They are the first datasets to provide sufficient training data for supervised learning of multiple fundamental matrix or planar homography fitting.
\item We achieve state-of-the-art results for vanishing point estimation, fundamental matrix estimation and homography estimation on multiple datasets. 
\item PARSAC is significantly faster than its competitors. Requiring just five milliseconds per image for vanishing point estimation, it is five times faster than the second fastest method, and 25 times faster than the fastest competitor with comparable accuracy.
\end{itemize}

\section{Related Work}
\subsection{Multi-Model Fitting}
Robust model fitting is a technique widely used for estimating geometric models from data contaminated with outliers.
The most commonly used approach, RANSAC~\cite{ransac1981}, randomly samples minimal sets of observations to generate model hypotheses, and selects the hypothesis with the largest consensus set, \ie observations which are inliers.
While effective in the single-instance case, RANSAC fails if multiple model instances are apparent in the data. 
In this case, we need to apply robust \emph{multi-model} fitting techniques.
Sequential RANSAC~\cite{vincent2001seqransac} fits multiple models sequentially by applying RANSAC repeatedly, removing inliers from the set of all observations after each iteration.
More recent methods, such as PEARL~\cite{isack2012energy}, instead optimise a global mixed-integer cost function~\cite{rosenhahn2023} initialised via a stochastic sampling, in order to fit multiple models simultaneously~\cite{pham2014interacting, amayo2018geometric}. 
Multi-X~\cite{barath2018multi} generalises this methodology to \emph{multi-class} problems, \ie cases where multiple models of possibly different types may fit the data.
As a hybrid approach, Progressive-X~\cite{barath2019progressive} guides hypothesis generation via intermediate estimates by interleaving sampling and optimisation steps. 
J- and T-Linkage~\cite{toldo2008robust,magri2014t} both use preference analysis~\cite{chin2009robust} to cluster observations agglomeratively, but use different methods to define the preference sets.
MCT~\cite{magri2019fitting} is a multi-class generalisation of T-Linkage, while RPA~\cite{magri2015robust} uses spectral instead of agglomerative clustering. 
Progressive-X+~\cite{barath2021progressive} combines Progressive-X with preference analysis and allows observations to be assigned to multiple models. It is faster and more accurate than its predecessor.
Fast-CP~\cite{ozbay2022fast} -- specifically tailored for fundamental matrix fitting -- utilises Christoffel polynomials in order to segment observations into clusters belonging to the same model.
Most recently, the authors of~\cite{farina2023quantum} utilise quantum computing for multi-model fitting.
While seminal in this regard, they make strong assumptions such as the absence of true outliers.
CONSAC~\cite{kluger2020consac} is the first approach to utilise deep learning for multi-model fitting.
Using a neural network to guide the hypothesis sampling, it discovers model instances sequentially and predicts new sampling weights after each step.
This technique provides high accuracy, but is computationally expensive due to its sequential operation and multi-hypothesis sampling.
While PARSAC also leverages deep learning, we are able to decouple the discovery of individual model instances, reducing run-time by two orders of magnitude.

\begin{figure*}	
\centering
\includegraphics[width=0.8\linewidth]{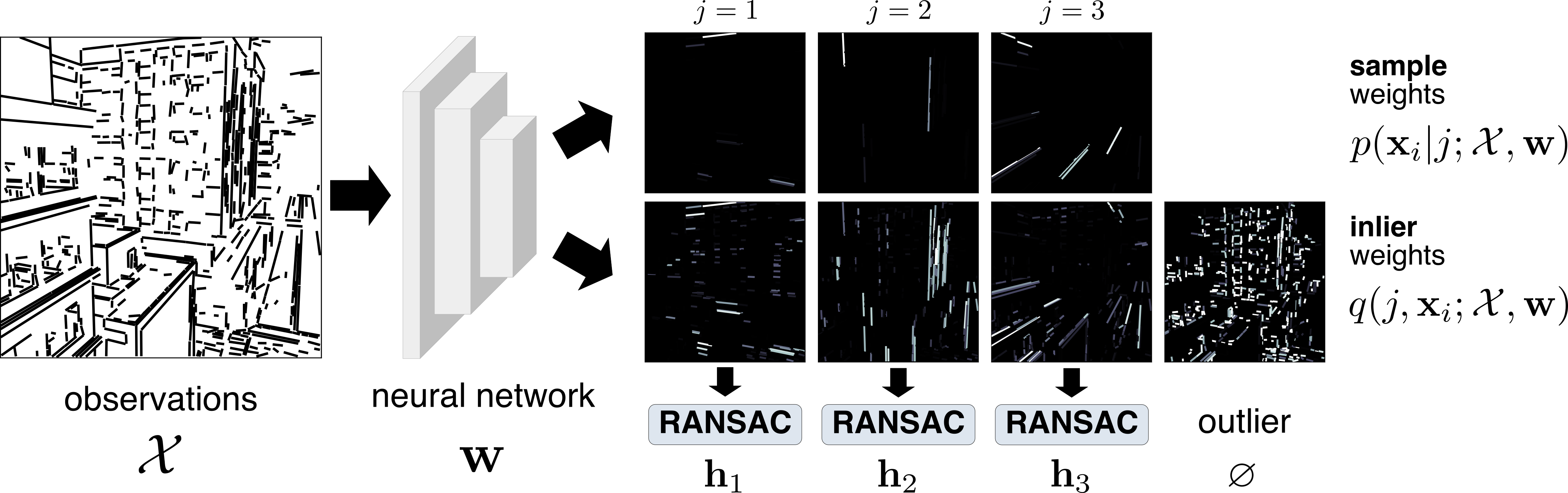}
\caption{ {PARSAC Overview:} 
Given observations $\obsvs$, \eg line segments or point correspondences, we predict sample weights $p$ and inlier weights $q$ for each observation and putative geometric model using a neural network. 
For each putative geometric model $j$, we independently sample model hypotheses in a RANSAC-like fashion, using the predicted sample weights.
We then select the best models which have the largest weighted inlier counts, using the predicted inlier weights.
An additional set of weights captures potential outliers.
}
\label{fig:sysfig}
\end{figure*}   

\subsection{Vanishing Point Estimation}
Although multiple vanishing point (VP) estimation can be described as a multi-model fitting problem, algorithms outside of this genre have also been designed to tackle this task specifically~\cite{antunes2013global, barinova2010geometric, kluger2017deep, lezama2014finding, simon2018contrario, tardif2009non, vedaldi2012self, wildenauer2012robust, xu2013minimum, zhai2016detecting, zhou2019neurvps, liu2021vapid, wu2021real, lin2022deep}. 
Unlike more general multi-model fitting methods, they usually exploit additional, domain-specific knowledge. 
Zhai et al. first predict a horizon line~\cite{kluger2020temporally} from the RGB image via a convolutional neural network (CNN), on which they then condition their VP estimates. 
Simon et al. also condition some VPs on the horizon line, as well as on the zenith VP. 
Kluger et al.  predict initial VP estimates via a CNN, and refine them using an expectation maximisation~\cite{dempster1977maximum} algorithm. 
Zhou et al.  propose a CNN with a conic convolution operator in order to find VPs for Manhattan-world scenes.
Lin et al.  incorporate a Hough transform and a Gaussian sphere mapping into their CNN in order to exploit geometric priors induced by the VP fitting problem.
General purpose robust multi-model fitting algorithms, such as our proposed PARSAC, do not rely on such domain-specific priors.

\section{Method}
\label{sec:method}
From a set of $\numobsvs$ observations $\obsv_i \in \obsvs$, contaminated with noise and outliers, we seek to detect $\nummodels$ instances of a model $\model$ that describe our data.
As visualised in Fig.~\ref{fig:sysfig}, PARSAC discovers the model instances $\models = (\model_1, \dots, \model_{\nummodels})$ in parallel, guided by a neural network with parameters $\nnw$:
\begin{compactenum}
\item For each observation $\obsv_i$ and putative model instance $\model_j$, the network predicts $\numpmodels \geq \nummodels$ sample weights $\sampleweight$ and inlier weights $\inlierweight$. 
\item We generate a set of putative model instances  $\pmodels = \{\model_1, \dots, \model_{\numpmodels}\}$ in parallel in a RANSAC-like fashion. The sample weights guide the sampling of hypotheses, and we perform hypothesis selection via \emph{weighted} inlier counting using the inlier weights.
\item We greedily rank the putative instances based on unique and overlapping inlier counts and discard models with an insufficient amount of inliers, giving us the final sequence of model instances $\models \subseteq \pmodels$.
\end{compactenum}

\paragraph{Sample and Inlier Weight Prediction}
For each observation $\obsv_i \in \obsvs$ and putative model $\model_j \in \pmodels$, we predict sample weights $\sampleweight$ and inlier weights $\inlierweight$. 
In addition, we predict outlier weights $\outlierweight$.
A neural network with parameters $\nnw$ computes two weight matrices $\mat{P}, \mat{Q}$ with size $\numobsvs \times \numpmodels+1$.
By normalising over the first or second dimension, respectively, we obtain the sample and inlier weights:
\begin{equation}
    \sampleweight = \frac{\mat{P}_{i,j}}{\sum_{k=1}^{\numobsvs} \mat{P}_{k,j} } \, , \quad
\end{equation}
\begin{equation}
    \inlierweight = \frac{\mat{Q}_{i,j}}{\sum_{k=1}^{{\numpmodels}+1} \mat{Q}_{i,k} } \, .
\end{equation}
Sample weights $\sampleweight$ determine how likely an observation $\obsv_i$ is to be sampled for generating a model hypothesis for the putative instance $\model_j$,
while inlier weights $\inlierweight$ predict whether an observation $\obsv_i$ is an inlier of putative instance $\model_j$.
The additional outlier weights:
\begin{equation}
    \outlierweight = q({\numpmodels}+1 , \obsv_i; \obsvs, \nnw) \, ,
\end{equation}
are not used for hypothesis sampling or selection, but allow the neural network to remove outliers from the putative models so that they do not interfere with the following steps.

\paragraph{Parallel Hypothesis Sampling and Selection}
For each putative model $\model_j$, we sample ${\numhypotheses}$ minimal sets of observations $\minimalset = \{\obsv_1, \dots, \obsv_C\} \subset \obsvs$ independently and in parallel, using the predicted sample weights $\sampleweight$.
A minimal solver $\minsolver$ then computes parameters of a model hypothesis $\hypothesis = \minsolver(\minimalset) $ for each minimal set.
We thus get a set of model hypotheses $\hypotheses_j = \{ \hypothesis_{j,1}, \dots,  \hypothesis_{j,{\numhypotheses}}\}$ for each putative model.
Given a distance metric $d(\obsv, \model)$ measuring the error between an observation $\obsv$ and a geometric model $\model$, we then compute the \emph{weighted} inlier counts of all hypotheses:
\begin{equation}
    \inliercount = \sum_{\obsv_i \in \obsvs} s(d(\obsv_i, \hypothesis_{j,k})) \cdot \inlierweight \, ,
    \label{eq:weighted_inlier_count}
\end{equation}
using the predicted inlier weights and an inlier scoring function $s(\cdot) \in [0, 1]$.
We implement $s(\cdot)$ as a soft inlier measure, with inlier threshold $\tau$ and softness parameter $\beta$, as proposed by~\cite{brachmann2018lessmore}.
During inference, we then select the hypothesis with the largest weighted inlier count:
\begin{equation}
    \model_j = \argmax_{\hypothesis_{j,k} \in \hypotheses_j} \inliercount \, .
    \label{eq:hyp_selection_argmax}
\end{equation}
Combining the results of all putative models, we obtain the set of putative model instances:
\begin{equation}
     \pmodels = \{\model_1, \dots, \model_{\numpmodels}\} \, .
\end{equation}
Using the weighted inlier counts to select the best hypothesis for each putative model is crucial.
If we were to use the unweighted inlier count instead, as in~\cite{kluger2020consac} or vanilla RANSAC, we would more likely select very similar hypotheses for every putative model, \ie the hypotheses with the largest consensus sets within all observations $\obsvs$.
We would thus ignore other less prominent but valid models that describe our data.
Instead, our weighted inlier count ensures that we get a more diverse set of model instances if our neural network is able to segment the observations in a sufficiently meaningful way.

\paragraph{Instance Ranking}
We extract the final sequence of model instances $\models$ from the putative models $\pmodels$ by greedily ranking the models in $\pmodels$.
Initially, the model sequence $\models = \emptyseq$ and the set of current inliers $\set{I} = \varnothing$ are empty.
Iteratively, we determine the number of unique inliers $I^{\text{u}}_{\model}$ and overlapping inliers $I^{\text{o}}_{\model}$ of each model  $\model \in \pmodels$ \wrt $\set{I}$:
\begin{equation}
    I^{\text{u}}_{\model} = | \set{I}_{\model} \setminus \set{I}| \, , \quad
    I^{\text{o}}_{\model} = | \set{I}_{\model} \cap \set{I}| \, ,
\end{equation}
with $\set{I}_{{\model}} = \{ \obsv \, | \, d(\obsv, \model) < \tau,\, \obsv \in \obsvs \}$ being the set of inliers for $\model$ using inlier threshold $\tau$.
We then select the model which maximises unique and minimises overlapping inliers: 
$ \model = \argmax_{\pmodel \in (\pmodels \setminus \models)} I^{\text{u}}_{\pmodel}-I^{\text{o}}_{\pmodel} $.

If the number of unique inliers is larger than the number of overlapping inliers by at least the minimal set size, \ie $I^{\text{u}}_{\model}-I^{\text{o}}_{\model} \geq C$, we append $\model$ to $\models$ and update $\set{I}$, and repeat the process: $ \models \leftarrow \models \frown (\model) \, , \quad \set{I} \leftarrow \set{I} \cup \set{I}_{\model} \, .$

\paragraph{Cluster Assignment}
After ranking model instances, we cluster observations $\obsv \in \obsvs$ by model affinity.
Starting with the first model $\model_1 \in \models$, we sequentially assign observation $\obsv$ to a model $\model_j$ if
either $d(\obsv, \model_j) < \tau$ and $\model_j \in \argmin_{\model \in \models} d(\obsv, \model)$, or
$d(\obsv, \model_j) < \tau_a$ and $\obsv$ is not yet assigned to a higher ranked model $\model_k$ with $k<j$.
Here, $\tau$ denotes the inlier threshold and $\tau_a$ is the assignment threshold $\tau_a > \tau$.
Observations which are not assigned to any model are labelled as outliers.
The result is a set of cluster labels $\labels$.

\subsection{Neural Network Training}
We seek to optimise neural network parameters $\nnw$ such that we obtain geometric model instances $\models$ which accurately describe our data $\obsvs$.
Similarly to~\cite{brachmann2016,brachmann2019neural,kluger2020consac,kluger2021cuboids}, we minimise the expected value of a task loss $\ell(\models)$ which measures the quality of the estimated model instances:
\begin{equation}
    \mathcal{L}(\nnw) = \mathbb{E}_{\models \sim p(\models|\obsvs ; \nnw)} \left[ \ell ({\models}) \right] \, .
    \label{eq:exp_task_loss}
\end{equation}
In order to facilitate this, we must turn the deterministic hypothesis selection of Eq.~\ref{eq:hyp_selection_argmax} into a probabilistic action $\model_j \sim p(\model_j|\hypotheses_j)$.
Otherwise, we would not be able to compute gradients of $\mathcal{L}$ \wrt inlier weights $\inlierweight$.
Based on~\cite{brachmann2019neural}, we model the likelihood of selecting a hypothesis $\hypothesis$ from a set of hypotheses $\hypotheses$ as the softmax over weighted inlier counts $I_{\text{w}}$:
\begin{equation}
    p(\model_j|\hypotheses_j; \obsvs, \nnw) = 
    \frac{\exp{(\alpha_s \cdot I_{\text{w}}(\model_j; \obsvs, \nnw) )}}{\sum_{\hypothesis_{j,k} \in \hypotheses_j} \exp{(\alpha_s \cdot \inliercount)}}  \, ,
    \nonumber
\end{equation}
with $\alpha_s$ being a scaling factor which controls how discriminating the resulting categorical distribution becomes.
We thus consider the joint probability of first sampling hypotheses sets $\hypsets = \{\hypotheses_1, \dots, \hypotheses_{\numpmodels}\}$ and then model instances $\models$:
\begin{equation}
\begin{split}
& p(\models,\hypsets |\obsvs, \nnw)  = p(\models | \hypsets; \obsvs, \nnw) \cdot  p(\hypsets | \obsvs, \nnw) \, , \\
& \text{with} \quad  p(\models | \hypsets; \obsvs, \nnw) = \prod_{j=1}^{\numpmodels} p(\model_j|\hypotheses_j; \obsvs, \nnw) \, , \\
& \text{with} \quad  p(\hypsets | \obsvs , \nnw) = \prod_{j = 1}^{\numpmodels} p(\hypotheses_j | \obsvs , \nnw) \,  ,\\
& \text{with} \quad  p(\hypotheses_j | \obsvs , \nnw)  = \prod_{\hypothesis_{j,k} \in \hypotheses_j} p(\hypothesis_{j,k} | \obsvs , \nnw) \, , \\ 
& \text{with} \quad  p(\hypothesis_{j,k} | \obsvs , \nnw)  = \prod_{\obsv_i \in \minimalset_{j,k}} \sampleweight \,  
.
\nonumber
\end{split}
\end{equation}
Using~\cite{schulman2015gradient}, we hence define the gradients of the expected task loss \wrt network parameters $\nnw$ as:
\begin{equation}
    \frac{\partial}{\partial \nnw}  \mathcal{L}(\nnw) = \mathbb{E}_{\models,\hypsets} \left[ \ell ({\models}) \frac{\partial}{\partial \nnw} \log p(\models,\hypsets |\obsvs, \nnw) \right] \, ,
    \label{eq:exp_task_loss_grad}
\end{equation}
which we approximate by first drawing $\numsamples$ samples of hypotheses sets $\hypotheses_j \sim p(\hypotheses_j | \obsvs , \nnw)$ for each putative model $j$, and then sampling $\numhypsamples$ samples of $\models \sim p(\models  | \hypsets; \obsvs, \nnw)$:
\begin{equation}
     \frac{\partial}{\partial \nnw}  \mathcal{L}(\nnw) \approx 
    \frac{1}{\numsamples \numhypsamples} \sum_{i=1}^{\numsamples}\sum_{j=1}^{\numhypsamples} 
    \ell ({\models_{ij}}) \frac{\partial}{\partial \nnw} \log p(\cdot) 
     \, ,
    \nonumber
\end{equation}
with $p(\cdot) = p(\models_{ij},\hypsets_i|\obsvs, \nnw)$.
As in~\cite{brachmann2019neural}, we subtract the mean of $\ell(\cdot)$ as a baseline to reduce the variance of the gradient.
\paragraph{Task Loss}
The type of loss we use depends on the task we seek to solve, as well as the type of ground truth data available.
If we want to optimise the parameters of our models $\models$ \wrt a set of ground truth models $\gtmodels$, we utilise a task specific loss $\ell_{\text{s}}(\model, \gtmodel)$ which measures the error between each ground truth and estimated model.
Using this loss, we determine a cost matrix $\vec{C}$, with $C_{ij} = \ell_{\text{s}}(\model_i, \gtmodel_j)$.
If $|\models| > |\gtmodels|$, we only consider the first $|\gtmodels|$ models in $\models$.
We then define the final task loss as the minimal assignment cost via the Hungarian method~\cite{kuhn1955hungarian} $h(\cdot)$:
\begin{equation}
    \ell(\models) = h(\vec{C}).
\end{equation}
Alternatively, if our ground truth consists of cluster labels $\gtlabels$, we compute the misclassification error (ME) \wrt our predicted cluster assignments $\labels$ which arise from $\models$:
\begin{equation}
    \ell(\models) = \mathrm{ME}(\labels,\gtlabels) \, .
\end{equation}
Eq.~\ref{eq:exp_task_loss_grad} does not require the task loss $\ell(\cdot)$ to be differentiable.

\begin{figure*}	
    \newlength{\imgwidth}
    \setlength{\imgwidth}{0.245\linewidth}
    \begin{subfigure}{0.5\textwidth}
    \centering
        \includegraphics[width=\imgwidth]{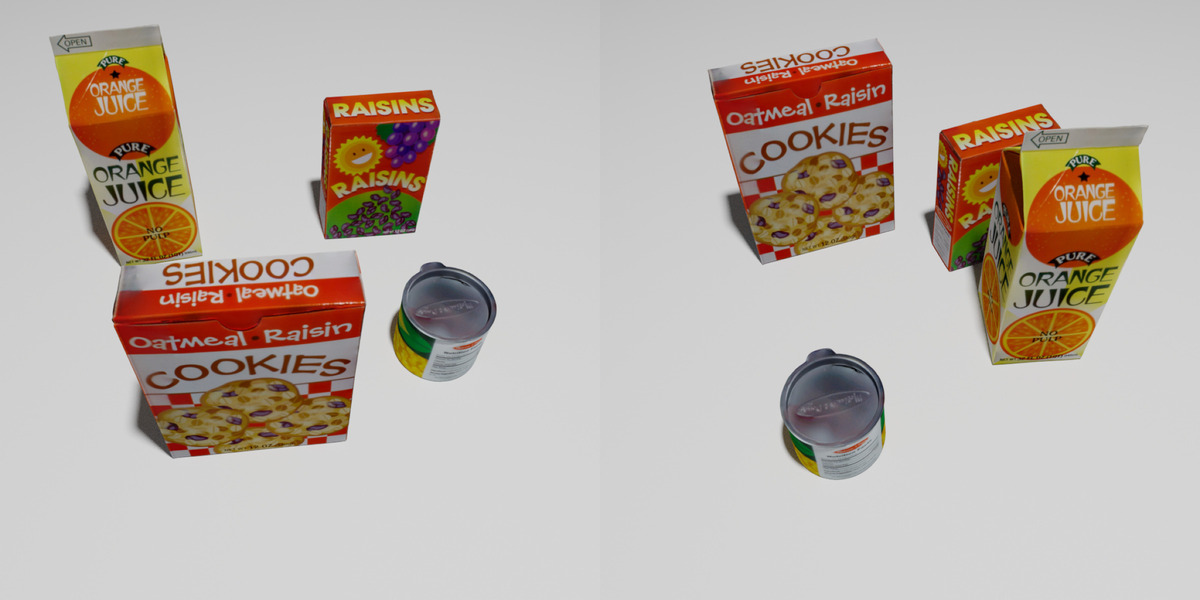}%
        \includegraphics[width=\imgwidth]{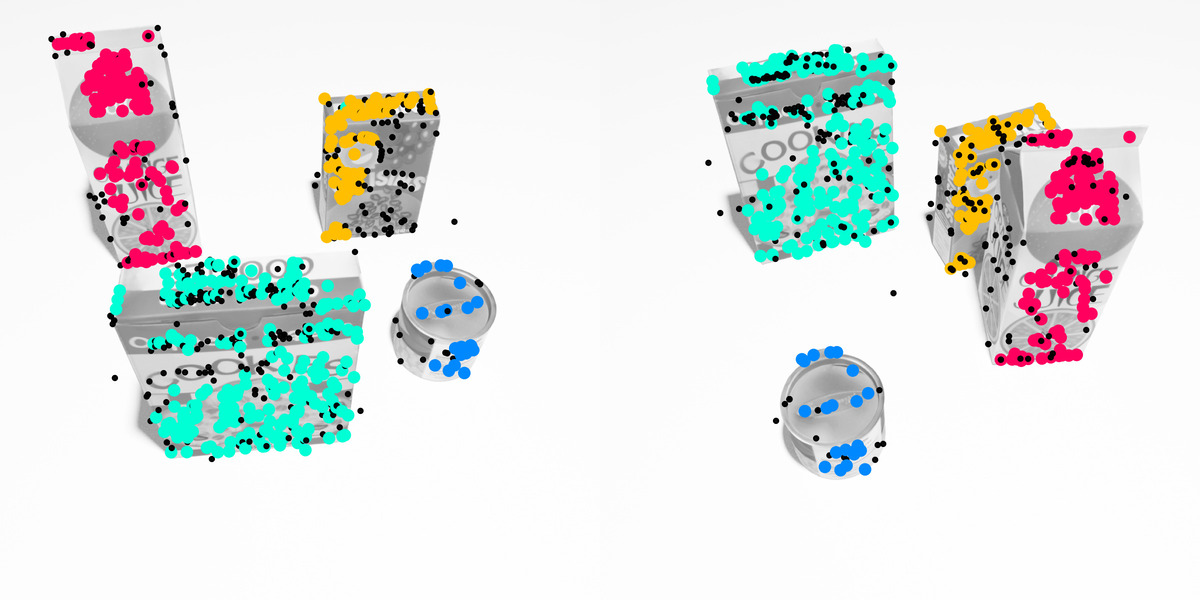}%
        \caption{HOPE-F Dataset}
        \label{fig:hope_dataset_examples}
    \end{subfigure}%
    \begin{subfigure}{0.5\textwidth}
    \centering
        \includegraphics[width=\imgwidth]{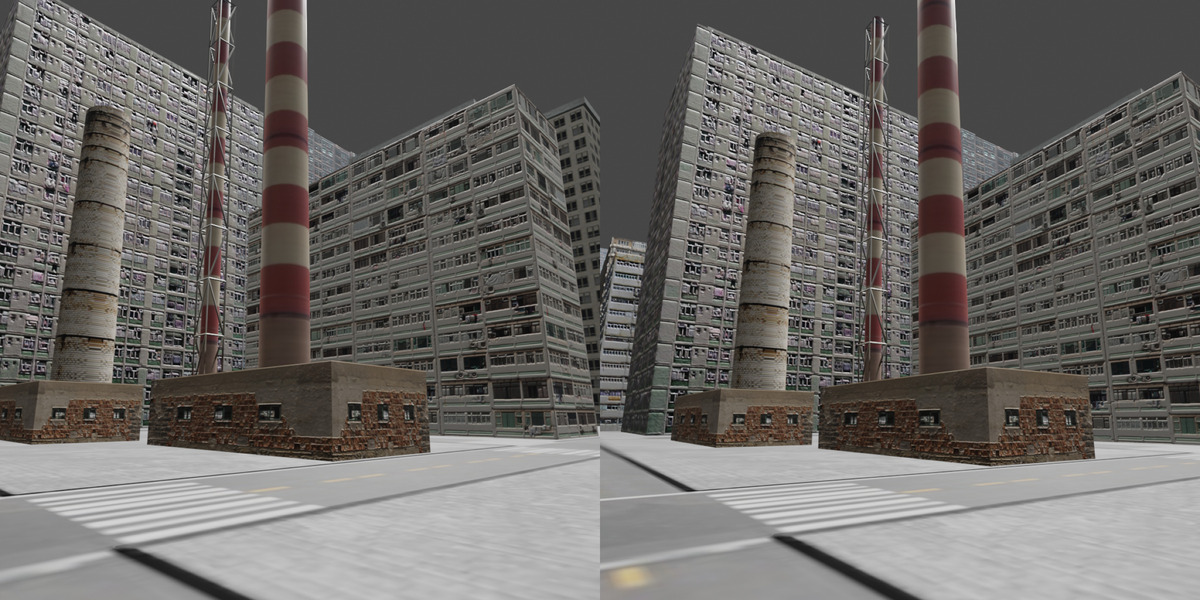}%
        \includegraphics[width=\imgwidth]{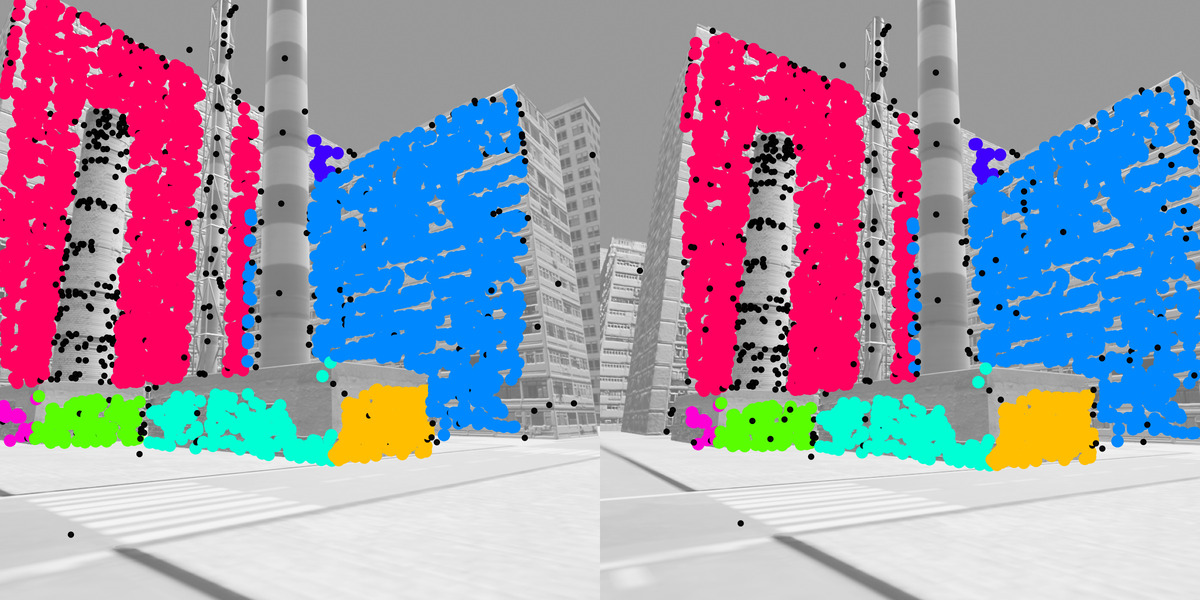}%
        \caption{Synthetic Metropolis Homographies (SMH) Dataset}
        \label{fig:city_dataset_examples}
    \end{subfigure}
    \caption{We propose two new datasets: HOPE-F for fundamental matrix fitting and SMH for homography fitting. For each dataset we show one example image pair: the left pair of each example shows the RGB images, the right pair visualises the pre-computed SIFT key-points, colour coded by ground truth label. Please refer to the \supp for additional examples.
    }
		\label{fig:dataset_examples}
\end{figure*}

\begin{table}
\setlength\tabcolsep{.26em}
\renewcommand{\arraystretch}{1.2} 
	\begin{center}
	\begin{tabular}{|l|c|c|c|c|c|}
        \cline{2-6}
        \multicolumn{1}{c|}{} &
        \makecell{scenes\\(train)} & \makecell{scenes\\(test)} & 
	\multicolumn{1}{c|}{instances} &
	\multicolumn{1}{c|}{points} & 
        \multicolumn{1}{c|}{outliers} \\
        \hline
        \textbf{HOPE-F} &
        3600 & 400 & 1--4 & 10--2k & 0--90\% \\
         Adelaide-F &
        0 & 19 & 1--4 & 165--360 & 27--73\% \\
         \hline
        \textbf{SMH} &
        44112 & 3890 & 1--32 & 12--16k & 1--96\% \\
        Adelaide-H &
        0 & 19 & 1--7 & 106--2k & 6--76\% \\
	\hline
	\end{tabular}
    \end{center}
	\caption{
{We compare our new HOPE-F and SMH datasets with AdelaideRMF~\cite{wong2011dynamic}.  Our new datasets contain significantly more scenes, with higher numbers of annotated model instances, as well as larger ranges of key-point correspondences and outlier ratios per scene.} 
}
	\label{tab:datasets}
\end{table}

\section{Multi-Model Fitting Datasets}
\label{sec:mmf_datasets}

A variety of tasks can be solved with robust multi-model fitting algorithms, and each task requires data with different modalities for training, parameter tuning and evaluation.
Early benchmark datasets for vanishing point estimation, such as the York Urban Dataset (YUD)~\cite{denis2008efficient} with 102 images, are relatively small.
This makes them unsuitable for contemporary deep learning methods, which require large amounts of training data.
Moreover, evaluation on such small datasets does not allow us to draw well-founded conclusions \wrt the performance of a method under a larger variety of conditions.
The SceneCity Urban 3D (SU3) dataset~\cite{zhou2019learning}, on the other hand, contains 23000 synthetic outdoor images showing a 3D model of a city rendered from various perspectives.
They adhere to the Manhattan world assumption and are annotated with three vanishing points each.
Complementary to this, the NYU-VP dataset~\cite{kluger2020consac} augments 1449 real-world non-Manhattan indoor images of the NYU Depth v2~\cite{silberman2012indoor} dataset with vanishing point annotations.
Its authors also presented YUD+, which adds additional non-Manhattan vanishing point annotations to YUD.
For fundamental matrix and homography estimation, however, the amount of data available is still very limited.
The AdelaideRMF~\cite{wong2011dynamic} dataset contains just 19 scenes for fundamental matrices and 19 for homographies.
Even though it is very small and was published over a decade ago, it is still the default benchmark for these two tasks.
We therefore present two new synthetic datasets: \emph{HOPE-F} for fundamental matrix fitting, and \emph{Synthetic Metropolis Homographies} for homography fitting.

\subsection{HOPE-F}
\label{sec:hope_dataset}
For fundamental matrix fitting, we construct a new dataset inspired by the scenes found in Adelaide-F.
We use a subset of 22 textured 3D meshes from the HOPE dataset~\cite{tyree2022hope} which depict household objects.
For each scene, we select between one and four of these meshes and place them on a 3D plane at randomised positions.
We then render two images of the scene using one virtual camera with randomised pose and focal length.
Before rendering the second image, we randomise the object positions again.
With known intrinsic camera parameters $\vec{K}$ and relative object poses $\vec{R}_i, \vec{t}_i$, $i \in \{1, \dots, 4\}$, we compute the ground truth fundamental matrices $\vec{F}_i$ for each image pair.
We then compute SIFT~\cite{Lowesift} key-point feature correspondences for each image pair.
We assign each correspondence either to one of the ground truth fundamental matrices, or to the outlier class.
These assignments represent the ground truth cluster labels.
Via this procedure, we generate a total of $4000$ image pairs with key-point features, of which we reserve $400$ as the test set.
Fig.~\ref{fig:hope_dataset_examples} shows an example image pair from this dataset, which we call \emph{HOPE-F}.
As Tab.~\ref{tab:datasets} shows, it is significantly larger than the Adelaide-F dataset.
Please refer to the \supp for additional details.

\subsection{Synthetic Metropolis Homographies (SMH)}
\label{sec:city_dataset}
For homography fitting, we construct a new dataset using a synthetic 3D model of a city.
Using the cars present in the 3D model as anchor points, we define seven camera trajectories, one of which is reserved for the test set.
Along each trajectory, we render sequences of images with varying step sizes and focal lengths.
To determine the ground truth homographies for each image pair, we first compute the normal forms $(\vec{n}_i, d_i)$ of the planes defined by each mesh polygon visible in the images. 
Using known camera intrinsics $\vec{K}$ and relative camera pose $\vec{R}, \vec{t}$, we compute ground truth homographies $\vec{H}_i$ for every plane in the scene for each image pair.
We then compute SIFT correspondences for each image pair and assign them either to one of the homographies, or to the outlier class.
Via this procedure, we generate a total of $48002$ image pairs with key-point feature correspondences, ground truth cluster labels and ground truth homographies.
Fig.~\ref{fig:city_dataset_examples} shows an example image pair from this dataset, which we call \emph{Synthetic Metropolis Homographies (SMH)}.
As Tab.~\ref{tab:datasets} shows, it is significantly larger than the Adelaide-H dataset with a much wider range of ground truth homographies. Please refer to the \supp for additional details.

\begin{table*}
\small
\setlength\tabcolsep{.39em}
\renewcommand{\arraystretch}{1.17} 
	\begin{center}
	\begin{tabular}{|l|c|c|cc|cc|cc|cc|cc|cc|}
	\hline
	\multicolumn{3}{|l|}{\textit{Datasets}} &
	\multicolumn{6}{c|}{SU3~\cite{zhou2019learning}} &
	\multicolumn{6}{c|}{YUD~\cite{denis2008efficient}} \\
        \hline
	\multicolumn{1}{|l|}{\multirow{2}{*}{\textit{Metrics}}} & 
	\multicolumn{2}{c|}{time}  & 
	\multicolumn{2}{c|}{\multirow{2}{*}{AUC @ $1\degree$}} &
	\multicolumn{2}{c|}{\multirow{2}{*}{AUC @ $3\degree$}} &
	\multicolumn{2}{c|}{\multirow{2}{*}{AUC @ $5\degree$}} &
	\multicolumn{2}{c|}{\multirow{2}{*}{AUC @ $3\degree$}} &
	\multicolumn{2}{c|}{\multirow{2}{*}{AUC @ $5\degree$}} &
	\multicolumn{2}{c|}{\multirow{2}{*}{AUC @ $10\degree$}} \\

        & \small{GPU} & \small{CPU} & \multicolumn{2}{c|}{} & \multicolumn{2}{c|}{} & \multicolumn{2}{c|}{} & \multicolumn{2}{c|}{} & \multicolumn{2}{c|}{} & \multicolumn{2}{c|}{}  \\
	\hline
	\multicolumn{15}{|c|}{{robust estimators (on pre-extracted line segments)}} \\

	\multicolumn{1}{|l|}{\textbf{PARSAC}}             
	& $\mathbf{4.91}$ & $\mathbf{11.59}$
	& $\underline{67.44} $ & $ \pm \mathbf{0.14}$
	  & $\underline{85.70} $ & $ \pm \underline{0.13}$ 
        & $90.17 $ & $ \pm \underline{0.11}$ 
        & $\mathbf{63.92} $ & $ \pm \mathbf{0.25}$ 
        & $\mathbf{75.90} $ & $ \pm \mathbf{0.15}$
        & $\mathbf{86.37} $ & $ \pm \mathbf{0.08}$ \\
        
	\multicolumn{1}{|l|}{CONSAC}      
	& $3941 $ & $2758 $
	& $51.58 $ & $ \pm \underline{0.18}$
	  & $78.52 $ & $ \pm \mathbf{0.10}$ 
        & $85.69 $ & $ \pm \mathbf{0.06}$ 
        & ${59.06} $ & $ \pm \underline{0.48}$ 
        & $71.33 $ & $ \pm \underline{0.44}$
        & $82.79 $ & $ \pm 0.29$ \\
     
	\multicolumn{1}{|l|}{J-Linkage} 
	& --- & $1139$
	& $47.66 $ & $ \pm 0.54$
	& $74.96 $ & $ \pm 0.49$
	& $83.01 $ & $ \pm 0.35$
	& $55.75 $ & $ \pm 2.53$
	& $68.69 $ & $ \pm 2.22$
	& $81.06 $ & $ \pm 1.52$ \\

	\multicolumn{1}{|l|}{T-Linkage}     
	& --- & $271.4 $
	& $40.57 $ & $ \pm 0.44$
	& $71.55 $ & $ \pm 0.26$
	& $80.90 $ & $ \pm 0.21$
	& $52.19 $ & $ \pm 1.91$
	& $66.10 $ & $ \pm 1.41$
	& $79.49 $ & $ \pm 0.87$ \\
        
	\multicolumn{1}{|l|}{Progressive-X}   
	& --- & $\underline{28.92} $
	& $63.14 $ & $ \pm 0.33$
	& $80.49 $ & $ \pm 0.40$
	& $84.82 $ & $ \pm 0.40$
	& $50.41 $ & $ \pm 0.84$
	& $60.10 $ & $ \pm 0.76$
	& $68.47 $ & $ \pm 0.72$ \\
        
	\multicolumn{15}{|c|}{{task-specific methods (full information)}} \\
  
	\multicolumn{1}{|l|}{NeurVPS}  
	& $766.0 $ & ---
	& $\mathbf{79.47} $ & 
	  & $\mathbf{92.57} $ & 
        & $\mathbf{95.44} $ & 
        & $50.63 $ & 
        & $63.12 $ &
        & $77.58 $ & \\
             
	\multicolumn{1}{|l|}{DeepVP}       
	& $\underline{130.3} $ & ---
	& $56.63 $ & 
	  & $84.05 $ &
        & $\underline{90.24} $ &  
        & $58.19 $ & 
        & ${72.25} $ & 
        & $\underline{84.98} $ &  \\
        
	\multicolumn{1}{|l|}{Contrario}   
	& --- & $767.2 $
	& $32.56 $ & $ \pm 0.19$
	  & $67.85 $ & $ \pm 0.17$ 
        & $77.72 $ & $ \pm \underline{0.11}$ 
        & $\underline{59.86} $ & $ \pm 0.59$ 
        & $\underline{72.61} $ & $ \pm 0.47$
        & $83.14 $ & $ \pm \underline{0.24}$ \\
        
	\hline
	\end{tabular}
    \end{center}
	\caption{
	{Manhattan-world VP estimation:} 
 Average AUC values (in \%, higher is better) and their standard deviations over five runs for vanishing point estimation on the SU3 and YUD datasets. We omit standard deviations for deterministic methods. Average run times are given in milliseconds. For robust estimators, we do not include time required for line segment detection: LSD~\cite{von2008lsd} requires $22.59$ ms on average.
}
	\label{tab:results_vp_manhattan}
\end{table*}

\begin{table*}
\small
\setlength\tabcolsep{.39em}
\renewcommand{\arraystretch}{1.17} 
	\begin{center}

    \begin{tabular}{|l|c|c|cc|cc|cc|cc|cc|cc|}
	\hline
	\multicolumn{3}{|l|}{\textit{Datasets}} &
\multicolumn{6}{c|}{NYU-VP~\cite{kluger2020consac}} &
	\multicolumn{6}{c|}{YUD+~\cite{kluger2020consac}} \\
        \hline
	\multicolumn{1}{|l|}{\multirow{2}{*}{\textit{Metrics}}} & 
	\multicolumn{2}{c|}{time}  & 
	\multicolumn{2}{c|}{\multirow{2}{*}{AUC @ $3\degree$}} &
	\multicolumn{2}{c|}{\multirow{2}{*}{AUC @ $5\degree$}} &
	\multicolumn{2}{c|}{\multirow{2}{*}{AUC @ $10\degree$}} &
	\multicolumn{2}{c|}{\multirow{2}{*}{AUC @ $3\degree$}} &
	\multicolumn{2}{c|}{\multirow{2}{*}{AUC @ $5\degree$}} &
	\multicolumn{2}{c|}{\multirow{2}{*}{AUC @ $10\degree$}} \\

        & \small{GPU} & \small{CPU} & \multicolumn{2}{c|}{} & \multicolumn{2}{c|}{} & \multicolumn{2}{c|}{} & \multicolumn{2}{c|}{} & \multicolumn{2}{c|}{} & \multicolumn{2}{c|}{}  \\
	\hline
	\multicolumn{15}{|c|}{{robust estimators (on pre-extracted line segments)}} \\

	\multicolumn{1}{|l|}{\textbf{PARSAC}}             
	& $\mathbf{5.16} $ & $\mathbf{9.38}$
	& $\underline{39.93} $ & $ \pm \mathbf{0.16}$
	  & $\underline{51.65} $ & $ \pm \mathbf{0.10}$ 
        & $\underline{64.58} $ & $ \pm \mathbf{0.13}$ 
        & $\mathbf{54.98} $ & $ \pm 0.63$ 
        & $\mathbf{65.48} $ & $ \pm \mathbf{0.37}$
        & $\mathbf{74.74} $ & $ \pm \underline{0.41}$ \\
        
	\multicolumn{1}{|l|}{CONSAC}             
	& $3843 $ & $2841$
	& $38.84 $ & $ \pm 0.34$
	  & $50.58 $ & $ \pm 0.37$ 
        & $64.25 $ & $ \pm 0.38$ 
        & $\underline{52.68} $ & $ \pm \mathbf{0.45}$ 
        & $\underline{63.73} $ & $ \pm 0.65$
        & $\underline{74.44} $ & $ \pm 0.76$ \\

	\multicolumn{1}{|l|}{J-Linkage}                    
	& --- & $1571 $
	& $30.61 $ & $ \pm 0.81$
	& $42.26 $ & $ \pm 0.80$
	& $56.58 $ & $ \pm 0.64$
	& $48.62 $ & $ \pm 1.29$
	& $60.40 $ & $ \pm 1.10$
	& $72.36 $ & $ \pm 0.82$ \\

	\multicolumn{1}{|l|}{T-Linkage}             
	& --- & $278.7 $
	& $30.93 $ & $ \pm 0.64$
	& $42.95 $ & $ \pm 0.68$
	& $57.75 $ & $ \pm 0.68$
	& $46.91 $ & $ \pm 0.66$
	& $59.45 $ & $ \pm 0.82$
	& $71.74 $ & $ \pm 0.54$ \\
        
	\multicolumn{1}{|l|}{Progressive-X}                    
	& --- & $\underline{26.15} $
	& $38.73 $ & $ \pm \underline{0.17}$
	& $49.25 $ & $ \pm \underline{0.17}$
	& $60.71 $ & $ \pm \underline{0.19}$
	& $50.13 $ & $ \pm 0.78$
	& $60.01 $ & $ \pm 0.69$
	& $68.53 $ & $ \pm 0.68$ \\
        
	\multicolumn{15}{|c|}{{task-specific methods (full information)}} \\
        
	\multicolumn{1}{|l|}{DeepVP}                    
	& $\underline{176.7} $ & --- 
	& $\mathbf{43.54} $ &
	  & $\mathbf{55.87} $ & 
        & $\mathbf{69.53} $ & 
        & $48.06 $ &
        & $59.57 $ &
        & $71.34 $ & \\

	\multicolumn{1}{|l|}{Contrario}                    
	& --- & $833.2 $
	& $35.91 $ & $ \pm 0.40$
	  & $47.61 $ & $ \pm 0.42$ 
        & $61.66 $ & $ \pm 0.38$ 
        & $51.50 $ & $ \pm \underline{0.52}$ 
        & $62.81 $ & $ \pm \underline{0.50}$
        & $72.55 $ & $ \pm \mathbf{0.38}$ \\
        
	\hline
	\end{tabular}
    \end{center}
	\caption{
		{Non-Manhattan VP estimation:} 
 Average AUC values (in \%, higher is better) and their standard deviations over five runs for vanishing point estimation on the NYU-VP and YUD+ datasets. We omit standard deviations for deterministic methods.  Average run times are given in milliseconds.  For robust estimators, we do not include time required for line segment detection: LSD~\cite{von2008lsd} requires $22.59$ ms on average.
}
	\label{tab:results_vp}
\end{table*}

\begin{table*}
\small
\setlength\tabcolsep{.6em}
\renewcommand{\arraystretch}{1.17} 
	\begin{center}
	\begin{tabular}{|l|c|c|cc|cc|cc|cc|}
	\hline
        \multirow{2}{*}{\textit{Fundamental Matrices}}  &
        \multicolumn{2}{c|}{time}  &
	\multicolumn{4}{c|}{{HOPE-F}} &
	\multicolumn{4}{c|}{{Adelaide-F~\cite{wong2011dynamic}}} \\
        & \small{GPU} & \small{CPU} & \multicolumn{2}{c|}{ME} & \multicolumn{2}{c|}{SE} & \multicolumn{2}{c|}{ME} & \multicolumn{2}{c|}{SE}  \\ 
	\hline

	\multicolumn{1}{|l|}{\textbf{PARSAC}}     
        & $\mathbf{12.25} $ & $\mathbf{19.81}$
	& $\mathbf{14.97} $ & $ \pm \mathbf{8.51}$
        & $\mathbf{3.07} $ & $ \pm \mathbf{3.39}$
        & ${9.83} $ & $ \pm \mathbf{4.17}$ 
        & $2.80 $ & $ \pm 2.37$ \\

        \multicolumn{1}{|l|}{Fast-CP}                    
	& -- & $\underline{33.65}$
	& ${24.59} $ & $ \pm \underline{13.3}$
        & $\underline{5.56} $ & $ \pm \underline{6.67}$
        & $\mathbf{4.92} $ & $ \pm \underline{5.23}$  
        & $\underline{1.43} $ & $ \pm \underline{1.31}$ \\
        
	\multicolumn{1}{|l|}{Progressive-X+}      
 	& --  & $67.38 $
	& $43.23 $ & $ \pm 15.7$
        & $9.72 $ & $ \pm 15.8$
        & $\underline{6.57} $ & $ \pm 6.64$  
        & $\mathbf{0.84} $ & $ \pm \mathbf{0.88}$ \\
        
	\multicolumn{1}{|l|}{Progressive-X} 
 	& --  & $1043 $
	& $\underline{22.78} $ & $ \pm 15.6$
        & $29.9 $ & $ \pm 105$
        & $ 12.85$ & $ \pm 14.1$  
        & $2.19 $ & $ \pm 4.51$ \\
        
	\hline
	\hline

        \multirow{2}{*}{\textit{Homographies}} & 
        \multicolumn{2}{c|}{time}  &
	\multicolumn{4}{c|}{{SMH}} &
	\multicolumn{4}{c|}{Adelaide-H~\cite{wong2011dynamic}} \\
         & \small{GPU} & \small{CPU}  & \multicolumn{2}{c|}{ME} & \multicolumn{2}{c|}{TE} & \multicolumn{2}{c|}{ME} & \multicolumn{2}{c|}{TE}  \\ 
	\hline

	\multicolumn{1}{|l|}{\textbf{PARSAC}}     
	& $\mathbf{64.00} $ & $1758$
	& $\mathbf{20.50} $ & $ \pm \underline{15.5}$
        & $\mathbf{1.81} $ & $ \pm \mathbf{20.8}$
        & $8.63 $ & $ \pm 8.01$
        & $5.34 $ & $ \pm 7.36$ \\

	\multicolumn{1}{|l|}{CONSAC}     
	& $\underline{4314} $ & $58913$
        & $33.45 $ & $ \pm 18.5$
	& $3.34 $ & $ \pm  25.7 $
        & $\mathbf{5.66} $ & $ \pm \mathbf{7.05}$
        & $\underline{3.44} $ & $ \pm 7.44$ \\
        
	\multicolumn{1}{|l|}{Progressive-X+}      
	& -- & $\mathbf{501.4} $
	& $52.69 $ & $ \pm 18.3$  
        & $7.19 $ & $ \pm 39.3$
        & $\underline{7.38} $ & $ \pm \underline{7.64}$
        & $3.74 $ & $ \pm \underline{5.13}$  \\
        
	\multicolumn{1}{|l|}{Progressive-X} 
	& -- & $\underline{674.2} $
	& $\underline{20.60} $ & $ \pm \mathbf{15.3}$  
        & $25.5 $ & $ \pm  113$
        & $8.05 $ & $ \pm 10.0$
        & $\mathbf{3.14} $ & $ \pm \mathbf{3.81}$  \\
        
	\hline
	\end{tabular}
    \end{center}
	\caption{
		{Fundamental Matrix and Homography Estimation:}  Average misclassification errors (ME, in \%, lower is better), Sampson errors (SE, in pixel, lower is better), transfer errors (TE, in pixel, lower is better), and their standard deviations over five runs for fundamental matrix fitting on our new HOPE-F dataset and the Adelaide dataset, and for homography fitting on our new SMH dataset and the Adelaide dataset. Average run time per scene is given in milliseconds. We do not include time required for feature extraction: SIFT~\cite{Lowesift} requires $106.3$ ms on average.
}
	\label{tab:results_fh}
\end{table*}

\section{Experiments}
\label{sec:experiments}
For sampling and inlier weight prediction, we implement a neural network based on~\cite{kluger2020consac}. 
Please refer to the \supp for implementation details, a description of all evaluation metrics, and additional experimental results and discussions.
We compute the results for all competitors using code provided by the respective authors, unless stated otherwise.
For non-deterministic approaches, we report mean and standard deviation over five runs for all metrics.
In addition, we measure average computation times with and without GPU acceleration, if applicable.
We mark best results in \textbf{bold} and second best results with \underline{underline}.

\subsection{Vanishing Point Estimation}
For vanishing point estimation, we use the cross product as our minimal solver with subsequent inlier weighted SVD refinement.
We evaluate on four datasets: SU3, YUD, YUD+ and NYU-VP.
The first two represent Manhattan-world scenarios, while the latter two contain non-Manhattan scenes.
We compare against the robust multi-model fitting methods CONSAC~\cite{kluger2020consac}, J- and T-Linkage~\cite{toldo2008robust,magri2014t}, Progressive-X~\cite{barath2019progressive}, and the task-specific VP estimators NeurVPS~\cite{zhou2019neurvps}, DeepVP~\cite{lin2022deep} and Contrario-VP~\cite{simon2018contrario}.
For J-/T-Linkage, we use the code provided by~\cite{kluger2020consac}.
We adopt the evaluation metric proposed by~\cite{kluger2020consac}, \ie we measure the angle between predicted and ground truth vanishing directions, and calculate the area under the recall curve (AUC) for multiple upper bounds. 

\paragraph{Manhattan World}
For the Manhattan-world scenario, we train PARSAC on SU3 and evaluate on both SU3 and YUD.
As Tab.~\ref{tab:results_vp_manhattan} shows, PARSAC outperforms all methods but NeurVPS~\cite{zhou2019neurvps} and DeepVP~\cite{lin2022deep} by large margins on SU3.
Compared to DeepVP, our method is on par \wrt \aucfive~but more than ten percentage points better \wrt \aucone, \ie we are able to estimate the vanishing points more accurately.
NeurVPS performs substantially better than our method on SU3, but ranks second-to-last on YUD.
This indicates that NeurVPS -- also trained on SU3 -- is not able to generalise well beyond its training domain.
PARSAC, on the other hand, clearly outperforms all competitors on YUD.
In addition, it is more than five times faster than the fastest competitor -- Progressive-X~\cite{barath2019progressive} -- which achieves lower AUC values on both datasets.
Our method is also more than $26$ times faster than DeepVP, which is the second fastest competitor.

\paragraph{Non-Manhattan World}
For the non-Manhattan scenario, we train our method on NYU-VP for evaluation on the same, and train it on SU3 for evaluation on YUD+.
As Tab.~\ref{tab:results_vp_manhattan} shows, DeepVP outperforms PARSAC on NYU-VP, whereas PARSAC outperforms DeepVP on YUD+.
While it is not obvious which method is more accurate overall, PARSAC has a clear advantage \wrt computation time, being $34$ times faster.
The only other competitive method, CONSAC, is almost three orders of magnitude slower.
We did not evaluate NeurVPS, as it is designed for the Manhattan-world case only.

\subsection{Fundamental Matrix Estimation}
For fundamental matrix fitting, we use the seven point algorithm~\cite{hartley2003multiple} as our minimal solver without subsequent refinement.
We compare PARSAC with Fast-CP~\cite{ozbay2022fast}, Progressive-X+~\cite{barath2021progressive} and Progressive-X~\cite{barath2019progressive}. 
CONSAC~\cite{kluger2020consac} has no implementation for F-matrix fitting available.
We evaluate on our new HOPE-F dataset as well as on the Adelaide~\cite{wong2011dynamic} dataset.
In Tab.~\ref{tab:results_fh}, we report misclassification errors (ME) and Sampson errors (SE).
PARSAC outperforms all competitors on the HOPE-F dataset by large margins, achieving an average ME which is $9.6$ percentage points lower than the best competitor (Fast-CP), with an SE which is $44\%$ lower.
It is more than twice as fast as Fast-CP and more than three times faster than Progressive-X+.
On Adelaide, our method performs worse than Fast-CP and Progressive-X+, albeit with a smaller ME margin of $4.9$ percentage points but a comparatively large SE.

\subsection{Homography Estimation}
For homography fitting, we use the four point DLT~\cite{hartley2003multiple} as our minimal solver without subsequent refinement.
We compare PARSAC with Progressive-X+~\cite{barath2021progressive}, Progressive-X~\cite{barath2019progressive} and CONSAC~\cite{kluger2020consac}.
Fast-CP~\cite{ozbay2022fast} has no implementation for homography fitting available.
We evaluate on our new SMH dataset as well as on the Adelaide~\cite{wong2011dynamic} dataset.
In Tab.~\ref{tab:results_fh}, we report misclassification errors (ME) and transfer errors (TE).
PARSAC achieves an ME which is roughly on par with Progressive-X, although the latter has a slight edge on Adelaide.
CONSAC and Progressive-X+ achieve lower MEs on Adelaide, but perform significantly worse on SMH.
PARSAC has the lowest TE on SMH by a large margin, but falls behind its competitors on Adelaide.
These baselines, however, are roughly between $10$ and $100$ times slower than PARSAC.

\paragraph{Limitations}
PARSAC requires a GPU to utilise its full potential.
The number of putative model instances must be set before training, and instances beyond this cannot be found.

\section{Conclusion}
We proposed a new robust multi-model fitting algorithm which decouples the estimation of individual instances of a geometric model. 
PARSAC uses a neural network to predict multiple sets of sample and inlier weights in a single forward pass.
This enables us to process all model instances in parallel and in real-time, yielding a significant speed-up compared to previous approaches.
PARSAC can be used for a variety of computer vision problems, such as finding vanishing points, fundamental matrices and homographies, and achieves results superior to or competitive with state-of-the-art on multiple datasets.
In addition, we contribute two new datasets for fundamental matrix and homography fitting, which will benefit the development of multi-model estimators in the future.

\section*{Acknowledgements}
This work was supported by the Federal Ministry of Education and Research (BMBF), Germany under the AI Service Centre KISSKI (grant no. 01IS22093C), the Centre for Digital Innovations Lower Saxony (ZDIN) and the Deutsche Forschungsgemeinschaft (DFG) under Germany’s Excellence Strategy within the Cluster of Excellence PhoenixD (EXC 2122). 

\ifarxiv
\renewcommand*\appendixpagename{\Large Appendix}
\begin{appendices}
\input{appendix}
\end{appendices}

\fi

\bibliography{aaai24}

\end{document}

%% file: appendix.tex
In the first section of this appendix, we discuss reasons for performance gaps between PARSAC and its competitors \wrt accuracy on fundamental matrix and homography estimation (Sec.~\ref{sec:gap_adelaide}), as well as \wrt computation times (Sec.~\ref{sec:gap_times}).
We provide additional implementation details for PARSAC as well as our competitors in Sec.~\ref{sec:implementation_details}.
This includes pseudo-code, a listing of hyper-parameters, a description of the neural network architecture, definitions of residual functions and minimal solvers, information about the hardware used for our experiments, and the parameter optimisation methodology we used for our competitors.
In Sec.~\ref{sec:evaluation_metrics}, we provide a description of all evaluation metrics used in our experiments.
Sec.~\ref{sec:selfsupervised} contains experimental results for self-supervised learning of PARSAC.
We furthermore provide several ablation studies \wrt our weighted inlier counting (Sec.~\ref{sec:ablation_weighted}), feature generalisation (Sec.~\ref{sec:ablation_feature}), the number of model instances (Sec.~\ref{sec:ablation_instances}), robustness to noise and outliers (Sec.~\ref{sec:ablation_robustness}) and parameter sensitivity (Sec.~\ref{sec:ablation_sensitivity}).
Lastly, Sec.~\ref{sec:dataset_details} contains additional details about our proposed new datasets for fundamental matrix and homography fitting: HOPE-F and Synthetic Metropolis Homographies. 
Source code for both training and evaluation of PARSAC, pre-trained neural network models, and datasets are available online\footnote{\texttt{https://github.com/fkluger/parsac}}.

\section{Discussion: Performance Gap}
\subsection{SMH and HOPE-F vs. Adelaide}
\label{sec:gap_adelaide}
Tab.~\ref{tab:results_fh} shows that PARSAC performs best on our new synthetic datasets -- SMH and HOPE-F -- for homography and fundamental matrix estimation.
All competitors, with the exception of Progressive-X on SMH, perform significantly worse on these datasets.
On Adelaide-H and Adelaide-F, on the other hand, PARSAC is beaten by its competitors.
We trained PARSAC on our new synthetic datasets and used Adelaide only for evaluation, while the competitors were designed with the Adelaide dataset in mind. 
It is thus expectable that PARSAC performs better on the synthetic datasets in relative terms. 
The performance of PARSAC on Adelaide is still competitive, which suggests that PARSAC generalises well between datasets. 
The performance of our competitors on the synthetic datasets indicates that, for the most part, they are not able to generalise well, and have possibly been overfit to the Adelaide dataset. 
Another factor which may contribute to the relatively worse performance of PARSAC on Adelaide, is the fact that we do not apply any refinement -- such as IRLS (Progressive-X+) or EM (CONSAC) -- to fundamental matrices and homographies, but evaluate the models computed by the minimal solvers.

\subsection{Computation Times}
\label{sec:gap_times}
As we emphasise in Sec.~\ref{sec:experiments}, PARSAC is significantly faster than its competitors.
Implementation differences \wrt optimisation, programming languages and hardware acceleration may account for differences in computation times.
In addition, there are intrinsic reasons why PARSAC is faster:
\begin{enumerate}
    \item It uses a light-weight neural network to predict sample weights for a relatively small number of input features. NeurVPS and DeepVP, in comparison, utilise larger and more complex CNN architectures which operate on RGB images directly, which results in significantly larger inference times.
    \item Being guided by a neural network, its hypothesis sampling is very efficient. Other methods -- such as Contrario, J-Linkage and T-Linkage -- sample hypotheses uniformly and thus require a much larger number of samples.
    \item It can compute and select all hypotheses in parallel, if the available hardware allows for it. Other methods -- such as CONSAC, J-Linkage, T-Linkage, Progressive-X, Progressive-X+ and Fast-CP -- estimate models iteratively and thus cannot be parallelised to the same degree as PARSAC. 
\end{enumerate}

\input{algo/parsac}
\input{algo/ranking}
\input{algo/clustering}

\section{Implementation Details}
\label{sec:implementation_details}
We provide a pseudo-code listing of PARSAC in Alg.~\ref{algo:parsac}. 
Please note that all for-loops in Alg.~\ref{algo:parsac} can be fully parallelised.
Alg.~\ref{algo:parsac:instance_ranking} and Alg.~\ref{algo:parsac:cluster_assignment} show the instance ranking and cluster assignment steps of PARSAC, respectively.
In Sec.~\ref{sec:network_arch}, we provide a detailed description of the neural network architecture used in our experiments.
We explain the residual functions, minimal solvers and model refinement techniques that we used for each application in Sec.~\ref{sec:residuals_solvers}.
Sec.~\ref{sec:hardware_timing} describes the hardware used in our experiments, as well as the resources required to train PARSAC for each application.
Lastly, we provide details regarding the parameters for our competitors used in our experiments in Sec.~\ref{sec:competitors}.

\subsection{Network Architecture}
\label{sec:network_arch}

\begin{figure*}	
\centering
\includegraphics[width=0.97\linewidth]{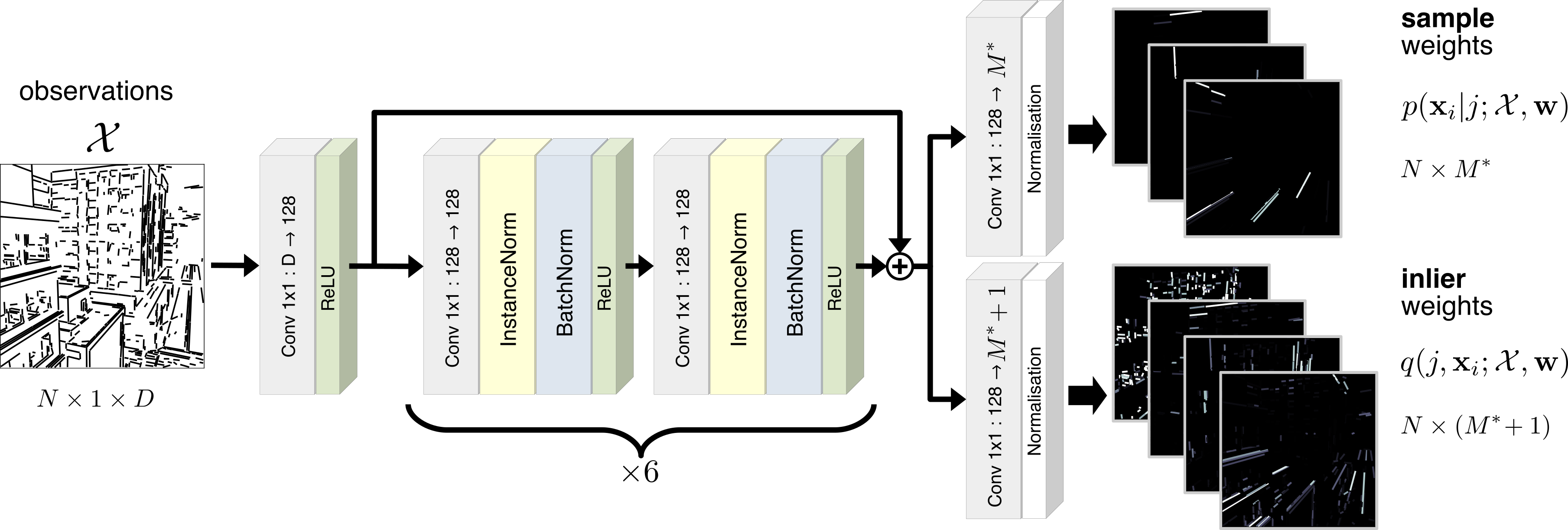}
\caption{ {Neural network architecture:}
We feed observations $\obsv \in \obsvs$ of dimension $D$ into our network as a tensor of size $\numobsvs \times 1 \times D$.
The network consists of linear $1\times 1$ convolutional layers interleaved with instance normalisation \cite{ulyanov2016instance}, batch normalisation \cite{ioffe2015batch} and ReLU \cite{he2015delving} layers which are arranged as residual blocks~\cite{he2016deep}. The architecture is based on \cite{kluger2020consac}. 
}
\label{fig:netarch}
\end{figure*}

In order to predict sample and inlier weights, we utilise a neural network which is based on the architecture used in~\cite{kluger2020consac}.
Fig.~\ref{fig:netarch} gives an overview of the network structure.
We stack observations $\obsv \in \obsvs$ -- each being a vector with dimension $D$ -- into a tensor of size $\numobsvs \times 1 \times D$, which is fed into the network.
Following a single convolutional layer ($1\times1$, $128$ channels) with ReLU~\cite{he2015delving} activation function, our network contains six residual blocks~\cite{he2016deep}. 
Each residual block is composed of two series of convolutions ($1\times1$, $128$ channels), instance normalisation \cite{ulyanov2016instance}, batch normalisation \cite{ioffe2015batch} and ReLU activation.
Two separate convolutional layers ($1\times1$, $\numpmodels$ and $\numpmodels+1$ channels) with log-sigmoid activation compute the unnormalised logarithmic sample weights $\log \mat{P}$ and inlier weights $\log \mat{Q}$, respectively. 
After normalisation, we get the logarithmic sample weights $\log p(\obsv_i|j;\obsvs,\vec{w})$ and inlier weights $\log q(j,\obsv_i;\obsvs,\vec{w})$:
\begin{equation}
    \log p(\obsv_i|j;\obsvs,\nnw) = \log {\mat{P}_{i,j}} - \log {\sum_{k=1}^{\numobsvs} \mat{P}_{k,j} } \, ,
\end{equation}
\begin{equation}
    \log q(j,\obsv_i;\obsvs,\nnw) = \log {\mat{Q}_{i,j}} - \log {\sum_{k=1}^{\numpmodels+1} \mat{Q}_{i,k} } \, .
\end{equation}
The network architecture is order invariant \wrt observations $\obsvs$ since it only uses $1 \times 1$ convolutions. 
We implement the architecture using PyTorch~\cite{paszke2017automatic} version 1.13.1.

\subsection{Parameters}
Tab.~\ref{tab:parameters} provides a listing of all user-definable parameters used in our experiments.
For vanishing point estimation and fundamental matrix estimation, we used the same parameters for all datasets.
On the Adelaide-H dataset for homography fitting, we used different parameters during inference for optimal results.

\input{tables/parameters}

\subsubsection{Dimensionality of Observations}
During training, we use a fixed number of observations $\numobsvs = 512$. 
If the actual number of observations in a data sample is larger, we randomly select a subset of observations.
If it is smaller, we duplicate observations in order to pad the unfilled entries in the tensor.
During inference, we use the actual number of observations $\numobsvs$ for each data sample.
The number of dimensions $D$ per observation $\obsv$ depends on the application, but is always $D=4$ in our experiments:
\begin{compactitem}
    \item Vanishing points: $\obsv$ is a line segment parametrised via centroid $(x,y)$, length $l$ and angle $\alpha$.
    \item Fundamental matrices and homographies: $\obsv$ is a point correspondence pair $(x_1, y_1, x_2, y_2)$.
\end{compactitem}

\subsubsection{Training Procedure}
We train our network for $N_e$ epochs using the Adam optimiser with a learning rate of $10^{-4}$.
After $N_{lr}$ epochs, we reduce the learning rate by a factor of ten.
We repeat each training run five times with different random seeds, and select the network weights which achieved the highest AUC for vanishing point detection or the lowest misclassification error for fundamental matrix and homography estimation, respectively, on the validation set.

\subsection{Residual Functions, Minimal Solvers and Refinement}
\label{sec:residuals_solvers}
The type of minimal solver $\minsolver(\minimalset)$, which computes a model hypothesis $\hypothesis$ from a minimal set of observations $\minimalset$, and the type of residual function $d(\obsv,\model)$, which measures the distance between an observation $\obsv$ and a model $\model$, both depend on the application.

\subsubsection{Vanishing Points}
Each observation $\obsv$ is a line segment which can be defined by two homogeneous points $(x_1, y_1, 1)$ and $(x_2, y_2, 1)$, and its resulting line in normal form $\vec{l} = (n_1, n_2, d)$.
Each model is a vanishing point in homogeneous coordinates: $\model = \vec{v} = (x_v, y_v, 1)$.
We define the residual function $d(\obsv,\model)$ via the cosine of the angle between $\obsv$ and the constrained line arising from $\obsv$ and $\vec{v}$, as described in the appendix of~\cite{kluger2020consac}.
In order to compute a new vanishing point hypothesis $\hypothesis$ from a minimal set of observations $\minimalset = \{\obsv_1, \obsv_2\}$, we simply use the cross product:
\begin{equation}
    \hypothesis = \minsolver(\minimalset) =  \vec{l}_1 \times \vec{l}_2 \, .
\end{equation}
After we have selected a set of vanishing points $\pmodels = \{\model_1, \dots, \model_{\numpmodels} \}$, we refine each vanishing point by computing the least squares solution of the homogeneous linear equation system $\vec{A}\vec{\tilde{h}} = 0$ with:
\begin{equation}
    \vec{A} = 
    \begin{pmatrix}
        \vec{l}_1 \cdot s(d(\obsv_1, \model)) \\
        \vdots \\
        \vec{l}_{\numobsvs} \cdot s(d(\obsv_{\numobsvs}, \model)) 
    \end{pmatrix}\, ,
\end{equation}
using the soft inlier function $s(\cdot)$ mentioned in Sec.~\ref{sec:method}.

\subsubsection{Fundamental Matrices}
Each observation $\obsv$ is a pair of point correspondences $(x_1, y_1)$ and $(x_2, y_2)$, while each model is a $3\times 3$ fundamental matrix $\model = \vec{F}$.
We define the residual function $d(\obsv,\model)$ as the square root of the Sampson distance~\cite{hartley2003multiple}.
In order to compute a new hypothesis, we use the seven point algorithm~\cite{hartley2003multiple}.
We do not apply any refinement to these hypotheses later on.

\subsubsection{Homographies}
Each observation $\obsv$ is a pair of point correspondences $(x_1, y_1)$ and $(x_2, y_2)$, while each model is a $3\times 3$ planar homography $\model = \vec{H}$.
We define the residual function $d(\obsv,\model)$ as the symmetric transfer error~\cite{hartley2003multiple}.
In order to compute a new hypothesis, we use the four point DLT~\cite{hartley2003multiple}.
We do not apply any refinement to these hypotheses later on.

\subsection{Hardware, Timing and Resource Usage}
\label{sec:hardware_timing}
In order to ensure reproducibility of our time measurements, we conducted all inference experiments on the same workstation with no other processes running simultaneously.
The workstation is running Ubuntu Linux 20.04.5 LTS and has an i7-7820X CPU, an RTX 3090 GPU and 64GB of RAM.
For fair and nuanced comparisons, we report times with and without GPU acceleration separately.
Most methods do not use GPU acceleration, including J-Linkage, T-Linkage, Progressive-X(+), Fast-CP and Contrario.
NeurVPS and DeepVP failed to run CPU-only. 
We thus left the corresponding entries in Tabs.~\ref{tab:results_vp_manhattan}-\ref{tab:results_fh} empty.

Neural network training was conducted on a compute cluster consisting of nodes with a variety of hardware configurations.
Each training run required only a single GPU and utilised 24GB of GPU RAM at most.
Approximate duration of one training run of our network for each dataset on the previously mentioned workstation:
\begin{itemize}
    \item SU3: 36 hours 
    \item NYU-VP: 20 hours 
    \item HOPE-F: 87 hours 
    \item SMH: 190 hours
\end{itemize}

\subsection{Competitors}
\label{sec:competitors}
In order to provide a fair comparison, we optimised the parameters of the baseline methods on the same datasets as PARSAC.
We describe the methodology used for parameter optimisation for each competitor below.
\paragraph{CONSAC, DeepVP and NeurVPS}
For evaluation on SU3, YUD and YUD+, we trained each method on the training set of the SU3 dataset according to the specifications of the respective authors~\cite{kluger2020consac, lin2022deep, zhou2019neurvps}.
For evaluation on NYU-VP, we trained each method apart from NeurVPS on the NYU-VP training set.
For evaluation on the SMH dataset, we trained CONSAC on the SMH training set using an inlier threshold of $10^{-6}$.
During evaluation, we set the maximum number of model instances to 16, because a larger value would trigger an out-of-memory error.
For evaluation on Adelaide-H, we used the parameters provided by the authors.

\paragraph{Progressive-X, Progressive-X+, J-Linkage, T-Linkage}
For these methods, we used the state-of-the-art hyper-parameter tuning method SMAC3~\cite{JMLR:v23:21-0888} to optimise their parameters with a budget of 100 iterations.
For evaluation on the SU3, YUD and YUD+ datasets, we have optimised the AUC @ $5\degree$ of each method on the training set of the SU3 dataset.
For evaluation on the NYU-VP dataset, we have optimised the AUC @ $10\degree$ of each method on the training set of NYU-VP.
For evaluation on the HOPE-F and SMH datasets, we have optimised the misclassification errors of each method on the respective training sets.
Approximate times required for parameter optimisation on our workstation:
\begin{tabular}{l|c|c|c|c}
          & SU3      & NYU-VP  & HOPE-F  & SMH     \\
\hline
Prog-X    & 15 h & 1 h  & 92 h  & 750 h \\
\hline
Prog-X+   & --       & --      & 6 h & 556 h \\
\hline
J-Linkage & 570 h  & 43 h  & --      & --      \\
\hline
T-Linkage & 150 h   & 10 h & --      & --     
\end{tabular}

\noindent For Adelaide-F and Adelaide-H, we used the parameters provided by the authors.

\paragraph{Contrario}
The code provided by the authors is implemented with MATLAB and does not lend itself to be optimised with SMAC3, which has a Python interface.
We thus optimised the three most sensitive parameters -- $\theta_z$, $\theta_h$ and $\theta_{con}$ according to~\cite{simon2018contrario} -- by testing five different parameters for each, yielding 125 trials in total.
We multiplied the default value of each parameter with factors of $\frac{1}{2}$, $\frac{6}{8}$, $1$, $\frac{6}{4}$ and $2$.
For evaluation on the SU3, YUD and YUD+ datasets, we have optimised the AUC @ $5\degree$ of each method on the training set of the SU3 dataset.
For evaluation on the NYU-VP dataset, we have optimised the AUC @ $10\degree$ of each method on the training set of NYU-VP.

\paragraph{Fast-CP} 
This method does not have user-definable parameters which could be optimised.

\section{Evaluation Metrics}
\label{sec:evaluation_metrics}
\subsection{Vanishing Point Error}
Given a set of ground truth vanishing points $\gtmodels = \{ \gt{\vec{v}}_1, \dots, \gt{\vec{v}}_{\numgtmodels} \}$ and estimated vanishing points $\models = ( {\vec{v}}_1, \dots, {\vec{v}}_{\nummodels} )$, we assume that $\models$ is sorted by significance, \ie ${\vec{v}}_1$ is the most important and ${\vec{v}}_{\nummodels}$ is the least important vanishing point.
First, we compute the vanishing direction $\vec{d}$ for each vanishing point using known intrinsic camera parameters $\vec{K}$:
\begin{equation}
    \vec{d} = \vec{K}\inverse \vec{v} \, .
\end{equation}
Then, we compute the angle error $\theta_{ij}$ between each ground truth vanishing direction $\gt{\vec{d}}_i$ and estimated vanishing direction ${\vec{d}}_j$:
\begin{equation}
    \theta_{ij} = \arccos\left( \left| \frac{\gt{\vec{d}}_i \tran {\vec{d}}_j}{\|\gt{\vec{d}}_i\| \cdot \|{\vec{d}}_j\|} \right| \right) \, .
\end{equation}
In order to find a matching between $\gtmodels$ and $\models$, we construct a cost matrix $\vec{C}$ of size $\numgtmodels \times \tilde{M}$ with entries $C_{ij}=\theta_{ij}$, only using the $\tilde{M} = \min (\numgtmodels, \nummodels)$ most important vanishing point estimates from $\models$.
Using $\vec{C}$, the Hungarian method~\cite{kuhn1955hungarian} yields a matching between $\models$ and $\gtmodels$, giving us one error $\theta_{i}$ for each ground truth vanishing point.
If a ground truth vanishing point $\gt{\vec{v}}_i$ remains unmatched, \ie if $\nummodels<\numgtmodels$, we set the corresponding error to $\theta_{i}=90\degree$.
We gather the angle errors for all ground truth vanishing points of all images within a dataset and compute the relative area under the recall curve up to a cutoff value $\theta_c$ (AUC @ $\theta_c$). 
This metric was proposed in~\cite{kluger2020consac}.

\subsection{Misclassification Error}

\input{tables/results_self}

Given ground truth cluster labels $\gtlabels = \{\gt{y}_1, \dots, \gt{y}_{\numobsvs}\}$ and estimated cluster labels $\labels = \{{y}_1, \dots, {y}_{\numobsvs}\}$ for observations $\obsvs = \{\obsv_1, \dots, \obsv_{\numobsvs} \}$, we compute the misclassification error:
\begin{equation}
    \mathrm{ME}(\labels, \gtlabels) = \frac{1}{\numobsvs} \sum_{i=1}^{\numobsvs} 1-[y_i=\gt{y}_i] \, ,
\end{equation}
with $[\cdot]$ being the Iverson bracket. 
This is the most commonly evaluation metric for multiple fundamental matrix and homography estimation~\cite{barath2016multi,barath2019progressive,barath2021progressive,kluger2020consac,magri2014t,magri2015robust,magri2016multiple,magri2019fitting}.

\subsection{Sampson Error}
For fundamental matrix estimation, we propose a new evaluation metric based on the square root of the Sampson distance~\cite{hartley2003multiple} between a fundamental matrix $\vec{F}$ and a point correspondence $(\vec{p}^a, \vec{p}^b)$:
\begin{equation}
\begin{split}
    & d_{\text{S}}(\vec{F}, \vec{p}^a, \vec{p}^b) = \\
    & \sqrt{ \frac{({\vec{p}^b}\tran \vec{F} \vec{p}^a)^2}{ (\vec{F} \vec{p}^a)_1^2 + (\vec{F} \vec{p}^a)_2^2 + (\vec{F}\tran \vec{p}^b)_1^2 + (\vec{F}\tran \vec{p}^b)_2^2 } } \, ,
\end{split}
\end{equation}
where $(\vec{a})_i$ represents the $i$-th entry of vector $\vec{a}$, with $\vec{p}^a, \vec{p}^b$ being 2D points in homogeneous coordinates.
This is a first-order approximation of the geometric reprojection error.
Given a set of ground truth fundamental matrices $\gtmodels = \{ \gt{\vec{F}}_1, \dots, \gt{\vec{F}}_{\numgtmodels} \}$ and estimated fundamental matrices $\models = ( {\vec{F}}_1, \dots, {\vec{F}}_{\nummodels} )$, we assume that $\models$ is sorted by significance, \ie $\vec{F}_1$ is the most important and ${\vec{F}}_{\nummodels}$ is the least important fundamental matrix.
We compute the minimum Sampson error of each observation $\obsv = (\vec{p}^a, \vec{p}^b) \in \obsvs$  using the $\tilde{M} = \min (\numgtmodels, \nummodels)$ most important estimated fundamental matrices:
\begin{equation}
    e_{\text{S}}(\obsv, {\models}) = \min_{j \in \{1, \dots, \tilde{M} \}} d_{\text{S}}({\vec{F}}_j, \vec{p}^a, \vec{p}^b) \, .
\end{equation}
If $\models = \emptyseq$, \ie no fundamental matrices were found by a method, we assume $\models = ( \vec{I} )$, with $\vec{I}$ being the identity matrix.
In order to avoid large outlier errors skewing the results, we clip the maximum value of $e_{\text{S}}$ to $\max(W,H)$, with $W$ and $H$ being the image width and height, respectively.
We then compute the mean of $e_{\text{S}}$ for all observations which are inliers according to the ground truth cluster labels $\gtlabels$:
\begin{equation}
    \mathrm{SE}(\obsvs, \models, \gtlabels) = \frac{\sum_{i=1}^{\numobsvs} e_{\text{S}}(\obsv_i, \models) \cdot  \iverson{\gt{y}_i>0}}{\sum_{i=1}^{\numobsvs} \iverson{\gt{y}_i>0}} \, ,
\end{equation}
with $\gt{y}_i>0$ indicating that observation $\obsv_i$ is not an outlier.
This metric gauges the geometric consistency between observations $\obsvs$ and estimated fundamental matrices $\models$.

\subsection{Transfer Error}
For homography estimation, we propose a new evaluation metric based on the square root of the symmetric transfer error~\cite{hartley2003multiple} between a homography $\vec{H}$ and a point correspondence $(\vec{p}^a, \vec{p}^b)$:
\begin{equation}
\begin{split}
     d_{\text{T}}(\vec{H}, \vec{p}^a, \vec{p}^b) = 
     \sqrt{ 
    d(\vec{p}^a, \vec{H}\inverse \vec{p}^b)^2 + d(\vec{p}^b, \vec{H} \vec{p}^a)^2
    } \, ,
\end{split}
\end{equation}
with $d(\vec{p},\vec{q})$ being the Euclidean distance between points $\vec{p}$ and $\vec{q}$, and $\vec{p}^a, \vec{p}^b$ being 2D points in homogeneous coordinates.
Given  a set of ground truth homographies $\gtmodels = \{ \gt{\vec{H}}_1, \dots, \gt{\vec{H}}_{\numgtmodels} \}$ and a sequence of estimated homographies $\models = ( {\vec{H}}_1, \dots, {\vec{H}}_{\nummodels} )$, we assume that $\models$ is sorted by significance, \ie ${\vec{H}}_1$ is the most important and ${\vec{H}}_{\nummodels}$ is the least important homography.
We compute the minimum transfer error of each observation $\obsv = (\vec{p}^a, \vec{p}^b) \in \obsvs$  using the $\tilde{M} = \min (\numgtmodels, \nummodels)$ most important estimated homographies:
\begin{equation}
    e_{\text{T}}(\obsv, \models) = \min_{j \in \{1, \dots, \tilde{M}\}} d_{\text{T}}({\vec{H}}_j, \vec{p}^a, \vec{p}^b) \, .
\end{equation}
If $\models = \emptyseq$, \ie no homographies were found by a method, we assume $\models = ( \vec{I} )$, with $\vec{I}$ being the identity matrix.
In order to avoid large outlier errors skewing the results, we clip the maximum value of $e_{\text{T}}$ to $\max(W,H)$, with $W$ and $H$ being the image width and height, respectively.
We then compute the mean of $e_{\text{T}}$ for all observations which are inliers according to the ground truth cluster labels $\gtlabels$:
\begin{equation}
    \mathrm{TE}(\obsvs, \models, \gtlabels) = \frac{\sum_{i=1}^{\numobsvs} e_{\mathrm{T}}(\obsv_i, \models) \cdot  \iverson{\gt{y}_i>0}}{\sum_{i=1}^{\numobsvs} \iverson{\gt{y}_i>0}} \, ,
\end{equation}
with $\gt{y}_i>0$ indicating that observation $\obsv_i$ is not a ground truth outlier.
This metric gauges the geometric consistency between observations $\obsvs$ and estimated homographies $\models$.

\section{Self-Supervised Learning}
\label{sec:selfsupervised}
\input{tables/results_unweighted}

In addition to supervised training, PARSAC can also be trained in a self-supervised fashion.
This means that we do not require ground truth labels, but use a task loss function that measures the quality of fit between our estimated models and our input data.
Similar to~\cite{kluger2020consac}, our task loss $\ell(\models)$ does not need to be differentiable.
We can thus, for example, use the negative inlier count as our loss for self-supervised learning:
\begin{equation}
    \ell(\models) = -|\set{I}| \, ,
    \label{eq:inlier_loss}
\end{equation}
with $\set{I}$ being the set of inlier observations \wrt final models $\models = ( \model_1, \dots, \model_{\nummodels} )$.
This is the approach used by~\cite{kluger2020consac}, but we noticed that it leads to sub-optimal results with PARSAC.
Simply maximising the inlier count encourages PARSAC to find as many model instances as possible, even if they are very similar, thus leading to over-segmentation of the data.
Instead, we emphasise inliers of the most prominent models by applying an exponential weighting to the inlier scores with $\gamma = 0.3$:
\begin{equation}
    \ell(\models) = - \sum_{i=1}^{\numobsvs} \sum_{j=1}^{\nummodels} \rho_{i, j} \cdot {\gamma}^j \, ,
\end{equation}
\begin{equation}
    \rho_{i, j} = \max_{\model \in \{ \model_1, \dots, \model_j \} } s(d(\obsv_i, \model)) \, .
\end{equation}
In Tab.~\ref{tab:results_self}, we compare self-supervised PARSAC using this weighted inlier loss against self-supervised PARSAC with the vanilla inlier-based loss (Eq.~\ref{eq:inlier_loss}).
We also compare both with supervised PARSAC, as well as with CONSAC~\cite{kluger2020consac}.
We evaluate for vanishing point estimation on SU3 and NYU-VP, and for fundamental matrix estimation on HOPE-F and Adelaide-F~\cite{wong2011dynamic}.
Self-supervised PARSAC with the weighted loss clearly outperforms self-supervised PARSAC with the unweighted inlier loss.
On SU3, it also performs better than CONSAC -- both supervised and self-supervised -- and is on par with self-supervised CONSAC on NYU-VP.

\section{Weighted Inlier Counting}
\label{sec:ablation_weighted}
We evaluate the importance of our weighted inlier counting for hypothesis selection (cf. Eq.~\ref{eq:weighted_inlier_count}) by comparing it with regular unweighted inlier counting:
\begin{equation}
    I(\hypothesis_{j,k};\obsvs) = \sum_{\obsv \in \obsvs} s(d(\obsv, \hypothesis_{j,k}))\, .
\end{equation}
In Tab.~\ref{tab:results_unweighted}, we report the performance using either weighted or unweighted inlier counting for vanishing point detection on SU3, fundamental matrix estimation on HOPE-F, and homography estimation on SMH.
As these results show, weighted inlier counting is crucial for PARSAC to achieve optimal accuracy.

\section{Feature Generalisation}
\label{sec:ablation_feature}
\input{tables/results_deeplsd}
To test how well PARSAC generalises to other types of features, we evaluate vanishing point detection using line segments detected with DeepLSD~\cite{pautrat2023deeplsd} instead of LSD~\cite{von2008lsd}. 
We show the results in Tab.~\ref{tab:results_deeplsd}.
The first row reports mean AUCs for PARSAC trained and evaluated with LSD, i.e. results from Tabs.~\ref{tab:results_vp_manhattan}-\ref{tab:results_vp} for reference.
The second row shows results for PARSAC trained with LSD but evaluated with DeepLSD. 
We observe a small or moderate decrease in AUC on all datasets, but the results are still competitive.
If we train PARSAC with DeepLSD features but test it with LSD features (row 3), performance gets a bit better for all datasets but NYU-VP, and yet not quite as good as PARSAC when trained and tested with LSD.
Lastly, when both trained and tested with DeepLSD (row 4), performance decreases a bit on NYU-VP but stays nearly the same on all other datasets. 
We also compare Progressive-X using either LSD or DeepLSD. 
As Tab.~\ref{tab:results_deeplsd} shows, Progressive-X performs similarly with LSD or DeepLSD features if optimised using LSD (rows 5-6).
Curiously, it achieves higher AUCs when optimised with DeepLSD but tested with LSD (row 7).
Overall, PARSAC still achieves the highest AUCs regardless of which features are used.

\section{Number of Model Instances}
\label{sec:ablation_instances}
\input{tables/results_instances}

The maximum number of model instances which PARSAC can discover is limited by the number of putative models $\numpmodels$. 
To investigate the influence of $\numpmodels$, we conducted an ablation study on the SU3 dataset. 
All scenes in SU3 have exactly three vanishing points. 
We vary $\numpmodels$ between 2 and 16 and report mean AUC values in Tab.~\ref{tab:results_instances}. 
The results are consistent for $\numpmodels > 3$, i.e. there is no penalty \wrt accuracy if we set $\numpmodels$ very high, apart from an increased computation time. 
If $\numpmodels$ is too small, e.g. $\numpmodels < 3$ for SU3, some models are simply not found, leading to a lower AUC.

\section{Robustness to Noise and Outliers}
\label{sec:ablation_robustness}
To assess the robustness of our method, we conduct additional experiments on the SU3 dataset for vanishing point detection, on Adelaide-F for fundamental matrices, and on Adelaide-H for homographies. 
We compare against Progressive-X, Progressive-X+, T-Linkage and CONSAC. 

\subsection{Noise}
\begin{figure}	
    \setlength{\imgwidth}{\linewidth}
    \begin{subfigure}{\imgwidth}
    \centering
        \includegraphics[width=\imgwidth]{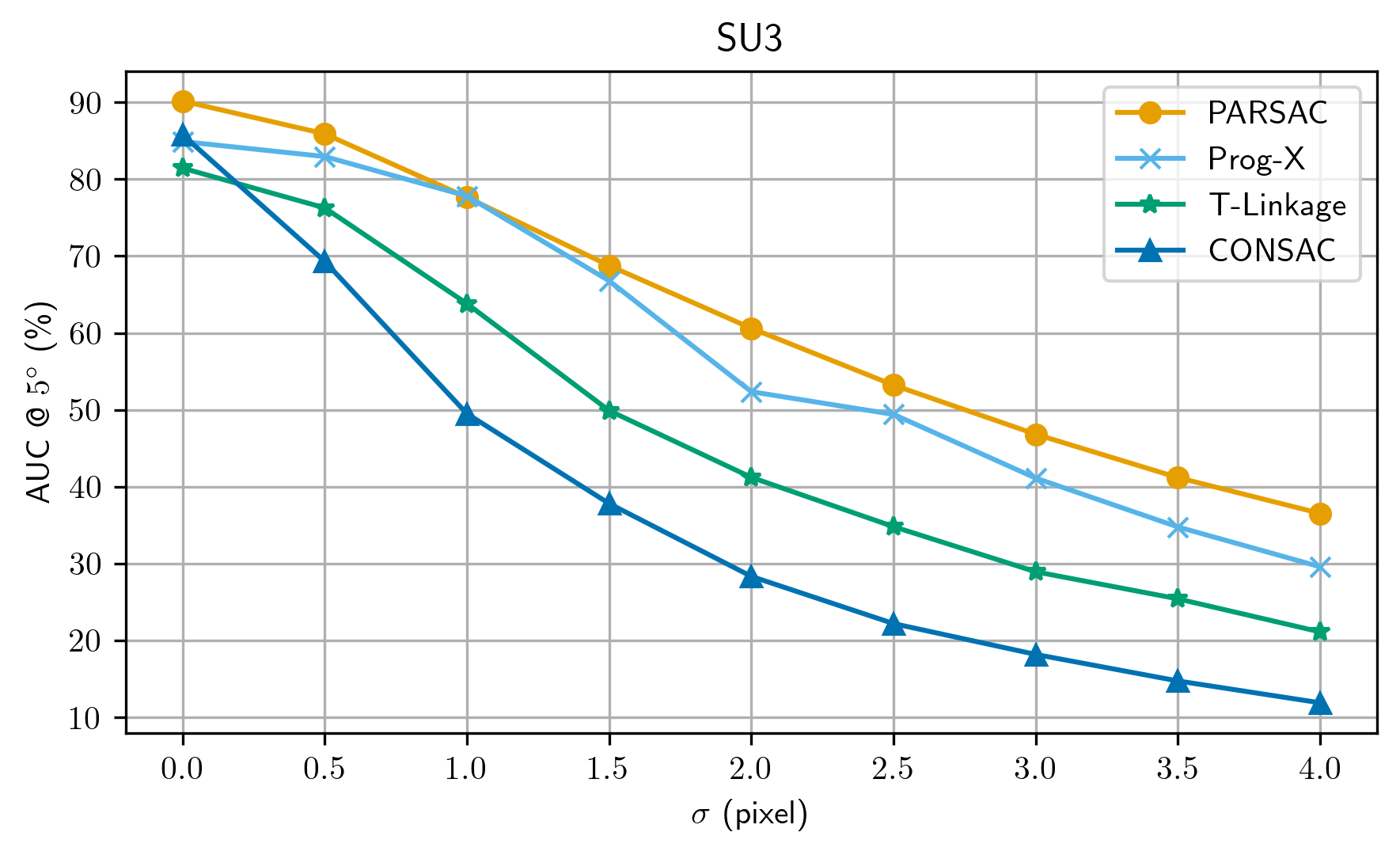}
        \caption{Vanishing point estimation on SU3}
        \label{fig:noise_vp}
    \end{subfigure}
    \begin{subfigure}{\imgwidth}
    \centering
        \includegraphics[width=\imgwidth]{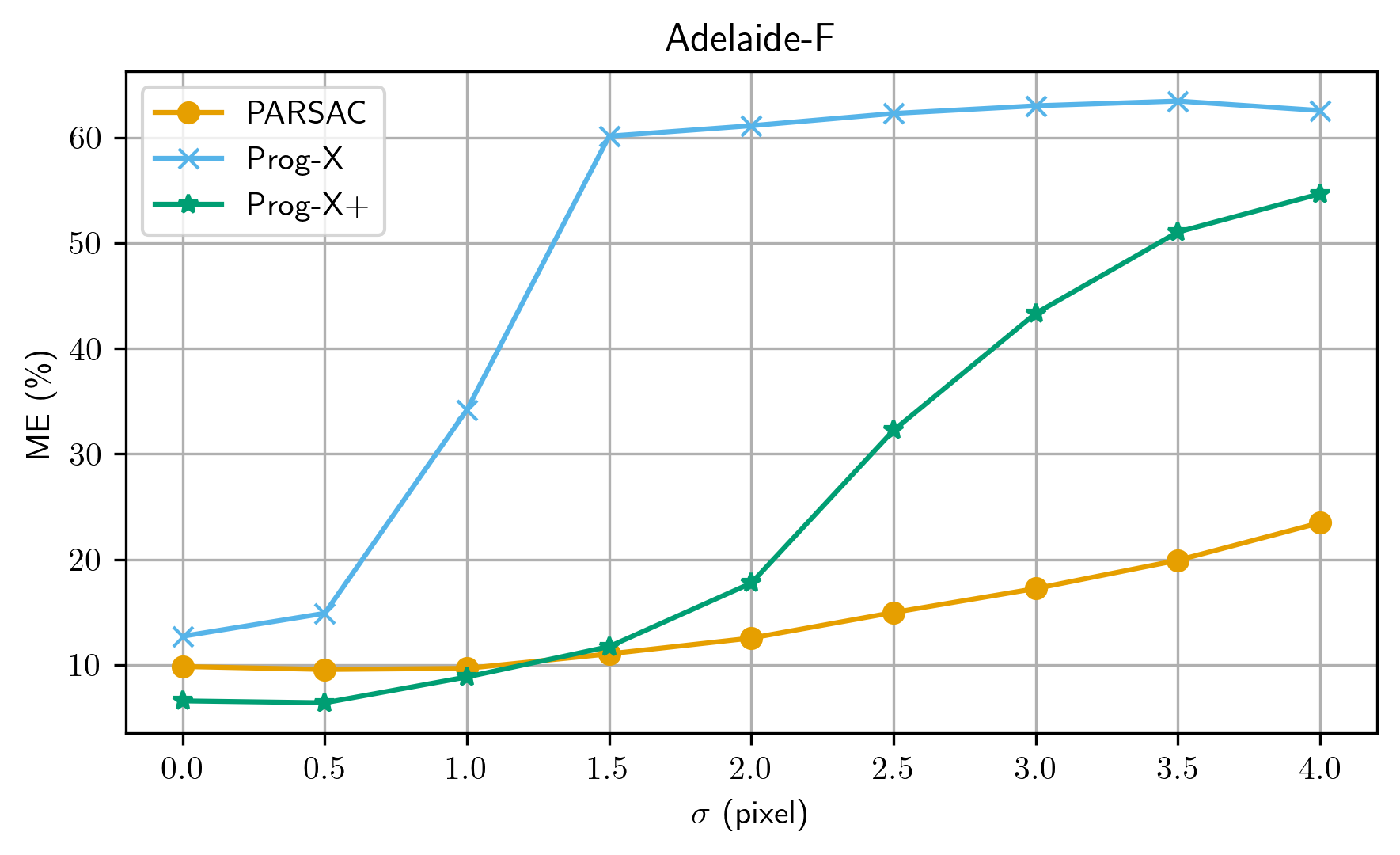}
        \caption{Fundamental matrix estimation on Adelaide-F}
        \label{fig:noise_f}
    \end{subfigure}
    \begin{subfigure}{\imgwidth}
    \centering
        \includegraphics[width=\imgwidth]{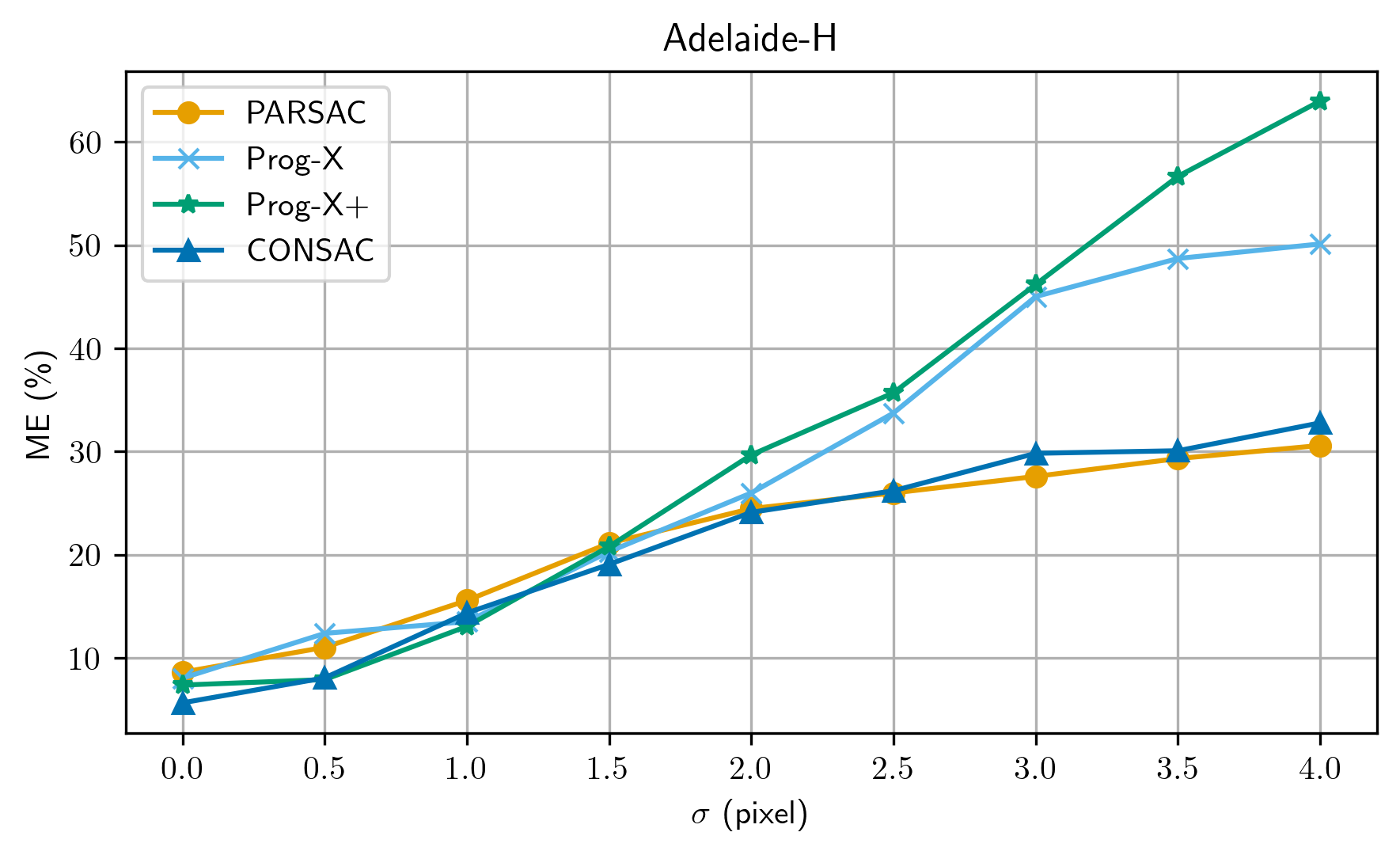}
        \caption{Homography estimation on Adelaide-H}
        \label{fig:noise_h}
    \end{subfigure}
    
    \caption{{Robustness to Noise: } We add Gaussian noise with varying standard deviation $\sigma$ to our input features and evaluate the AUC @ $5\degree$ (higher is better) for vanishing point estimation, as well as the misclassification error (ME, lower is better) for fundamental matrix and homography estimation. We compare PARSAC against Progressive-X, Progressive-X+, T-Linkage and CONSAC.
    }
		\label{fig:noise}
\end{figure}

We add Gaussian noise with zero mean and varying standard deviation $\sigma$ to our input features, \ie line segment end point coordinates and SIFT keypoint coordinates. 
Fig.~\ref{fig:noise_vp} shows AUC values for vanishing point estimation on SU3.
All methods exhibit a steady decline in AUC as $\sigma$ increases, but PARSAC retains the highest accuracy at all times.
CONSAC, on the other hand, degrades most rapidly.
For F-matrix estimation on Adelaide-F, Fig.~\ref{fig:noise_f} shows that the misclassification errors (ME) of Progressive-X and Progressive-X+ increase rapidly after $\sigma=0.5$ and $\sigma=2.0$, respectively.
By comparison, PARSAC shows only a moderate increase in ME after $\sigma=2.0$.
In Fig.~\ref{fig:noise_h}, all methods show a similar steady increase in ME up to $\sigma=1.5$ for homography estimation on Adelaide-H.
Beyond this point, the performance of Progressive-X and Progressive-X+ decreases further, while PARSAC and CONSAC level off.

\subsection{Outliers}
\begin{figure}	
    \setlength{\imgwidth}{\linewidth}
    \begin{subfigure}{\imgwidth}
    \centering
        \includegraphics[width=\imgwidth]{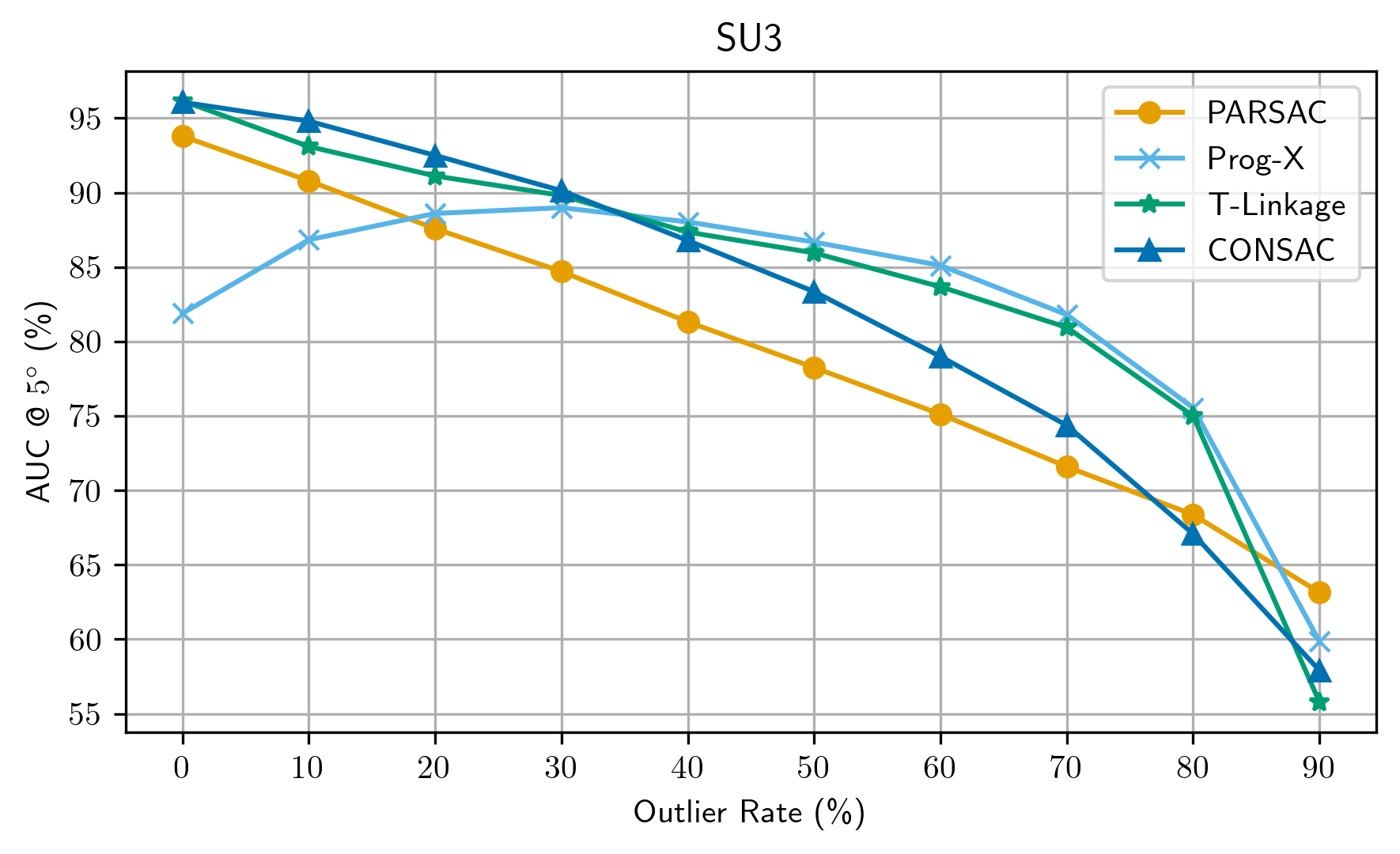}
        \caption{Vanishing point estimation on SU3}
        \label{fig:outlier_vp}
    \end{subfigure}
    \begin{subfigure}{\imgwidth}
    \centering
        \includegraphics[width=\imgwidth]{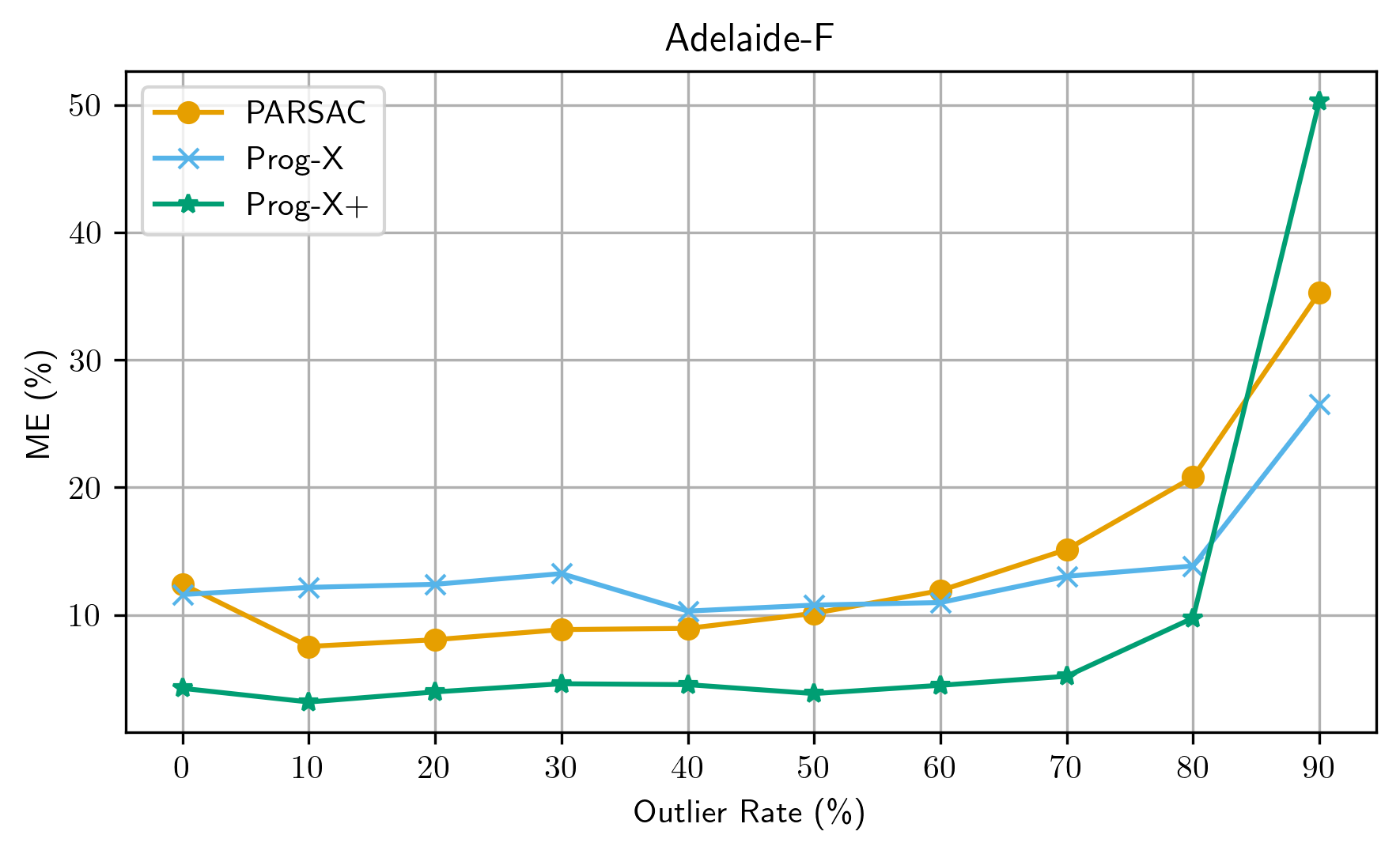}
        \caption{Fundamental matrix estimation on Adelaide-F}
        \label{fig:outlier_f}
    \end{subfigure}
    \begin{subfigure}{\imgwidth}
    \centering
        \includegraphics[width=\imgwidth]{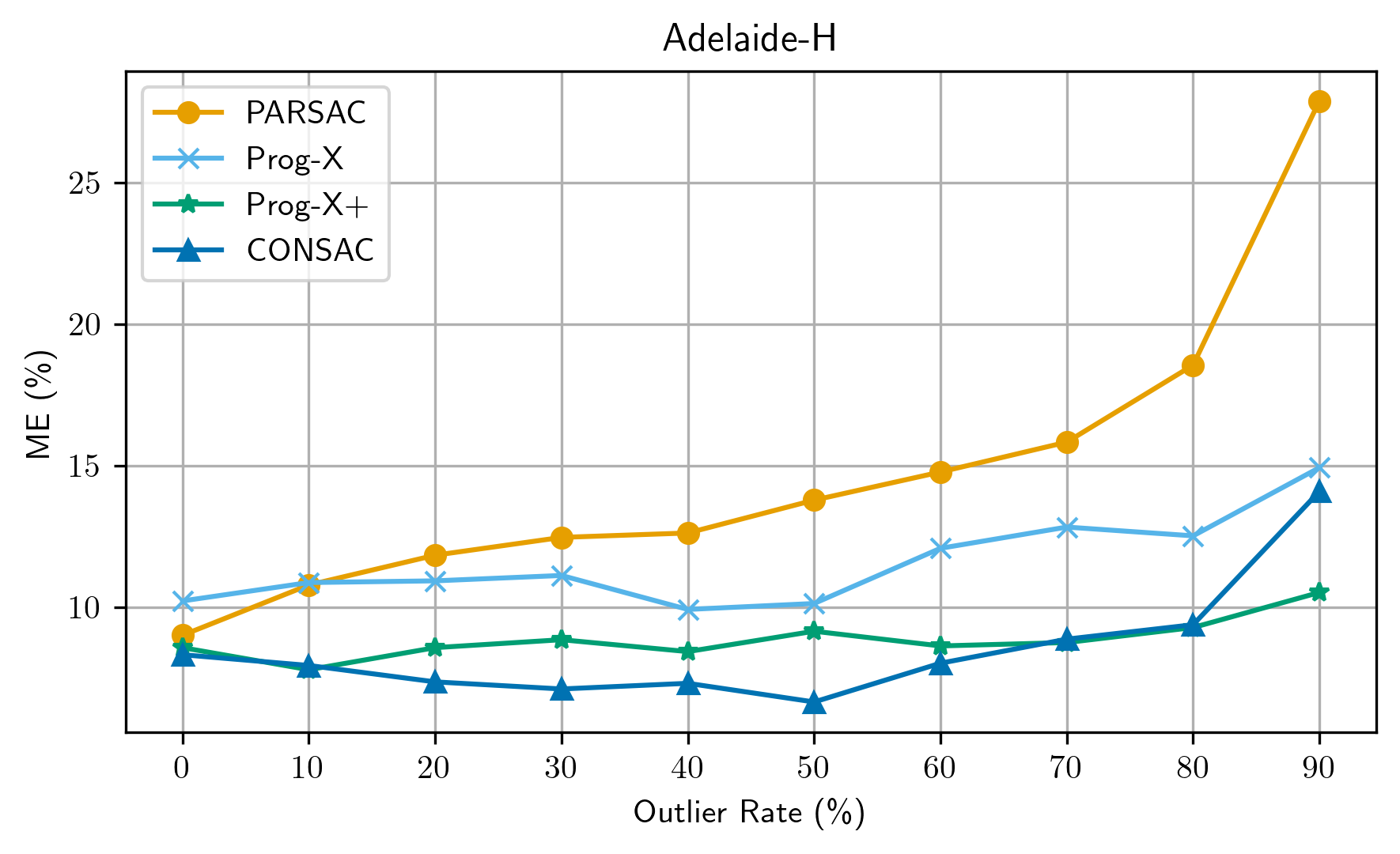}
        \caption{Homography estimation on Adelaide-H}
        \label{fig:outlier_h}
    \end{subfigure}
    
    \caption{{Robustness to Outliers: } We add synthetic outliers with varying outlier rates to our input features and evaluate the AUC @ $5\degree$ (higher is better) for vanishing point estimation, as well as the misclassification error (ME, lower is better) for fundamental matrix and homography estimation. We compare PARSAC against Progressive-X, Progressive-X+, T-Linkage and CONSAC.
    }
		\label{fig:outlier}
\end{figure}

We first remove all outliers from the input features, then add synthetic outliers with varying outlier rates. 
For vanishing point estimation, we define outliers as line segments with an angle of more than $0.5\degree$ to any constrained line arising from the centre of the line segment and a ground truth vanishing point.
We generate synthetic outlier lines by sampling one point within the image frame uniformly for each line, and then sampling the two line segment end points uniformly within a $100\times 100$ pixel square centred on the previously sampled point.
For fundamental matrices and homographies, outliers are defined by the ground truth class labels.
We sample synthetic outliers uniformly within the image frame. 
We calculate the misclassification errors only for ground truth inliers, as varying outlier rates would skew the results otherwise.
On SU3, PARSAC shows a linear decline in AUC with an increasing outlier rate (cf. Fig.~\ref{fig:outlier_vp}). 
Curiously, Progressive-X performs best at an outlier rate of 30\%, and is relatively stable between 10-50\% outlier rates. 
CONSAC and T-Linkage are also very stable up to around 70\% outlier rate. 
On Adelaide-F (Fig.~\ref{fig:outlier_f}), PARSAC remains stable up to a 60\% outlier rate, but degrades after that. 
Progressive-X and Progressive-X+ degrade a bit later. 
On Adelaide-H (Fig.~\ref{fig:outlier_h}), Progressive-X, Progressive-X+ and CONSAC are stable across the whole range, while PARSAC shows a steadily increasing ME.

\section{Parameter Sensitivity}
\label{sec:ablation_sensitivity}
PARSAC has four user-definable parameters for inference:
\begin{itemize}
    \item Number of model hypotheses $\numhypotheses$
    \item Inlier threshold $\tau$
    \item Inlier softness $\beta$
    \item Assignment threshold $\tau_a$
\end{itemize}
We analyse the sensitivity of PARSAC \wrt each parameter while using fixed values for all other parameters (Tab.~\ref{tab:parameters}). 
For vanishing point estimation, we use the validation sets of NYU-VP and SU3, and report the AUC @ $10\degree$ and AUC @ $5\degree$, respectively.
For fundamental matrix estimation, we use Adelaide-F and the validation set of HOPE-F, and report the misclassification error (ME) and Sampson error (SE).
For homography estimation, we use Adelaide-H and the validation set from SMH, and report the misclassification error (ME) and transfer error (TE).

\subsection{Number of Model Hypotheses}
\begin{figure}	
    \setlength{\imgwidth}{\linewidth}
    \begin{subfigure}{\linewidth}
    \centering
        \includegraphics[width=\imgwidth]{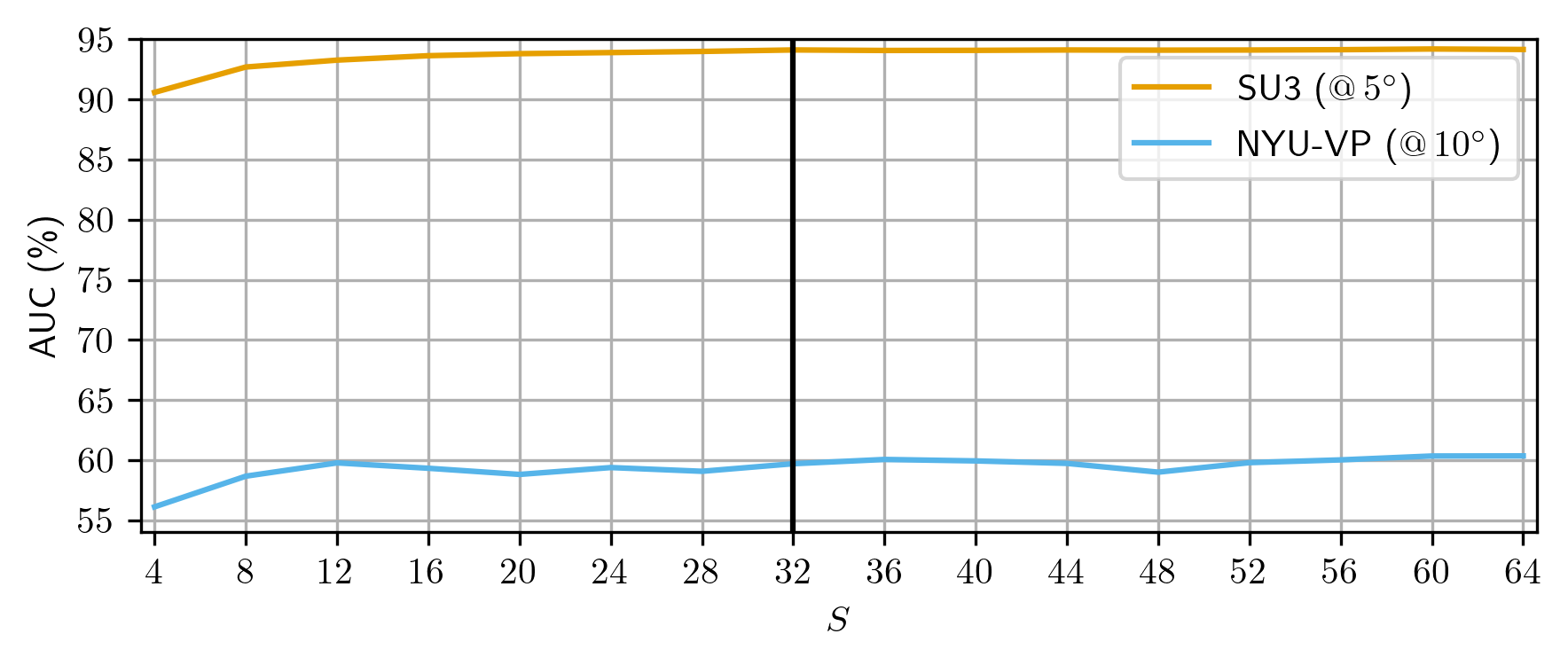}
        \caption{Vanishing point estimation}
        \label{fig:sensitivity_S:vp}
    \end{subfigure}
    \begin{subfigure}{\linewidth}
    \centering
        \includegraphics[width=\imgwidth]{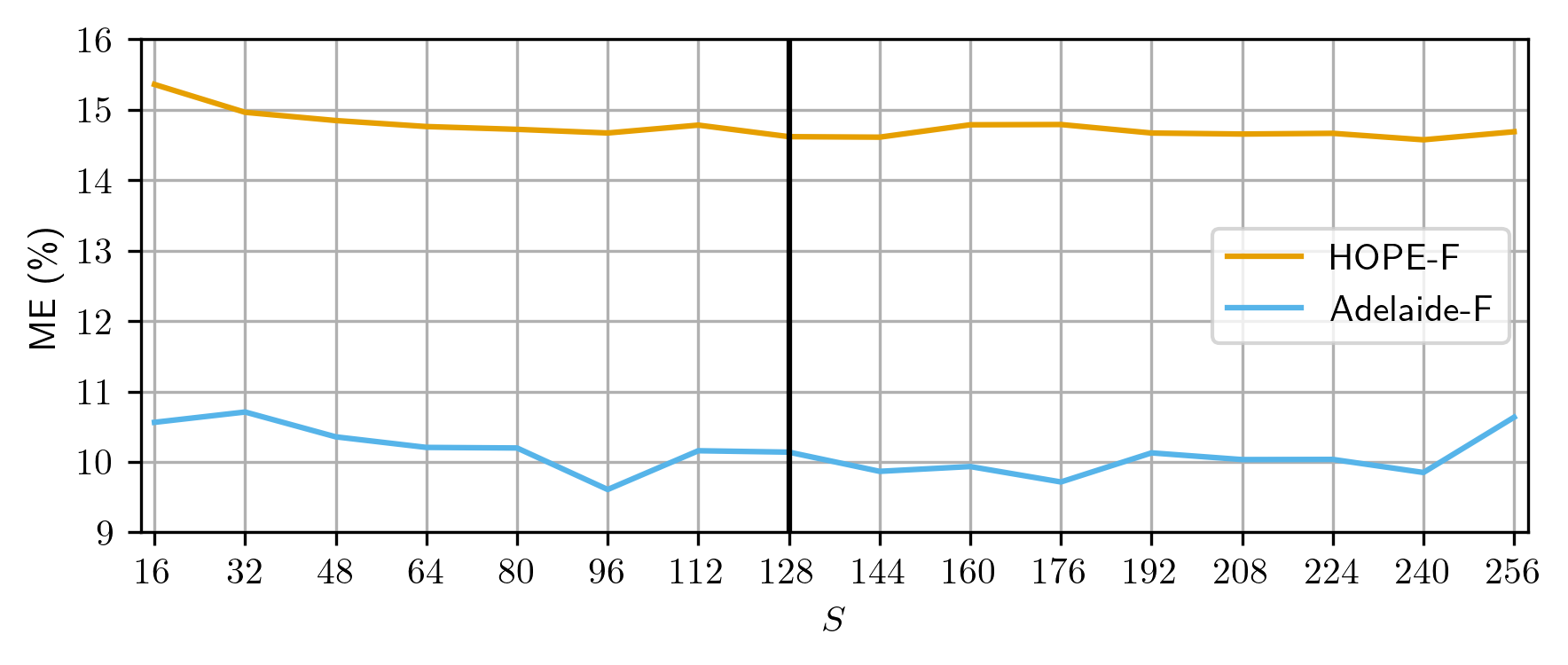}
        \caption{Fundamental matrix estimation (misclassification error)}
        \label{fig:sensitivity_S:fun_ME}
    \end{subfigure}
    \begin{subfigure}{\linewidth}
    \centering
        \includegraphics[width=\imgwidth]{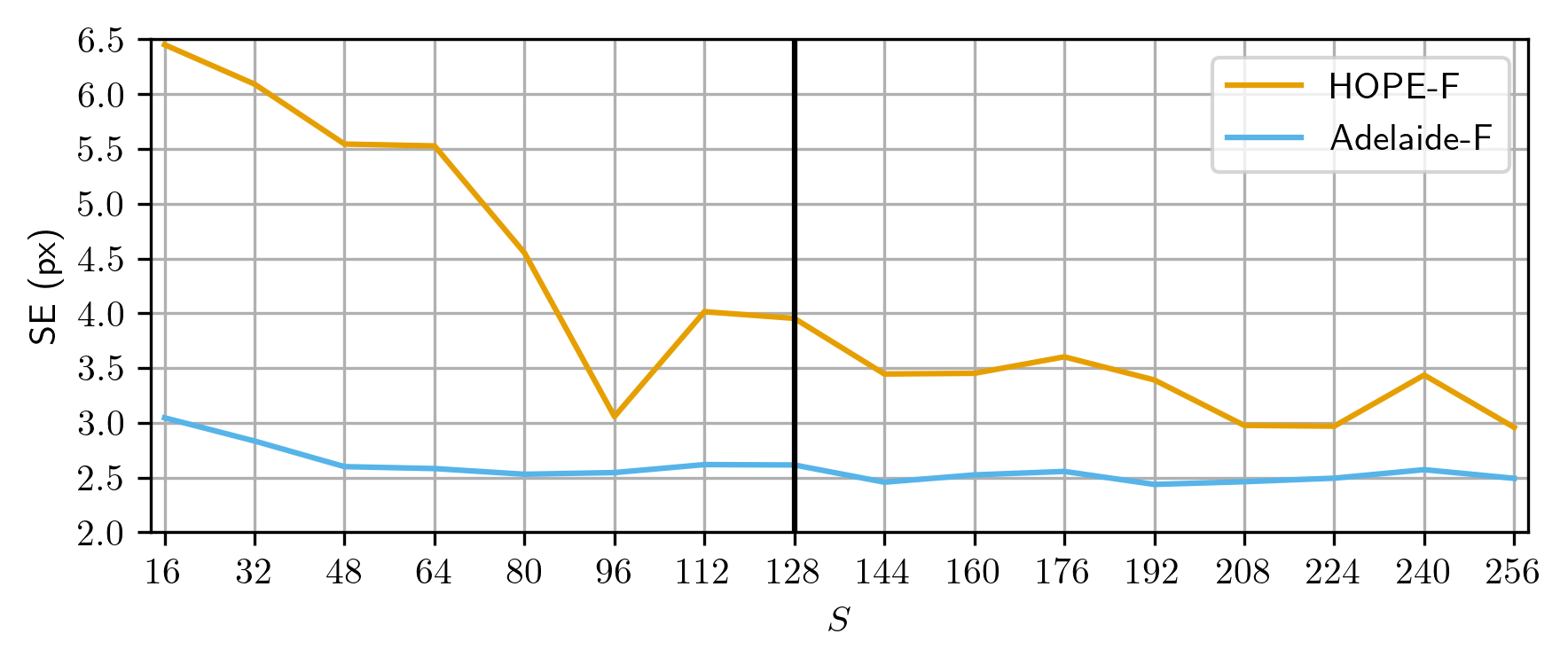}
        \caption{Fundamental matrix estimation (Sampson error)}
        \label{fig:sensitivity_S:fun_SE}
    \end{subfigure}
    \begin{subfigure}{\linewidth}
    \centering
        \includegraphics[width=\imgwidth]{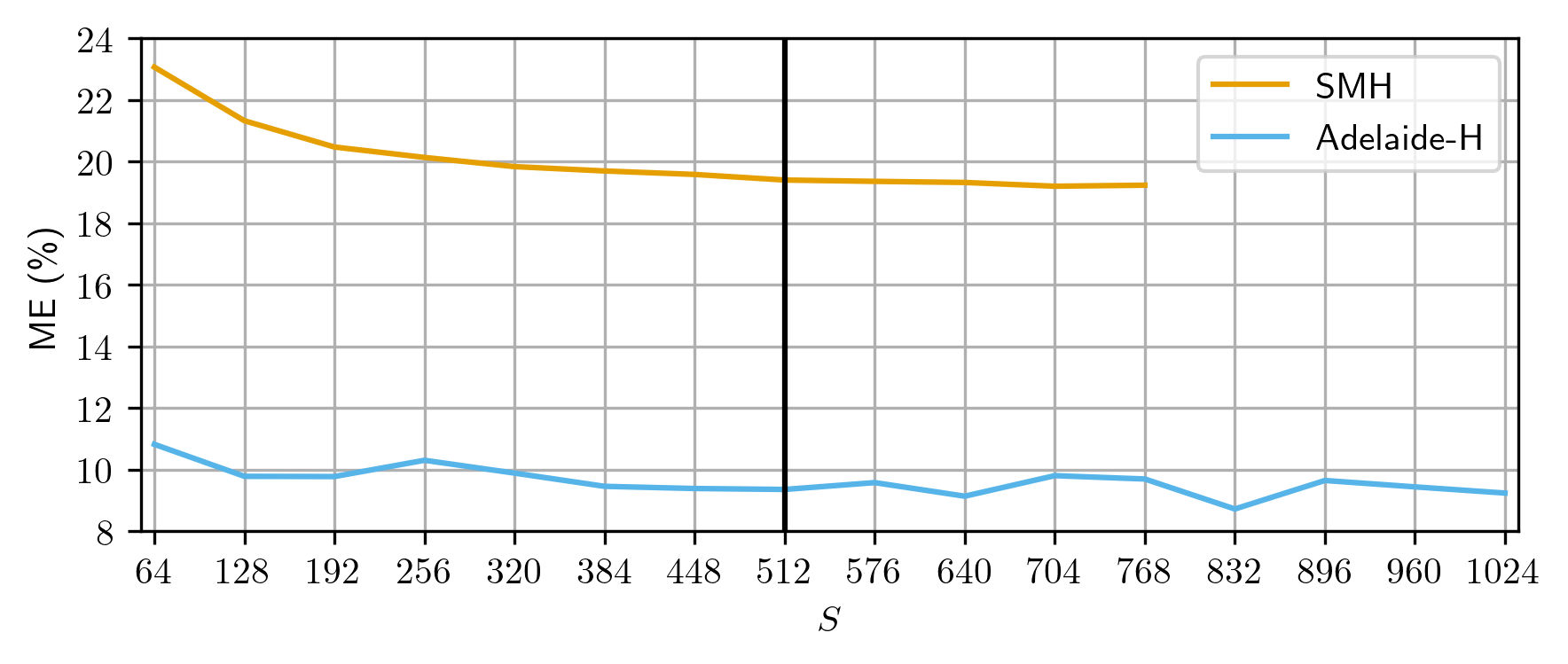}
        \caption{Homography estimation (misclassification error)}
        \label{fig:sensitivity_S:hom_ME}
    \end{subfigure}
    \begin{subfigure}{\linewidth}
    \centering
        \includegraphics[width=\imgwidth]{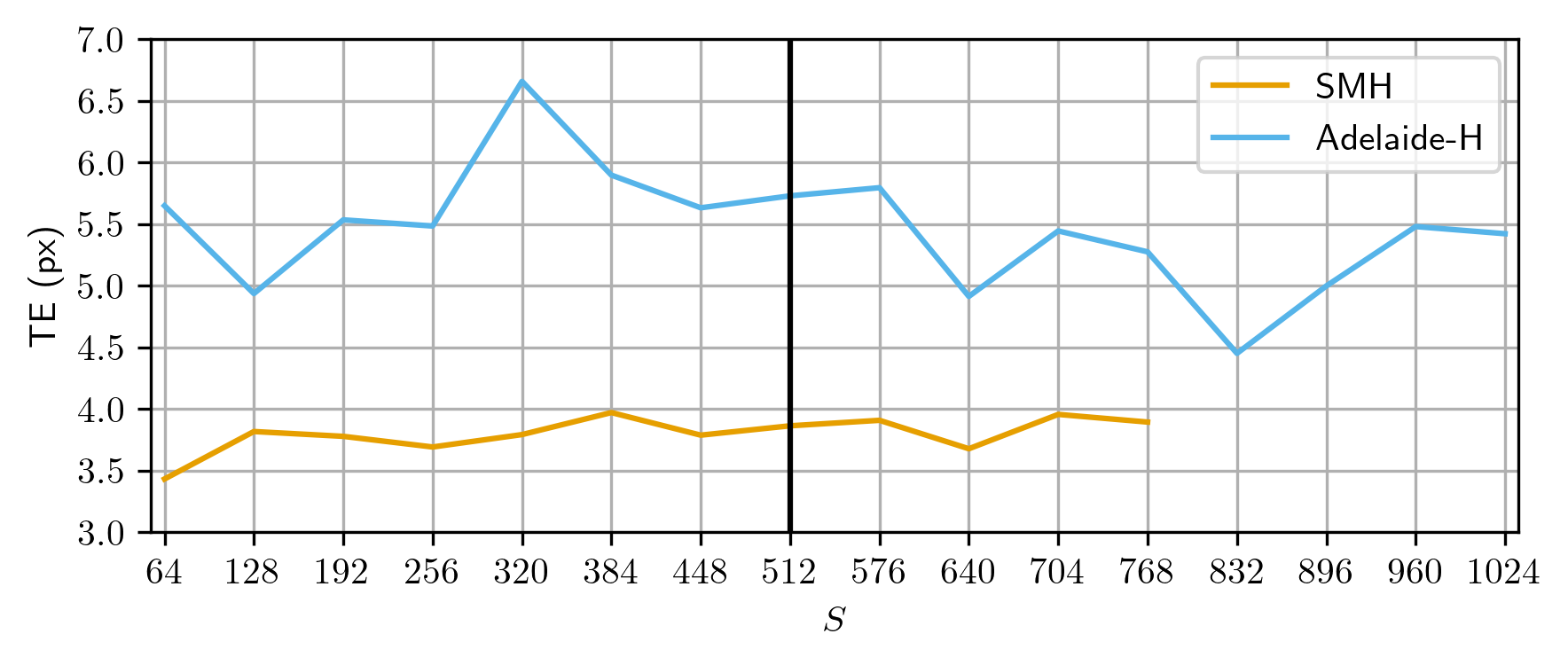}
        \caption{Homography estimation (transfer error)}
        \label{fig:sensitivity_S:hom_TE}
    \end{subfigure}
    \caption{Sensitivity of PARSAC \wrt the number of hypotheses $S$ for vanishing points (a), fundamental matrices (b-c) and homographies (d-e). Cyan coloured vertical lines indicate the values we used for our main experiments. }
    \label{fig:sensitivity_S}
\end{figure}

As Fig.~\ref{fig:sensitivity_S} shows, the accuracy of PARSAC generally increases when we increase the number of model hypotheses $\numhypotheses$, albeit with diminishing returns.
This is to be expected, as a larger number of hypotheses increases the likelihood of sampling and then selecting models with larger inlier counts.

\subsection{Inlier Threshold}
PARSAC exhibits clear optima for each metric and dataset \wrt the inlier threshold $\tau$.
For vanishing point estimation (Fig.~\ref{fig:sensitivity_tau:vp}), the AUC on SU3 deviates by $1.6\%$ from the maximum at most across a range of values for $\tau$ covering two orders of magnitude.
On NYU-VP, the AUC deviates by up to $11\%$.
For homography and fundamental matrix estimation, the errors vary more significantly across the tested range of values.
For fundamental matrix estimation (Figs.~\ref{fig:sensitivity_tau:fun_ME}-\ref{fig:sensitivity_tau:fun_SE}), errors are minimal on both HOPE-F and Adelaide-F near $\tau = 10^{-2}$.
On the other hand, for homography estimation (Figs.~\ref{fig:sensitivity_tau:hom_ME}-\ref{fig:sensitivity_tau:hom_TE}) we observe different global minima on SMH and Adelaide-H.
Additionally, on SMH, the minima for ME and TE are at different values of $\tau$, and we thus had to select an inlier threshold which yields a compromise between the two metrics.

\subsection{Inlier Softness}
\begin{figure}	
    \setlength{\imgwidth}{\linewidth}
    \begin{subfigure}{\linewidth}
    \centering
        \includegraphics[width=\imgwidth]{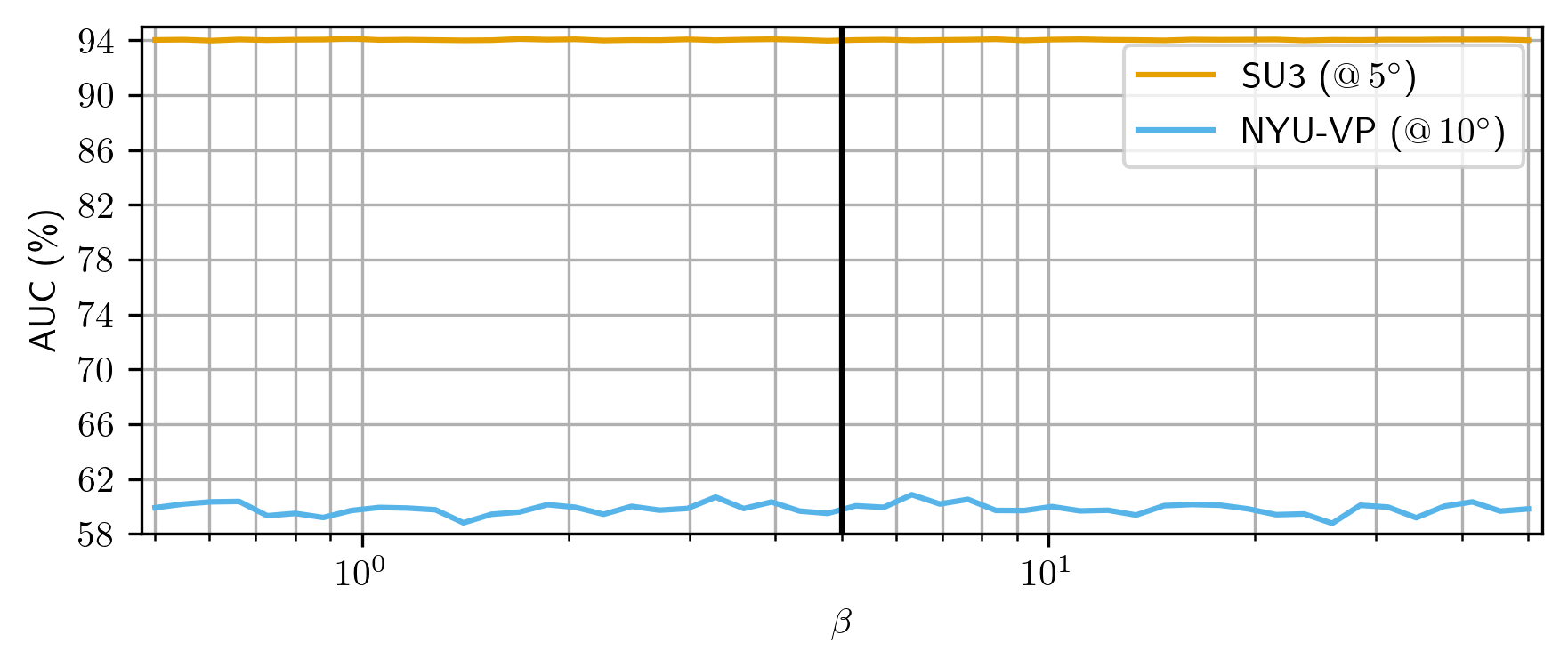}
        \caption{Vanishing point estimation}
        \label{fig:sensitivity_beta:vp}
    \end{subfigure}
    \begin{subfigure}{\linewidth}
    \centering
        \includegraphics[width=\imgwidth]{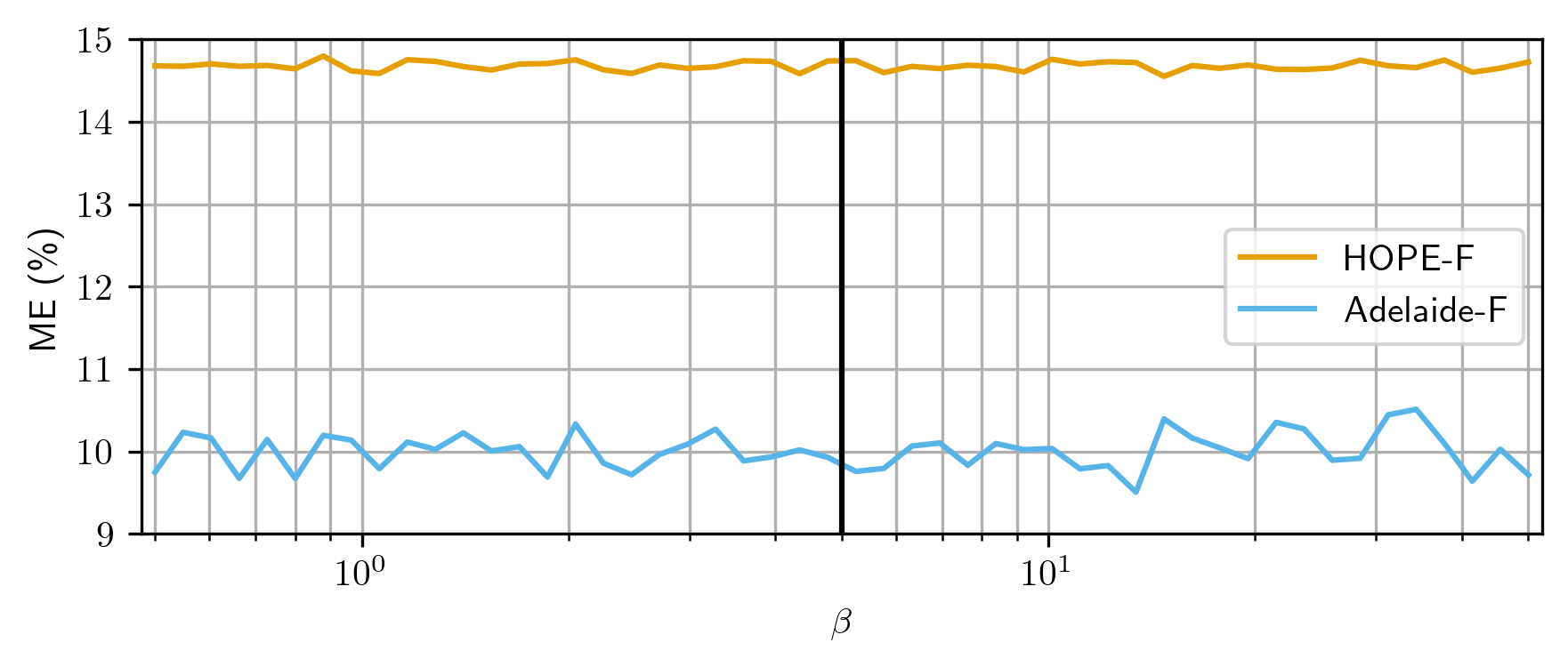}
        \caption{Fundamental matrix estimation (misclassification error)}
        \label{fig:sensitivity_beta:fun_ME}
    \end{subfigure}
    \begin{subfigure}{\linewidth}
    \centering
        \includegraphics[width=\imgwidth]{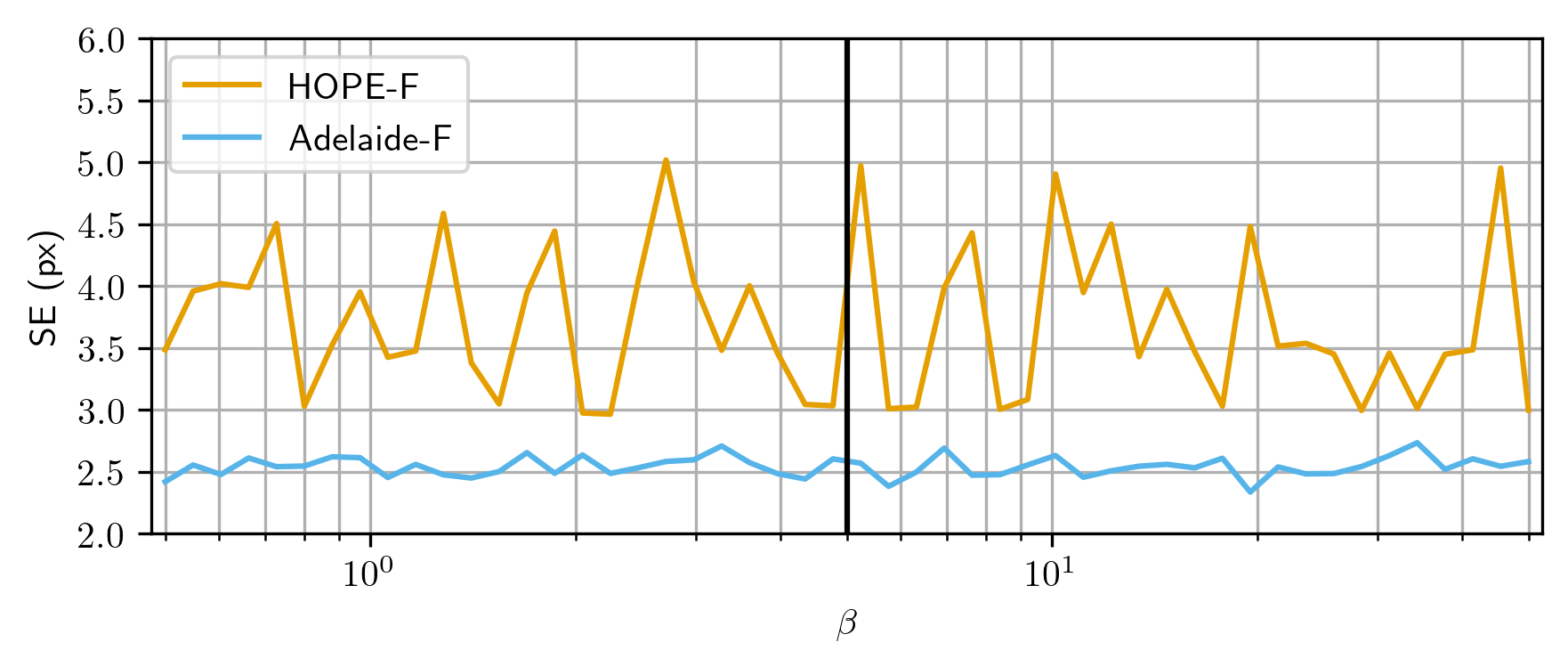}
        \caption{Fundamental matrix estimation (Sampson error)}
        \label{fig:sensitivity_beta:fun_SE}
    \end{subfigure}
    \begin{subfigure}{\linewidth}
    \centering
        \includegraphics[width=\imgwidth]{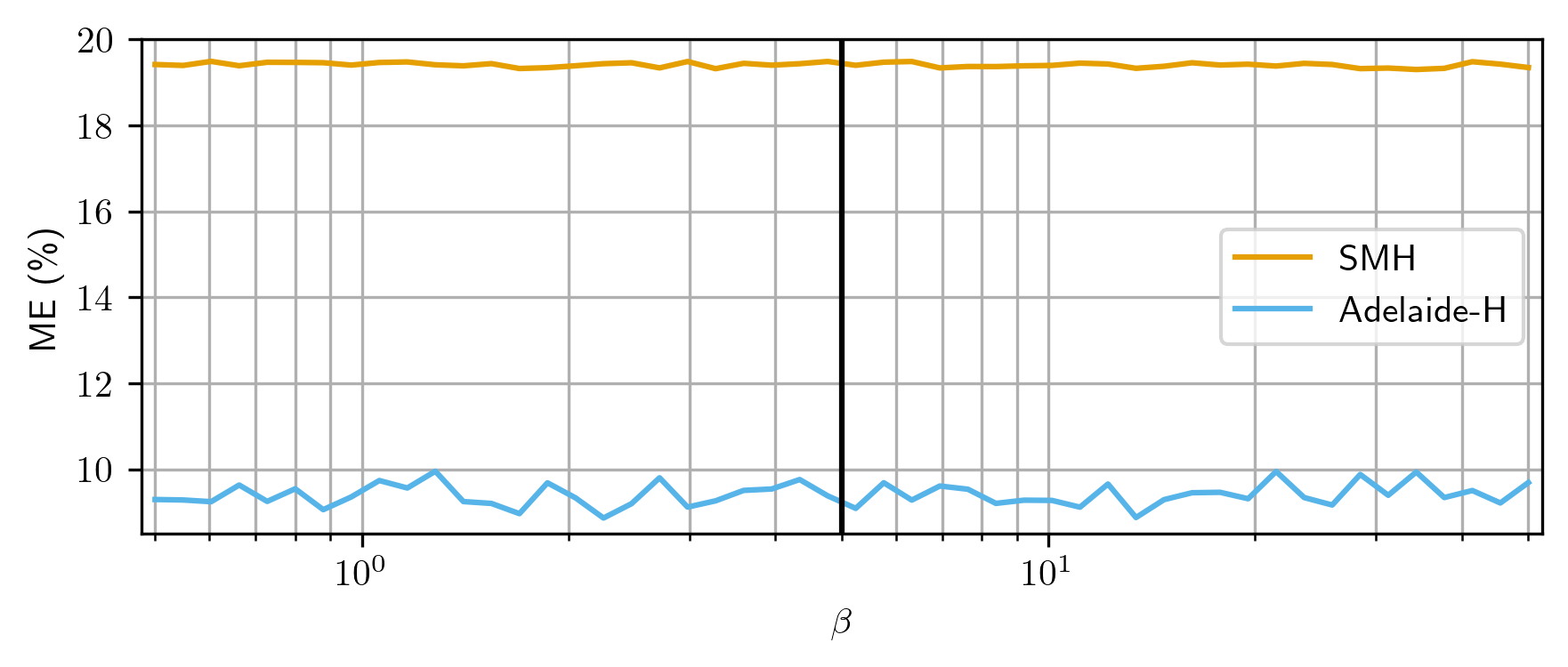}
        \caption{Homography estimation (misclassification error)}
        \label{fig:sensitivity_beta:hom_ME}
    \end{subfigure}
    \begin{subfigure}{\linewidth}
    \centering
        \includegraphics[width=\imgwidth]{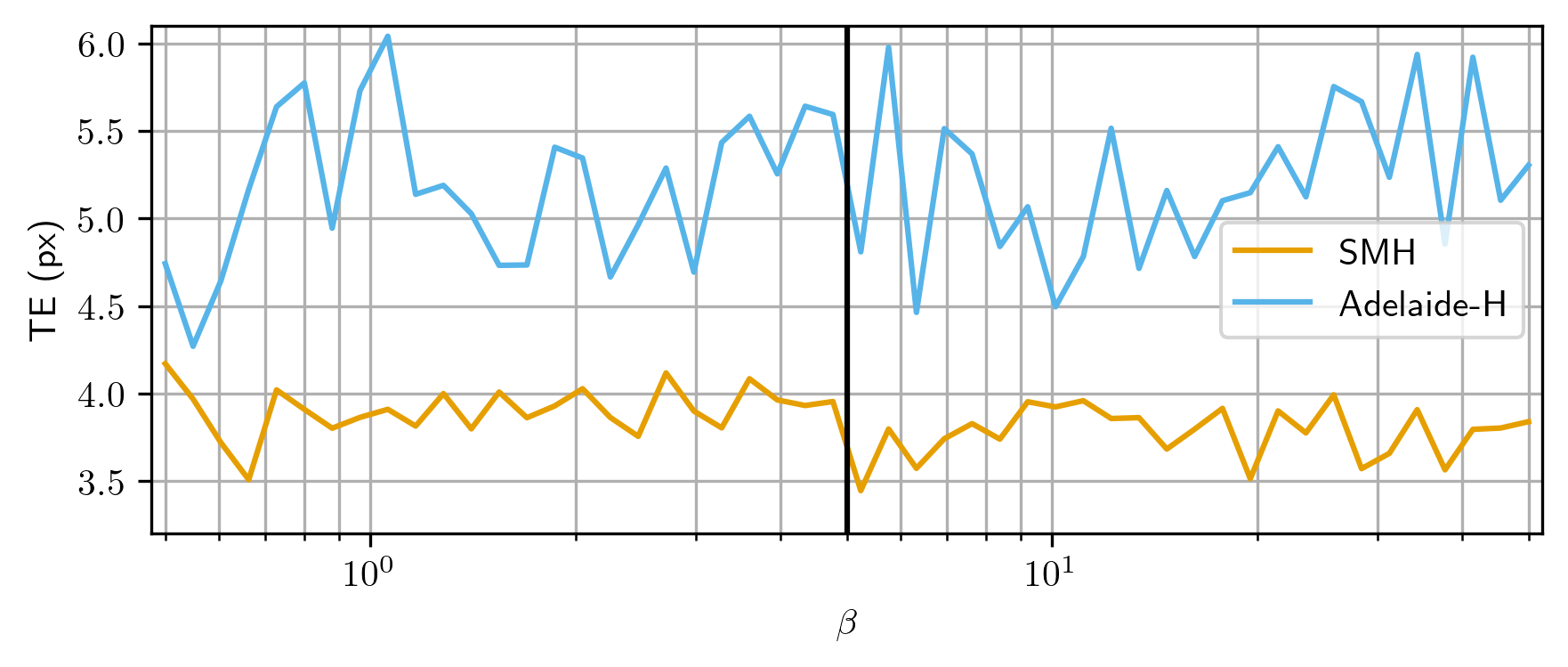}
        \caption{Homography estimation (transfer error)}
        \label{fig:sensitivity_beta:hom_TE}
    \end{subfigure}
    \caption{Sensitivity of PARSAC \wrt the inlier softness $\beta$ for vanishing points (a), fundamental matrices (b-c) and homographies (d-e).  Black vertical lines indicate the values we used for our main experiments (cf. Tab.~\ref{tab:parameters}).}
    \label{fig:sensitivity_beta}
\end{figure}

As Fig.~\ref{fig:sensitivity_beta} shows, PARSAC is not sensitive to the value of the inlier softness parameter $\beta$ within the range of values that we have tested.

\subsection{Assignment Threshold}
\begin{figure}	
    \setlength{\imgwidth}{\linewidth}
    \begin{subfigure}{\linewidth}
    \centering
        \includegraphics[width=\imgwidth]{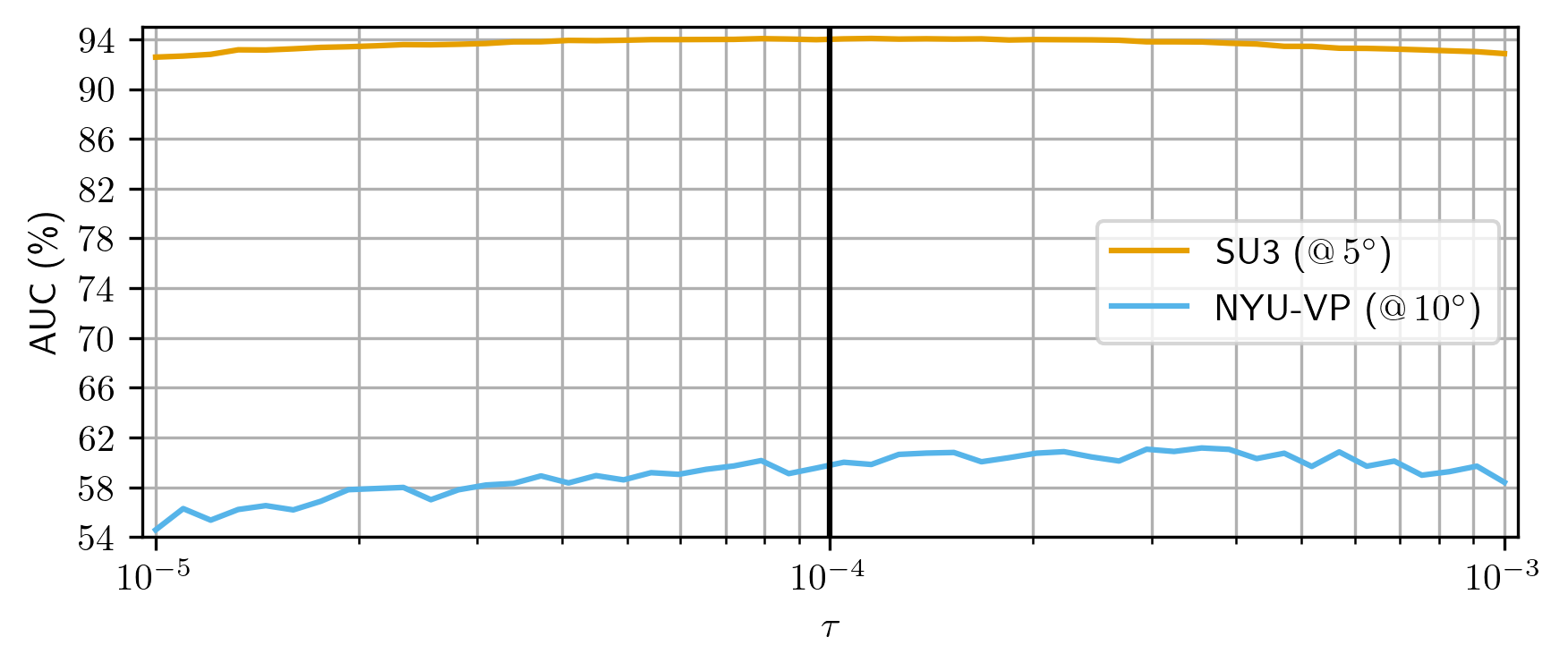}
        \caption{Vanishing point estimation}
        \label{fig:sensitivity_tau:vp}
    \end{subfigure}
    \begin{subfigure}{\linewidth}
    \centering
        \includegraphics[width=\imgwidth]{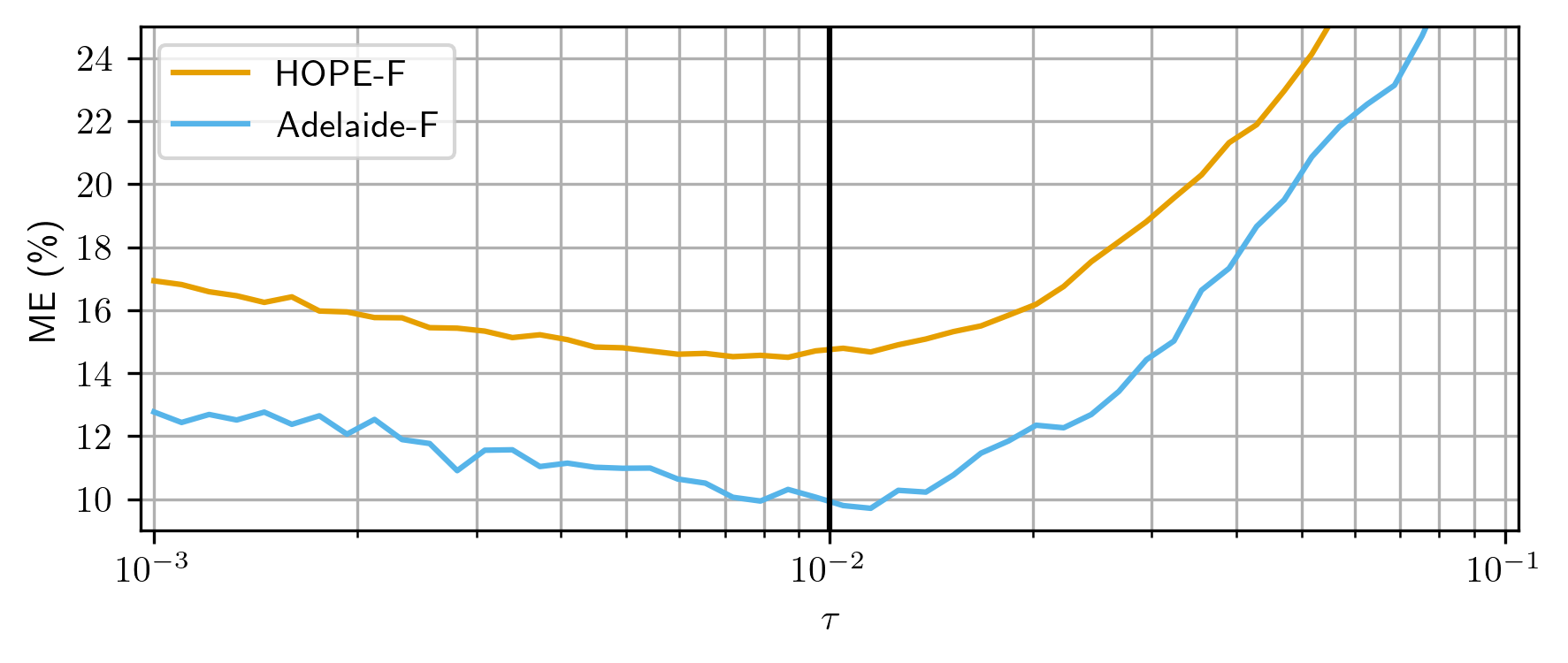}
        \caption{Fundamental matrix estimation (misclassification error)}
        \label{fig:sensitivity_tau:fun_ME}
    \end{subfigure}
    \begin{subfigure}{\linewidth}
    \centering
        \includegraphics[width=\imgwidth]{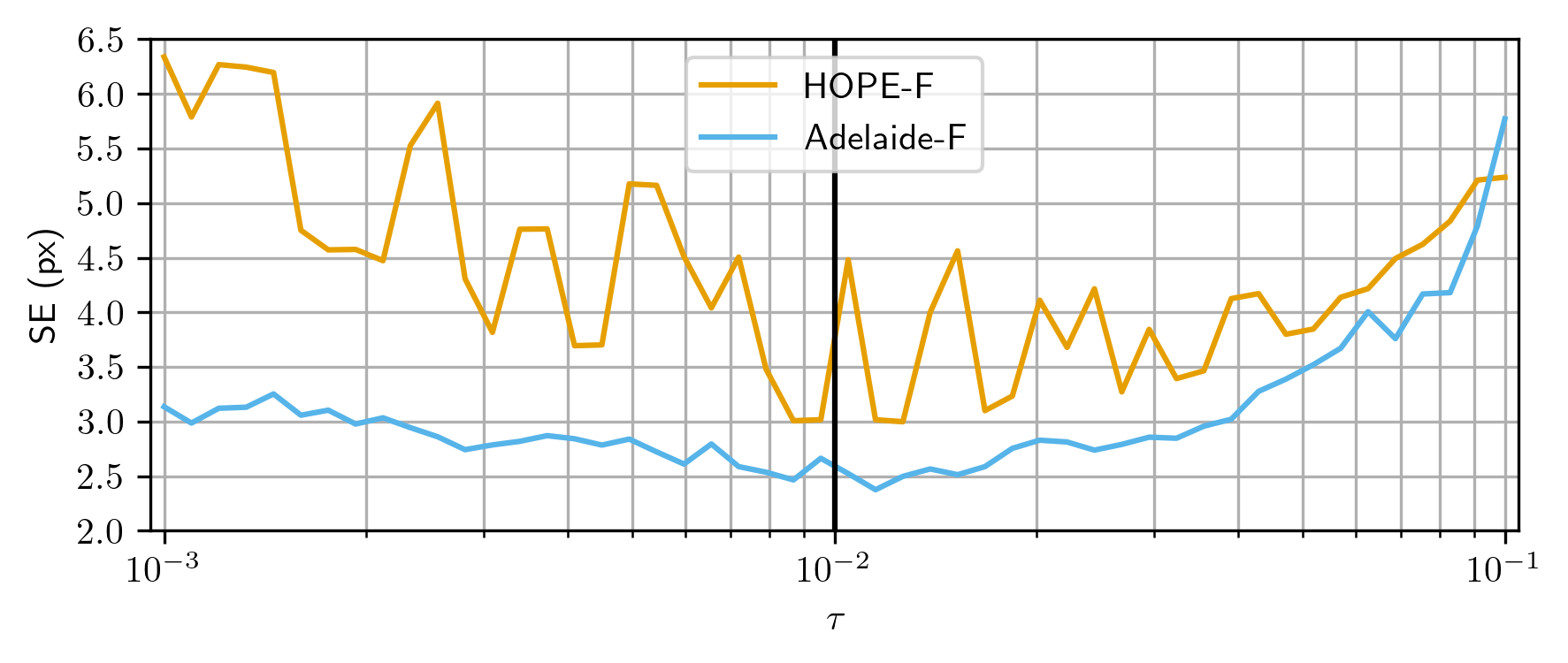}
        \caption{Fundamental matrix estimation (Sampson error)}
        \label{fig:sensitivity_tau:fun_SE}
    \end{subfigure}
    \begin{subfigure}{\linewidth}
    \centering
        \includegraphics[width=\imgwidth]{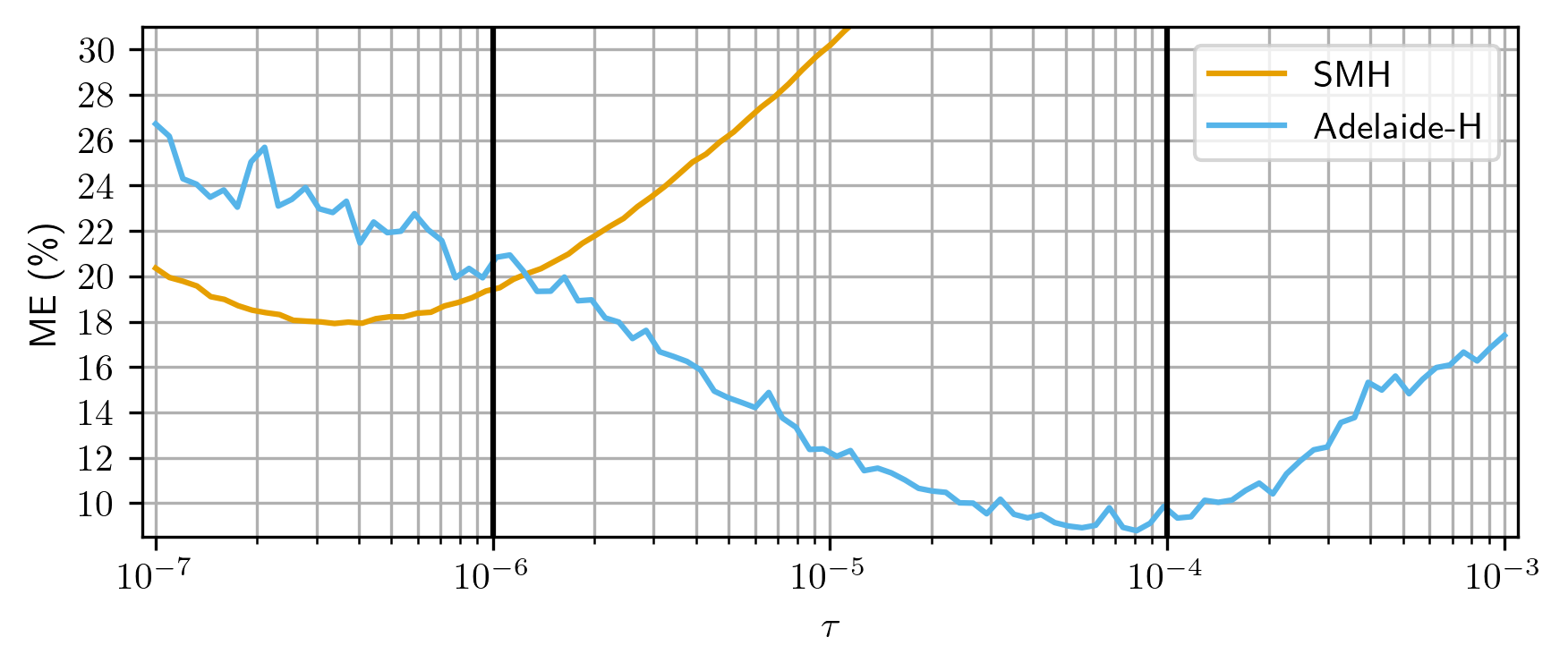}
        \caption{Homography estimation (misclassification error)}
        \label{fig:sensitivity_tau:hom_ME}
    \end{subfigure}
    \begin{subfigure}{\linewidth}
    \centering
        \includegraphics[width=\imgwidth]{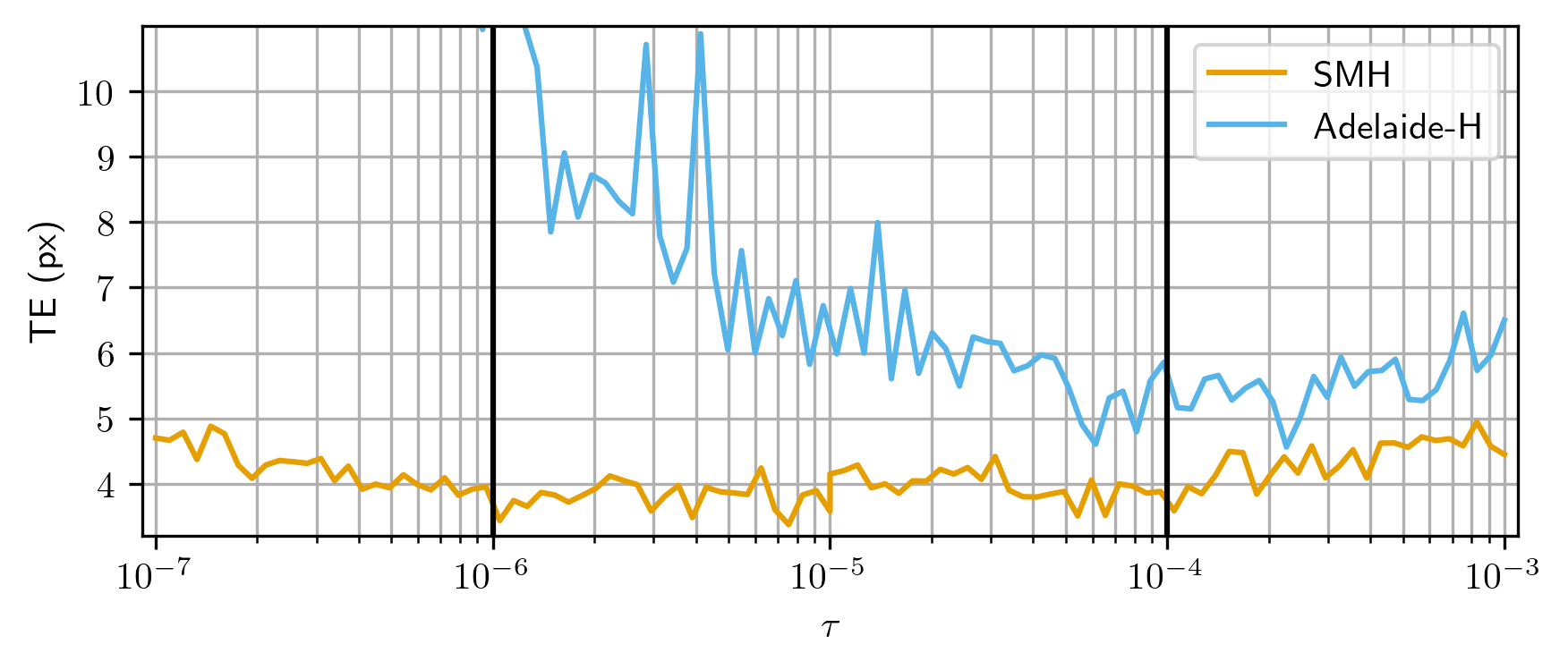}
        \caption{Homography estimation (transfer error)}
        \label{fig:sensitivity_tau:hom_TE}
    \end{subfigure}
    \caption{Sensitivity of PARSAC \wrt the inlier threshold $\tau$ for vanishing points (a), fundamental matrices (b-c) and homographies (d-e). Black vertical lines indicate the values we used for our main experiments (cf. Tab.~\ref{tab:parameters}).}
    \label{fig:sensitivity_tau}
\end{figure}

The assignment threshold $\tau_a$ does not affect the model parameters; it only affects the cluster assignment of the observations.
In Fig.~\ref{fig:sensitivity_tau}, we thus only look at the misclassification errors (ME) for fundamental matrix and homography estimation.
On HOPE-F, the ME is nearly constant for $\tau_a < 0.02$.
On Adelaide-F, there is a minimum at around $\tau_a = 0.02$.
For homography estimation, the ME is minimal at $\tau_a = 4 \cdot 10^{-6}$ on SMH, while the optimal value for Adelaide-H is around $\tau_a = 4 \cdot 10^{-3}$.

\begin{figure}	
    \setlength{\imgwidth}{\linewidth}
    \begin{subfigure}{\linewidth}
    \centering
        \includegraphics[width=\imgwidth]{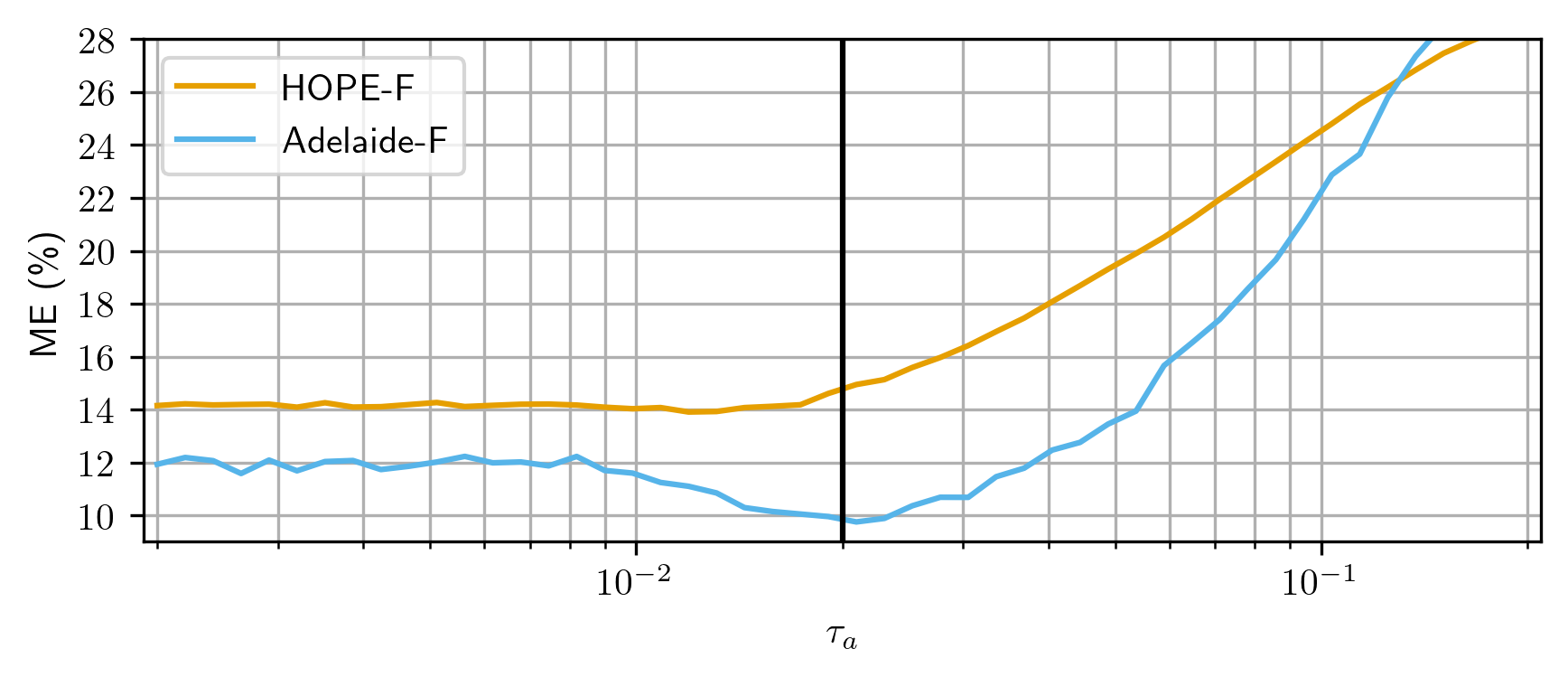}
        \caption{Fundamental matrix estimation (misclassification error)}
        \label{fig:sensitivity_taua:fun}
    \end{subfigure}
    \begin{subfigure}{\linewidth}
    \centering
        \includegraphics[width=\imgwidth]{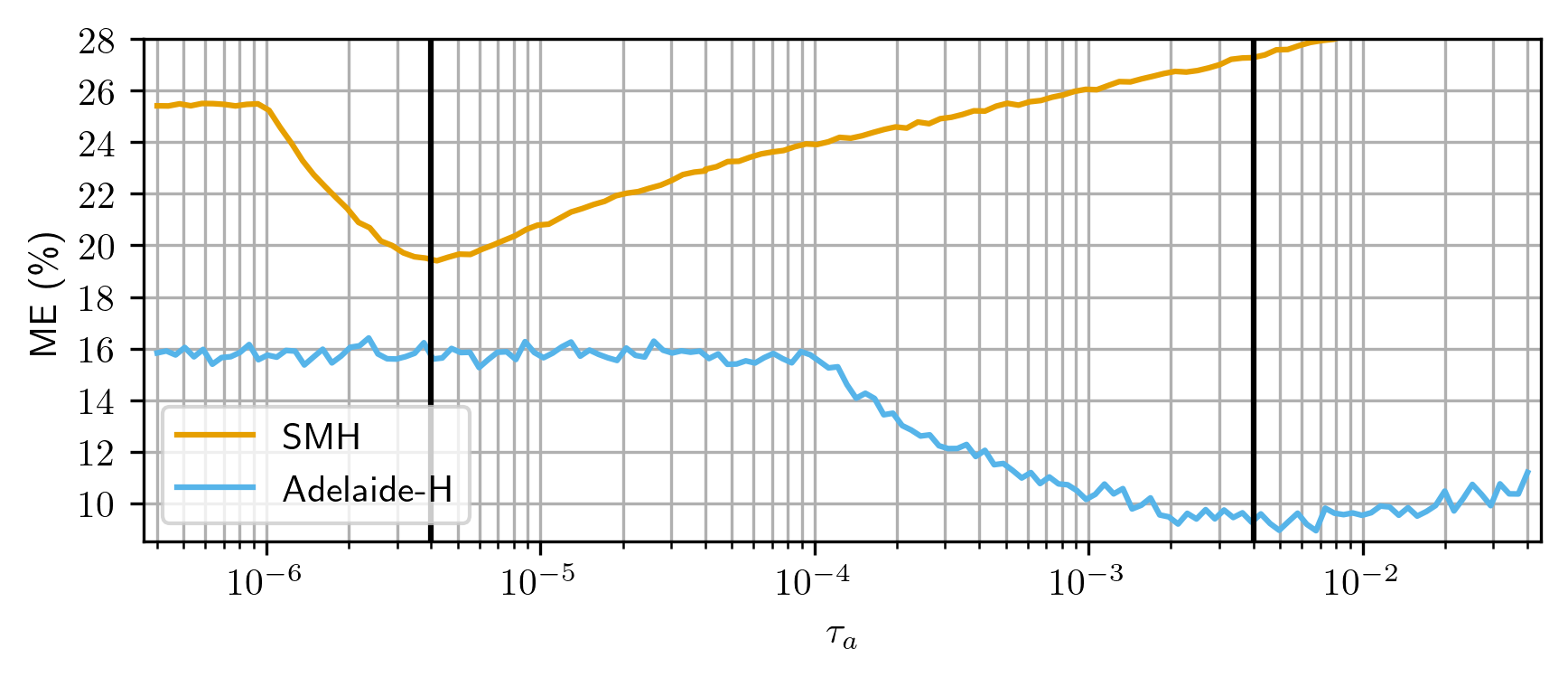}
        \caption{Homography estimation (misclassification error)}
        \label{fig:sensitivity_taua:hom}
    \end{subfigure}
    \caption{Sensitivity of PARSAC \wrt the assignment threshold $\tau_a$ for F-matrices (a) and homographies (b). Vertical lines indicate values used for our main experiments.}
    \label{fig:sensitivity_taua}
\end{figure}

\section{Datasets}
\label{sec:dataset_details}
In Sec.~\ref{sec:mmf_datasets}, we present two new synthetic datasets: HOPE-F for fundamental matrix fitting and Synthetic Metropolis Homography (SMH) for homography fitting.
We use Blender 3.4.1 to generate both datasets and render all images with a resolution of $1024 \times 1024$ pixel. 

\subsection{HOPE-F}
We use a subset of 22 textured 3D meshes from the HOPE dataset~\cite{tyree2022hope} which depict household objects:
\emph{AlphabetSoup}, \emph{Butter}, \emph{Cherries}, \emph{ChocolatePudding}, \emph{Cookies}, \emph{Corn}, \emph{CreamCheese}, \emph{GranolaBars}, \emph{GreenBeans}, \emph{MacaroniAndCheese}, \emph{Milk}, \emph{Mushrooms}, \emph{OrangeJuice}, \emph{Parmesan}, \emph{Peaches}, \emph{PeasAndCarrots}, \emph{Pineapple}, \emph{Popcorn}, \emph{Raisins}, \emph{Spaghetti}, \emph{Tuna} and \emph{Yogurt}.
We omit the remaining objects from the HOPE dataset due to a relative lack of texture.

For each scene, we select between one and four of these meshes and place them on a 3D plane at randomised positions.
Each object is set on one of four possible positions on a $2 \times 2$ grid on a fixed ground plane, with a randomised offset of up to $\pm 0.25$ in x- and y-directions.
We also apply a randomised rotation around the z-axis of the object of up to $\pm 10^{\circ}$.
We then render two images of the scene using one virtual camera with randomised pose and focal length.
We uniformly sample the focal length between $24\si{\milli\metre}$ and $40\si{\milli\metre}$, azimuth between $90^{\circ}$ and $270^{\circ}$, elevation between $25^{\circ}$ and $65^{\circ}$, and camera distance to the centre of the scene between three and five metres.
Before rendering the second image, we randomise the object positions again.

As we know the intrinsic camera parameters $\vec{K}$ and the relative object poses $\vec{R}_i, \vec{t}_i$, $i \in \{1, \dots, 4\}$, we can compute the ground truth fundamental matrices for each image pair: 
\begin{equation}
    \vec{F}_i = \vec{K}\negtran \crossmat{\vec{t}_i} \vec{R}_i \vec{K}\inverse \, .
\end{equation}
We then compute SIFT~\cite{Lowesift} keypoint features and match them between each image pair.
We assign each point correspondence to the ground truth fundamental matrix with the smallest Sampson distance if it is below a threshold of $2$ pixel, or to the outlier class otherwise.
These assignments represent the ground truth cluster labels.

\begin{figure}
\centering
\includegraphics[width=\linewidth]{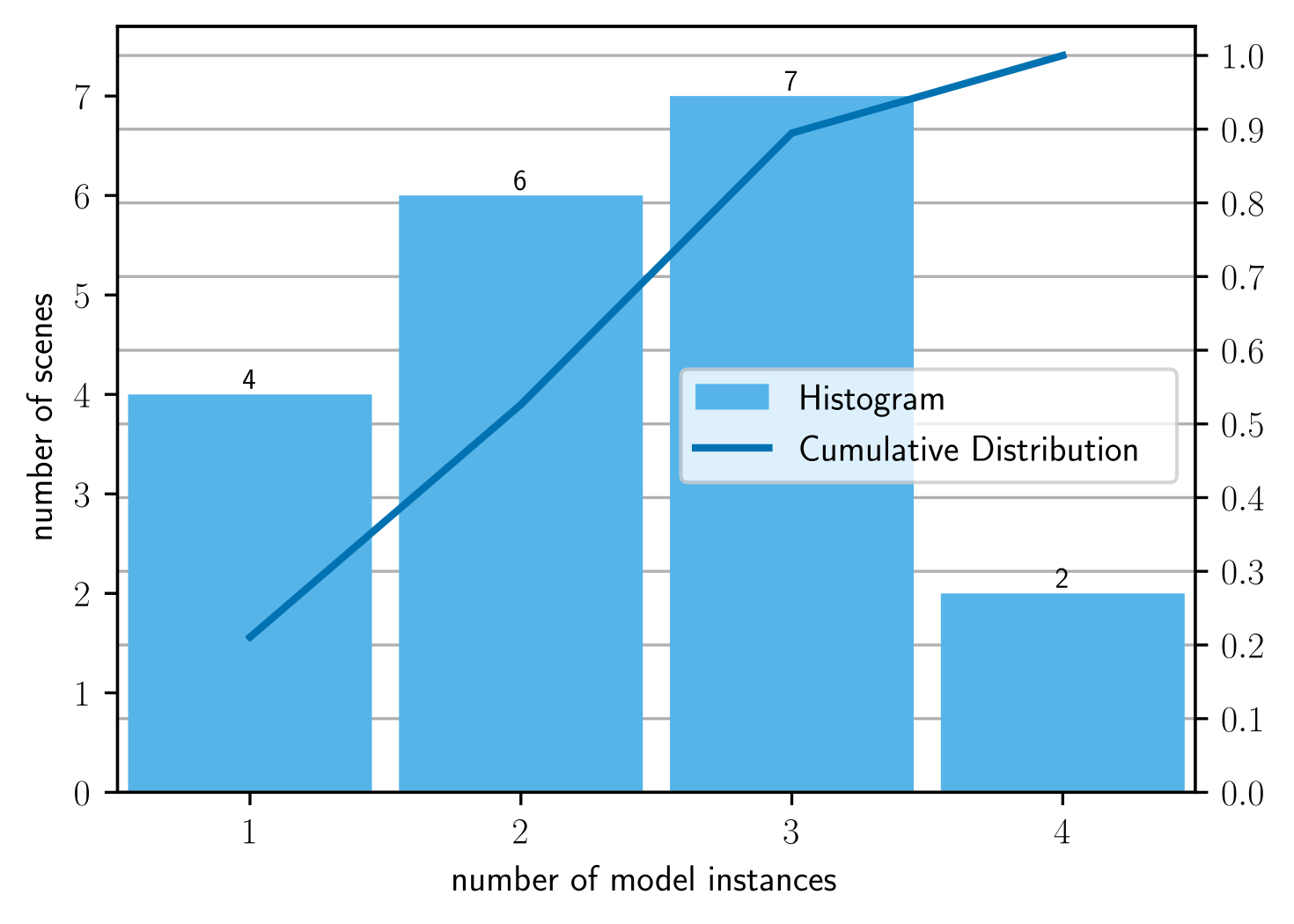}
\caption{Distribution of fundamental matrices in the Adelaide-F dataset. }
\label{fig:datasets:adelaidef_models}
\end{figure}

Via this procedure, we generate a total of $4000$ image pairs with keypoint features, of which we reserve $400$ as the test set.
The number of fundamental matrices per image pair is uniformly distributed between one and four.
In Adelaide-F, for comparison, most scenes in the dataset contain two or three fundamental matrices, as Fig.~\ref{fig:datasets:adelaidef_models} shows.

\begin{figure}
\centering
\includegraphics[width=\linewidth]{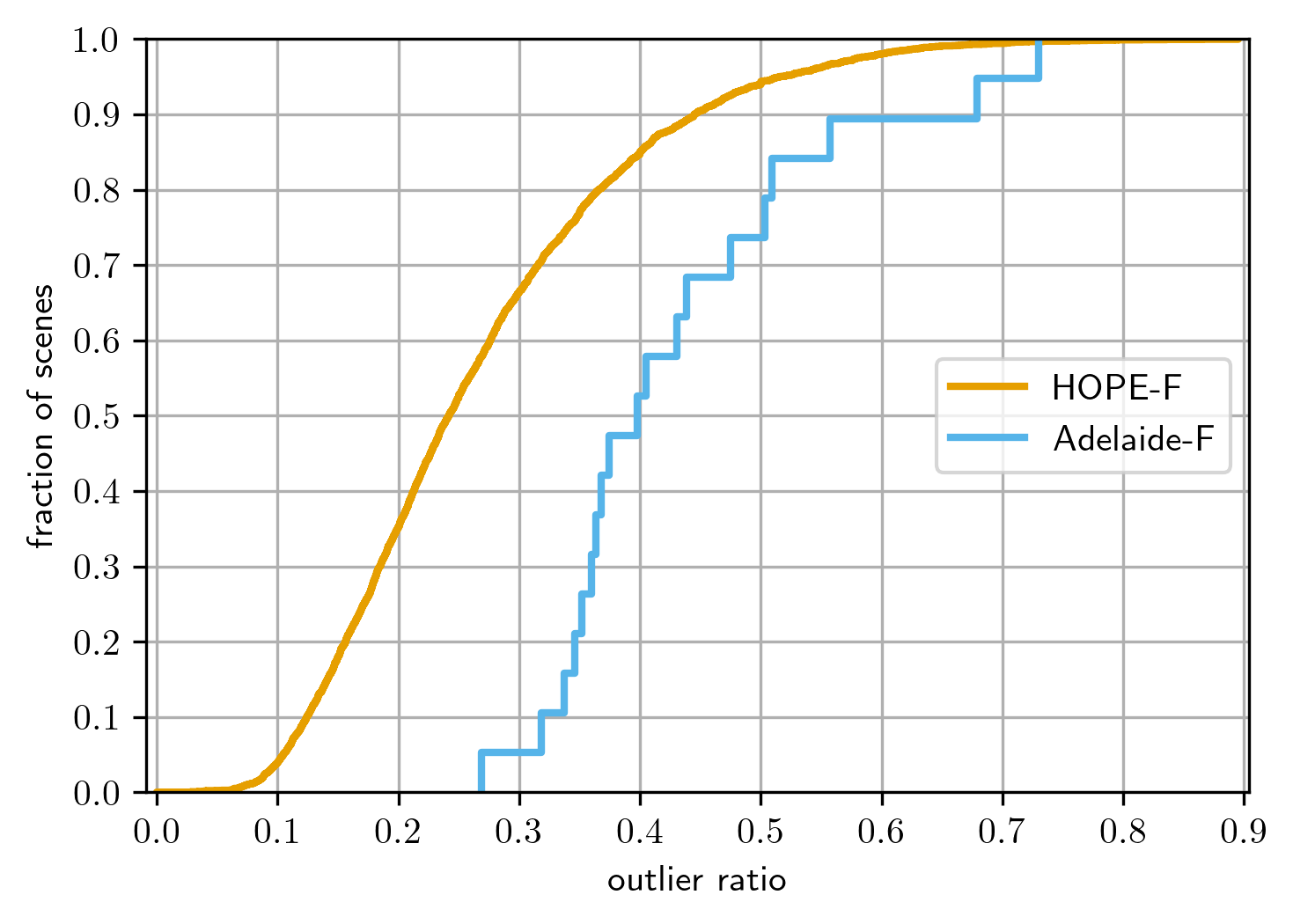}
\caption{Cumulative distributions of outlier ratios in the HOPE-F and Adelaide-F datasets. }
\label{fig:datasets:f_outliers}
\end{figure}

\begin{figure}
\centering
\includegraphics[width=\linewidth]{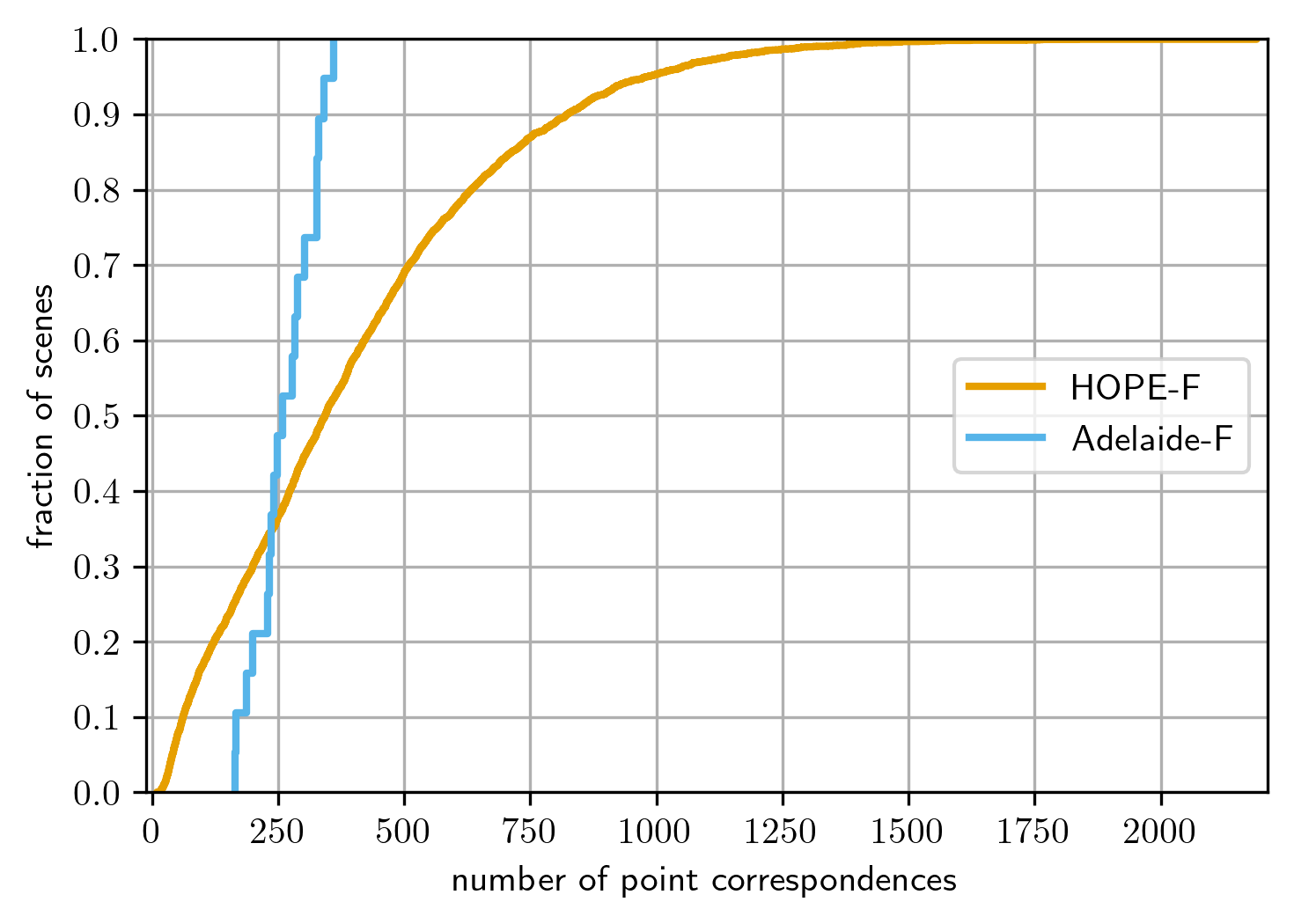}
\caption{Cumulative distributions of point correspondences per scene in the HOPE-F and Adelaide-F datasets. }
\label{fig:datasets:f_points}
\end{figure}

\begin{figure}
\centering
\includegraphics[width=\linewidth]{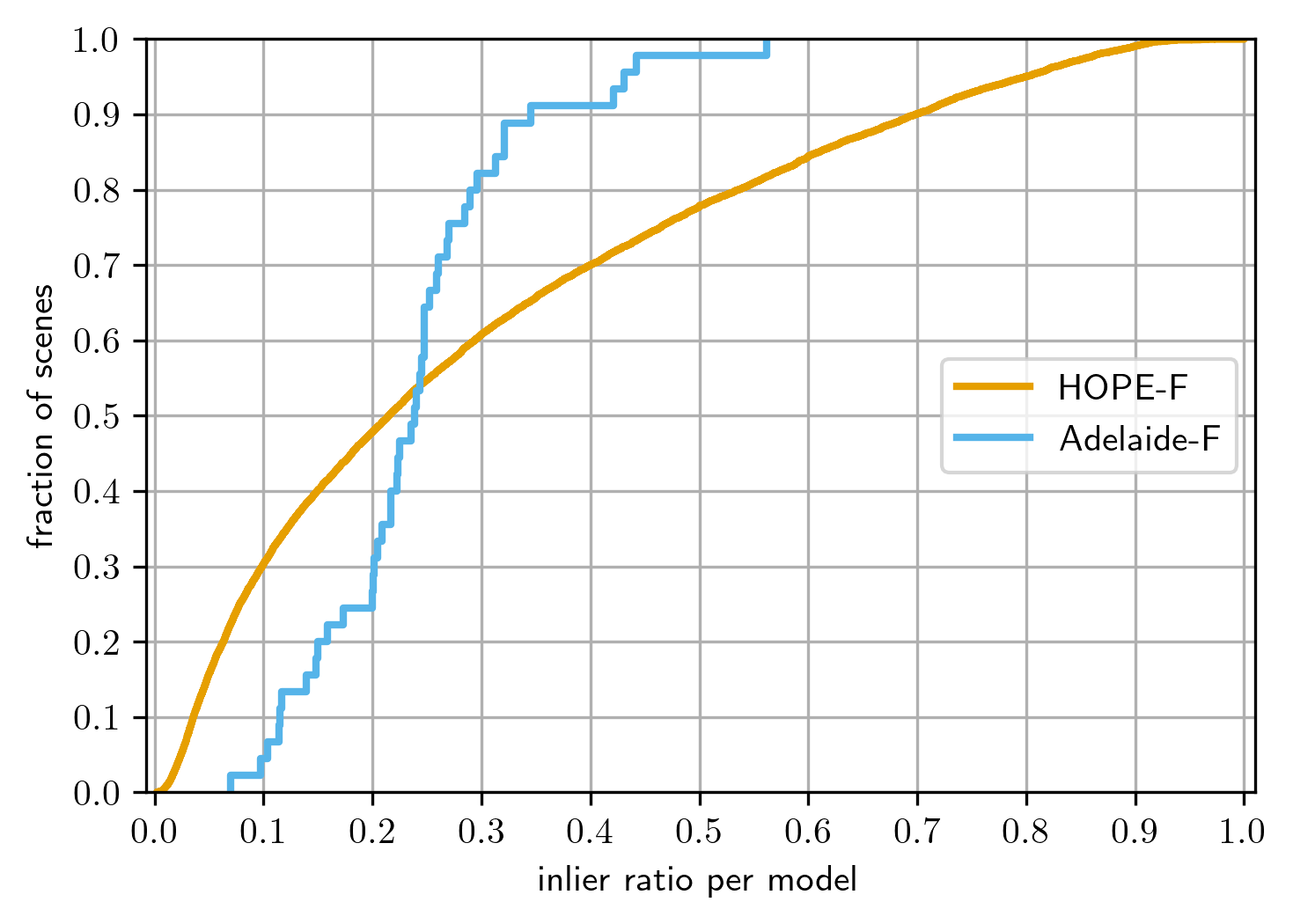}
\caption{Cumulative distributions of inlier ratios per fundamental matrix in the HOPE-F and Adelaide-F datasets. }
\label{fig:datasets:f_inliers}
\end{figure}

\begin{figure*}	
    \setlength{\imgwidth}{0.248\linewidth}
    
    \includegraphics[width=\imgwidth]{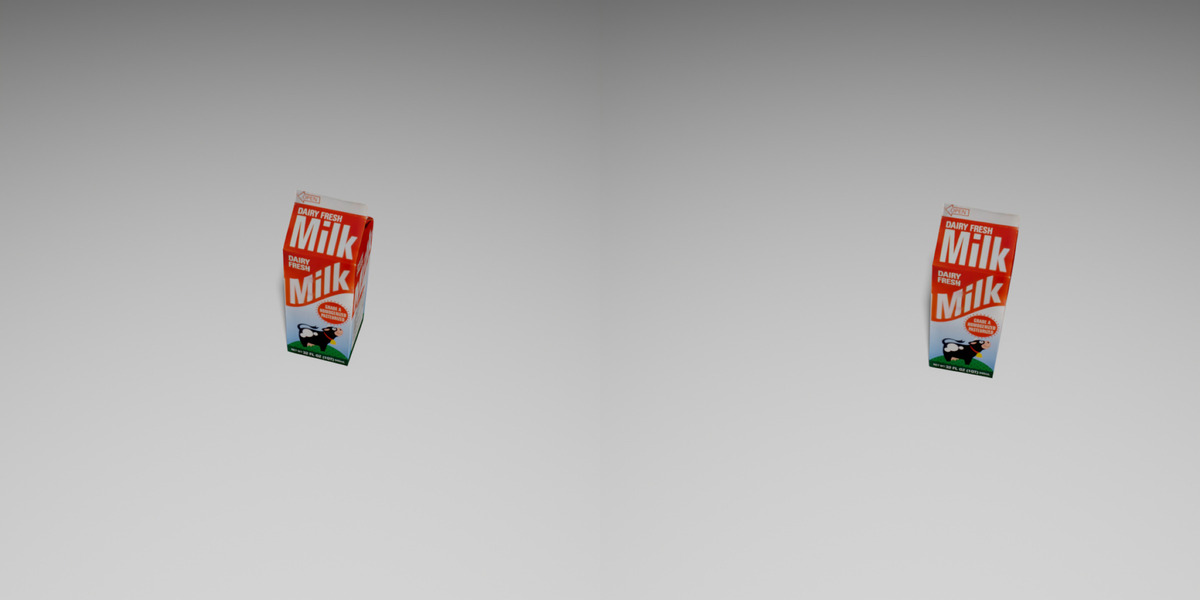}%
    \includegraphics[width=\imgwidth]{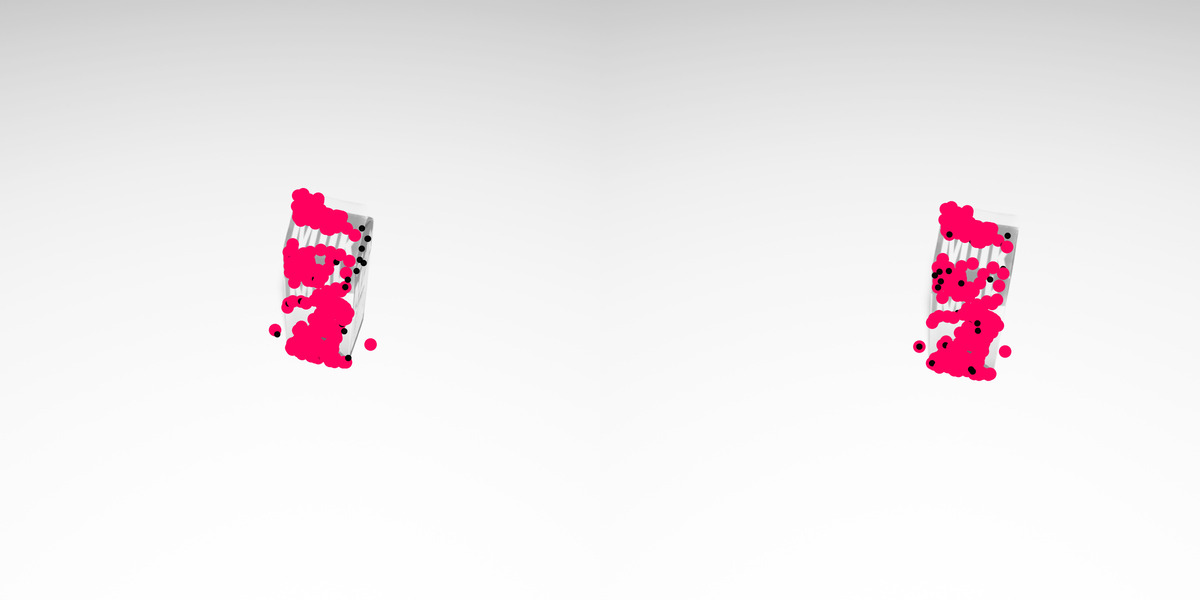}%
    \hfill%
    \includegraphics[width=\imgwidth]{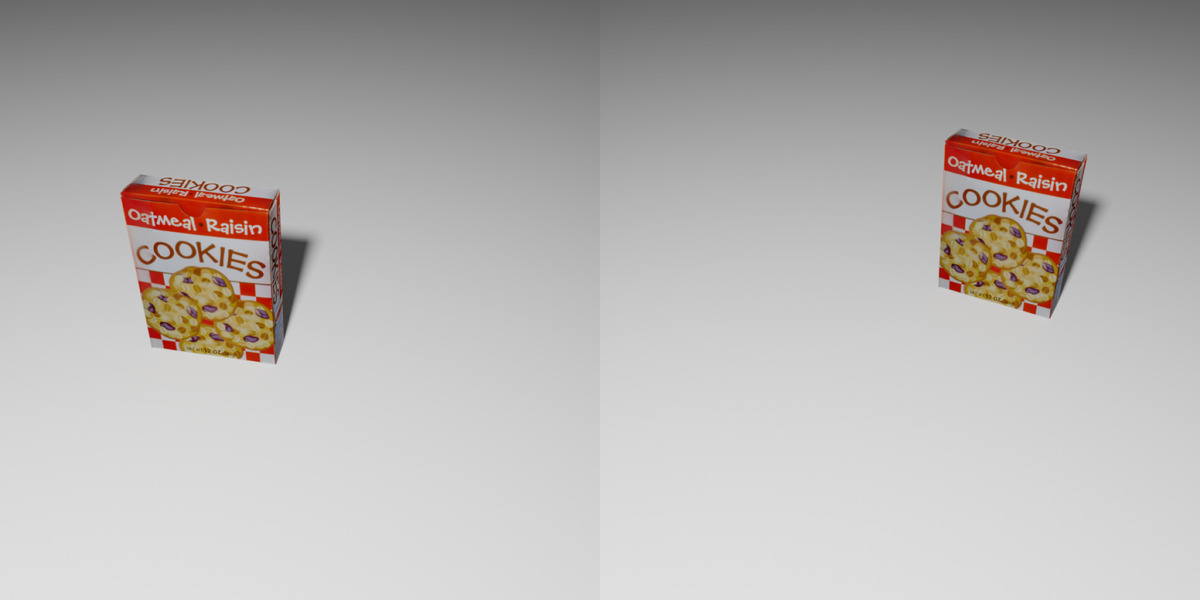}%
    \includegraphics[width=\imgwidth]{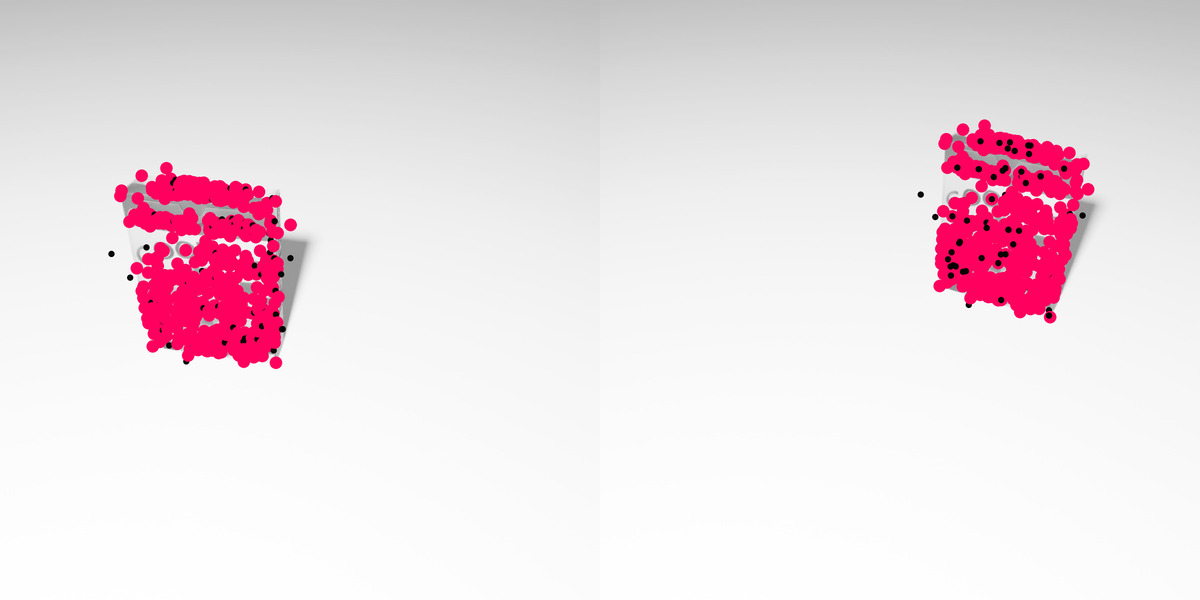}
    
    \includegraphics[width=\imgwidth]{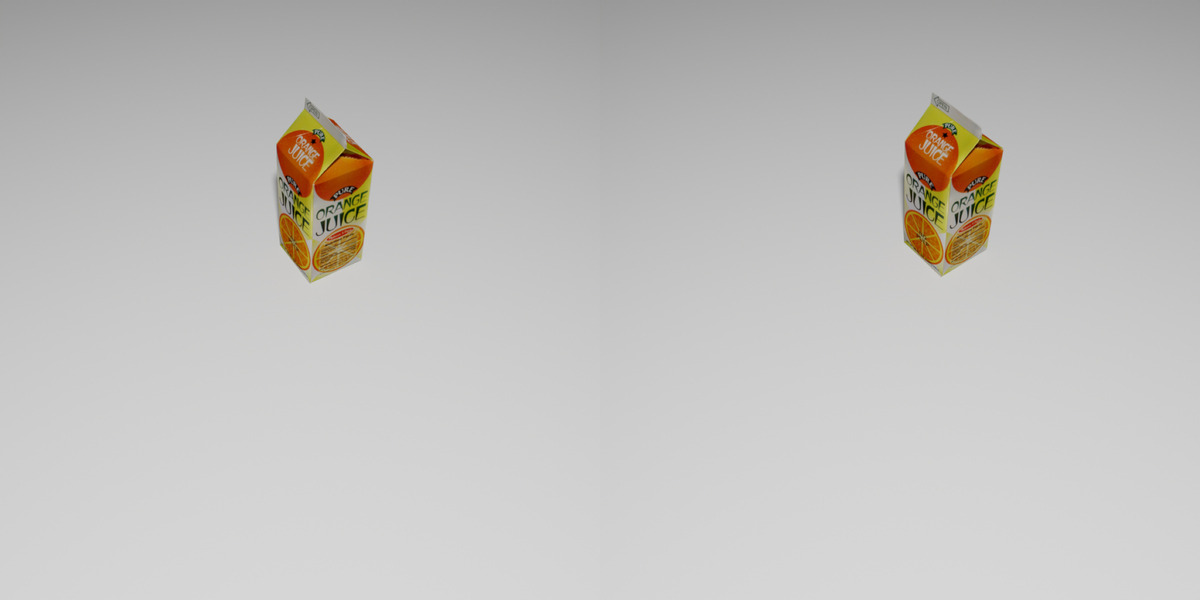}%
    \includegraphics[width=\imgwidth]{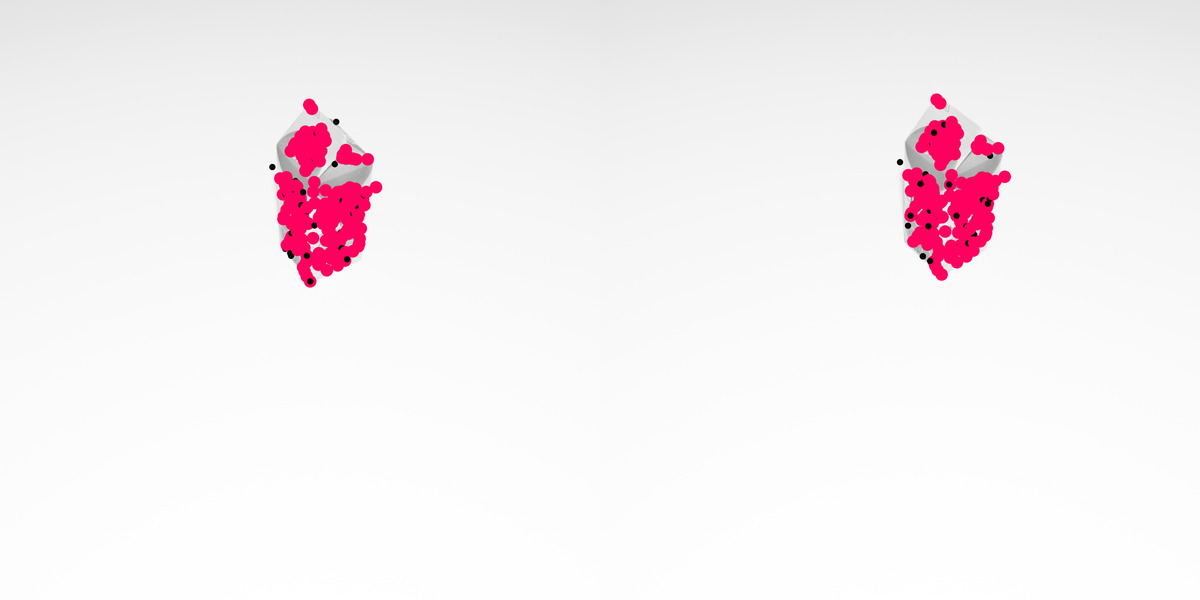}%
    \hfill%
    \includegraphics[width=\imgwidth]{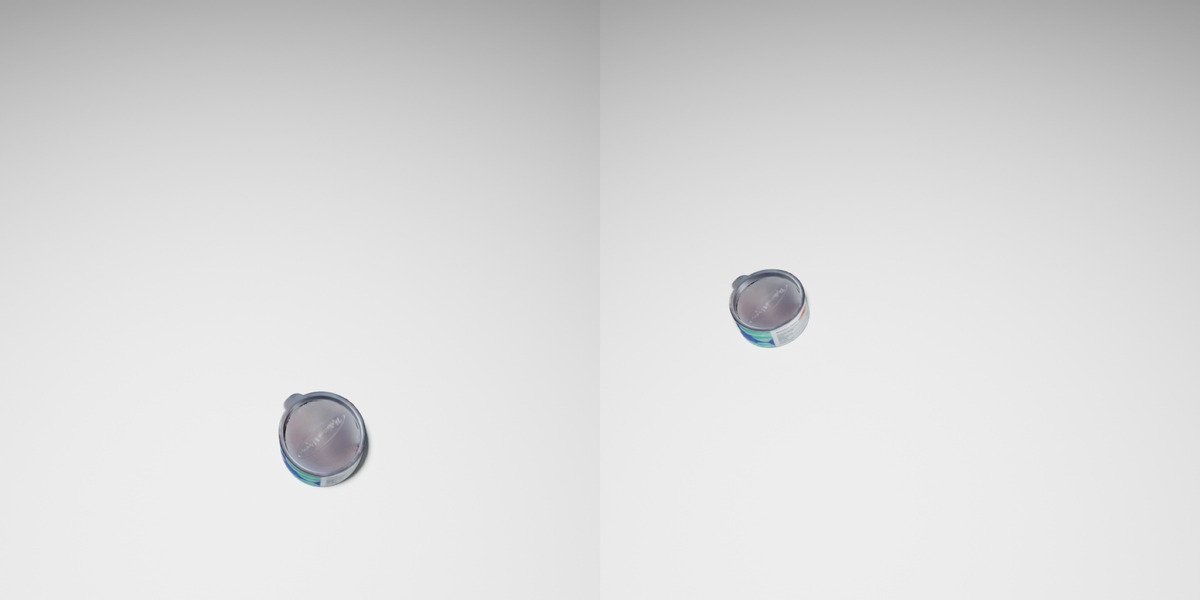}%
    \includegraphics[width=\imgwidth]{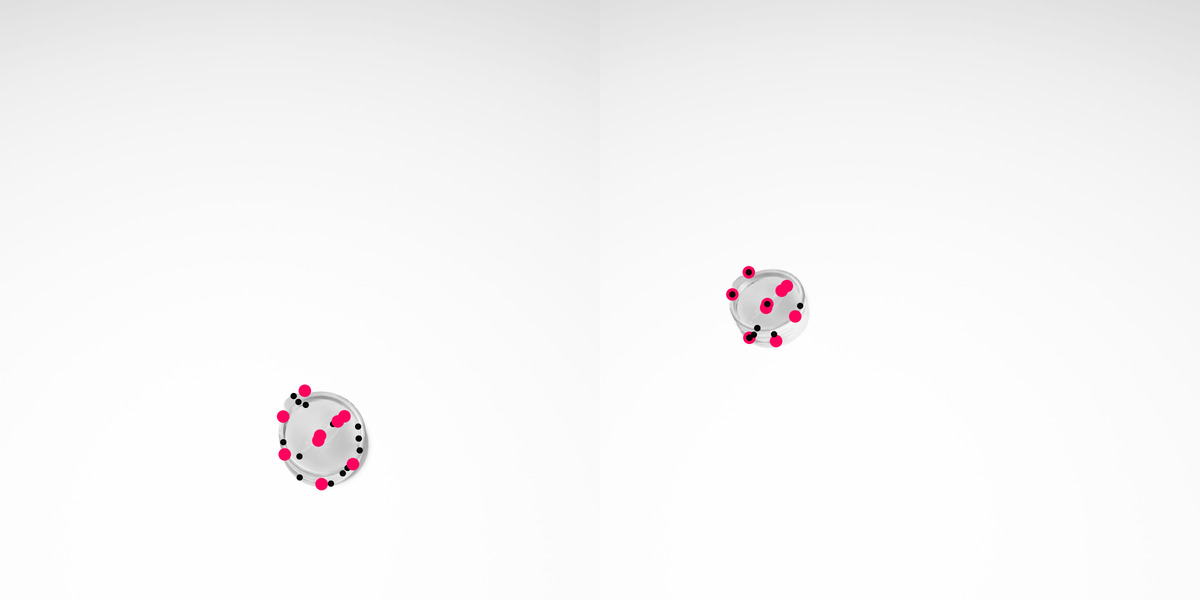}
    
    \includegraphics[width=\imgwidth]{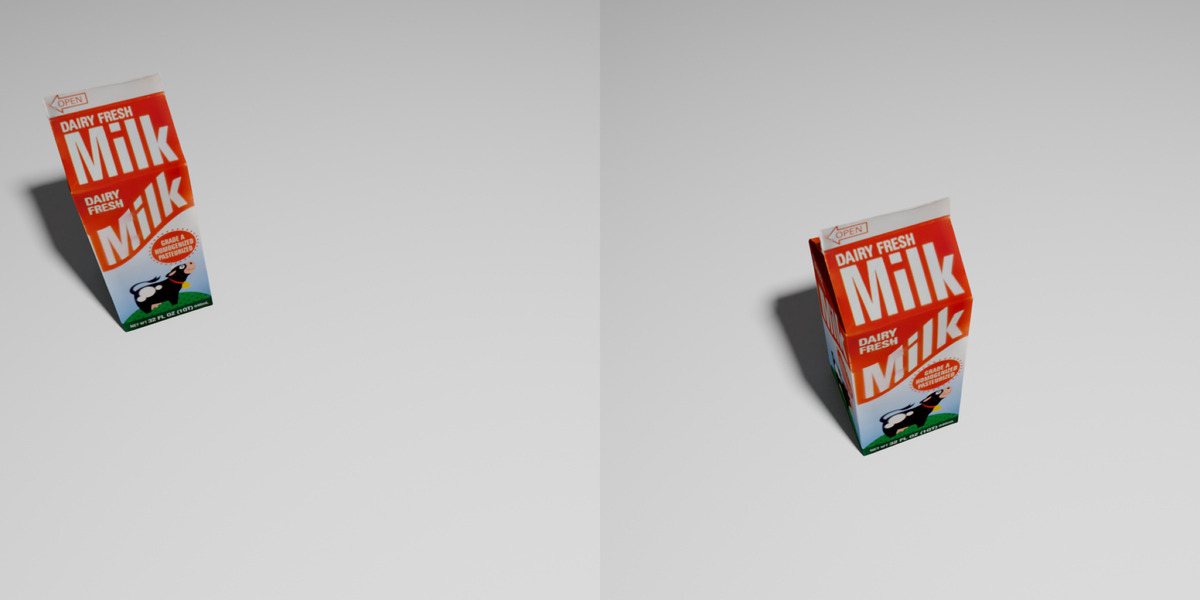}%
    \includegraphics[width=\imgwidth]{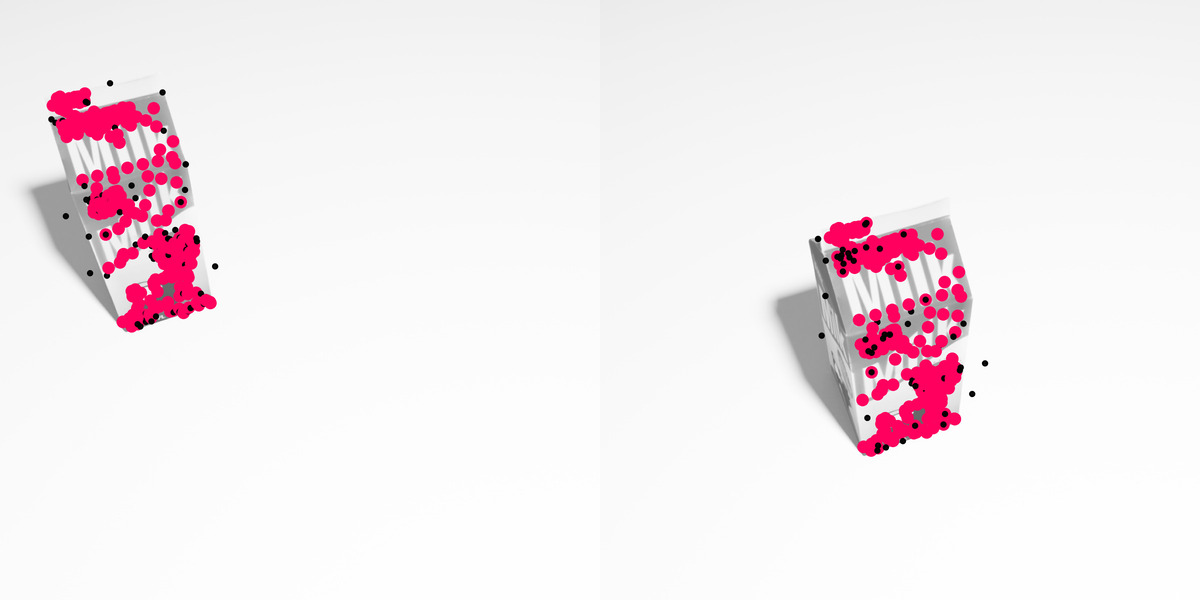}%
    \hfill%
    \includegraphics[width=\imgwidth]{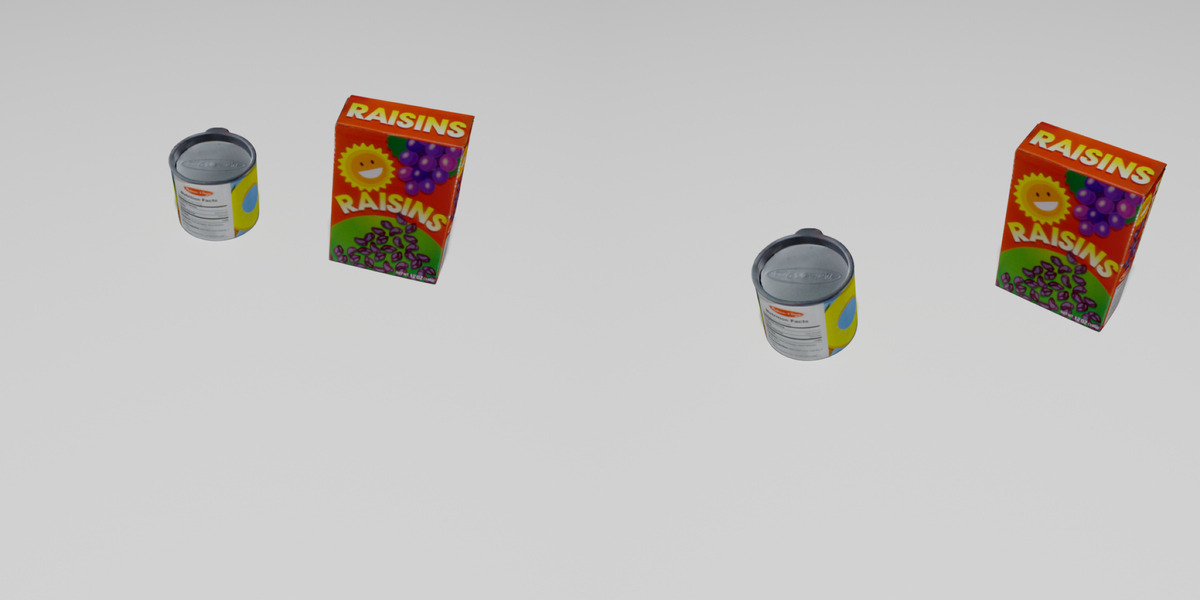}%
    \includegraphics[width=\imgwidth]{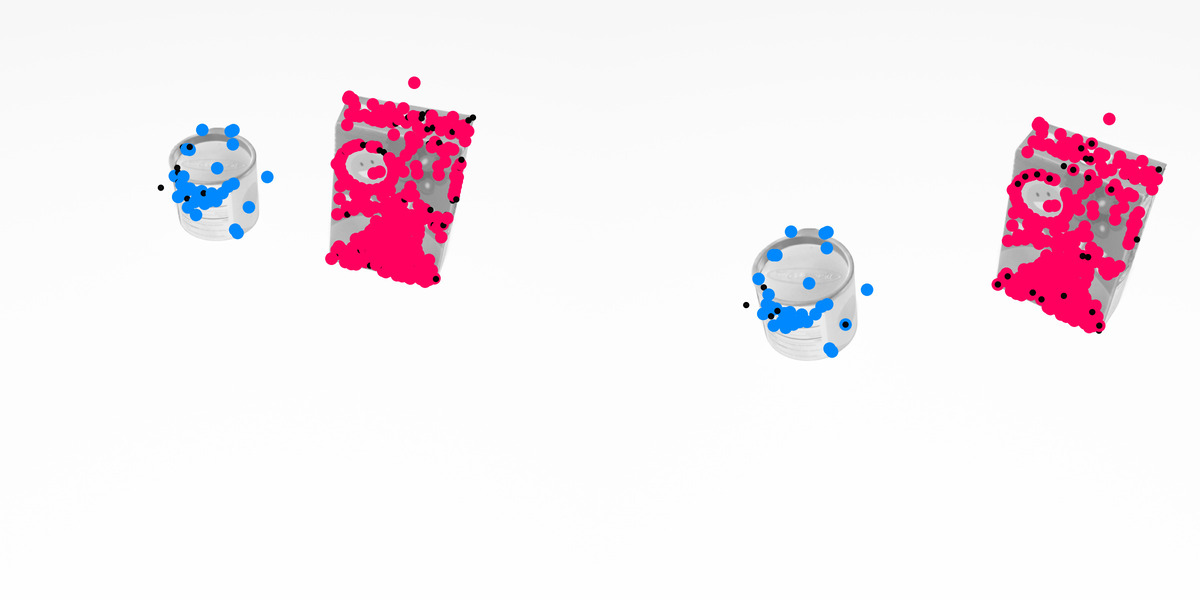}
    
    \includegraphics[width=\imgwidth]{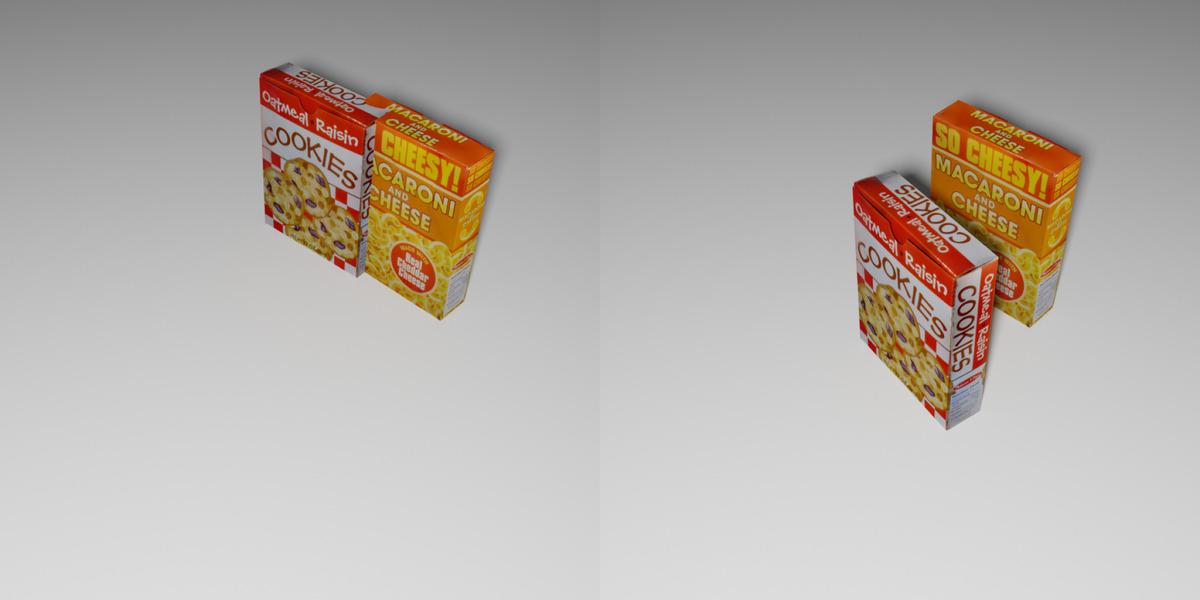}%
    \includegraphics[width=\imgwidth]{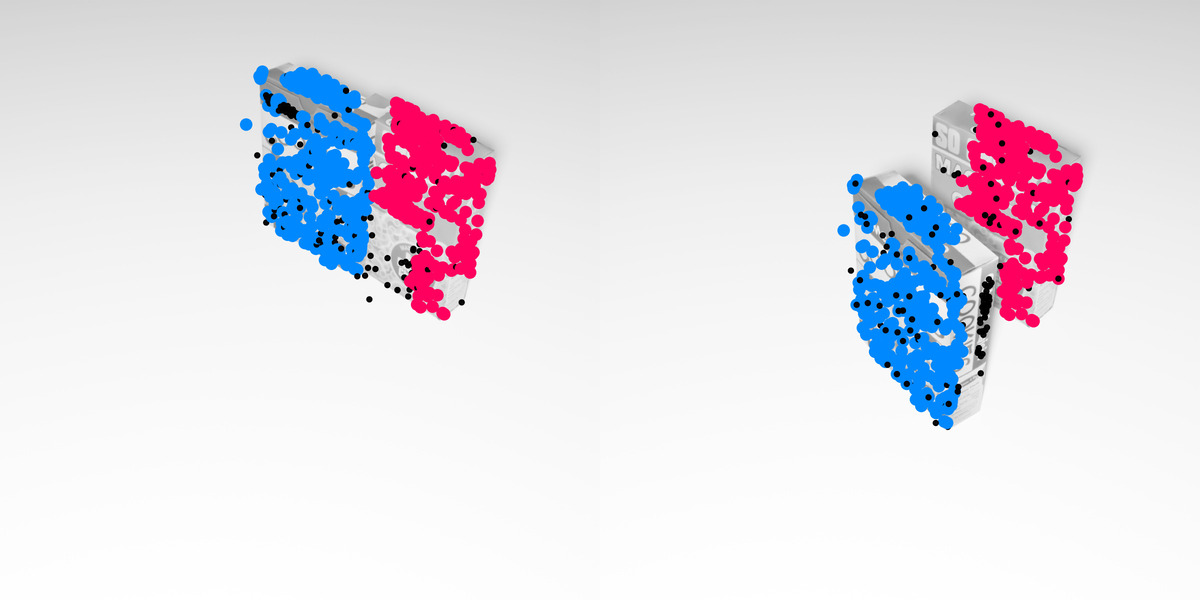}%
    \hfill%
    \includegraphics[width=\imgwidth]{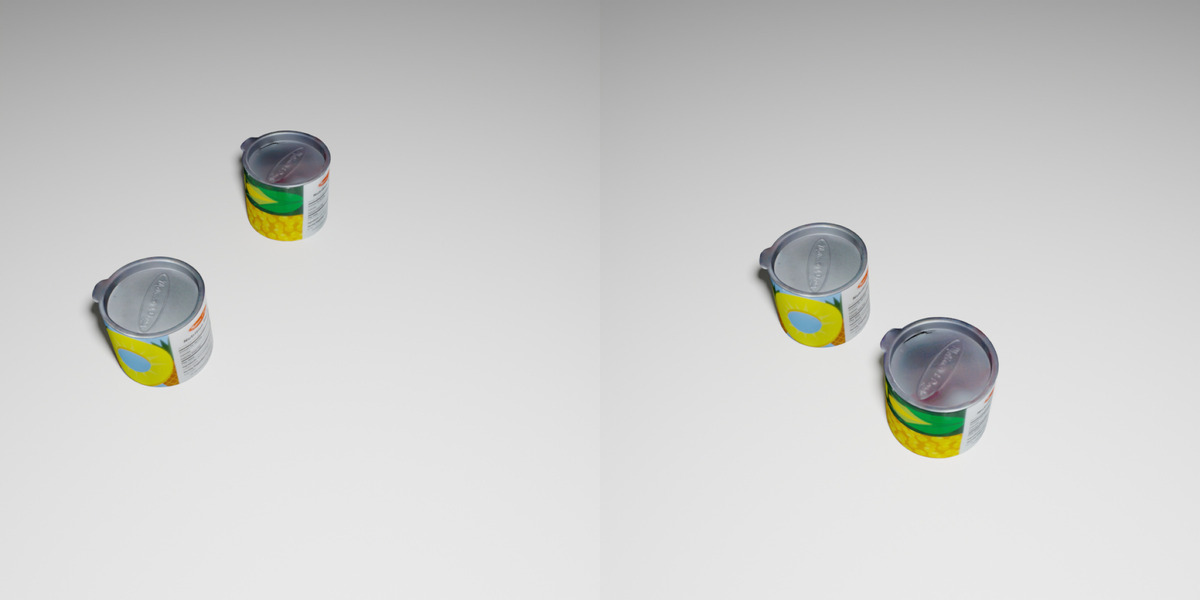}%
    \includegraphics[width=\imgwidth]{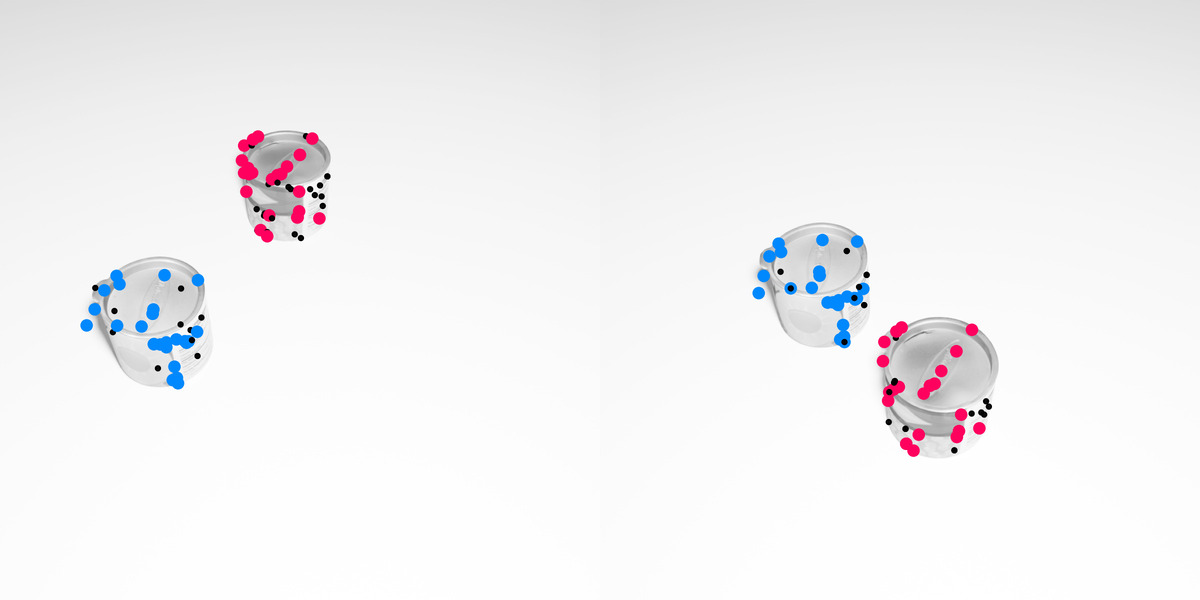}
    
    \includegraphics[width=\imgwidth]{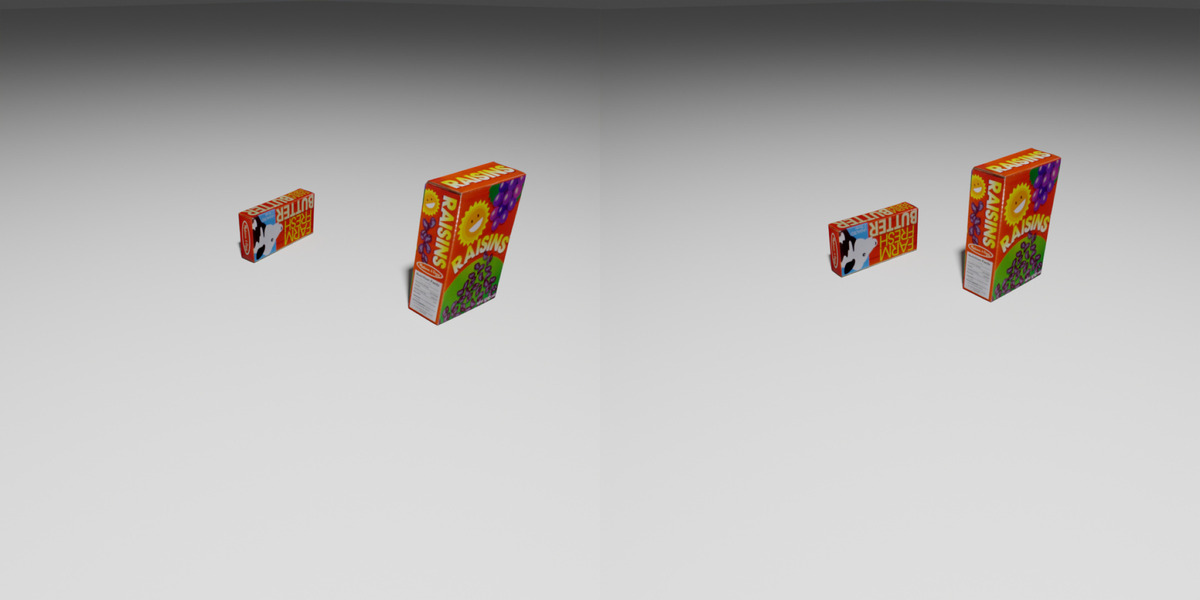}%
    \includegraphics[width=\imgwidth]{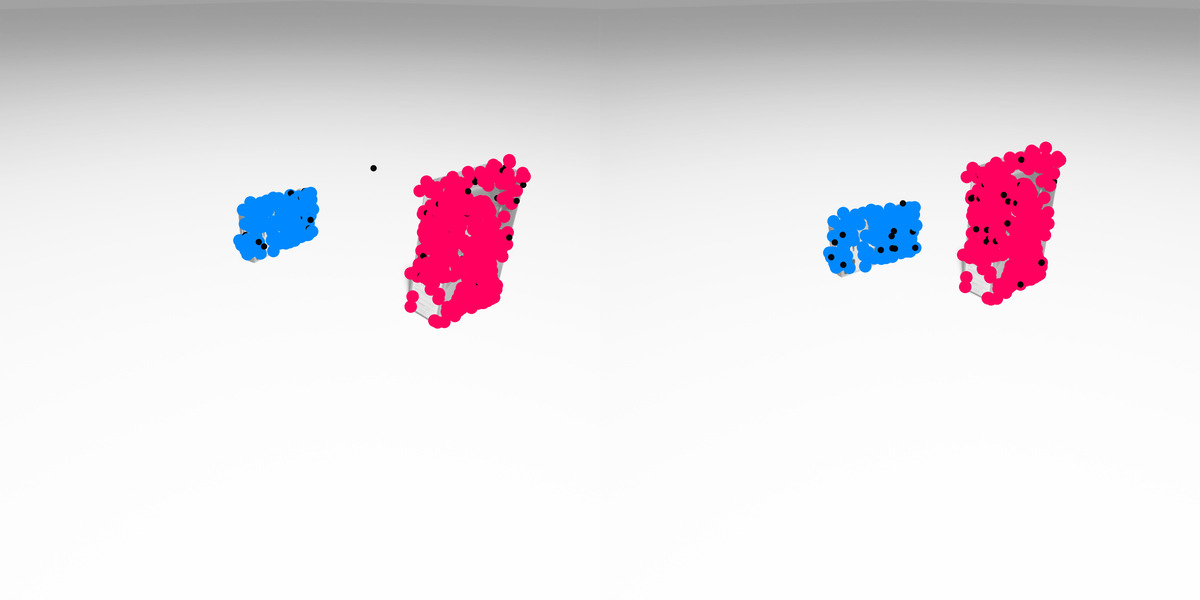}%
    \hfill%
    \includegraphics[width=\imgwidth]{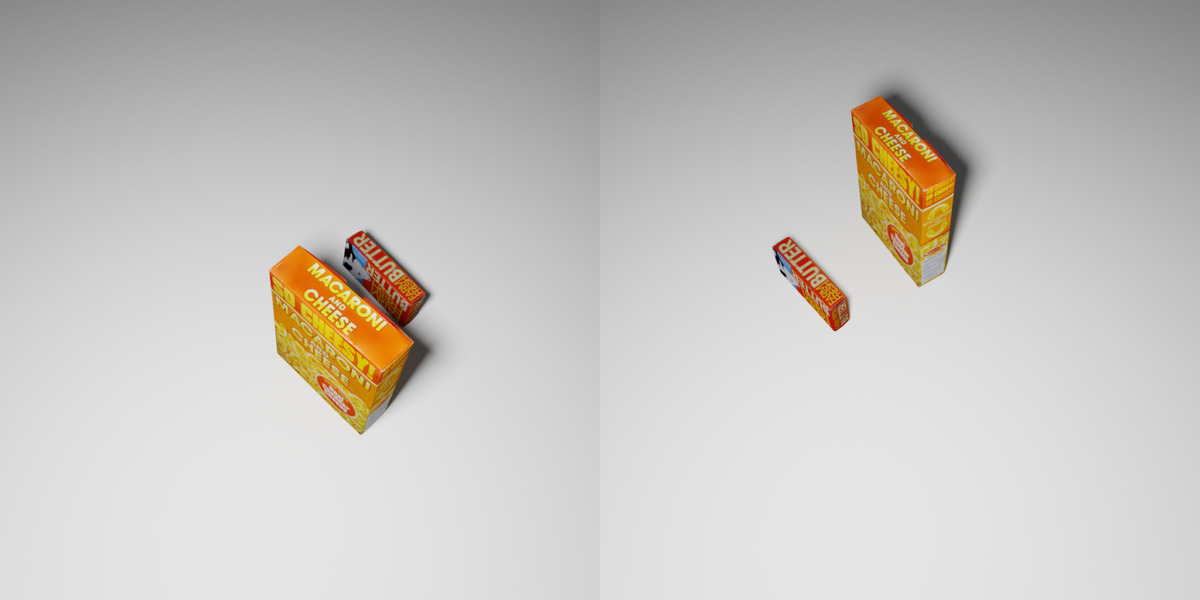}%
    \includegraphics[width=\imgwidth]{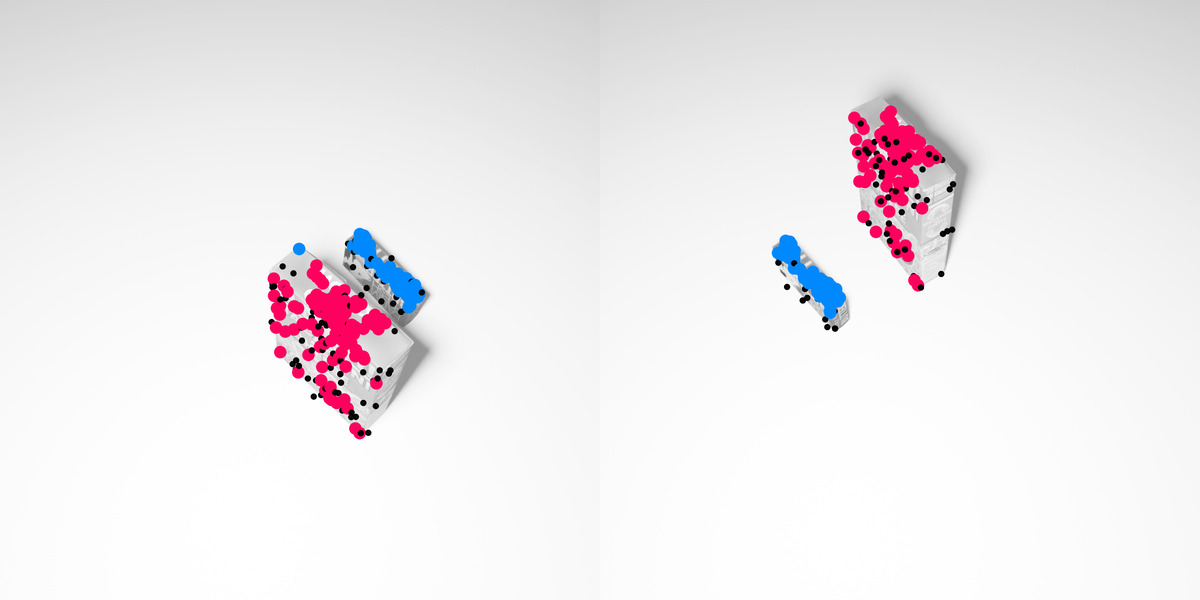}
    
    \includegraphics[width=\imgwidth]{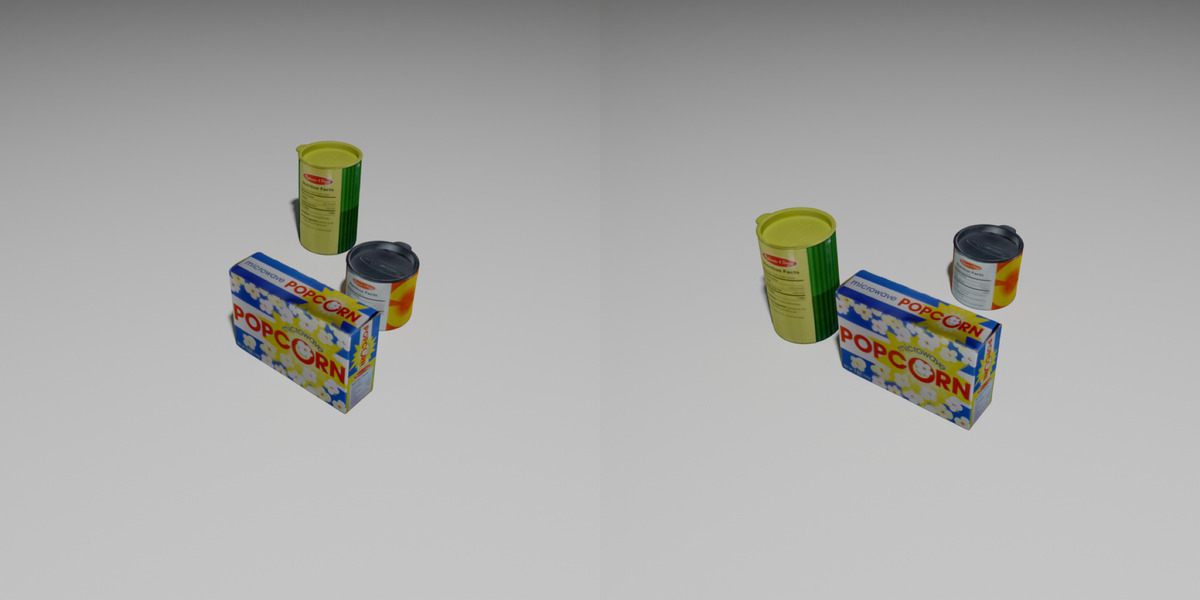}%
    \includegraphics[width=\imgwidth]{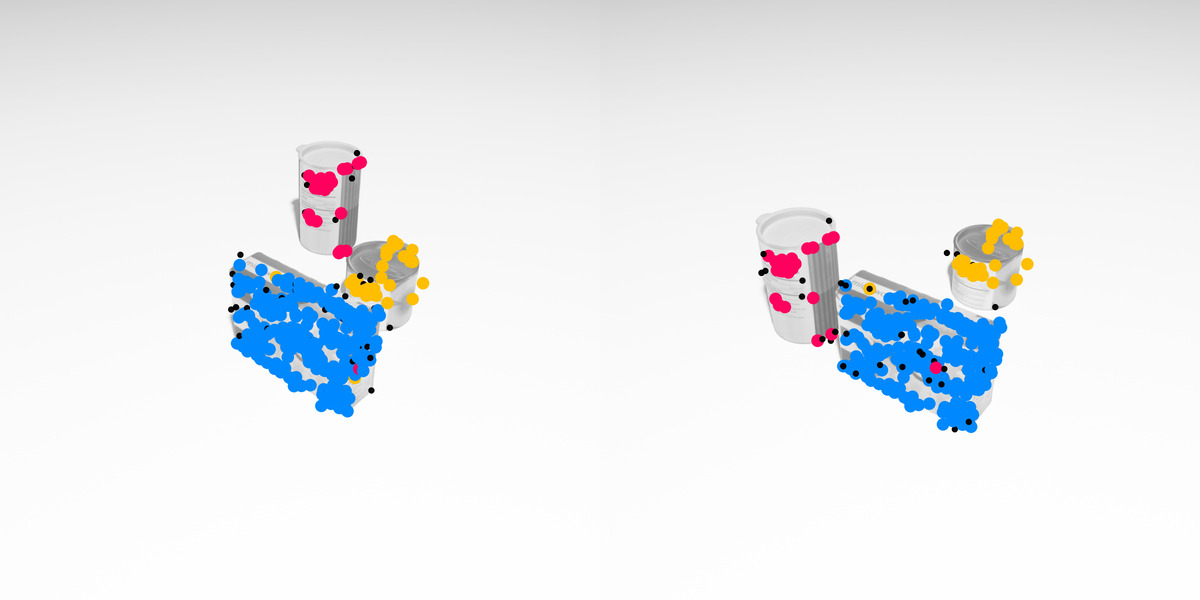}%
    \hfill%
    \includegraphics[width=\imgwidth]{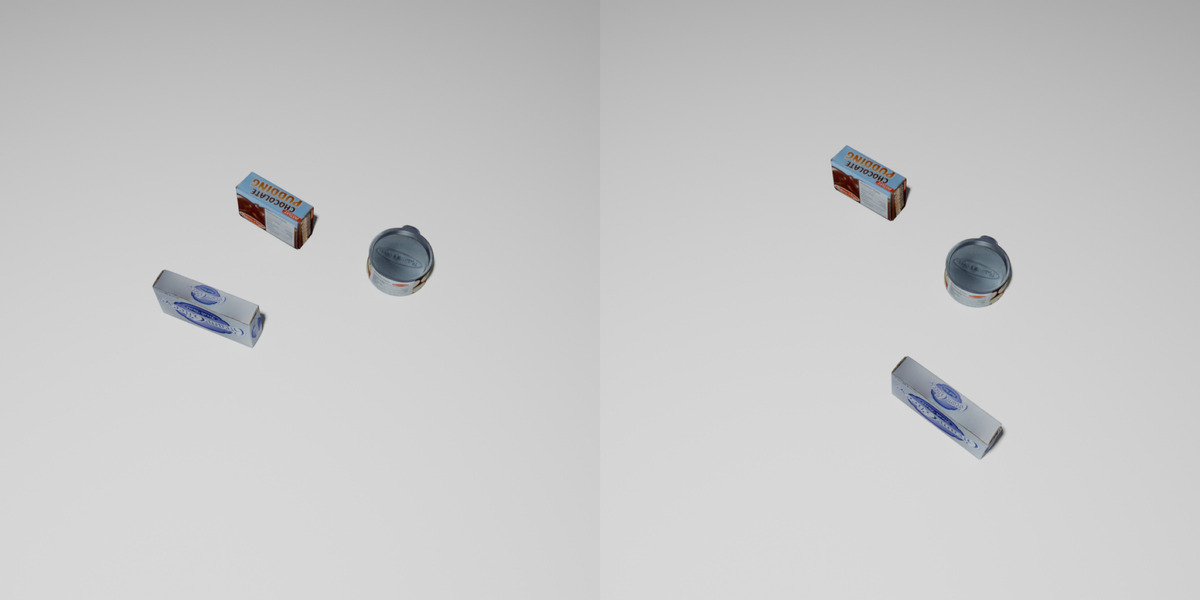}%
    \includegraphics[width=\imgwidth]{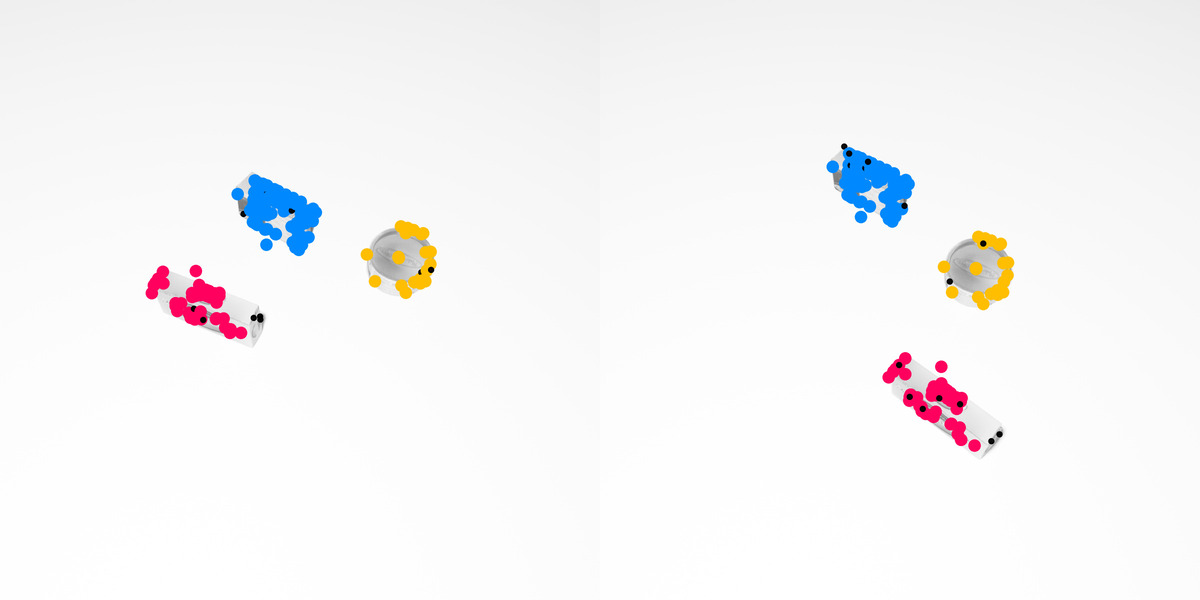}
    
    \includegraphics[width=\imgwidth]{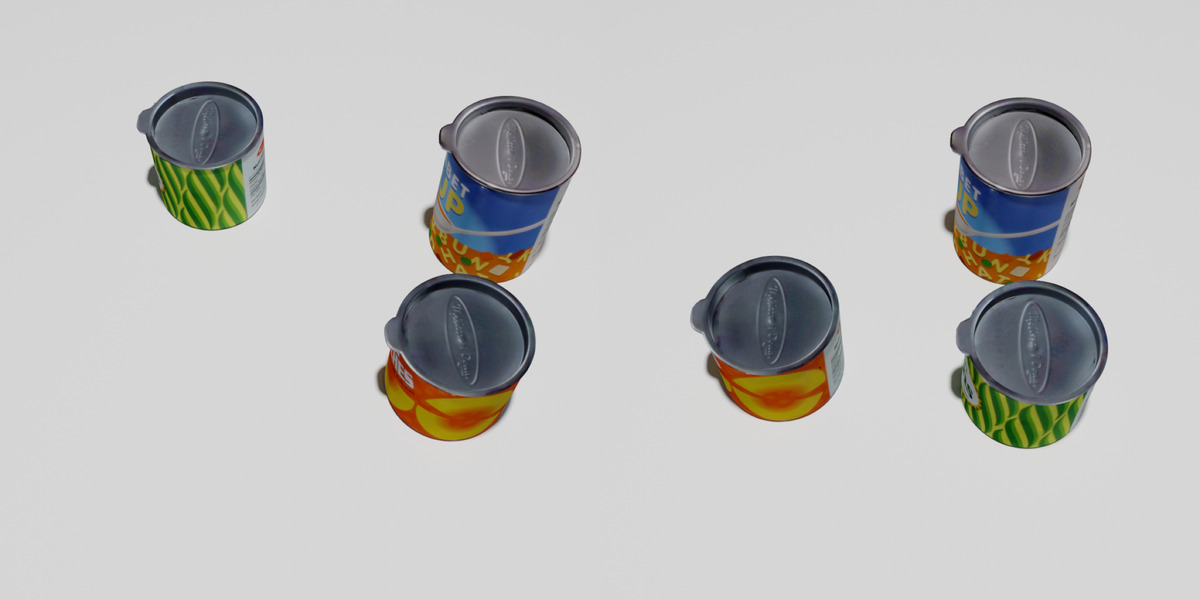}%
    \includegraphics[width=\imgwidth]{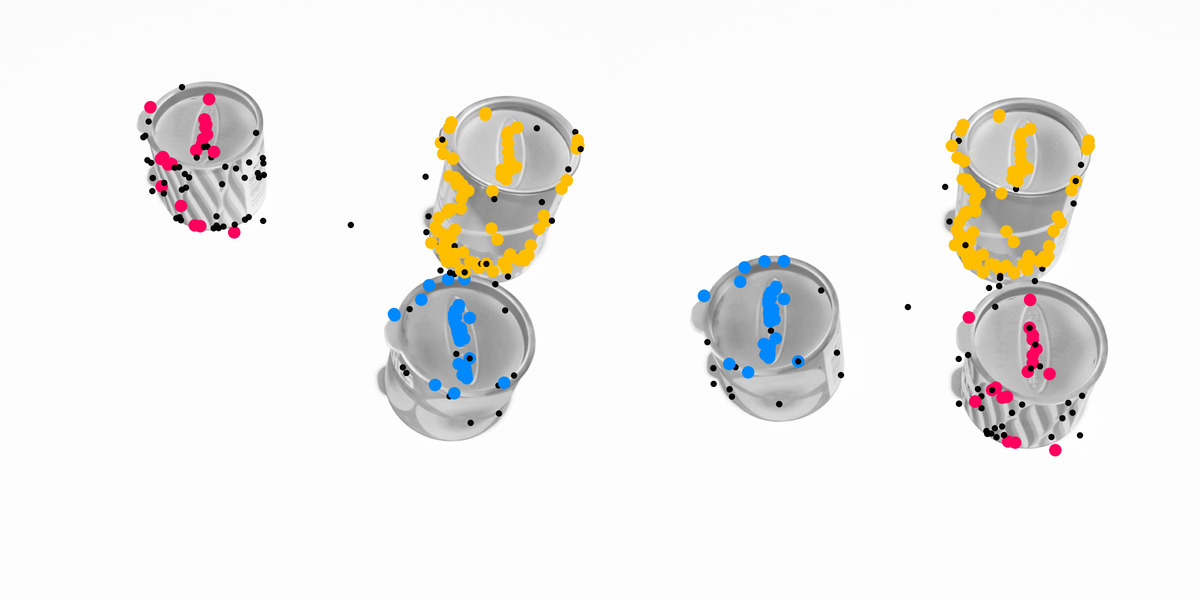}%
    \hfill%
    \includegraphics[width=\imgwidth]{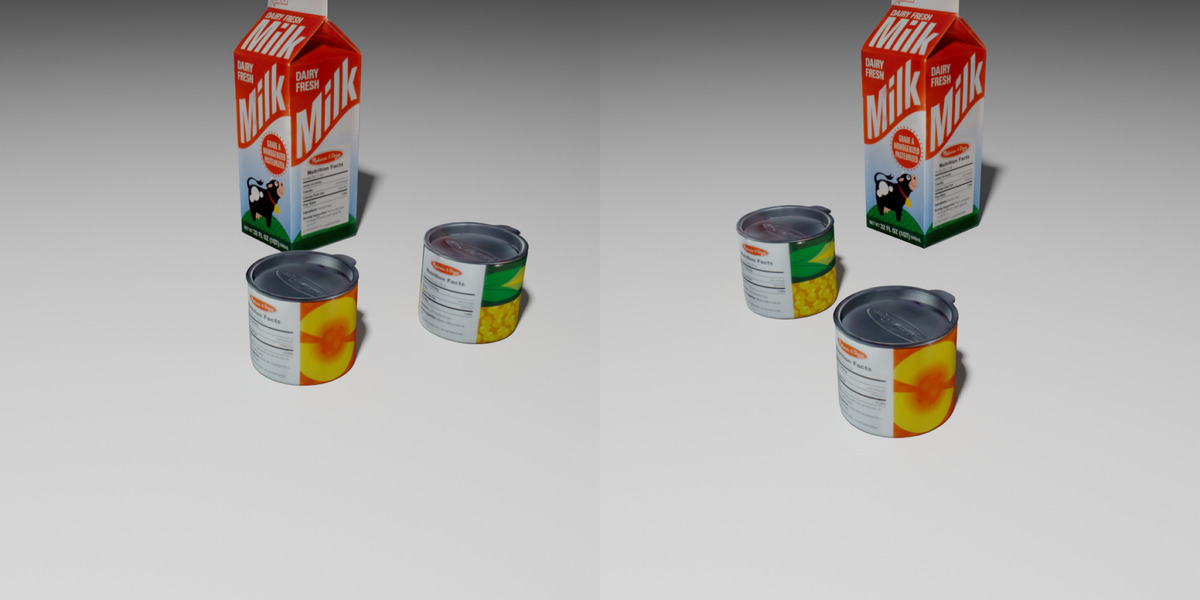}%
    \includegraphics[width=\imgwidth]{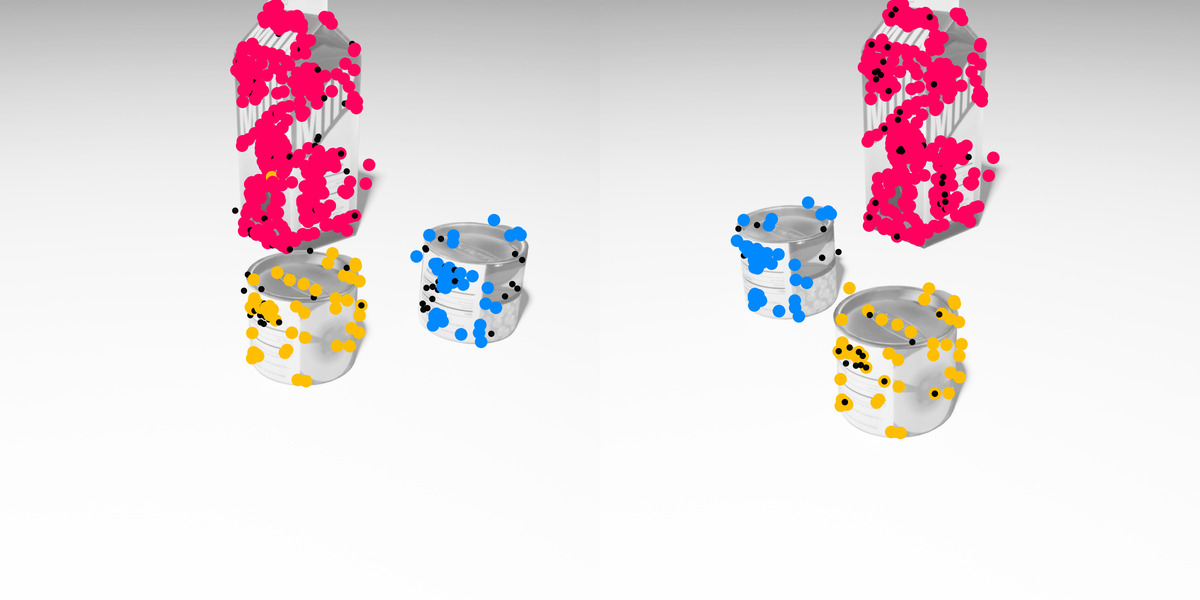}

    \includegraphics[width=\imgwidth]{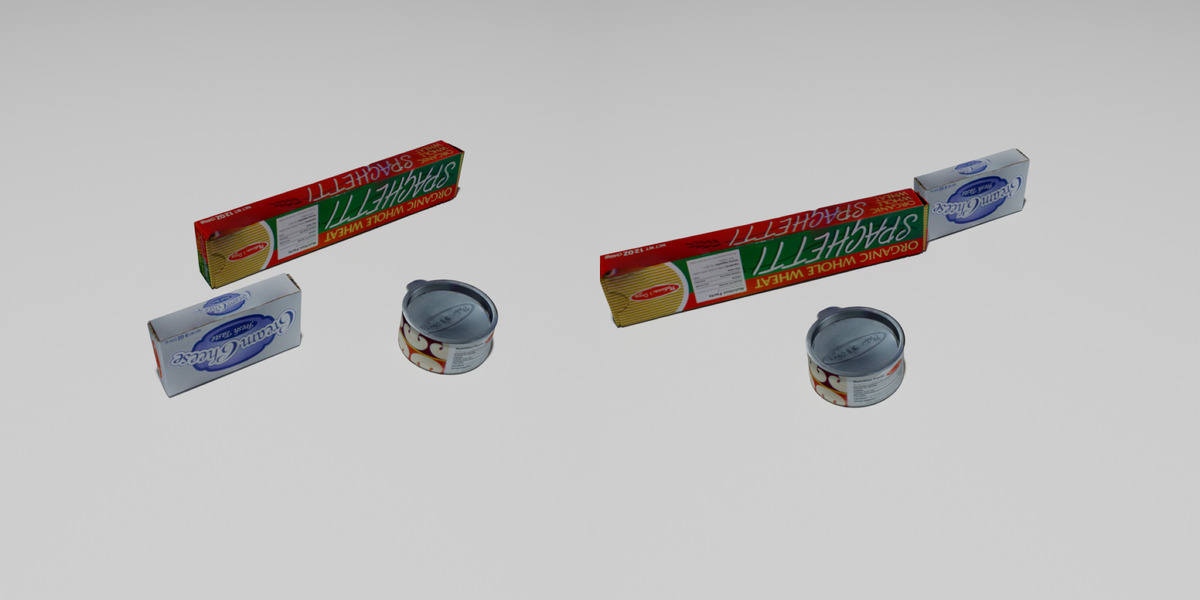}%
    \includegraphics[width=\imgwidth]{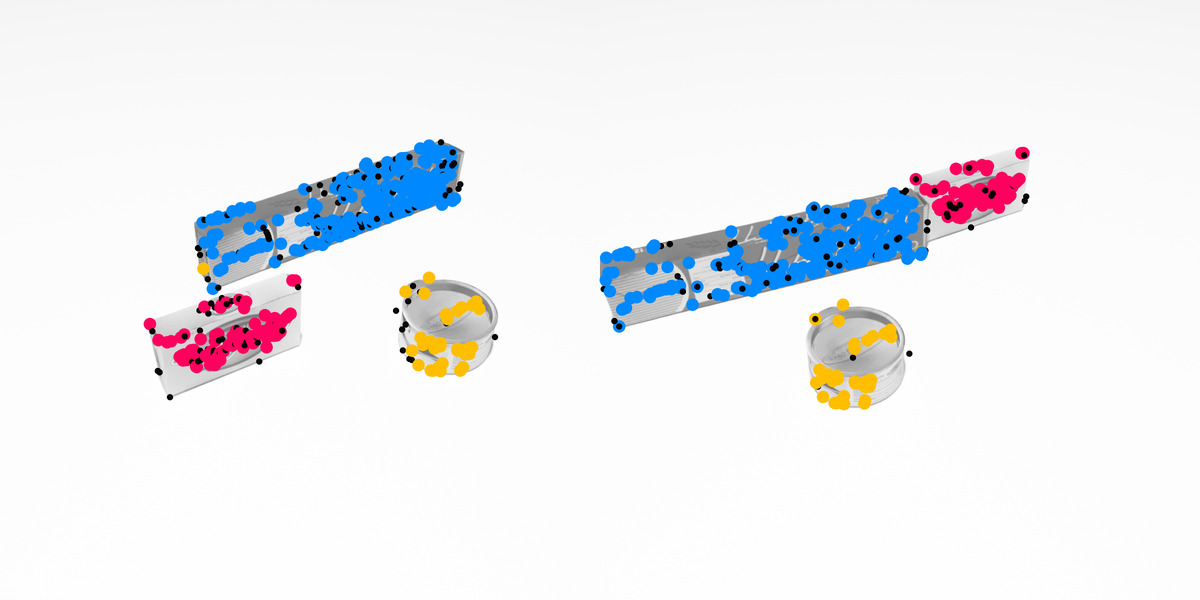}%
    \hfill%
    \includegraphics[width=\imgwidth]{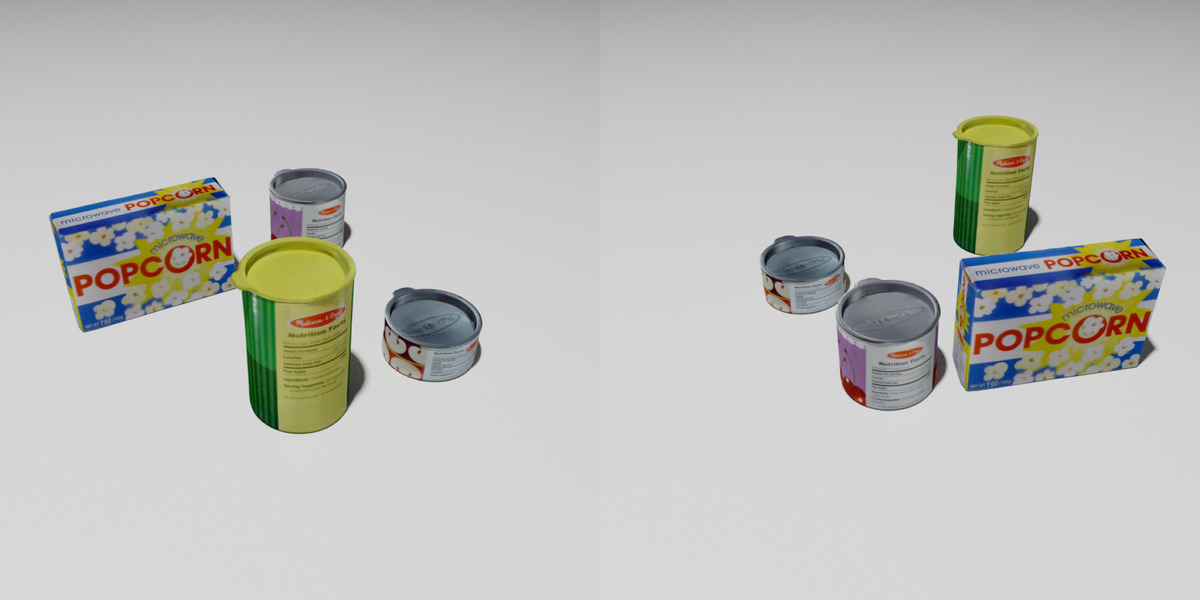}%
    \includegraphics[width=\imgwidth]{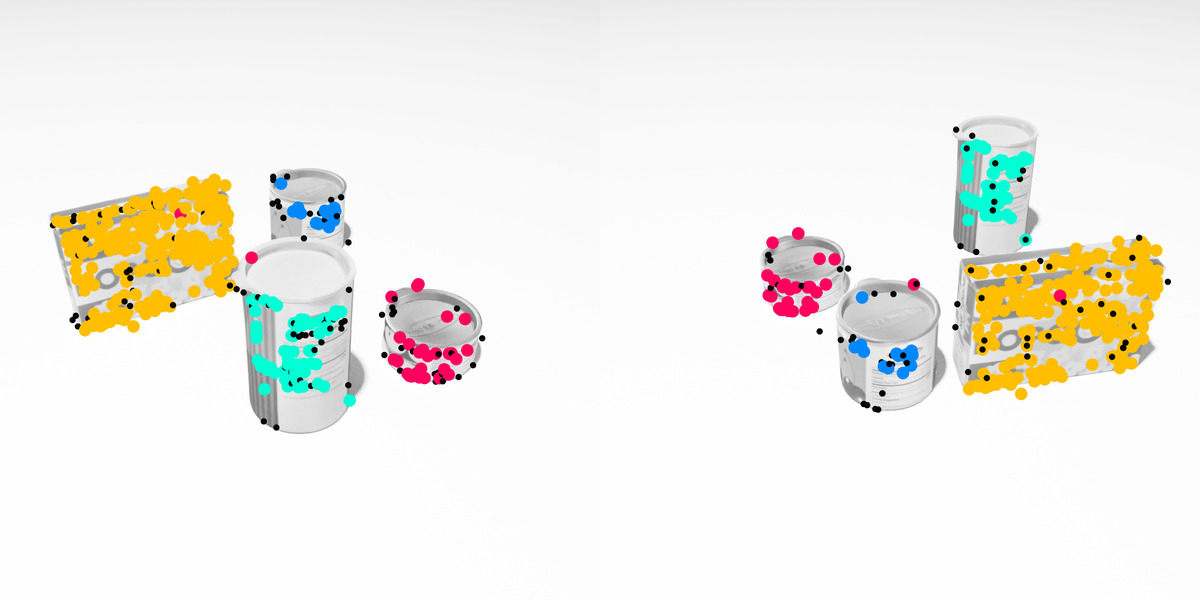}

    \includegraphics[width=\imgwidth]{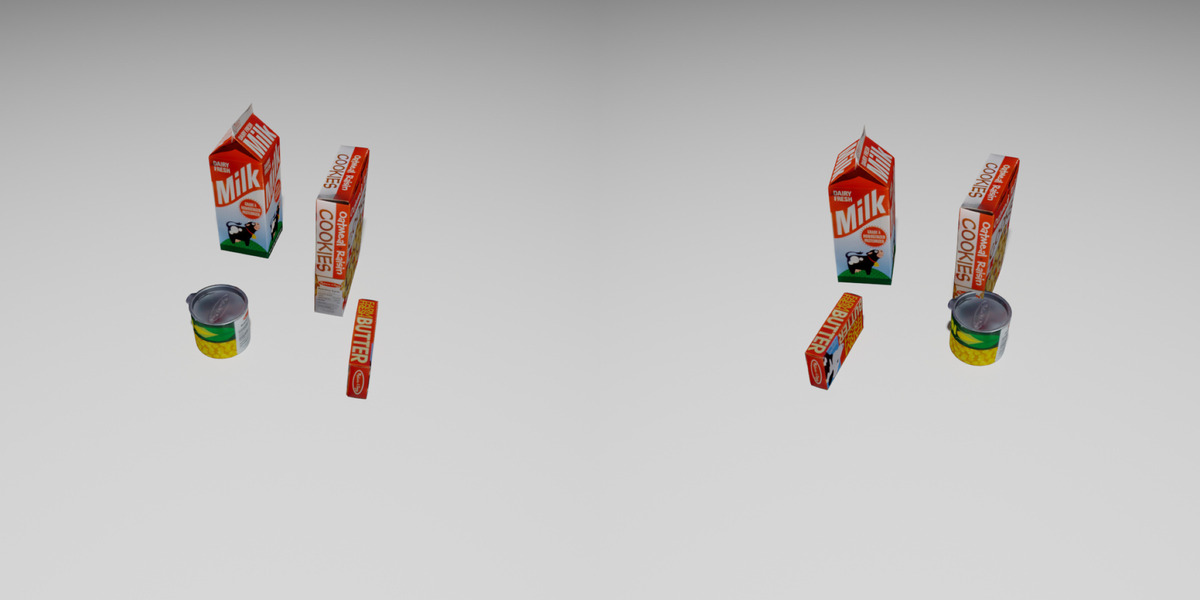}%
    \includegraphics[width=\imgwidth]{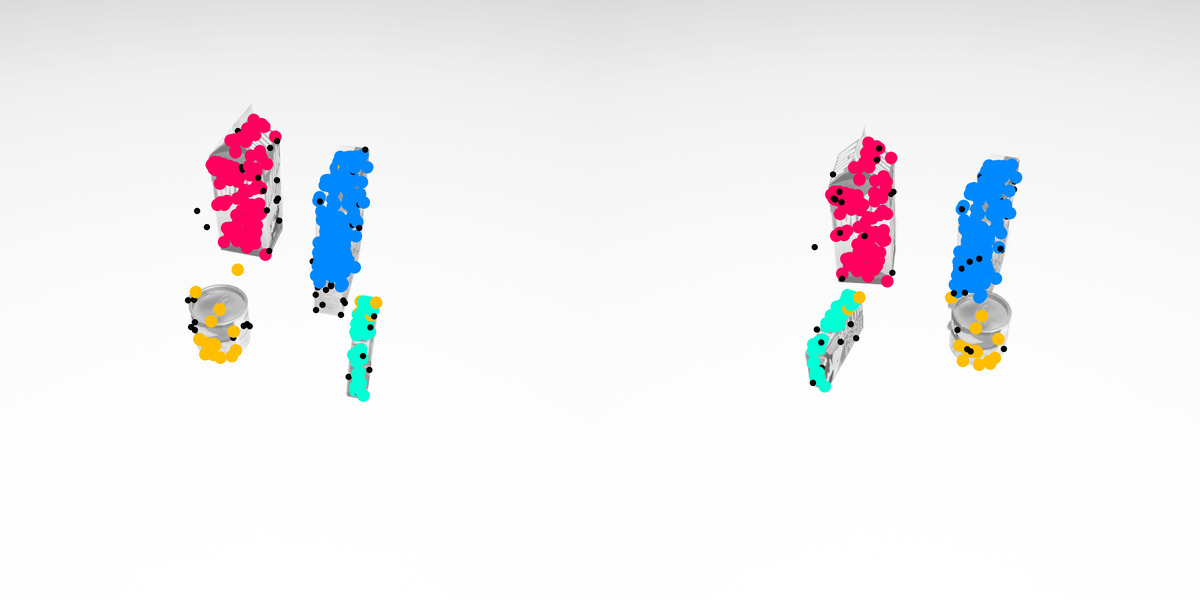}%
    \hfill%
    \includegraphics[width=\imgwidth]{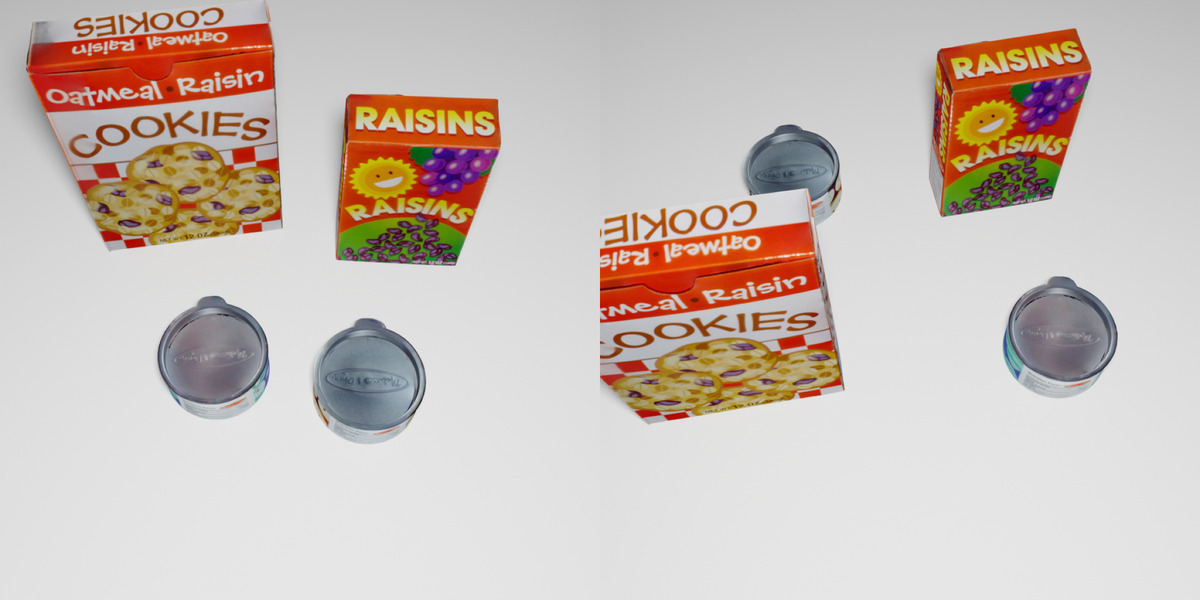}%
    \includegraphics[width=\imgwidth]{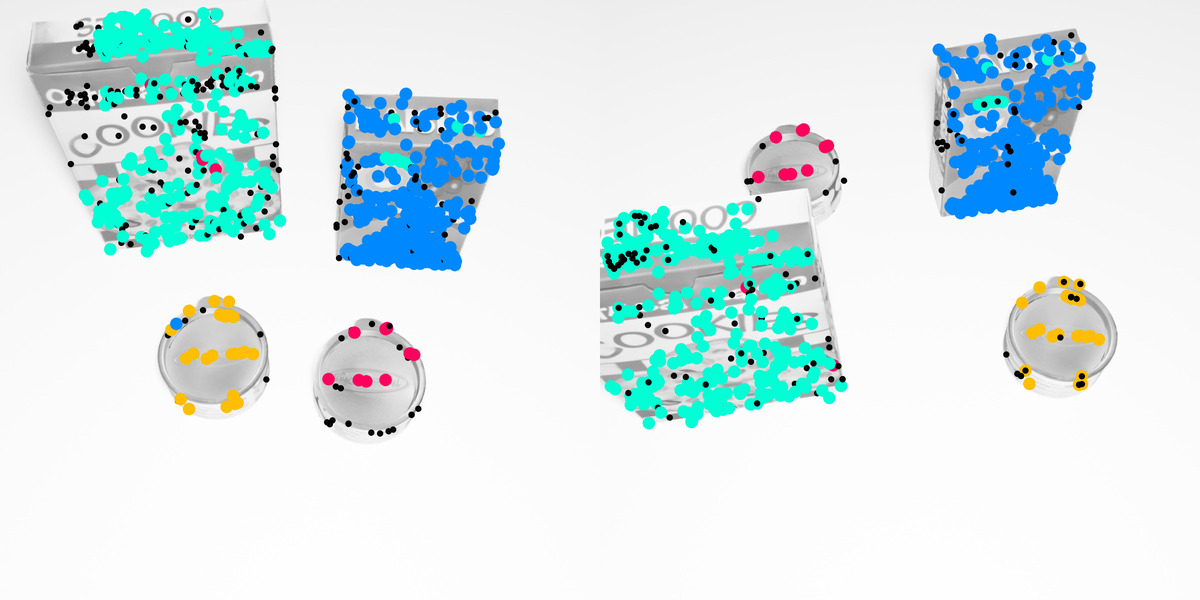}
    
    \includegraphics[width=\imgwidth]{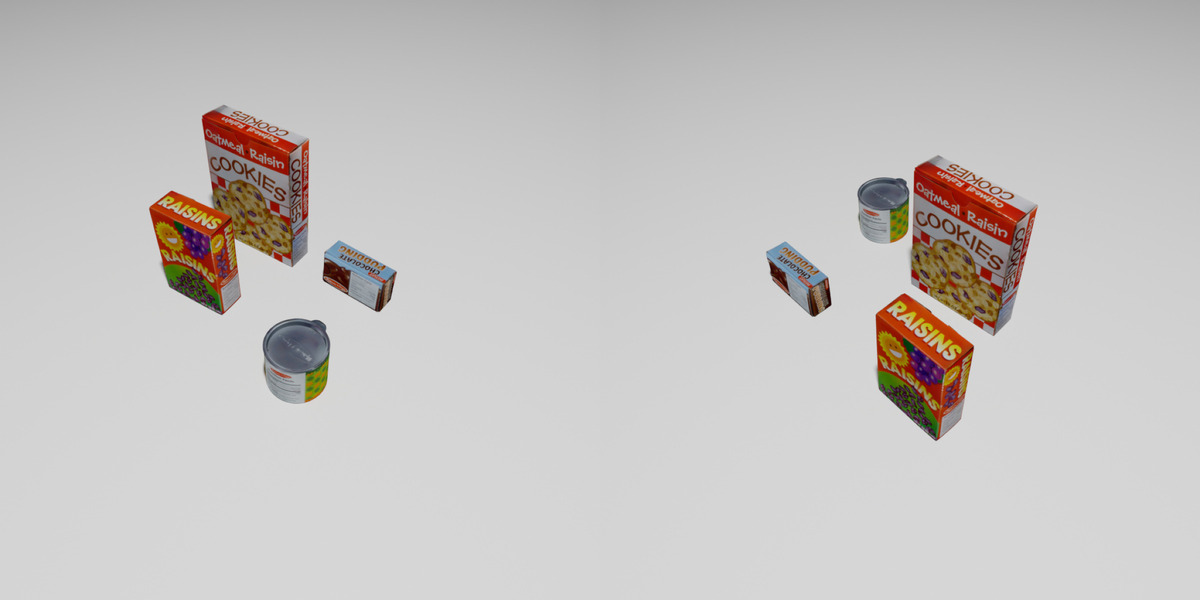}%
    \includegraphics[width=\imgwidth]{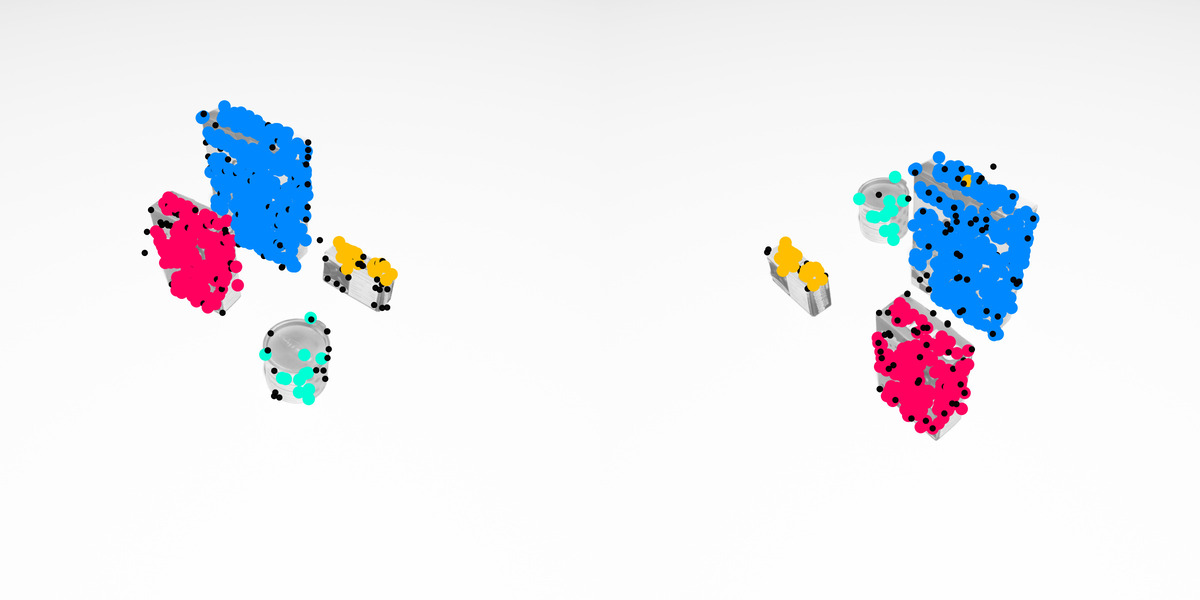}%
    \hfill%
    \includegraphics[width=\imgwidth]{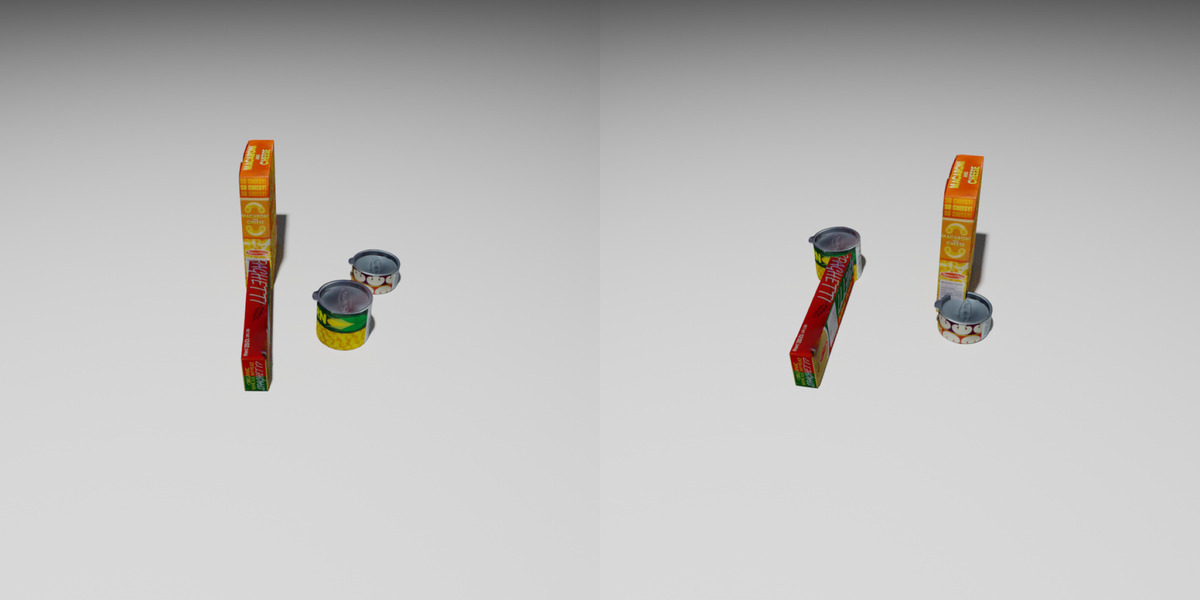}%
    \includegraphics[width=\imgwidth]{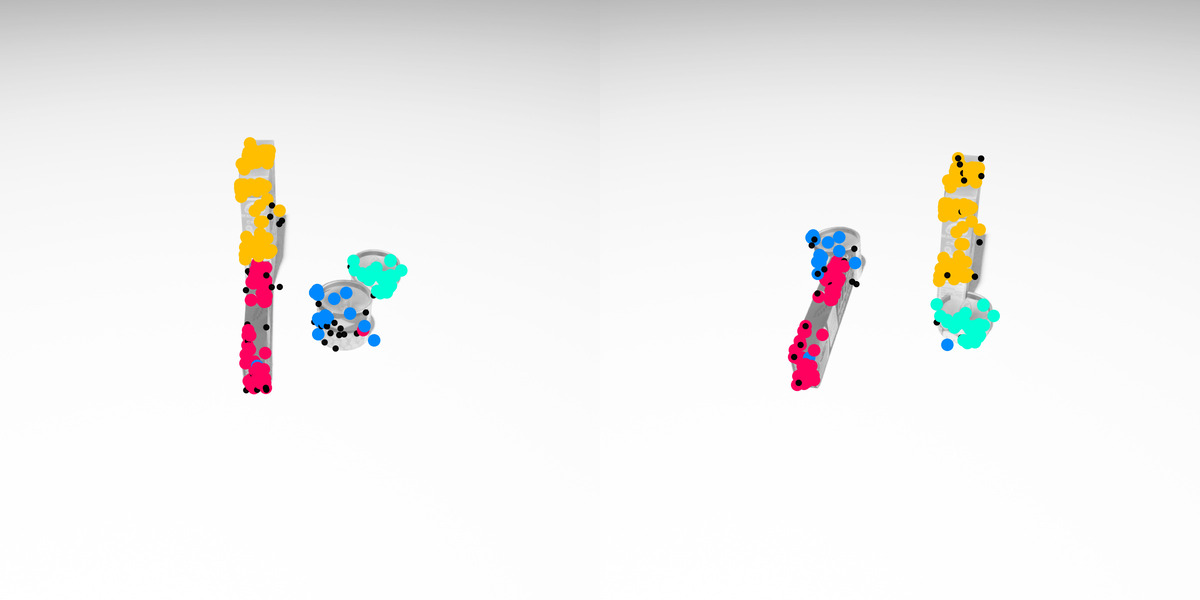}
    
    \caption{{HOPE-F:} We show additional examples from our new dataset for fundamental matrix fitting. The left pair of each example shows the RGB images, the right pair visualises the pre-computed SIFT keypoints, colour coded by ground truth label.}
		\label{fig:more_hope_examples}
\end{figure*}

Fig.~\ref{fig:datasets:f_outliers} compares the distribution of outlier ratios between HOPE-F and Adelaide-F.
The average outlier ratio of Adelaide-F is higher than that of HOPE-F, but the latter contains a broader range of outlier ratios per scene.
Fig.~\ref{fig:datasets:f_points} shows that the number of SIFT point correspondences per scene also varies more strongly in HOPE-F compared to Adelaide-F.
We furthermore look at the percentage of inliers for each individual fundamental matrix in Fig.~\ref{fig:datasets:f_inliers}.
The models in HOPE-F have inlier ratios between $0.2\%$ and $100\%$, compared to $7-56\%$ in Adelaide-F.
Fig.~\ref{fig:more_hope_examples} shows additional examples from the HOPE-F dataset.

\subsection{Synthetic Metropolis Homography}

For homography fitting, we construct a new dataset using a synthetic 3D model of a city~\footnote{\texttt{https://sketchfab.com/3d-models/city-\\1f50f0d6ec5a493d8e91d7db1106b324} \\ Created by Mateusz Woliński under the CC BY 4.0 DEED
license. The author does not allow the 3D model to be used in datasets for, in the development of, or as inputs to generative AI programs.}.
Using the positions of the cars present in the 3D model as anchor points, we define seven camera trajectories, one of which is reserved for the test set.
We remove the cars from the model before rendering.

We define the trajectories by first finding the shortest paths between two car positions which do not intersect any mesh polygons, then iteratively combining paths with the shortest distance between their respective start or end positions, until no paths can be combined without creating a trajectory which intersects a mesh polygon.

Along each trajectory, we render sequences of images with varying step sizes $\Delta_t \in \{1, 2, 4\}$ metres, focal lengths $f \in \{24, 30, 45\}\si{\milli\metre}$ and with the camera pointing either forward or turned to the right by $90^{\circ}$.
The camera height is fixed at $1.8\si{\metre}$.
Camera elevation starts at $100^{\circ}$ and is randomly tilted up or down by an angle of $\beta \sim \mathcal{N}(0,\,1)$ degrees but always clipped to a range between $90^{\circ}$ and $115^{\circ}$.

In order to determine the ground truth homographies for each image pair, we first compute the normal forms $(\vec{n}_i, d_i)$ of the planes arising from each mesh polygon visible in the images. 
Using the known camera intrinsics $\vec{K}$ and relative camera pose $\vec{R}, \vec{t}$, we compute ground truth homographies for every plane in the scene for each of the image pairs:
\begin{equation}
    \vec{H}_i = \vec{K} \left( \vec{R} - {d_i}\inverse \vec{t} \vec{n}_i\tran \right) \vec{K}\inverse \, .
\end{equation}
We then compute SIFT correspondences for each image pair.
For each SIFT keypoint, we determine the mesh polygon it originates from via ray casting.
If two corresponding keypoints lie on the same polygon and are inliers of the homography arising from that polygon, we assign this SIFT correspondence to said homography.
We define a correspondence as an inlier if the symmetric transfer error is below $1$ pixel.
Keypoint correspondences which do not fulfil the above criteria are marked as outliers.
Then, we cluster all correspondences together which lie on approximately coplanar polygons.
The planes of two polygons must have an angle of at most $2^{\circ}$ between them and an offset difference of at most $0.1\si{\metre}$ to be considered approximately coplanar.
Lastly, we remove all clusters which contain fewer than ten correspondences and mark these as outliers. 

\begin{figure*}
\centering
\includegraphics[width=\linewidth]{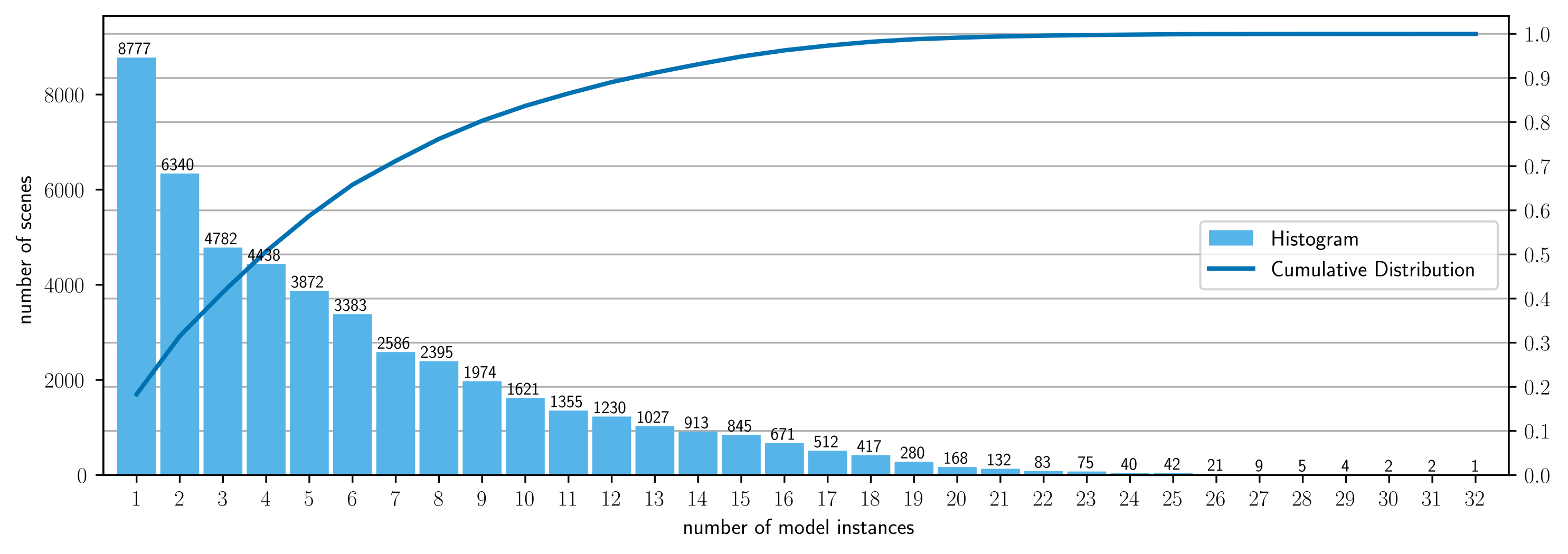}
\caption{Distribution of homographies in the Synthetic Metropolis Homography (SMH) dataset. }
\label{fig:datasets:smh_models}
\end{figure*}

Via this procedure, we generate a total of $48002$ image pairs with keypoint feature correspondences, ground truth cluster labels and ground truth homographies.
We additionally render ground truth depth maps for each image.
SMH is thus significantly larger than Adelaide-H and contains a much larger range of model instances.
As Fig.~\ref{fig:datasets:smh_models} shows, roughly 60\% of scenes in Adelaide-H contain only one or two models, with no scenes showing eight or more models.
In SMH (cf. Fig.~\ref{fig:datasets:smh_models}), on the other hand, roughly 60\% of scenes contain at least four models, and 30\% contain at least eight models.

\begin{figure}
\centering
\includegraphics[width=\linewidth]{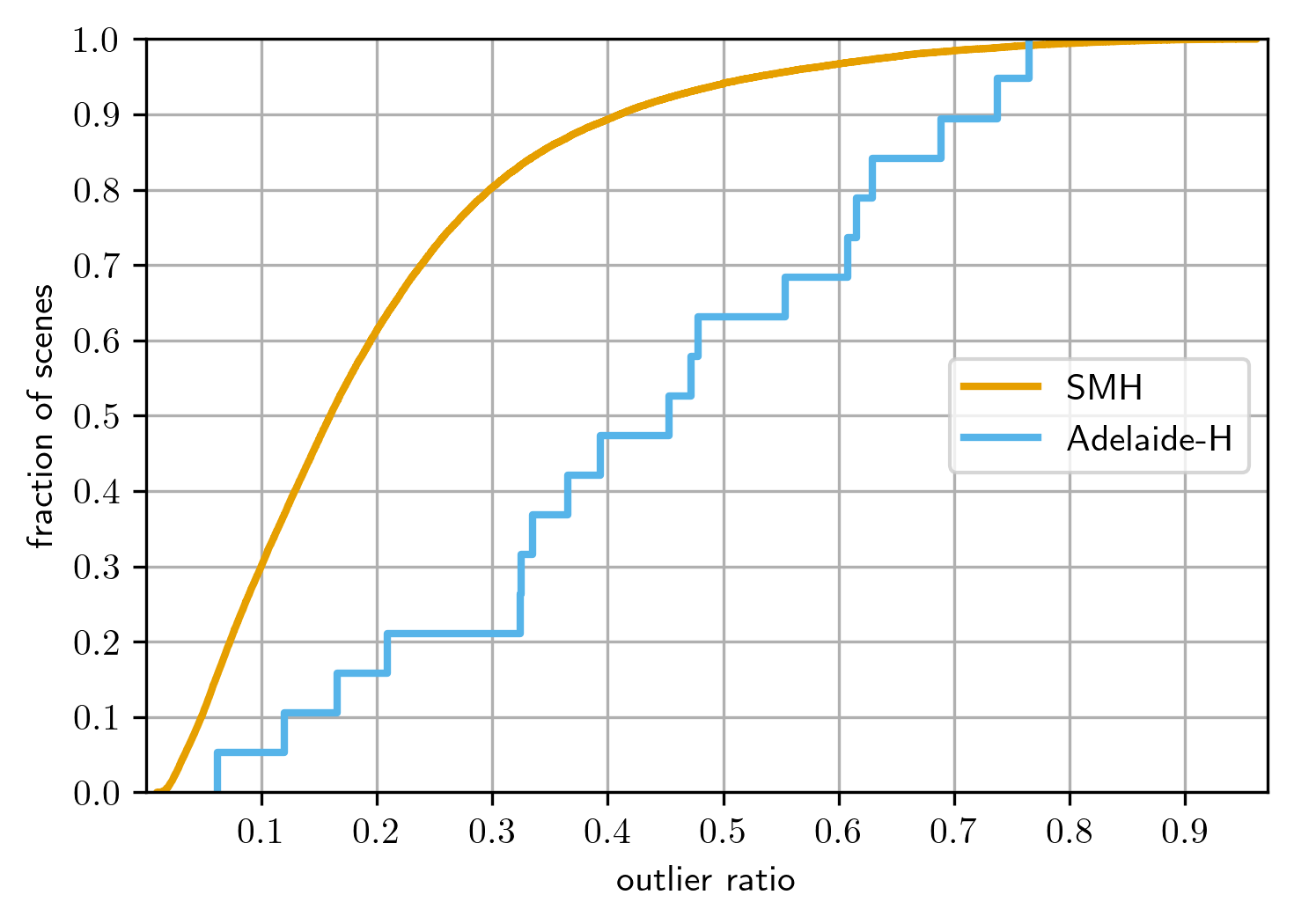}
\caption{Cumulative distributions of outlier ratios in the SMH and Adelaide-H datasets. }
\label{fig:datasets:h_outliers}
\end{figure}

\begin{figure}
\centering
\includegraphics[width=\linewidth]{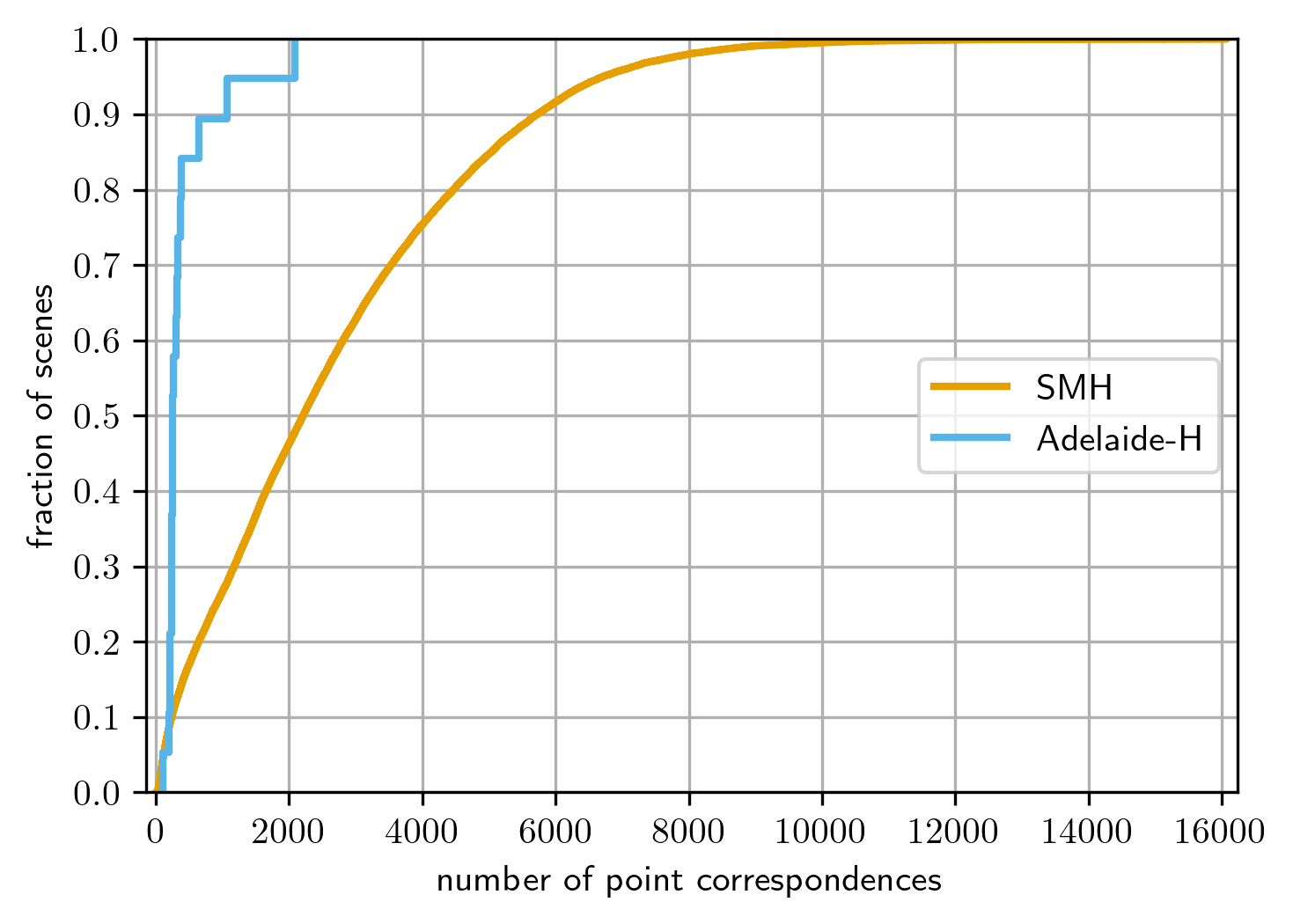}
\caption{Cumulative distributions of point correspondences per scene in the SMH and Adelaide-H datasets. }
\label{fig:datasets:h_points}
\end{figure}

\begin{figure}
\centering
\includegraphics[width=\linewidth]{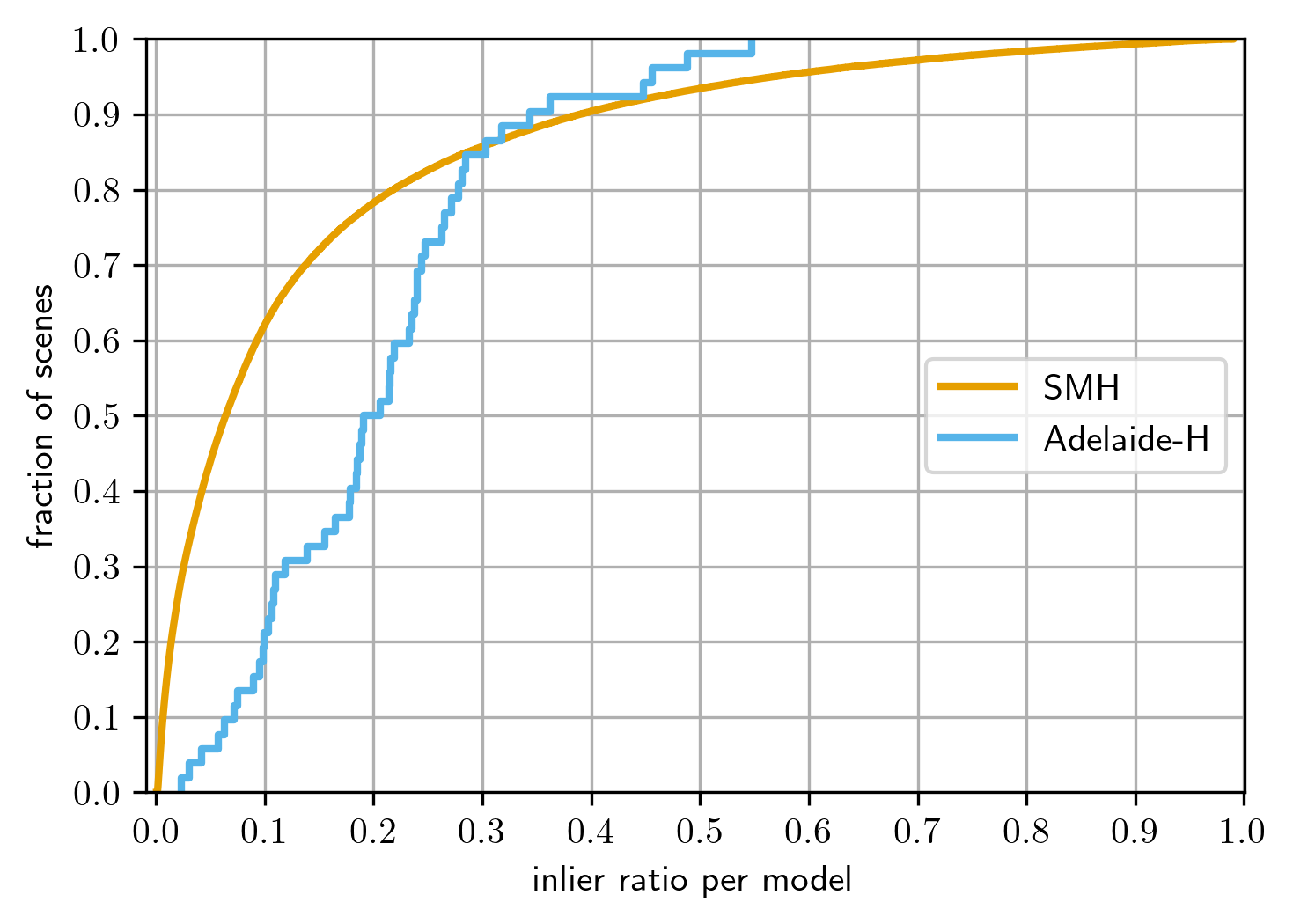}
\caption{Cumulative distributions of inlier ratios per fundamental matrix in the SMH and Adelaide-H datasets. }
\label{fig:datasets:h_inliers}
\end{figure}

\begin{figure*}	
    \setlength{\imgwidth}{0.248\linewidth}
    
    \includegraphics[width=\imgwidth]{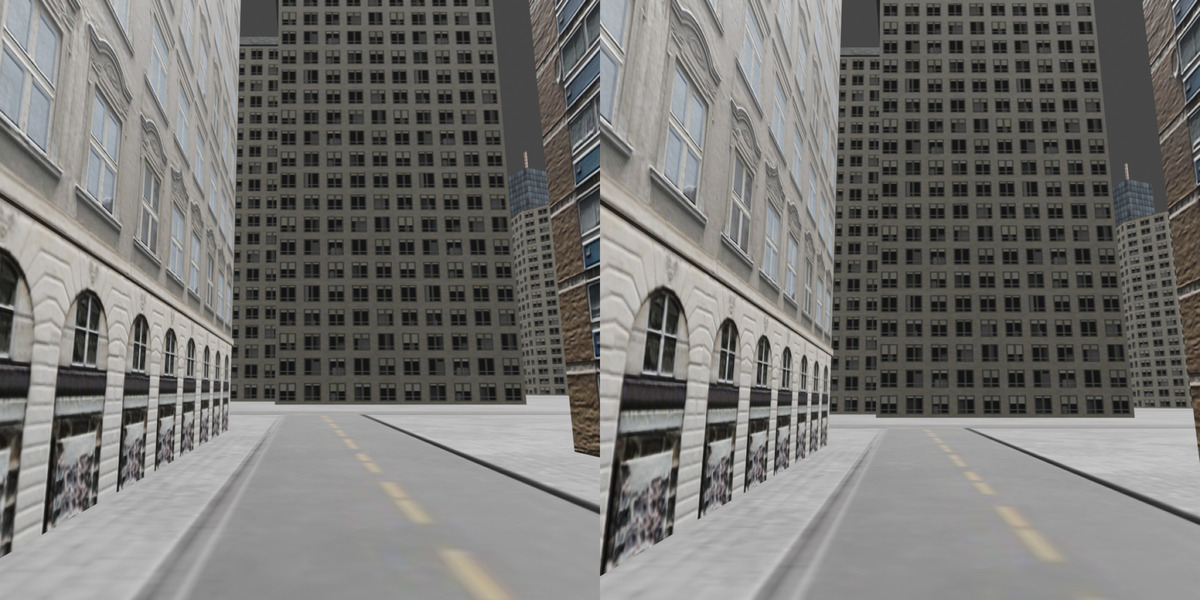}%
    \includegraphics[width=\imgwidth]{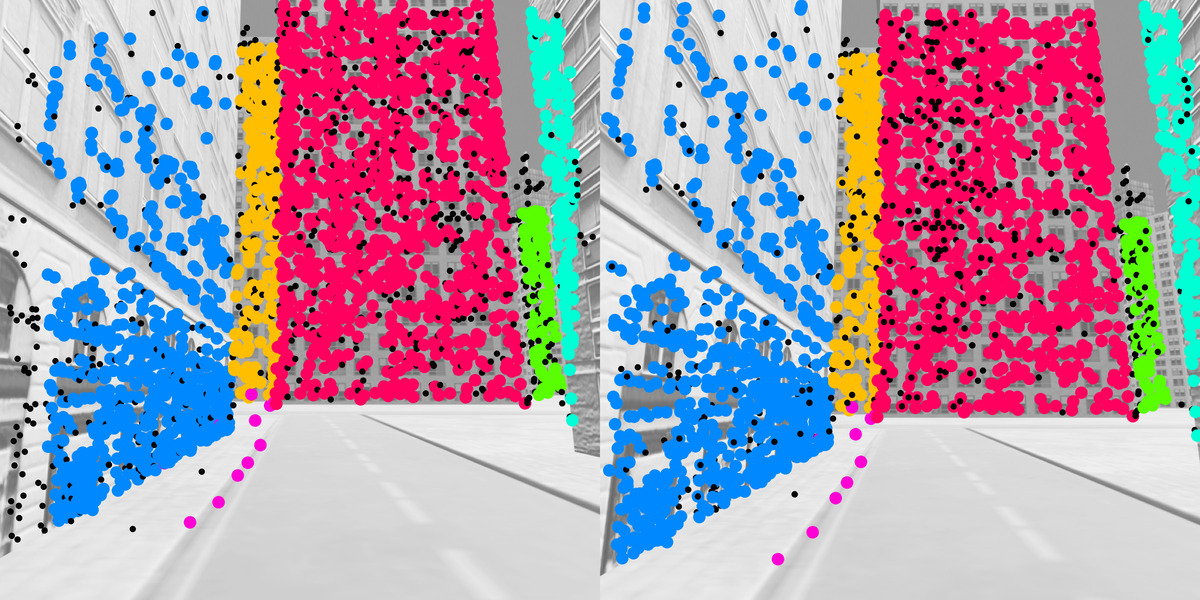}%
    \hfill%
    \includegraphics[width=\imgwidth]{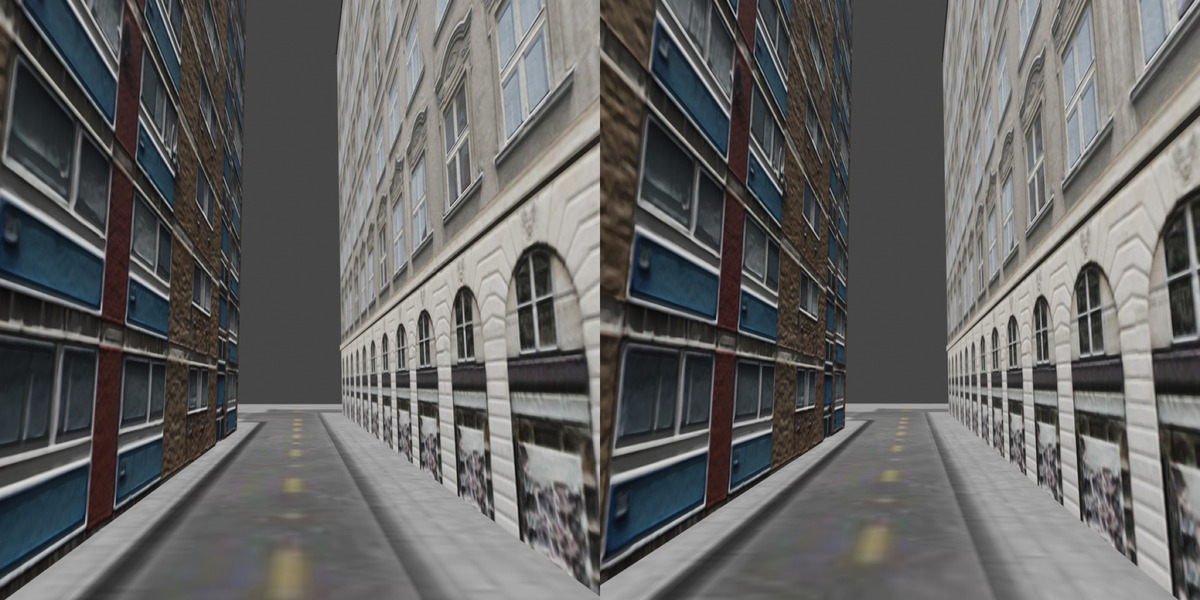}%
    \includegraphics[width=\imgwidth]{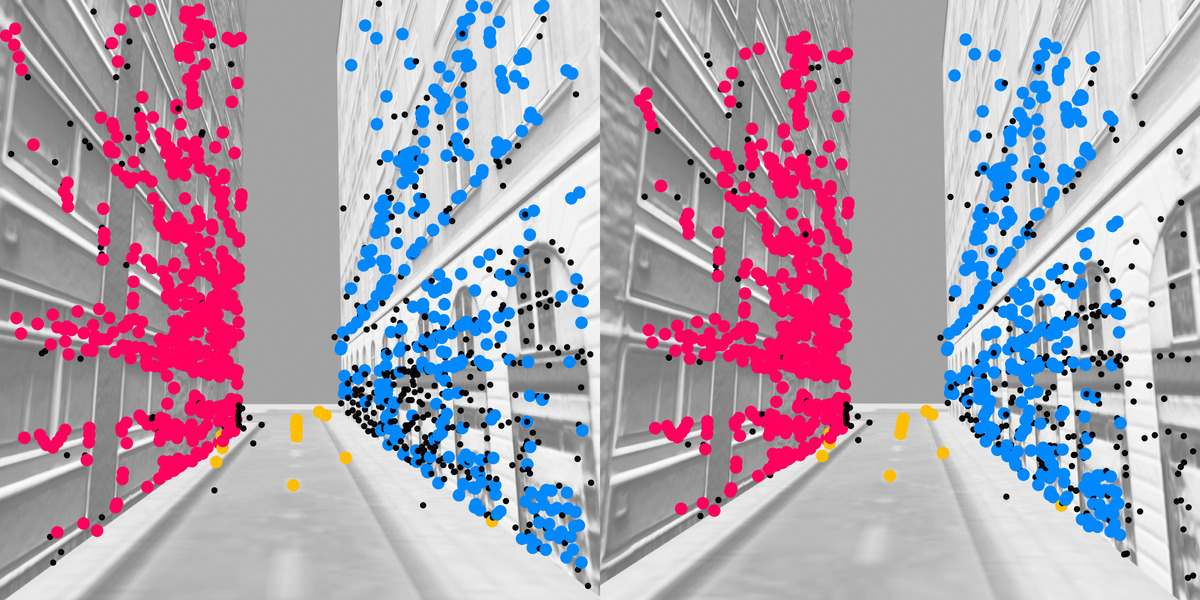}
    
    \includegraphics[width=\imgwidth]{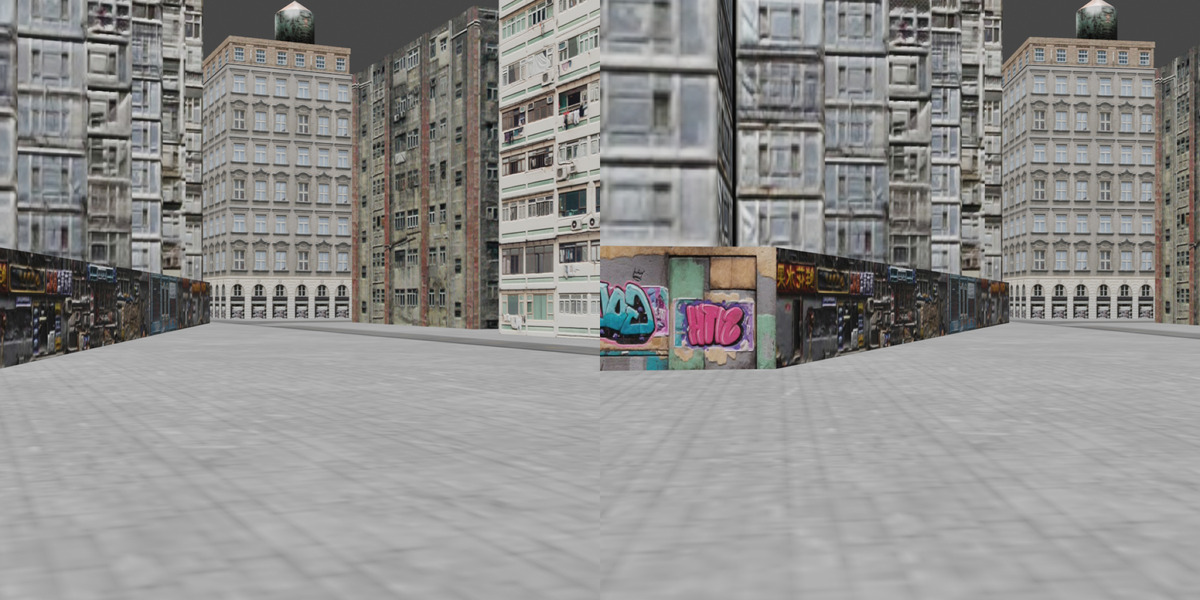}%
    \includegraphics[width=\imgwidth]{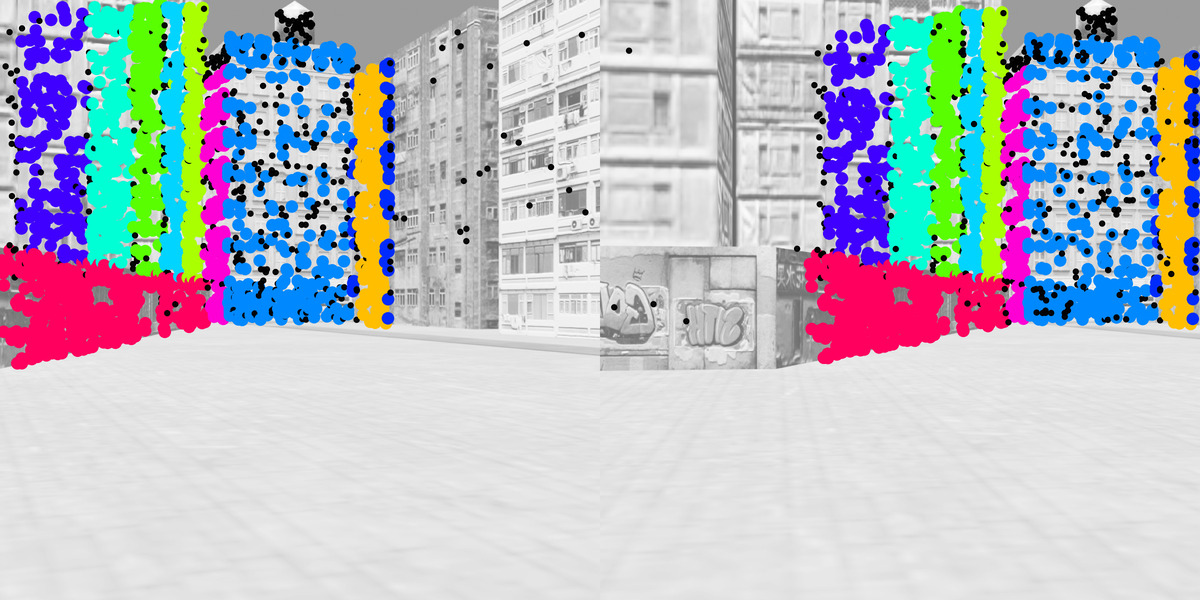}%
    \hfill%
    \includegraphics[width=\imgwidth]{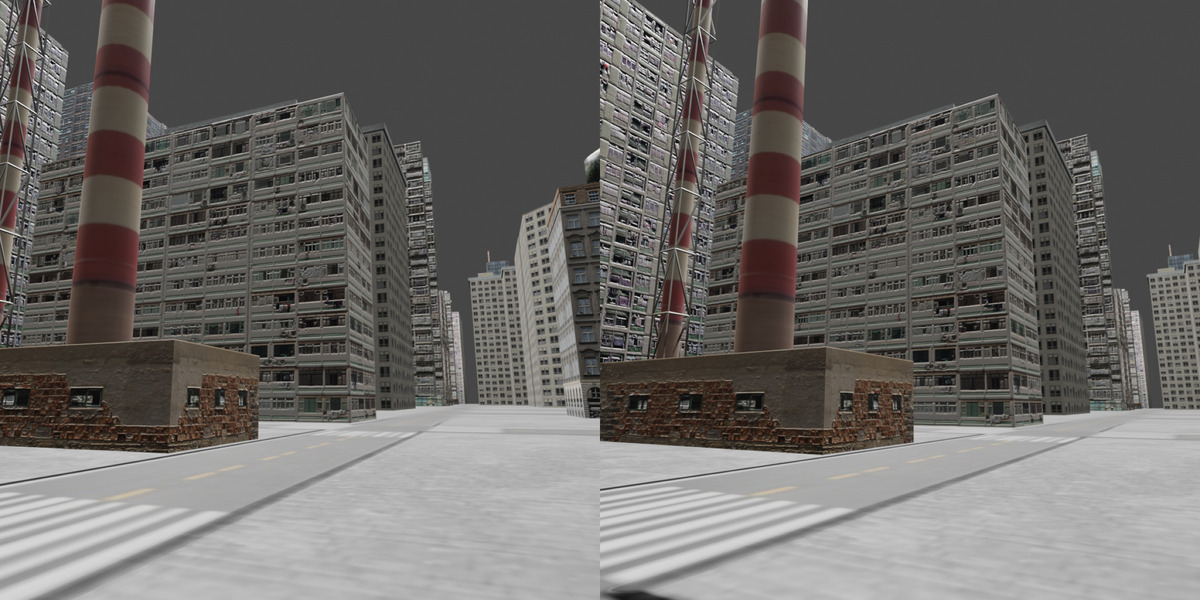}%
    \includegraphics[width=\imgwidth]{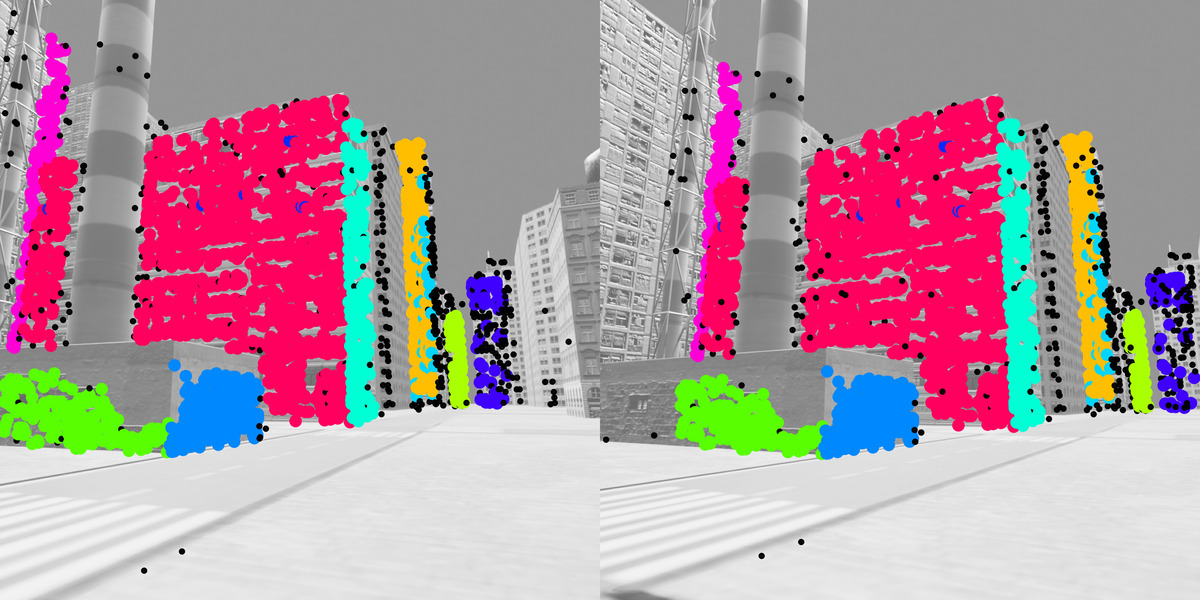}
    
    \includegraphics[width=\imgwidth]{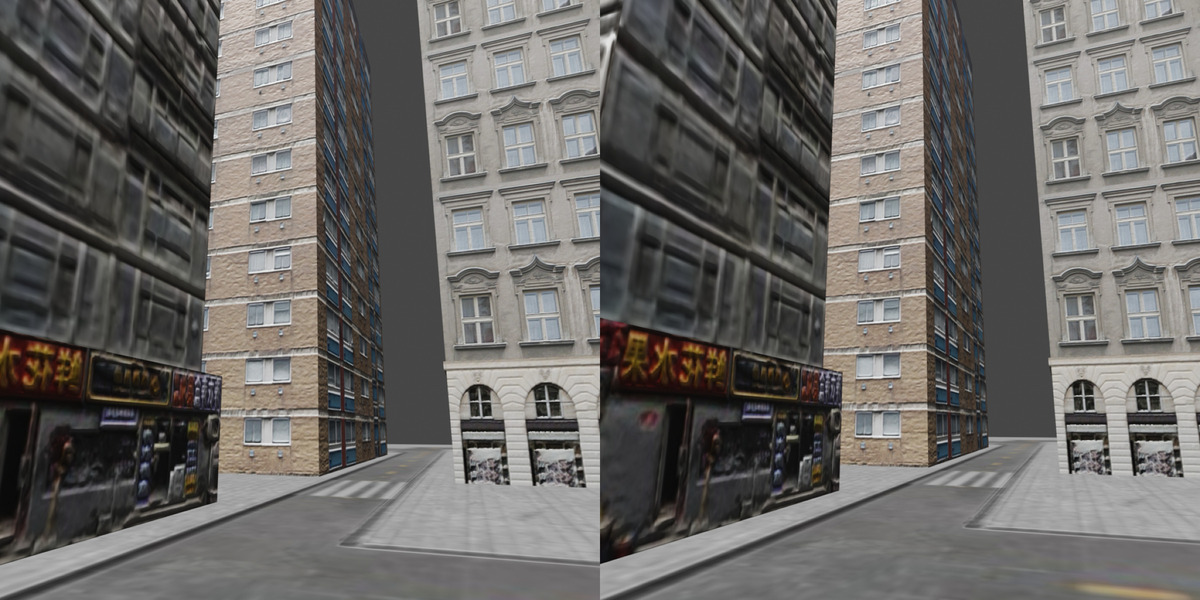}%
    \includegraphics[width=\imgwidth]{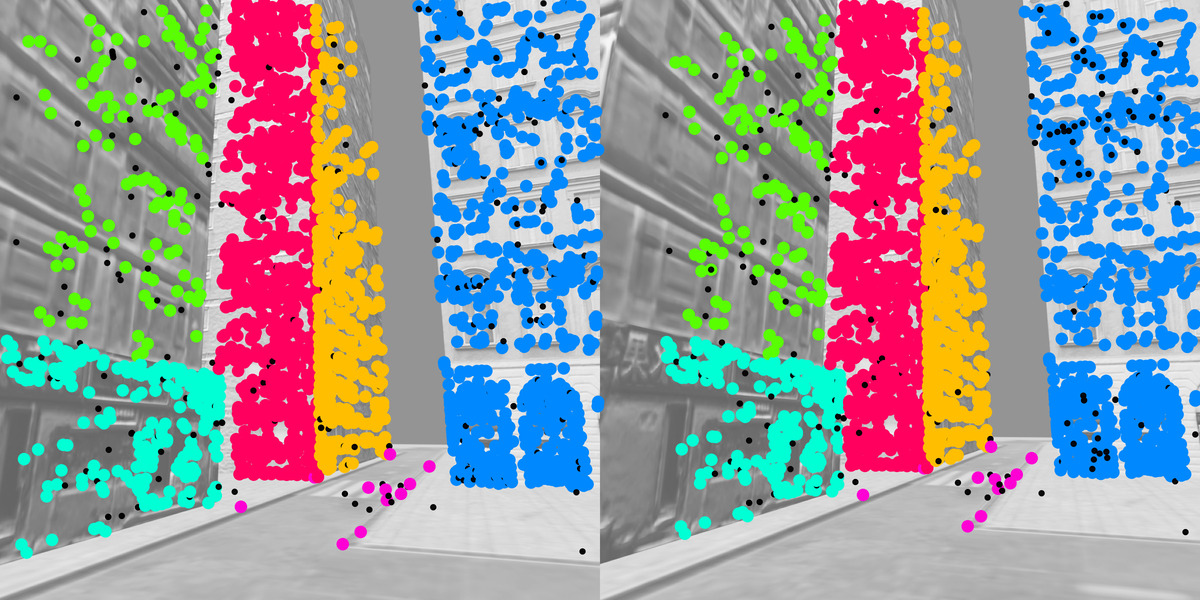}%
    \hfill%
    \includegraphics[width=\imgwidth]{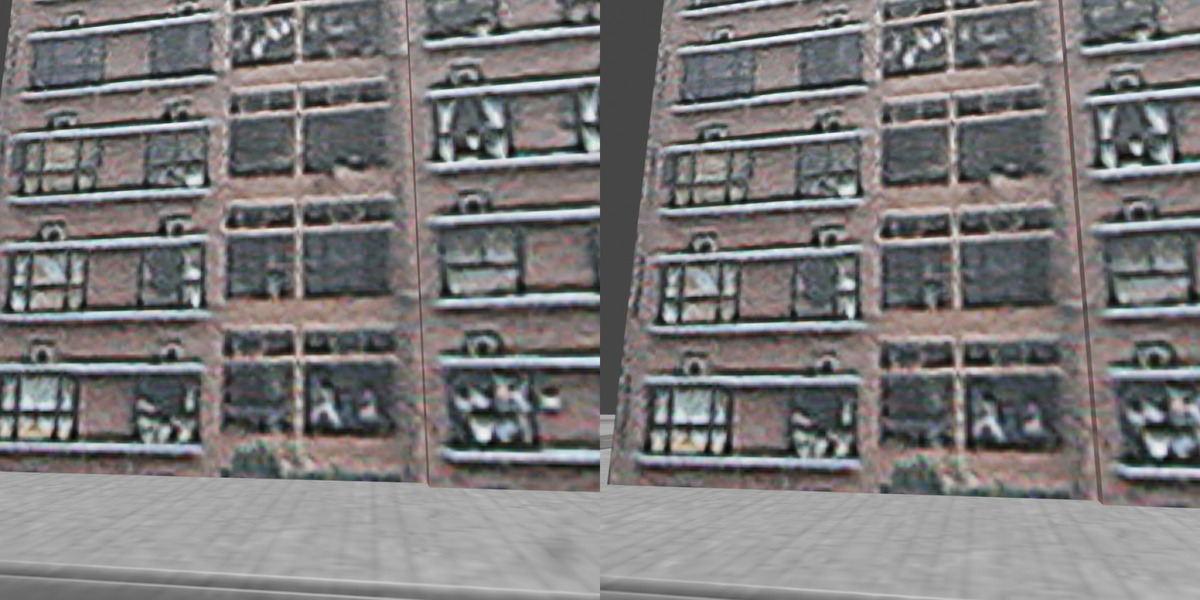}%
    \includegraphics[width=\imgwidth]{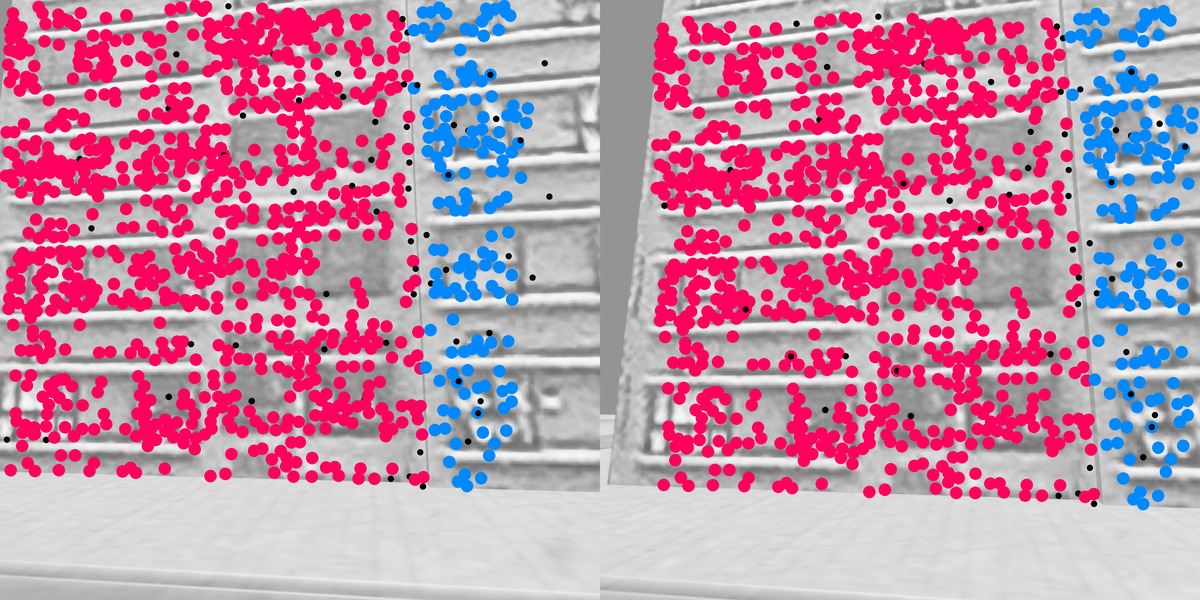}
    
    \includegraphics[width=\imgwidth]{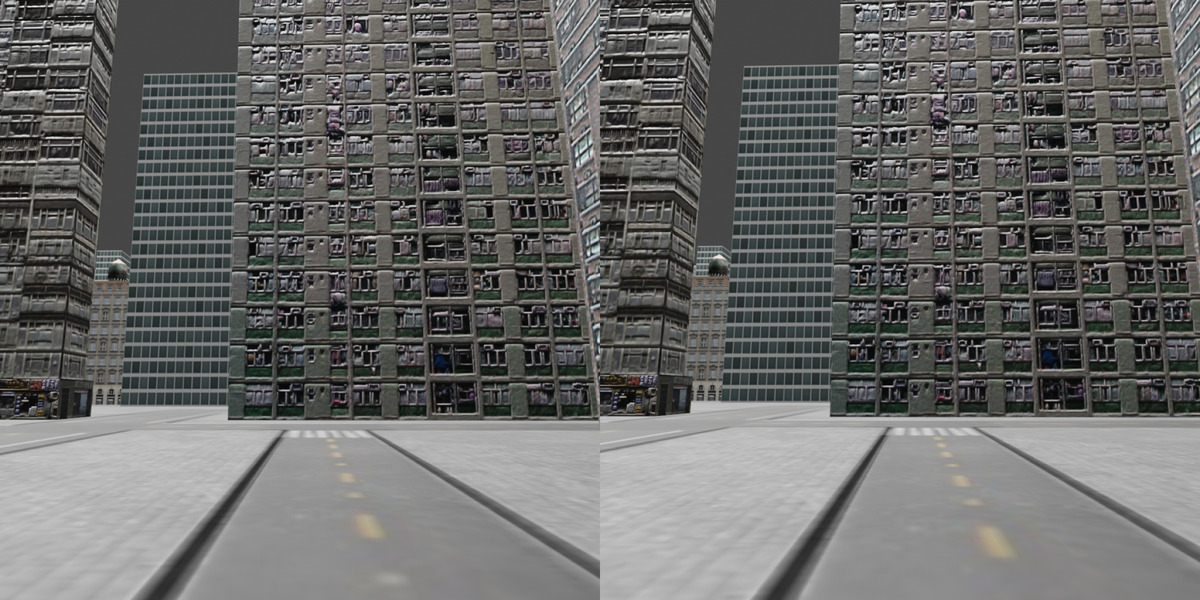}%
    \includegraphics[width=\imgwidth]{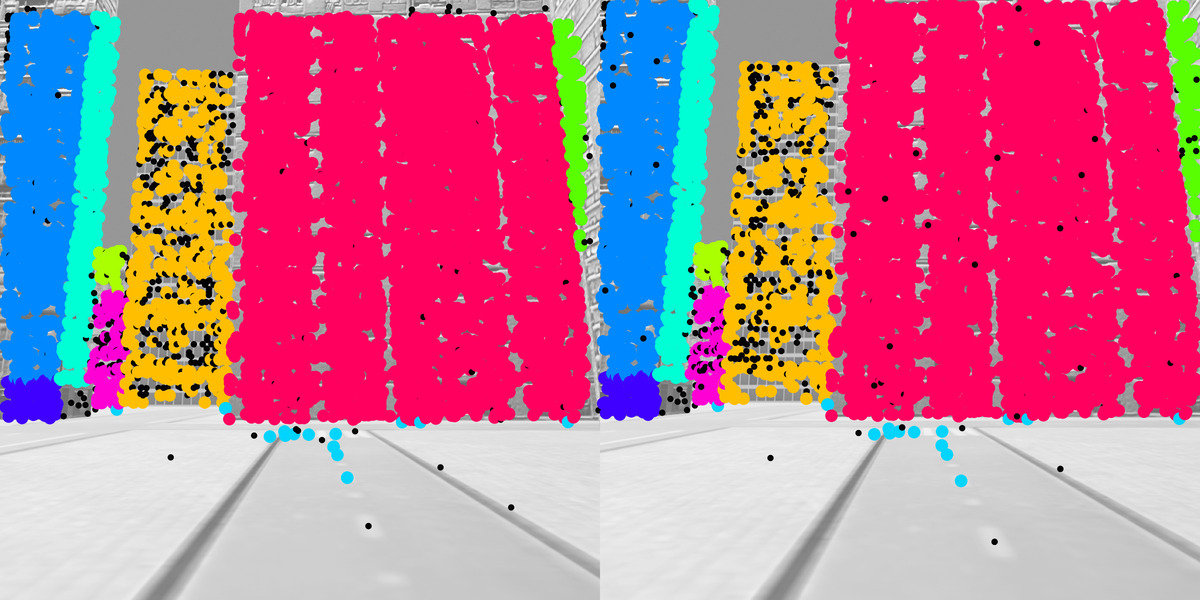}%
    \hfill%
    \includegraphics[width=\imgwidth]{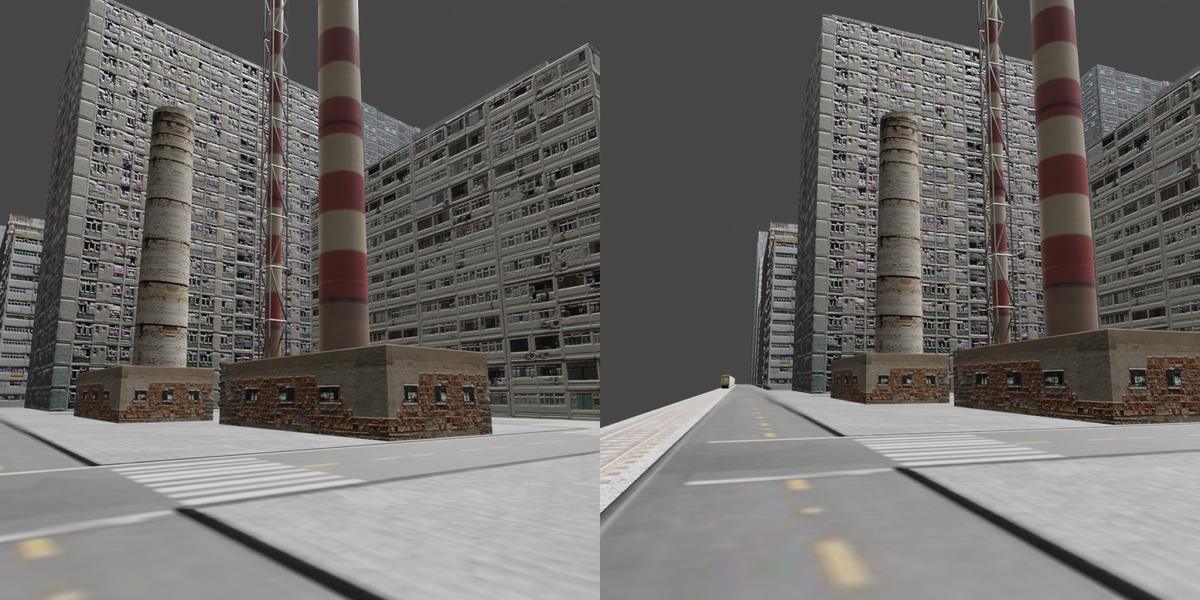}%
    \includegraphics[width=\imgwidth]{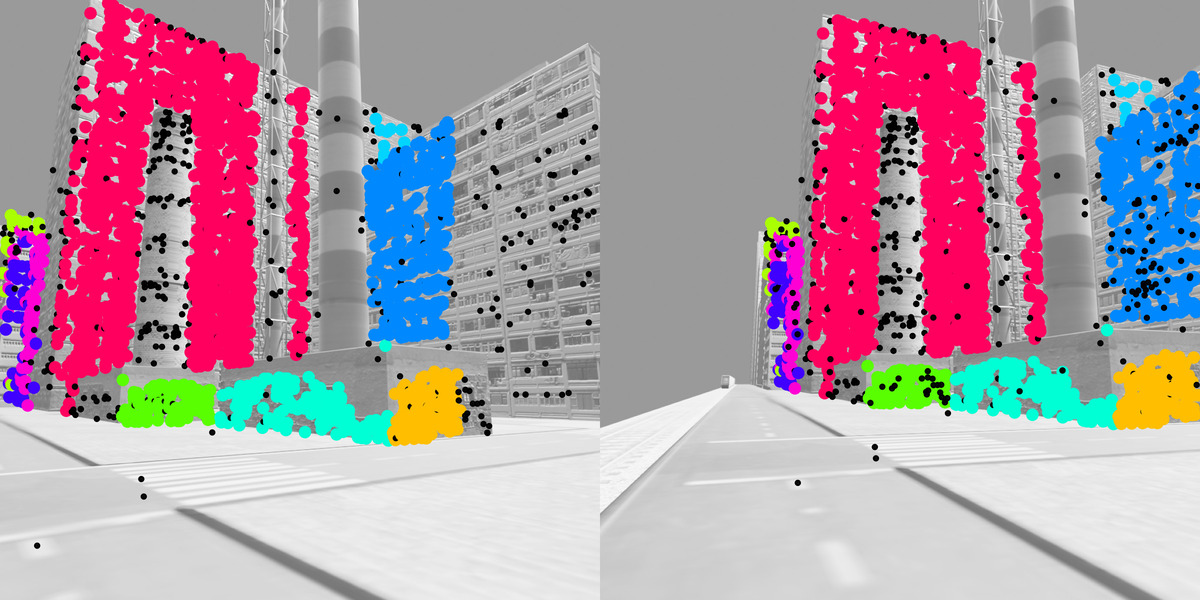}
    
    \includegraphics[width=\imgwidth]{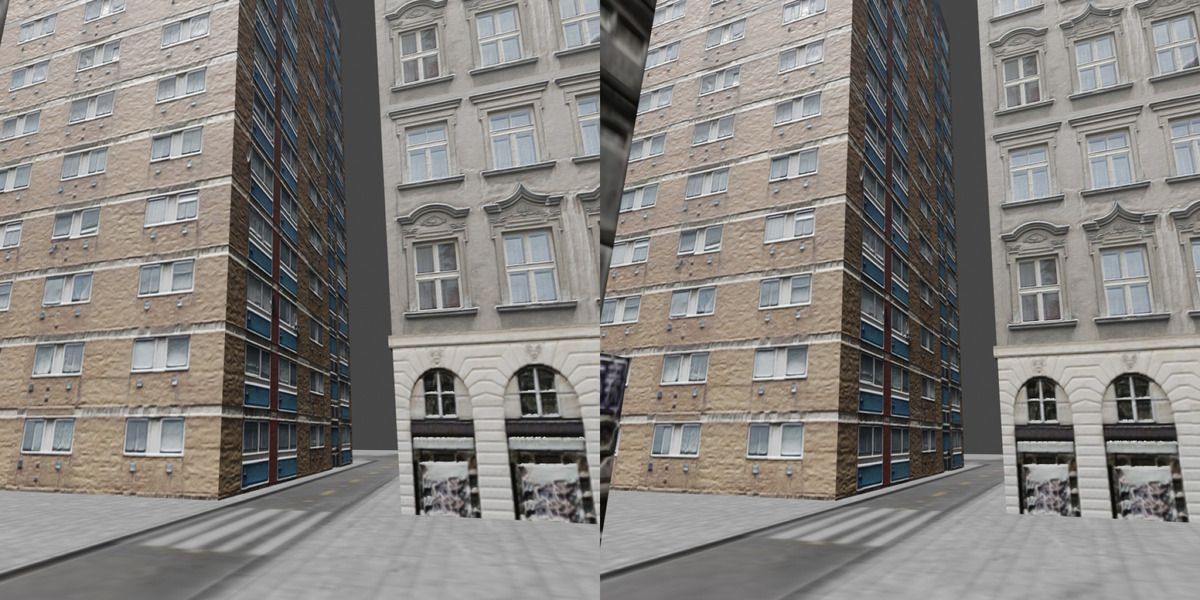}%
    \includegraphics[width=\imgwidth]{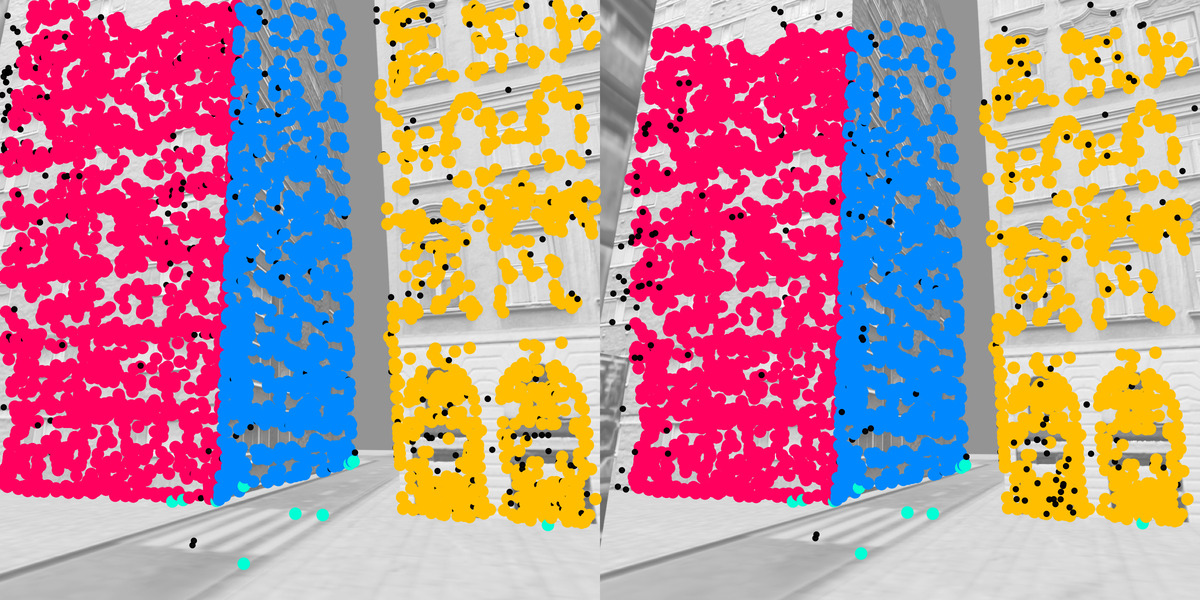}%
    \hfill%
    \includegraphics[width=\imgwidth]{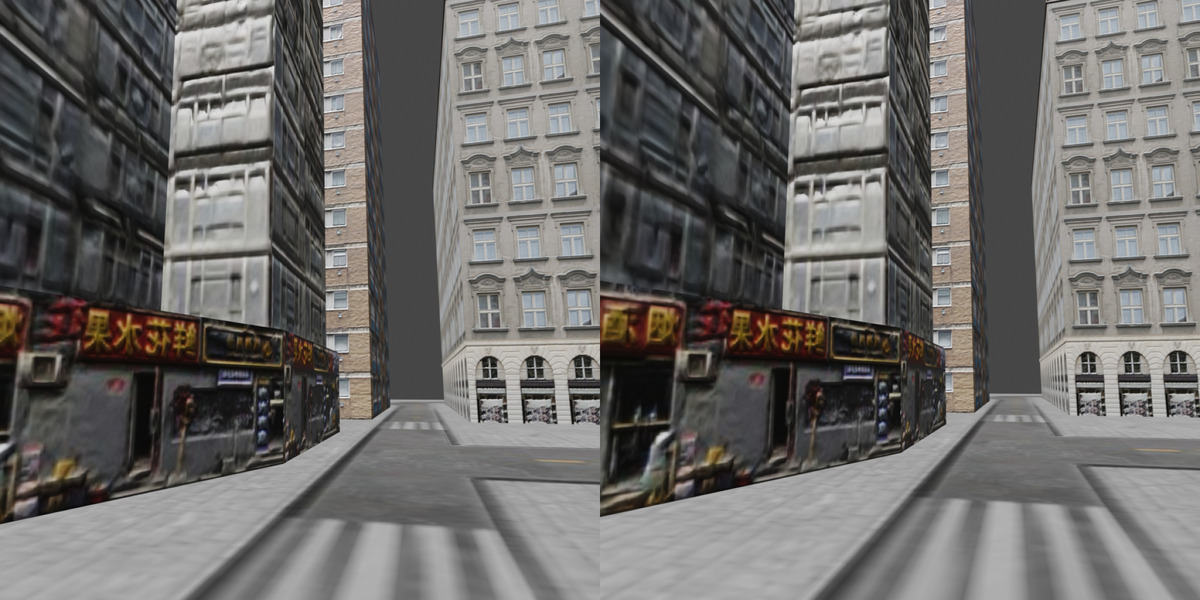}%
    \includegraphics[width=\imgwidth]{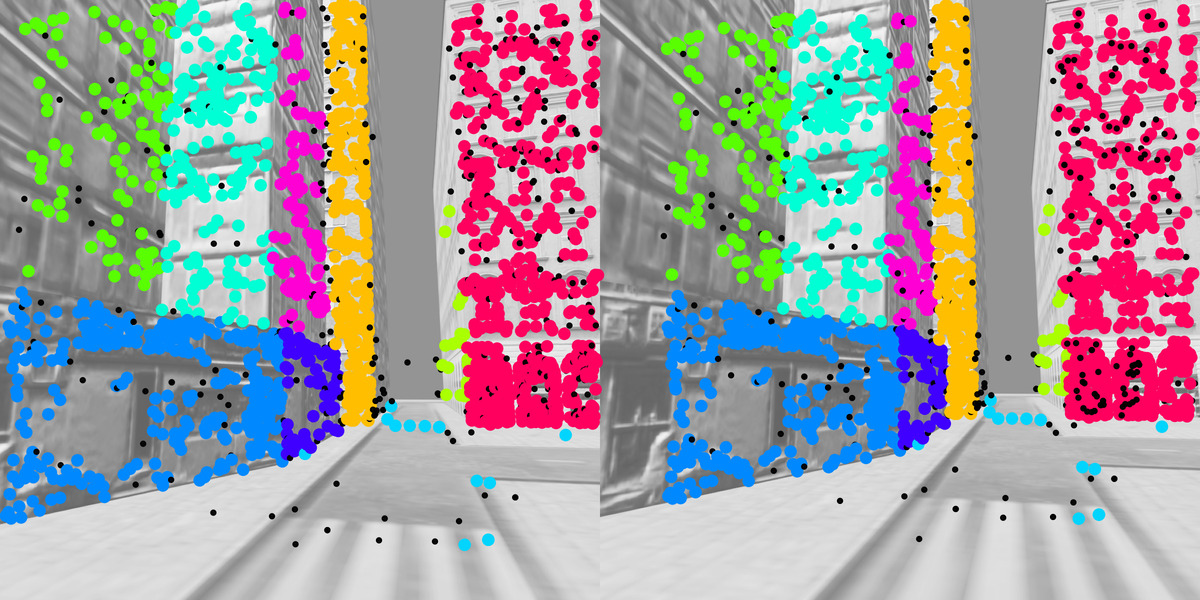}
    
    \includegraphics[width=\imgwidth]{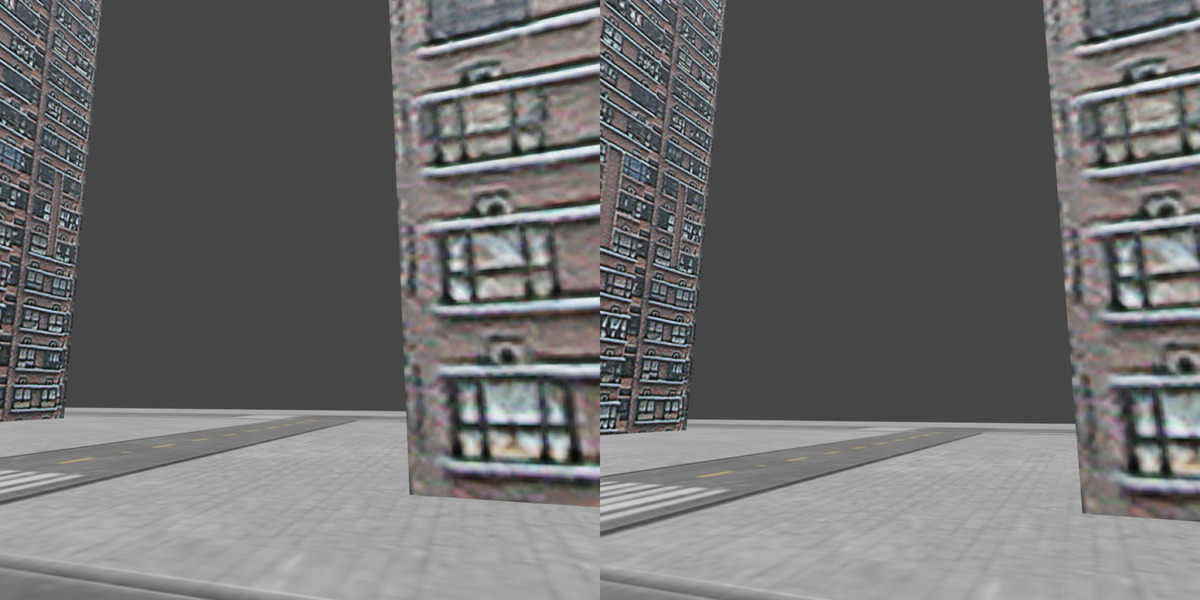}%
    \includegraphics[width=\imgwidth]{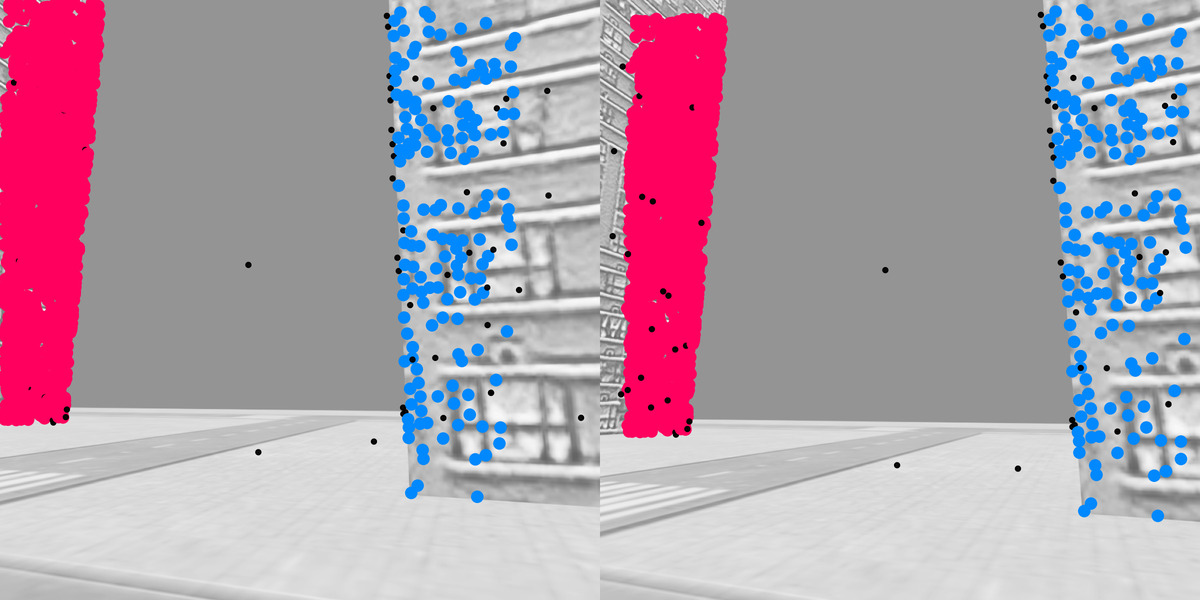}%
    \hfill%
    \includegraphics[width=\imgwidth]{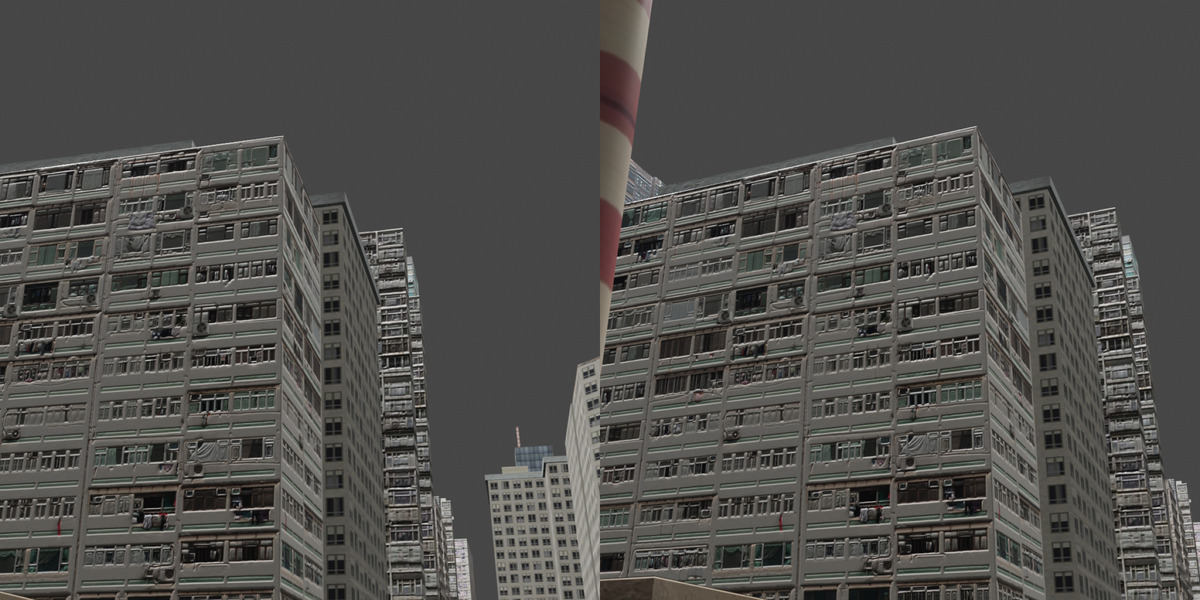}%
    \includegraphics[width=\imgwidth]{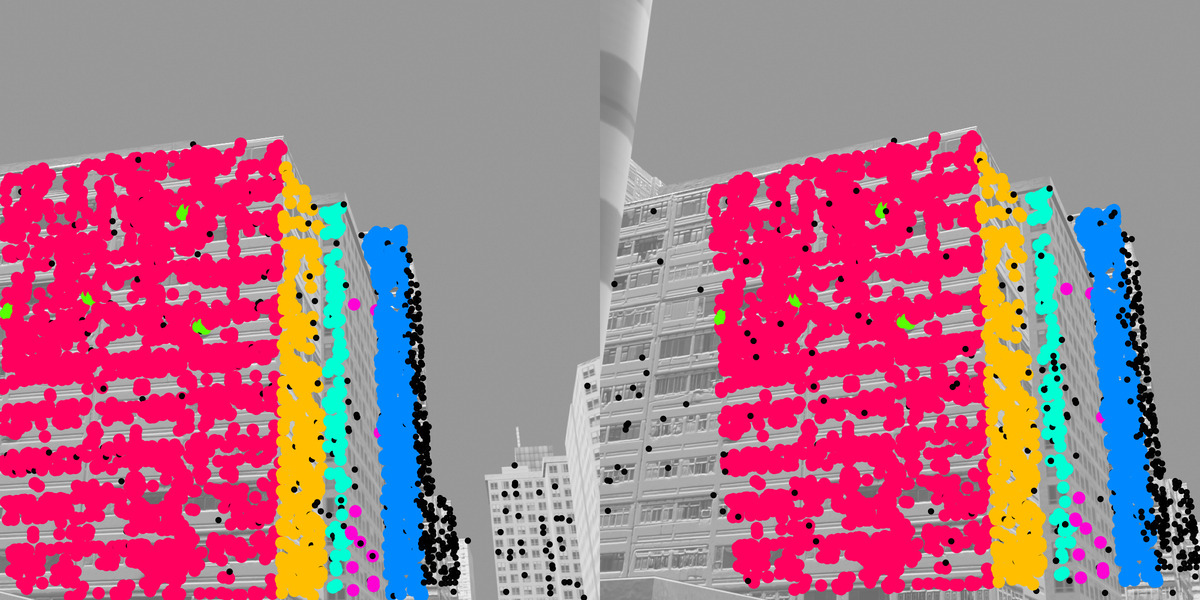}
    
    \includegraphics[width=\imgwidth]{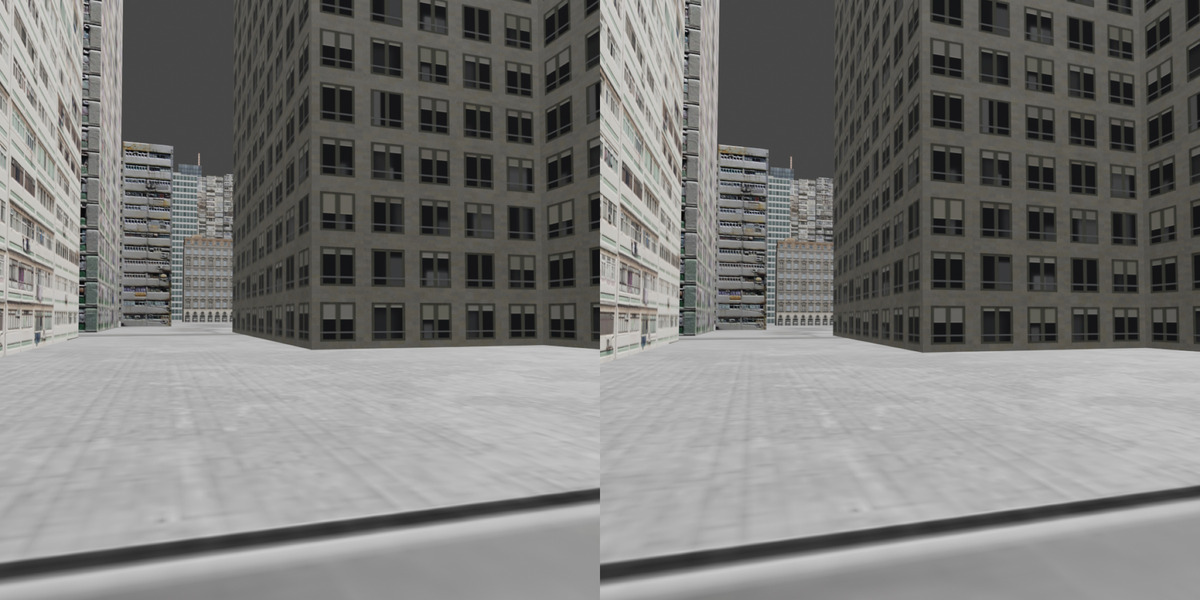}%
    \includegraphics[width=\imgwidth]{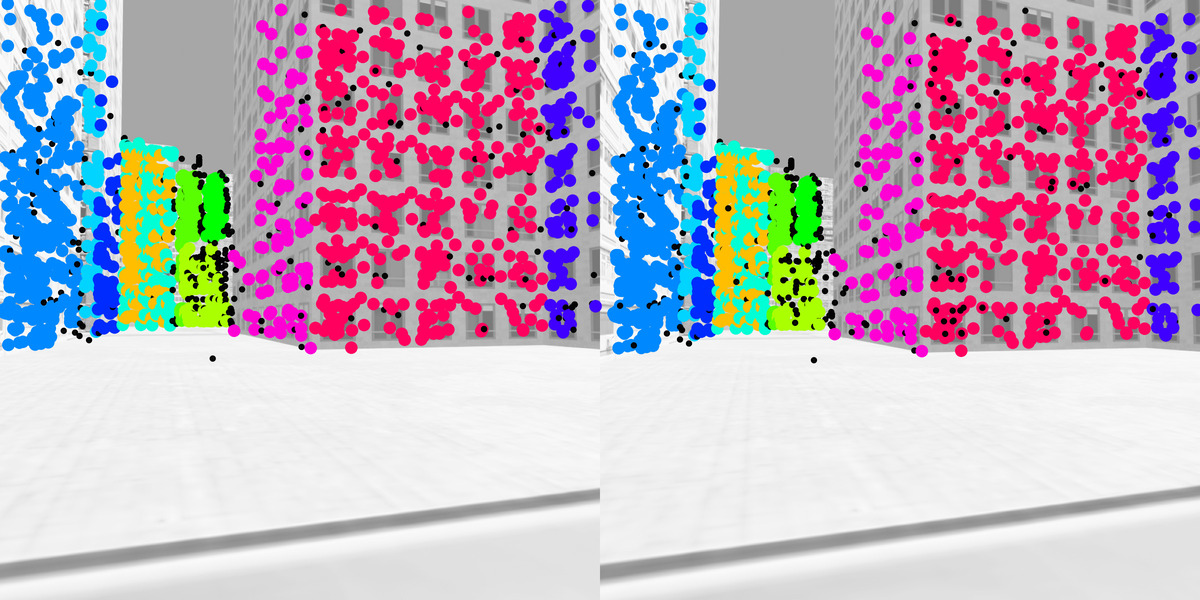}%
    \hfill%
    \includegraphics[width=\imgwidth]{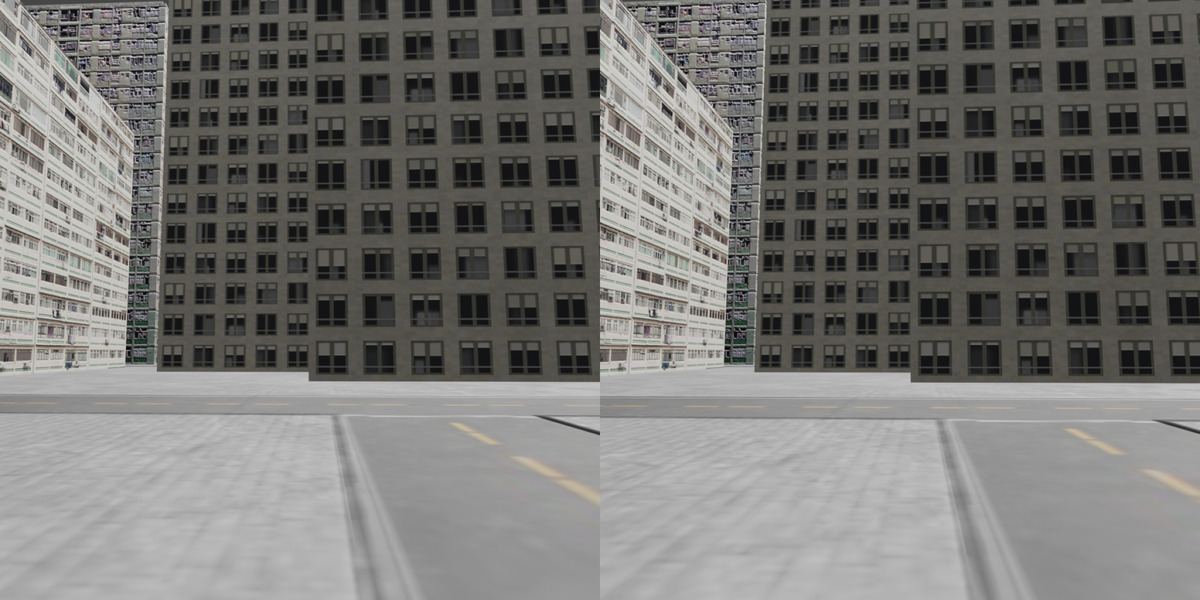}%
    \includegraphics[width=\imgwidth]{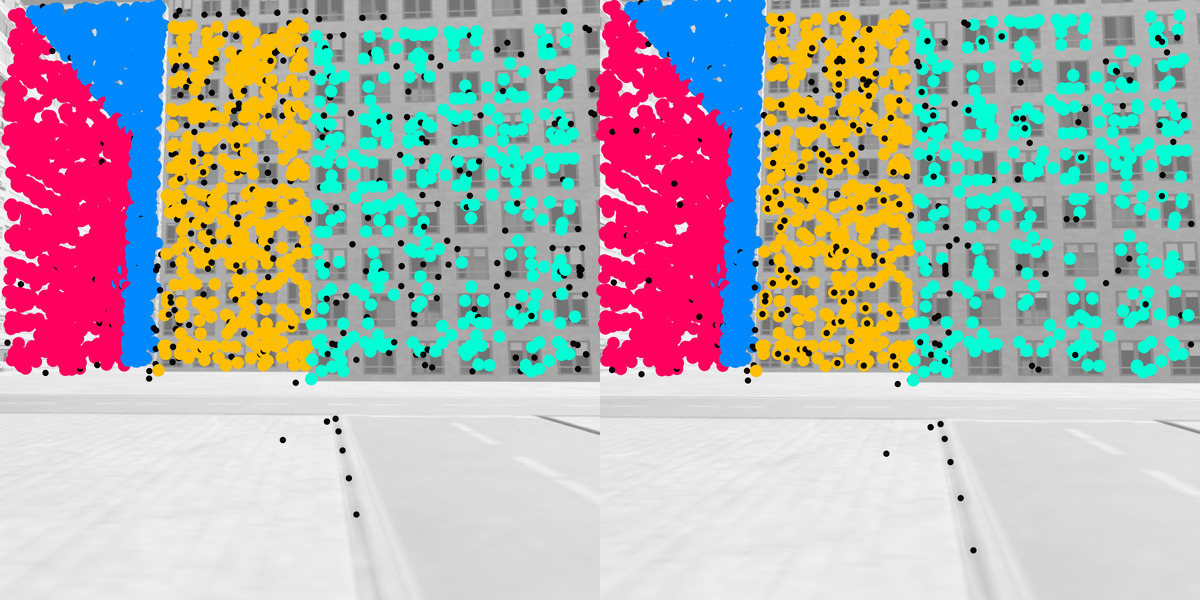}
    
    \includegraphics[width=\imgwidth]{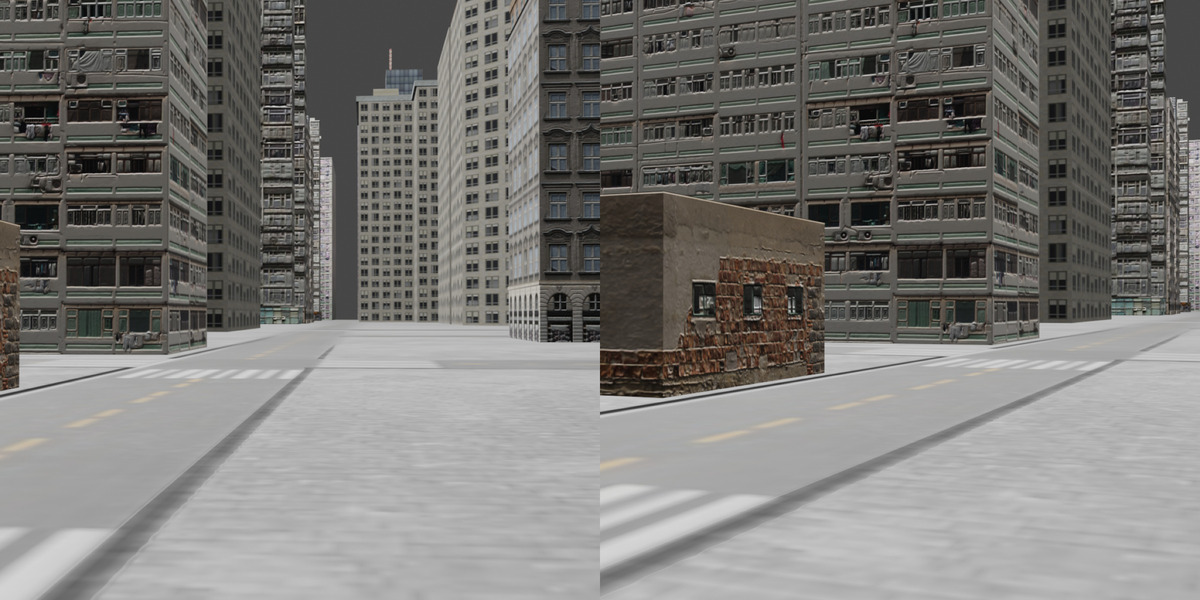}%
    \includegraphics[width=\imgwidth]{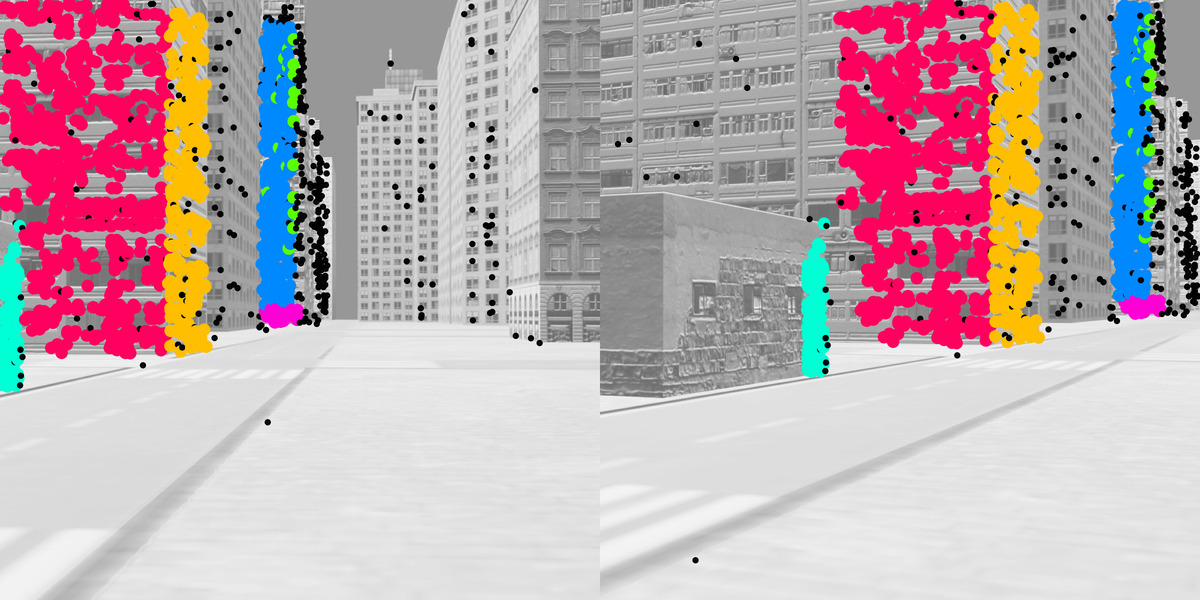}%
    \hfill%
    \includegraphics[width=\imgwidth]{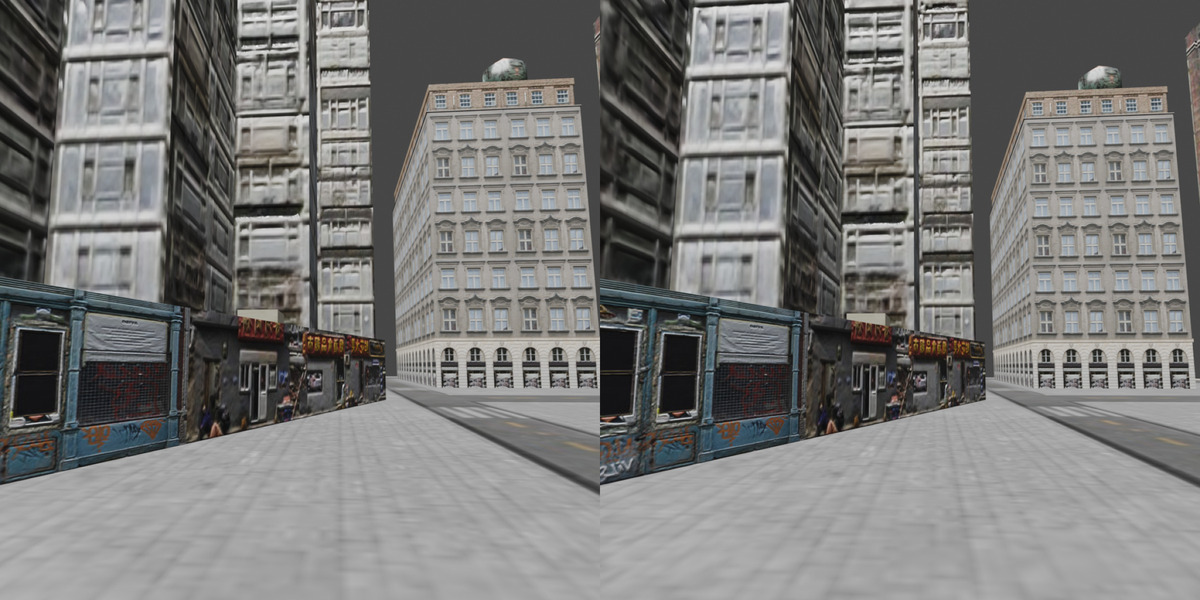}%
    \includegraphics[width=\imgwidth]{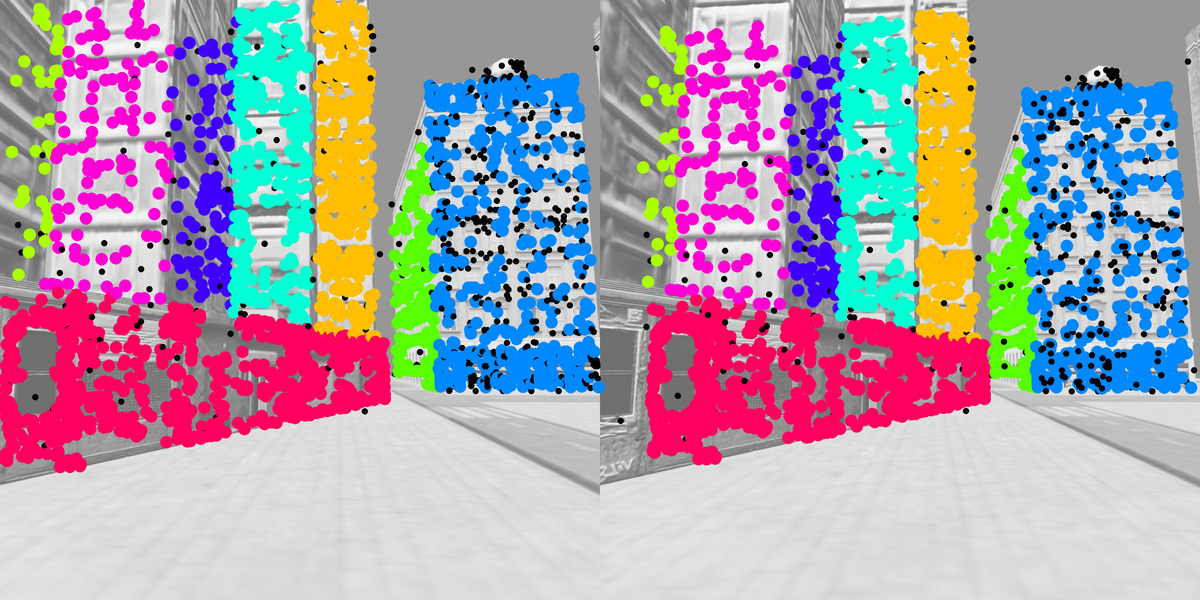}
    
    \includegraphics[width=\imgwidth]{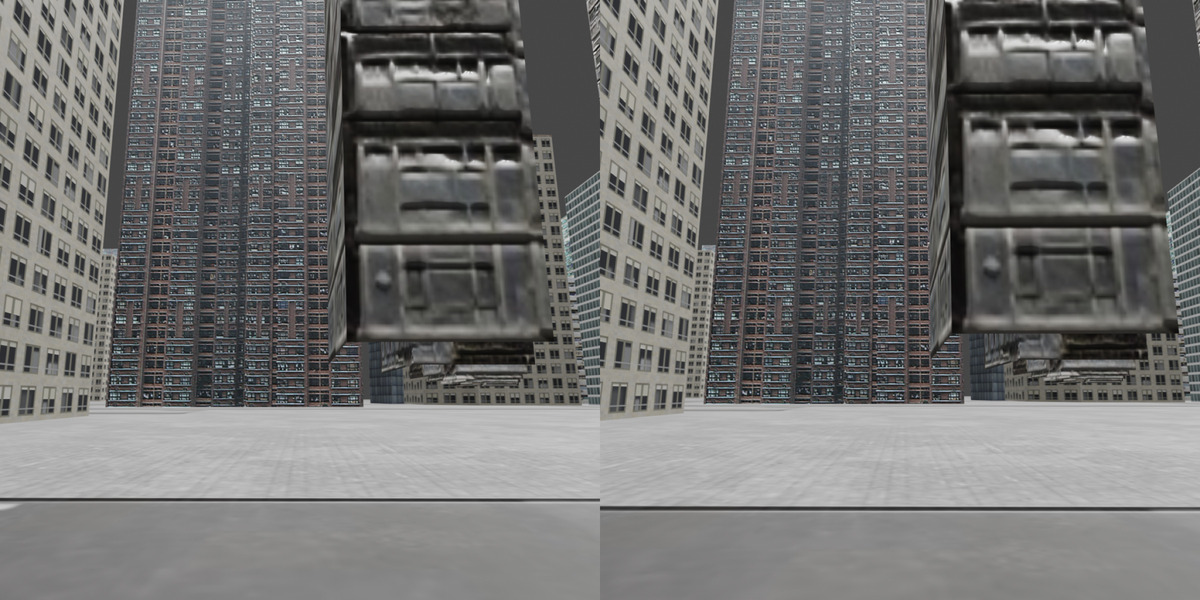}%
    \includegraphics[width=\imgwidth]{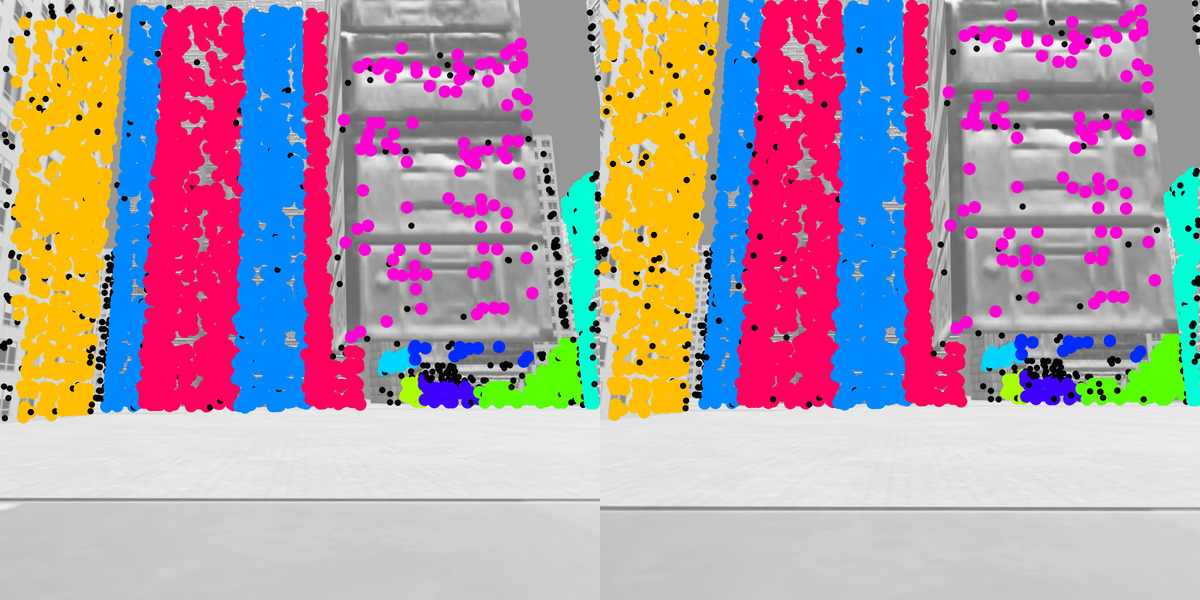}%
    \hfill%
    \includegraphics[width=\imgwidth]{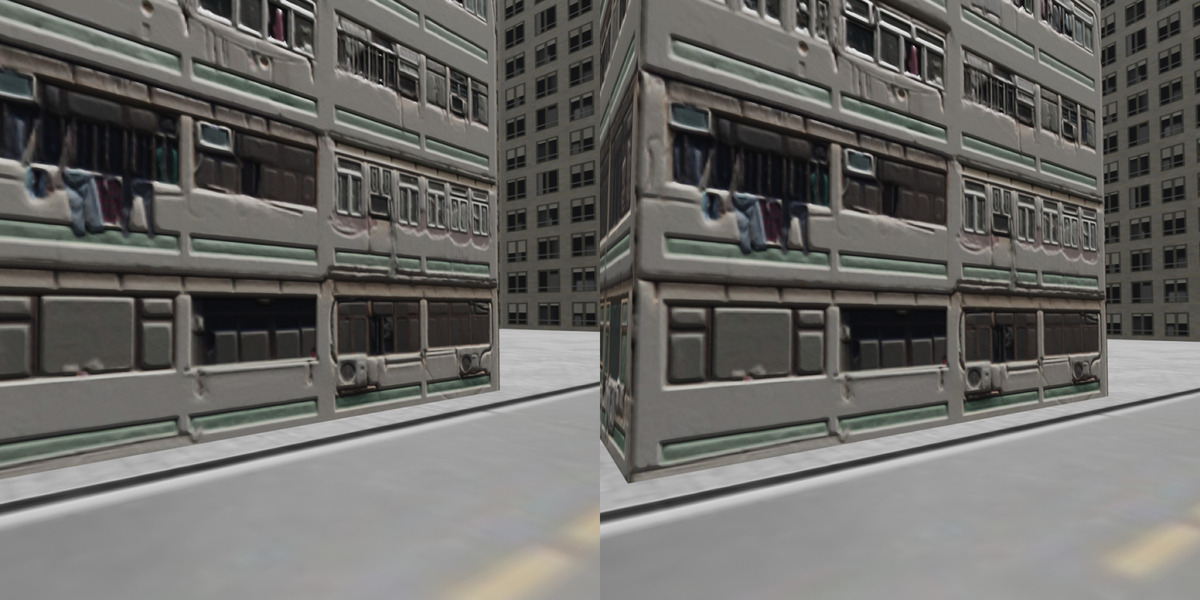}%
    \includegraphics[width=\imgwidth]{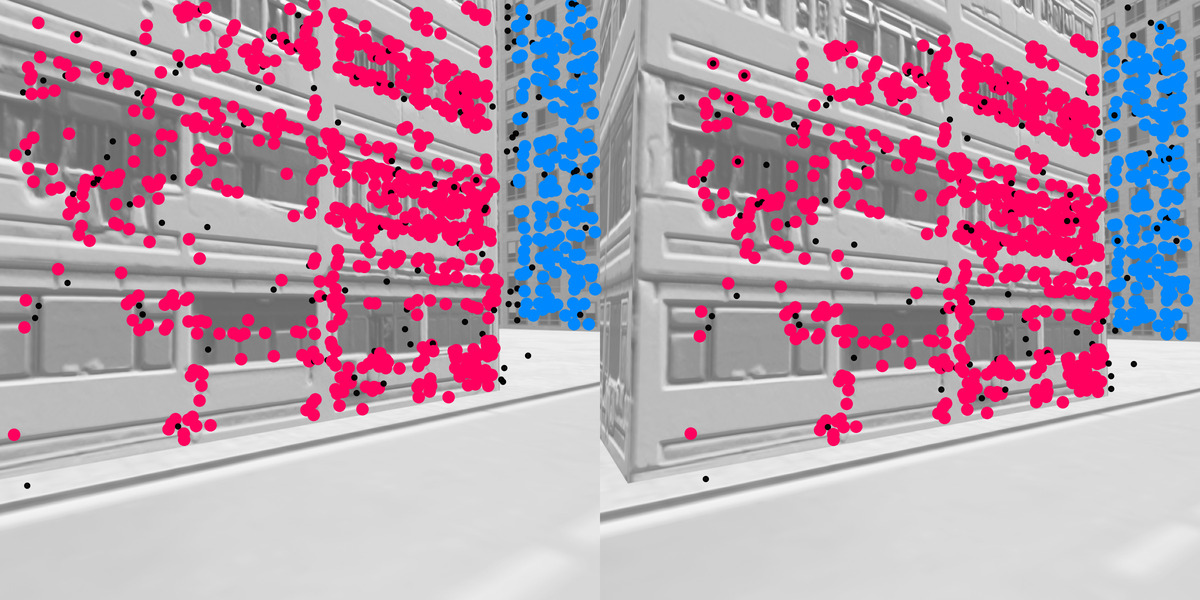}
    
    \includegraphics[width=\imgwidth]{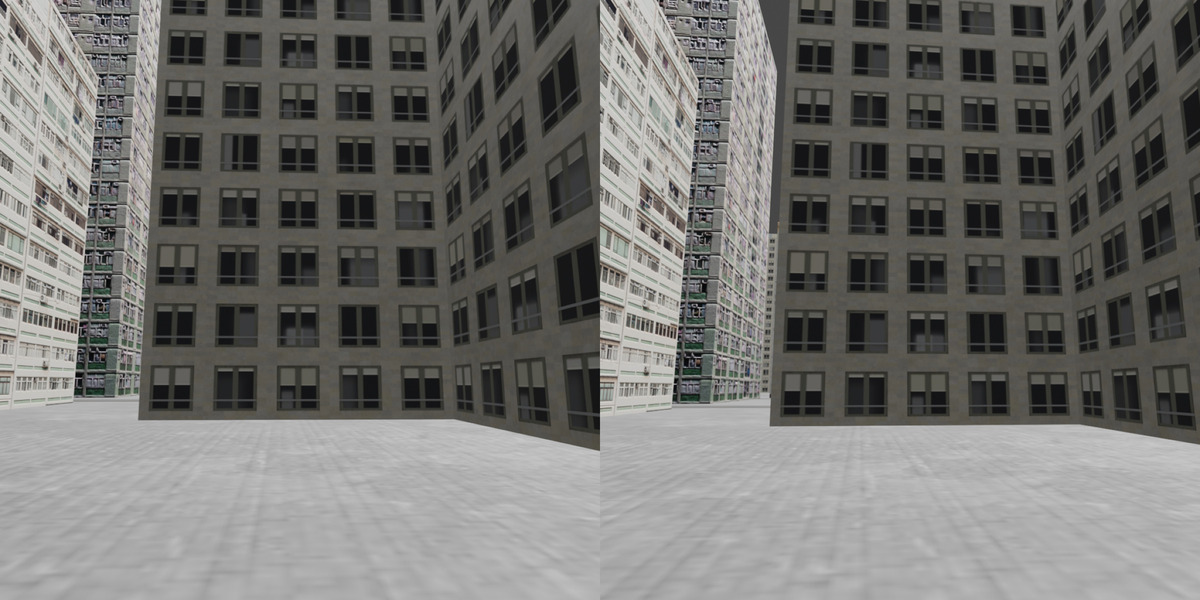}%
    \includegraphics[width=\imgwidth]{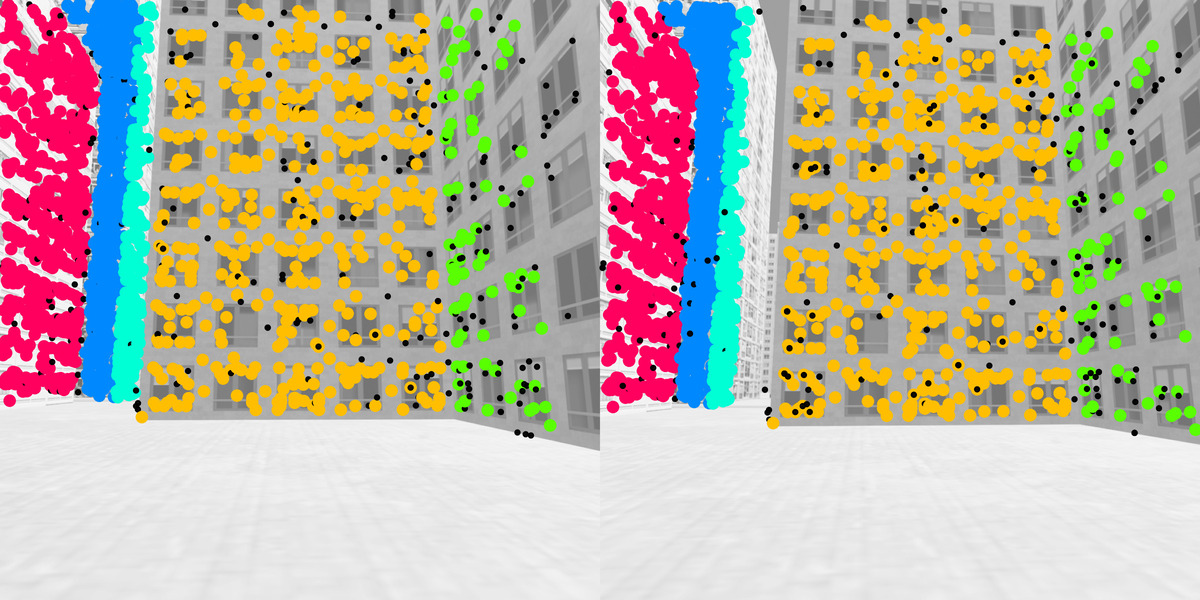}%
    \hfill%
    \includegraphics[width=\imgwidth]{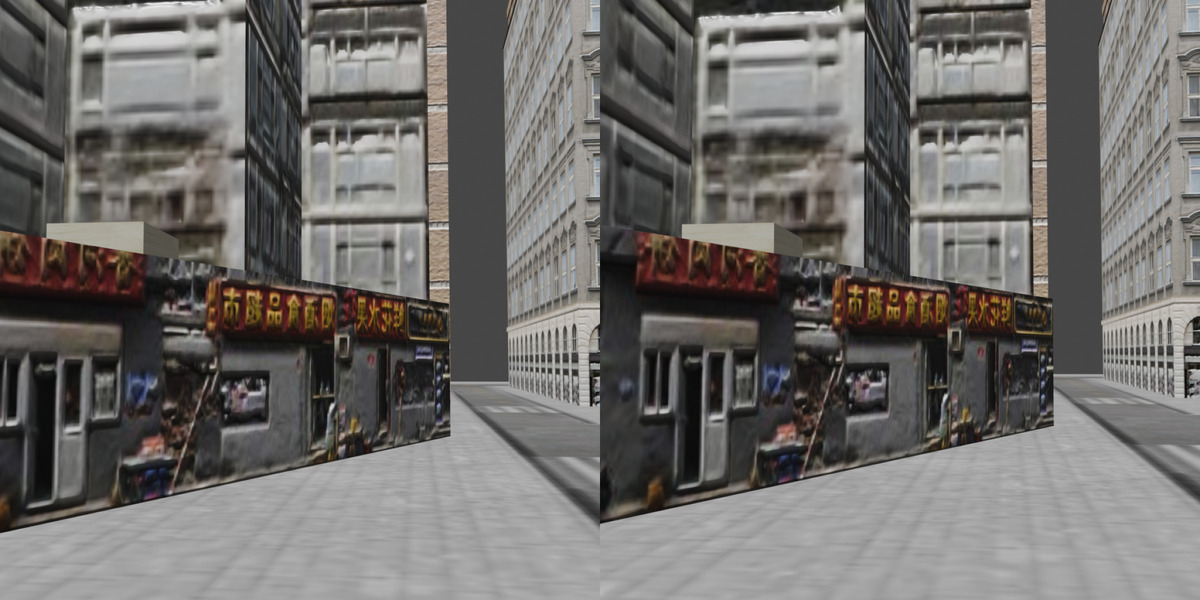}%
    \includegraphics[width=\imgwidth]{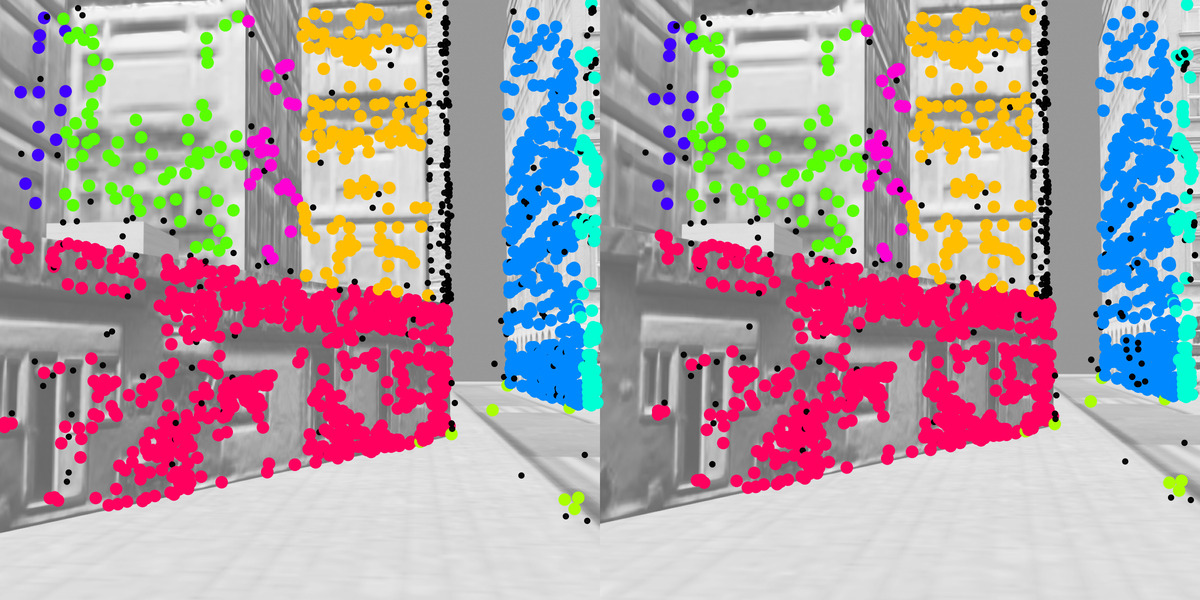}

    \caption{{SMH:} We show additional examples from our new dataset for homography fitting. The left pair of each example shows the RGB images, the right pair visualises the pre-computed SIFT keypoints, colour coded by ground truth label.}
		\label{fig:more_smh_examples}
\end{figure*}

Fig.~\ref{fig:datasets:h_outliers} compares the distribution of outlier ratios between SMH and Adelaide-H.
The average outlier ratio of Adelaide-H is higher than that of SMH, but the latter contains a broader range of outlier ratios per scene.
Fig.~\ref{fig:datasets:h_points} shows that the number of SIFT point correspondences per scene also varies more strongly -- and is much larger on average -- in SMH compared to Adelaide-H.
We furthermore look at the percentage of inliers for each individual fundamental matrix in Fig.~\ref{fig:datasets:h_inliers}.
The models in SMH have inlier ratios between $0.07\%$ and $99\%$, compared to $2-54\%$ in Adelaide-H.
Fig.~\ref{fig:more_smh_examples} shows additional example image pairs from the SMH dataset.

\begin{figure}
\centering
\includegraphics[width=\linewidth]{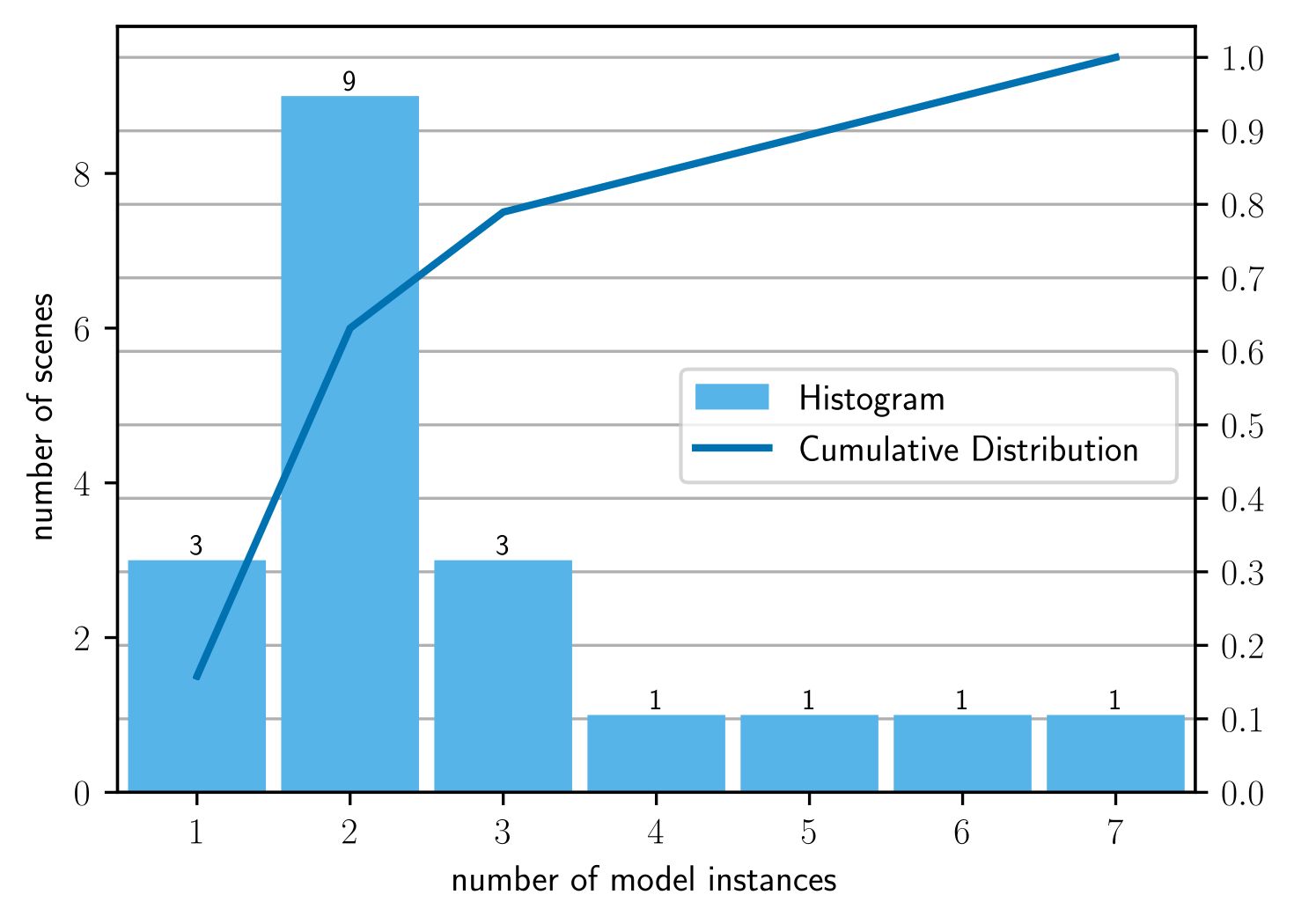}
\caption{Distribution of homographies in the Adelaide-H dataset.}
\label{fig:datasets:adelaideh_models}
\end{figure}

%% file: algo/parsac.tex
\begin{algorithm}
\SetAlgoLined
\DontPrintSemicolon
\caption{PARSAC}
\label{algo:parsac}
\SetKwInput{KwData}{Parameters}
\KwIn{$\obsvs$ -- set of observations}
\KwData{\mbox{$\nnw$ -- neural network parameters,} 
\mbox{$\numpmodels$ -- number of putative model instances,}  
\mbox{$S$ -- number of hypotheses,} 
\mbox{$\tau$ -- inlier threshold,} 
\mbox{$\beta$ -- inlier softness,}   
\mbox{$\tau_a$ -- assignment threshold,} 
\mbox{$C$ -- minimal set size} 
}
\KwOut{\mbox{$\models$ -- sequence of models, $\labels$ -- cluster labels}}

$\pmodels \leftarrow \varnothing$ \;


\For(\tcp*[f]{Find putative models}){$j\leftarrow 1$ \KwTo $\numpmodels$}{
    $\hypotheses \leftarrow \varnothing$ \;
    
    \For(\tcp*[f]{Generate hypotheses}){$l\leftarrow 1$ \KwTo $S$}{

        $\minimalset \leftarrow \varnothing$
        
        \For(\tcp*[f]{Sample minimal set}){$i\leftarrow 1$ \KwTo $C$}{
            $\obsv \sim \psweight$ \;
                
            $\minimalset \leftarrow \minimalset \cup \{\obsv\}$ \;
        }
        
        $\hypothesis \leftarrow \minsolver(\minimalset)$ \tcp*{Solve for hypothesis} 
        
        $\hypotheses \leftarrow \hypotheses \cup \{\hypothesis\} $ \;
    }
    \tcc{Select best hypothesis via weighted inlier counting:}
    
    $\model \leftarrow \argmax_{\hypothesis \in \hypotheses} 
        I_{\text{w}}(\hypothesis; \obsvs, \nnw, \tau, \beta)$ \;
    
    $\pmodels \leftarrow \pmodels \cup \{\model\} $ \;
    
}

$\models \leftarrow \text{InstanceRanking}(\pmodels, \obsvs, \tau, C)$ \tcp*{Alg.~\ref{algo:parsac:instance_ranking}}

$\labels \leftarrow \text{ClusterAssignment}(\models, \obsvs, \tau, \tau_a)$ \tcp*{Alg.~\ref{algo:parsac:cluster_assignment}}

\end{algorithm}

%% file: algo/ranking.tex
\begin{algorithm}
\SetAlgoLined
\SetNoFillComment
\DontPrintSemicolon
\setcounter{AlgoLine}{0}
\caption{Instance Ranking}
\label{algo:parsac:instance_ranking}
\SetKwInput{KwData}{Parameters}
\KwIn{\mbox{$\pmodels$ -- set of putative model instances, } \mbox{$\obsvs$ -- set of observations}}
\KwData{\mbox{$\tau$ -- inlier threshold, $C$ -- minimal set size}}
\KwOut{$\models$ -- sequence of model instances}

$\models \leftarrow \emptyseq$ \;

$\set{I} \leftarrow \varnothing$ \;

\While{$|\pmodels| > 0$}{
    
    $\vec{d} \leftarrow \vec{0}$ \;
    
    \For{$\model_j \in \pmodels$} {  
        \tcp{Inliers of putative model $\model_j$:}
        
        $\set{I}_{\model_j} \leftarrow \{ \obsv \, | \, d(\obsv, \model_j) < \tau, \obsv \in \obsvs \} \, ,$ \;

        \tcp{Unique and overlapping inliers:}

        $I^{\text{u}} \leftarrow | \set{I}_{\model_j} \setminus \set{I}|$ \;
        
        $I^{\text{o}} \leftarrow | \set{I}_{\model_j} \cap \set{I}|$ \;

        $d_j \leftarrow I^{\text{u}} - I^{\text{o}}$ \; 

    }

    $k \leftarrow \argmax_{j} d_j  $ \tcp*{Select best model} 
    
    \uIf{$d_k \geq C$}{
        \tcp{Append model to output:}
        $\models \leftarrow \models \frown (\model_k) $ \;

        \tcp{Remove model from input set:}
        $\pmodels \leftarrow \pmodels \setminus \{\model_k\} $ \;

        \tcp{Update set of inliers:}
        $\set{I} \leftarrow \set{I} \cup \{ \obsv \, | \, d(\obsv, \model_k) < \tau, \obsv \in \obsvs \} $
    }
    \Else{
        break \;
    }
    
}

\end{algorithm}

%% file: algo/clustering.tex
\begin{algorithm}
\SetAlgoLined
\SetNoFillComment
\DontPrintSemicolon
\setcounter{AlgoLine}{0}
\caption{Cluster Assigment}
\label{algo:parsac:cluster_assignment}
\SetKwInput{KwData}{Parameters}
\SetKw{And}{and}
\SetKw{Or}{or}
\KwIn{\mbox{$\models$ -- sequence of model instances, } \mbox{$\obsvs$ -- set of observations}}
\KwData{\mbox{$\tau$ -- inlier threshold, $\tau_a$ -- assigment threshold}}
\KwOut{$\labels$ -- set of cluster labels}

$\labels \leftarrow \varnothing$ \;

\For{$\obsv_i \in \obsvs$}{

    $y_i \leftarrow 0$ \tcp*{Initialise as outlier class}

    \tcc{Assign observation to model with smallest distance if it is below inlier threshold:}
    \uIf{$\min_{\model_j \in \models} d(\obsv_i,\model_j) < \tau$}{
        $y_i \leftarrow \argmin_{j \in \{1,\dots,|\models| \}} d(\obsv_i,\model_j) $
    }
    \Else{
    \For{$\model_j \in \models$} {  
        \tcc{Otherwise assign to first model with residual below assignment threshold:}
        \If{$d(\obsv_i,\model_j) < \tau_a$} {
            $y_i \leftarrow j$ \;

            break \;
        }
    }
    }
    $\labels \leftarrow \labels \cup \{y_i\} $    
        
}
\end{algorithm}

%% file: tables/parameters.tex
\begin{table*}[t]
\setlength\tabcolsep{.35em}
    \centering
    \begin{tabular}{c|lc|c|c|c|c|}
        \multicolumn{3}{c|}{}                           & vanishing & fundamental & \multicolumn{2}{c|}{homographies} \\
        \multicolumn{3}{c|}{}                           & points & matrices & SMH & Adelaide \\
        \hline
        \parbox[t]{3mm}{\multirow{3}{*}{\rotatebox[origin=c]{90}{general}}}
        &  minimal set size        & $C$      & $2$       & $7$             & \multicolumn{2}{c|}{$4$} \\
        &  putative instances      & $\numpmodels$& $8$       & $4$             & \multicolumn{2}{c|}{$24$} \\
        &  assignment threshold    & $\tau_a$ & -- & $2\cdot10^{-2}$ & $4\cdot 10^{-6}$& $4 \cdot 10^{-3}$\\
        \hline
        \parbox[t]{3mm}{\multirow{10}{*}{\rotatebox[origin=c]{90}{training}}}
        &  inlier threshold        & $\tau$   & $10^{-4}$ & $4 \cdot 10^{-3}$ & $10^{-6}$       & $10^{-4}$\\
        &  learning rate           &          & $10^{-4}$ & $10^{-4}$       & $10^{-4}$       & -- \\
        &  softmax scale factor    & $\alpha_s$& $1000$ & $1000$       & $1000$       & -- \\
        &  epochs                  & $N_e$    & $2000$    & $3000$          & $500$           & -- \\
        &  LR reduction            & $N_{lr}$ & $1500$    & $2500$          & $350$           & -- \\
        &  batch size              & $B$      & $64$      & $32$            & $4$             & -- \\
        &  hypotheses set samples  & $\numsamples$    & $8$       & $16$            & $8$             & -- \\
        &  model set samples       & $\numhypsamples$    & $64$      & $128$           & $64$            & -- \\
        &  model hypotheses        & $\numhypotheses$      & $32$      & $32$            & $32$            & -- \\
        &  max. input observations & $\numobsvs$& $512$     & $512$           & $512$           & -- \\
        \hline
        \parbox[t]{3mm}{\multirow{3}{*}{\rotatebox[origin=c]{90}{test}}}
        &  inlier threshold        & $\tau$   & $10^{-4}$ & $10^{-2}$ & $10^{-6}$       & $10^{-4}$\\
        &  model hypotheses        & $\numhypotheses$      & $32$      & $128$           & $512$           & $512$ \\
        &  max. input observations & $\numobsvs$& \multicolumn{4}{c|}{variable} \\
        
    \end{tabular}
    \vspace{.5em}
    \caption{{Hyper-parameters: } we distinguish between parameters used during both training and testing, only for training, and only for testing. Different applications require different parameters. } 
    \label{tab:parameters}
\end{table*}

%% file: tables/results_self.tex
\begin{table*}
\small
\renewcommand{\arraystretch}{1.2} 
	\begin{center}
	\begin{tabular}{|l|cc|cc|cc|cc|cc|cc|}
	\hline
	\multicolumn{1}{|l|}{\textit{Datasets}} &
	\multicolumn{2}{c|}{SU3} &
	\multicolumn{2}{c|}{NYU-VP} &
	\multicolumn{2}{c|}{HOPE-F} &
	\multicolumn{2}{c|}{Adelaide-F} \\
        \hline
	\multicolumn{1}{|l|}{\textit{Metrics}} & 
	\multicolumn{2}{c|}{AUC @ $5\degree$} &
	\multicolumn{2}{c|}{AUC @ $10\degree$} &
	\multicolumn{2}{c|}{ME} &
	\multicolumn{2}{c|}{ME} \\
	\hline

	\multicolumn{1}{|l|}{\textbf{PARSAC} (supervised)}      
	& $90.17 $ & $\scriptstyle \pm 0.11$      
	& $64.58 $ & $\scriptstyle \pm 0.13$      
	& $14.97 $ & $\scriptstyle \pm 8.51$      
	& $9.83 $ & $\scriptstyle \pm 4.17$      \\
 
	\multicolumn{1}{|l|}{\textbf{PARSAC} (self-supervised, weighted)}      
	& $86.14 $ & $\scriptstyle \pm 0.11$      
	& $61.31 $ & $\scriptstyle \pm 0.56$      
	& $16.73 $ & $\scriptstyle \pm 9.65$      
	& $11.47 $ & $\scriptstyle \pm 6.07$      \\
 
	\multicolumn{1}{|l|}{\textbf{PARSAC} (self-supervised, unweighted)}      
	& $85.15 $ & $\scriptstyle \pm 0.14$      
	& $59.06 $ & $\scriptstyle \pm 0.69$      
	& $18.93 $ & $\scriptstyle \pm 11.4$      
	& $11.99 $ & $\scriptstyle \pm 5.37$      \\
	\multicolumn{1}{|l|}{CONSAC~\cite{kluger2020consac} (supervised)}   
	& $85.69 $ & $\scriptstyle \pm 0.06$      
	& $64.25 $ & $\scriptstyle \pm 0.38$      
	& \multicolumn{2}{c|}{--}      
	& \multicolumn{2}{c|}{--}      \\
 
	\multicolumn{1}{|l|}{CONSAC~\cite{kluger2020consac} (self-supervised)}   
	& $82.88 $ & $\scriptstyle \pm 0.07$      
	& $61.61 $ & $\scriptstyle \pm 0.16$      
	& \multicolumn{2}{c|}{--}      
	& \multicolumn{2}{c|}{--}   \\  
 
	\hline
	\end{tabular}
    \end{center}
	\caption{
	{Self-supervised Learning:} 
 We report average AUC values (in \%, higher is better) and their standard deviations over five runs for vanishing point estimation on the SU3 and NYU-VP datasets. For fundamental matrix estimation, we report average misclassification errors (ME in \%, lower is better) and their standard deviations over five runs on HOPE-F and Adelaide-F.
}
	\label{tab:results_self}
\end{table*}

%% file: tables/results_unweighted.tex
\begin{table}
\small
\setlength\tabcolsep{.19em}
\renewcommand{\arraystretch}{1.2} 
	\begin{center}
	\begin{tabular}{|c|cc|cc|cc|cc|}
	\hline
	   weighted  & 
	\multicolumn{4}{c|}{SU3~\cite{zhou2019learning}} &
        \multicolumn{2}{c|}{HOPE-F} &
        \multicolumn{2}{c|}{SMH}\\
        \cline{2-9}
	\multicolumn{1}{|c|}{inliers} & 
	\multicolumn{2}{c|}{AUC @ $3\degree$} &
	\multicolumn{2}{c|}{AUC @ $5\degree$} &
	\multicolumn{2}{c|}{ME} &
	\multicolumn{2}{c|}{ME} \\
	\hline
	\cmark 
	  & $\mathbf{85.73} $ & $\scriptstyle \pm 0.08$ 
        & $\mathbf{90.21} $ & $\scriptstyle \pm 0.08$  
        & $\mathbf{14.97} $ & $\scriptstyle \pm 8.51$  
        & $\mathbf{20.50} $ & $\scriptstyle \pm 15.5$
        \\
	\xmark         
	  & $83.94 $ & $\scriptstyle \pm 0.05$ 
        & $88.97 $ & $\scriptstyle \pm 0.08$  
        & $16.26 $ & $\scriptstyle \pm 9.10$    
        & $21.59 $ & $\scriptstyle \pm 16.2$    \\
  
	\hline
	\end{tabular}
    \end{center}
	\caption{
	We compare the performance of PARSAC using either weighted or unweighted inlier counting. We report AUC values (in \%, higher is better) for vanishing point detection on SU3, as well as misclassification errors (ME in \%, lower is better) for fundamental matrix estimation on HOPE-F and for homography estimation on SMH.
}
	\label{tab:results_unweighted}
\end{table}

%% file: tables/results_deeplsd.tex
\begin{table*}
\small
\setlength\tabcolsep{.42em}
\renewcommand{\arraystretch}{1.2} 
	\begin{center}
	\begin{tabular}{|l|c|c|c|c|c|c|c|c|c|c|c|c|}
	\hline
	\multicolumn{1}{|l|}{\textit{Dataset}} &
	\multicolumn{3}{c|}{SU3 } &
	\multicolumn{3}{c|}{YUD } &
	\multicolumn{3}{c|}{NYU-VP} &
	\multicolumn{3}{c|}{YUD+} \\
        \hline
	\multicolumn{1}{|l|}{\textit{AUC}} & 
	@ $1\degree$ &
	@ $3\degree$ &
	@ $5\degree$ &
	@ $3\degree$ &
	@ $5\degree$ &
	@ $10\degree$ &
	@ $3\degree$ &
	@ $5\degree$ &
	@ $10\degree$ &
	@ $3\degree$ &
	@ $5\degree$ &
	@ $10\degree$ \\
	\hline

	\multicolumn{1}{|l|}{PARSAC\tssstar + LSD }             
	& $67.44 $ & $85.70 $ & $90.17 $
        & $63.92 $ & $75.90 $ & $86.37 $
        & $39.93 $ & $51.65 $ & $64.58 $
        & $54.98 $ & $65.48 $ & $74.74 $ \\
        
	\multicolumn{1}{|l|}{PARSAC\tssstar + DeepLSD }             
	& $63.60 $ & $83.09 $ & $88.12 $
        & $61.37 $ & $72.68 $ & $83.48 $
        & $39.79 $ & $51.05 $ & $63.81 $
        & $52.86 $ & $62.90 $ & $72.79 $ \\
\hline
	\multicolumn{1}{|l|}{PARSAC\tssstard + LSD }             
	& $64.89 $ & $84.64 $ & $89.53 $   
	& $61.80 $ & $74.02 $ & $84.70 $   
	& $39.28 $ & $50.91 $ & $63.53 $   
	& $53.44 $ & $64.12 $ & $73.80 $ \\
 
	\multicolumn{1}{|l|}{PARSAC\tssstard + DeepLSD }             
	& $63.95 $ & $84.30 $ & $89.45 $   
	& $62.98 $ & $74.62 $ & $84.22 $   
	& $38.23 $ & $49.27 $ & $61.61 $   
	& $53.59 $ & $63.70 $ & $73.43 $ \\
\hline
	\multicolumn{1}{|l|}{Progressive-X\tssdagger + LSD}          
	& $63.14 $ & $80.49 $ & $84.82 $  
        & $50.41 $ & $60.10 $ & $68.47 $
        & $38.73 $ & $49.25 $ & $60.71 $
        & $50.13 $ & $60.01 $ & $68.53 $ \\

	\multicolumn{1}{|l|}{Progressive-X\tssdagger + DeepLSD}          
	& $62.70 $ & $80.09 $ & $84.37 $  
        & $51.02 $ & $60.97 $ & $69.58 $
        & $39.31 $ & $49.53 $ & $60.79 $
        & $50.22 $ & $60.13 $ & $68.73 $ \\
\hline        
	\multicolumn{1}{|l|}{Progressive-X\tssddagger + LSD}          
	& $63.41 $ & $80.81 $ & $85.16 $  
        & $51.80 $ & $62.15 $ & $71.52 $
        & $39.88 $ & $49.88 $ & $61.36 $
        & $51.07 $ & $61.53 $ & $71.16 $ \\

	\multicolumn{1}{|l|}{Progressive-X\tssddagger + DeepLSD}          
	& $62.68 $ & $80.06 $ & $84.34 $  
        & $50.12 $ & $60.12 $ & $68.51 $
        & $39.08 $ & $49.32 $ & $60.54 $
        & $50.62 $ & $60.62 $ & $69.06 $ \\

	\hline
	\end{tabular}
    \end{center}
    
	\caption{
	{Generalisation from LSD to DeepLSD Features:} We report mean AUC values (in \%, higher is better) for PARSAC and Progressive-X. 
\tssstar Trained using LSD line segments. \tssstard Trained using DeepLSD line segments. \tssdagger Using parameters optimised with LSD line segments. \tssddagger Using parameters optimised with DeepLSD line segments. Training and parameter optimisation was performed on the SU3 training set for SU3 and YUD(+), and on the NYU-VP train set for NYU-VP. 
}
\label{tab:results_deeplsd}
\end{table*}

%% file: tables/results_instances.tex
\begin{table}
\small
\setlength\tabcolsep{.25em}
\renewcommand{\arraystretch}{1.1} 
	\begin{center}
	\begin{tabular}{|c|c|cc|cc|cc|}
	\hline
	   Model &
	\multicolumn{7}{c|}{SU3~\cite{zhou2019learning}} \\
        \cline{2-8}
	\multicolumn{1}{|l|}{ instances $\numpmodels$} & 
        time &
	\multicolumn{2}{c|}{AUC @ $1\degree$} &
	\multicolumn{2}{c|}{AUC @ $3\degree$} &
	\multicolumn{2}{c|}{AUC @ $5\degree$} \\
	\hline
	{$2$}         
        & $\mathbf{5.70} $
	& $49.21 $ & $\scriptstyle \pm 0.13$
	  & $59.41 $ & $\scriptstyle \pm 0.09$ 
        & $61.69 $ & $\scriptstyle \pm 0.07$  \\
	{$3$}               
        & $\underline{5.71} $       
	& $67.19 $ & $\scriptstyle \pm 0.15$
	  & $84.25 $ & $\scriptstyle \pm 0.06$ 
        & $88.57 $ & $\scriptstyle \pm 0.06$  \\
	{$4$}            
        & $5.75 $          
	& $\underline{67.79} $ & $\scriptstyle \pm 0.12$
	  & $85.47 $ & $\scriptstyle \pm 0.08$ 
        & $89.87 $ & $\scriptstyle \pm 0.07$  \\
	{$6$}            
        & $6.16 $          
	& $\mathbf{67.86} $ & $\scriptstyle \pm 0.04$
	  & $\underline{85.75} $ & $\scriptstyle \pm 0.04$ 
        & $90.14 $ & $\scriptstyle \pm 0.03$  \\
	{$8$}            
        & $6.42 $          
	& $67.45 $ & $\scriptstyle \pm 0.09$
	  & $85.73 $ & $\scriptstyle \pm 0.08$ 
        & $90.21 $ & $\scriptstyle \pm 0.08$  \\
	{$10$}           
        & $6.57 $           
	& $67.19 $ & $\scriptstyle \pm 0.09$
	  & $\mathbf{85.87} $ & $\scriptstyle \pm 0.11$ 
        & $\mathbf{90.41} $ & $\scriptstyle \pm 0.10$  \\
	{$12$}           
        & $6.98 $           
	& $66.79 $ & $\scriptstyle \pm 0.19$
	  & $85.73 $ & $\scriptstyle \pm 0.10$ 
        & $\underline{90.35} $ & $\scriptstyle \pm 0.08$  \\
	{$16$}               
        & $7.24 $       
	& $66.09 $ & $\scriptstyle \pm 0.09$
	  & $85.48 $ & $\scriptstyle \pm 0.06$ 
        & $90.26 $ & $\scriptstyle \pm 0.06$  \\
  
	\hline
	\end{tabular}
    \end{center}
	\caption{
	We vary the number of putative model instances $\numpmodels$ and evaluate its impact on AUC and computation time for vanishing point detection on the SU3 dataset. 
}
	\label{tab:results_instances}
\end{table}